\documentclass[11pt]{article}
\usepackage{amsmath}
\usepackage{amssymb}
\usepackage{amsthm}
\usepackage{amscd}
\usepackage{amsfonts}
\usepackage{dsfont}
\usepackage{graphicx}%
\usepackage{subcaption}
\usepackage{fancyhdr}
\usepackage{empheq}
\usepackage{bm}

\usepackage{cite}
\usepackage{color}
\usepackage[ruled]{algorithm2e}
\usepackage{ifthen}
\usepackage{algorithmic}
\newtheorem{lemma}{Lemma}
\newtheorem{theorem}{Theorem}

\newtheorem{definition}{Definition}

\newtheorem{assumption}{Assumption}
\newtheorem{remark}{Remark}

\usepackage{mathtools}
\DeclarePairedDelimiter\ceil{\lceil}{\rceil}
\DeclarePairedDelimiter\floor{\lfloor}{\rfloor}

\def\scr#1{{\cal #1}} 
\ifodd 1
\def\eq#1{\begin{equation}#1\end{equation}}
\newcommand{\R}{{\rm I\!R}}
\newcommand{\bbb}{\mathbb}

\newcommand{\diag}{{\rm diag}}
\else

\newcommand{\com}[1]{}
\newcommand{\clar}[1]{}
\newcommand{\response}[1]{}
\fi

\newcommand{\1}{\mathbf{1}}
\newcommand{\0}{\mathbf{0}}

\usepackage{hyperref}

\def\qed{ \rule{.08in}{.08in}}

\usepackage{parskip}
\title{Finite-Time Error Bounds for Distributed Linear Stochastic Approximation
}

\author{Yixuan Lin \hspace{.3in} Vijay Gupta \hspace{.3in} Ji Liu\thanks{%This research was supported in part by ONR MURI Grant N00014-16-1-2710, and in part by the Australian Research Council under grants DP-130103610 and DP-160104500, and Data61-CSIRO.
Y.~Lin is with the Department of Applied Mathematics and Statistics at Stony Brook University (\texttt{yixuan.lin.1@stonybrook.edu}).
V.~Gupta is with the Department of Electrical Engineering at University of Notre Dame
(\texttt{vgupta2@nd.edu}).
J.~Liu is with the Department of Electrical and Computer Engineering at Stony Brook University
(\texttt{ji.liu@stonybrook.edu}).
}
}

\begin{document}
\date{}

\maketitle

\thispagestyle{empty}

\vspace{-.15in}

\begin{abstract} 
This paper considers a novel multi-agent linear stochastic approximation algorithm driven by Markovian noise and general consensus-type interaction, in which each agent evolves according to its local stochastic approximation process which depends on the information from its neighbors. The interconnection structure among the agents is described by a time-varying directed graph. While the convergence of consensus-based stochastic approximation algorithms when the interconnection among the agents is described by doubly stochastic matrices (at least in expectation) has been studied, less is known about the case when the interconnection matrix is simply stochastic. For any uniformly strongly connected graph sequences whose associated interaction matrices are stochastic, the paper derives finite-time bounds on the mean-square error, defined as the deviation of the output of the algorithm from the unique equilibrium point of the associated ordinary differential equation. For the case of interconnection matrices being stochastic, the equilibrium point can be any unspecified convex combination of the local equilibria of all the agents in the absence of communication. Both the cases with constant and time-varying step-sizes are considered. In the case when the convex combination is required to be a straight average and interaction between any pair of neighboring agents may be uni-directional, so that doubly stochastic matrices cannot be implemented in a distributed manner, the paper proposes a push-sum-type distributed stochastic approximation algorithm and provides its finite-time bound for the time-varying step-size case by leveraging the analysis for the consensus-type algorithm with stochastic matrices and developing novel properties of the push-sum algorithm.
Distributed temporal difference learning is discussed as an illustrative application.
\end{abstract}

%\vspace{-.1in}

\section{Introduction}

%\vspace{-.1in}

The use of reinforcement learning (RL) to obtain policies that describe solutions to a Markov decision process (MDP) in which an autonomous agent interacting with an unknown environment aims to optimize its long term reward is now standard \cite{sutton2018reinforcement}. Multi-agent RL is useful when a team of agents interacts with an unknown environment or system and aims to collaboratively accomplish tasks involving distributed decision-making. 
{\color{black}
Distributed here implies that agents exchange information only with their neighbors according to a certain communication graph. Recently, many distributed algorithms for multi-agent RL have been proposed and analyzed \cite{zhang2019multi}. The basic result in such works is of the type that if the graph describing the communication among the agents is bi-directional (and hence can be represented by a doubly stochastic matrix), then an algorithm that builds on traditional consensus algorithms converges to a solution in terms of policies to be followed by the agents that optimize the sum of the utility functions of all the agents; further, both finite and infinite time performance of such algorithms can be characterized \cite{doan2019convergence,kaiqing}.

This paper aims to relax the assumption of requiring bi-directional communication among agents in a distributed RL algorithm. This assumption is arguably restrictive and will be violated due to reasons such as packet drops or delays, differing privacy constraints among the agents, heterogeneous capabilities among the agents in which some agents may be able to communicate more often or with more power than others, adversarial attacks, or even sophisticated resilient consensus algorithms being used to construct the distributed RL algorithm. A uni-directional communication graph can be represented through a (possibly time-varying) stochastic -- which may not be doubly stochastic -- matrix being used in the algorithm. As we discuss in more detail below, relaxing the assumption of a doubly stochastic matrix to simply a stochastic matrix in the multi-agent and distributed RL algorithms that have been proposed in the literature, however, complicates the proofs of their convergence and finite time performance characterizations. The main result in this paper is to provide a finite time bound on the mean-square error for a multi-agent linear stochastic approximation algorithm in which the agents interact over a time-varying directed graph characterized by a stochastic matrix. This paper, thus, extends the applicability of distributed and multi-agent RL algorithms presented in the literature to situations such as those mentioned above where bidirectional communication at every time step cannot be guaranteed. As we shall see, this extension is technically challenging and requires new proof techniques that may be of independent interest for the theory of distributed optimization and learning.
}

%\input{related_work}
%\paragraph{Related Work}
%We provide a detailed related literature review in Appendix~\ref{sec:relatework}.
%Reinforcement learning (RL) at its core is a collection of methods for determining approximate solutions to Markov decision processes (MDPs) in which a learning agent interacts with an unknown environment or system, receiving rewards based on the actions it chooses to perform, with the goal of finding a policy that maximizes its long-term reward. 

% \vspace{.1in}

%\paragraph{Related Work.}

%{\bf Related Work \;} 
%(See Appendix~\ref{relatedwork} for more related work) 

Stochastic approximation is a family of model-free stochastic algorithms tailored for seeing the extrema of unknown functions via noisy observations only \cite{robbins1951stochastic}. It is a key tool for designing and analyzing RL algorithms, including temporal difference (TD) learning as a special case~\cite{sutton2018reinforcement}. Convergence study of stochastic approximation based on ordinary differential equation (ODE) methods has a long history \cite{borkar2000ode}. Notable examples are~\cite{tsitsiklis1997analysis,dayan1992convergence} which prove asymptotic convergence of TD($\lambda$). Recently, finite-time performance of single-agent stochastic approximation and TD algorithms has been studied in \cite{dalal2018finite,lakshminarayanan2018linear,bhandari2018finite,Srikant,gupta2019finite,wang2017finite,ma2020variance,xu2019two,chen2020finite}; many other works have now appeared that perform finite-time analysis for other RL algorithms, see, e.g.,~\cite{zou,qu2020finite,wu2020finite,xu2019finite,weng2020mean,wang2020finite_gq,iclr_gq,chen2020explicit,wang2019multistep,dalal2018finite_2,2timeSA}, just to name a few.  
%%In this direction, our work provides both convergence results and finite time analysis of a distributed RL algorithm involving coupled stochastic approximation processes being carried out at multiple interacting nodes, without assuming either i.i.d. noise or a projection step. 
Many distributed multi-agent 
RL algorithms have been proposed in the literature \cite{zhang2019multi}. In this setting, each agent can receive information only from its neighbors, and no single agent can solve the problem alone or by `taking the lead'. 
%A backbone of almost all distributed RL algorithms proposed in the literature is the consensus-type interaction among the agents, dating back at least to~\cite{Ts3}. 
Many works have analyzed asymptotic convergence of such RL algorithms using ODE methods \cite{zhang2019distributed,kaiqing,wes,zhang2018networked,Yixuan}. This can be viewed as an application of ideas from distributed stochastic approximation~\cite{kushner87,stankovic2010decentralized,huang2012stochastic,stankovic2016multi,bianchi2013performance,stankovic2016distributed}. Finite-time performance guarantees for distributed RL have also been provided in works, most notably in~\cite{doan2019convergence,doan2019finite,wang2020decentralized,zhang2018finite,sun2020finite,zeng2020finite}.

The assumption that is the central concern of this paper and is made in all the existing finite-time analyses for distributed RL algorithms is that the consensus interaction is characterized by doubly stochastic matrices \cite{doan2019convergence,doan2019finite,wang2020decentralized,zhang2018finite,sun2020finite,zeng2020finite} at every time step, or at least in expectation \cite{bianchi2013performance}. 
In a realistic network, especially with mobile agents such as autonomous vehicles, drones, or robots, uni-directional communication is inevitable due to various reasons such as asymmetric communication and privacy constraints, non-zero communication failure probability between any two agents at any given time, and application of resilient consensus in the presence of adversary attacks \cite{vaidya2012iterative,leblanc2013resilient}, all leading to an interaction among the agents characterized by a stochastic matrix, which may further be time-varying. The problem of design of distributed RL algorithms with time-varying stochastic matrices and characterizing either their asymptotic convergence or finite time analysis remains open.
%{\color{red} Technical challenges in removing the assumption of doubly stochastic matrices are discussed in detail in \cite[Section 1]{lin2021finite}.}

{\color{black} 

As a step towards solving this problem, we propose a novel distributed stochastic approximation algorithm and provide its convergence analyses when a time-dependent stochastic matrix is being used due to uni-directional communication in a dynamic network. One of the first guarantees to be lost as the assumption of doubly stochastic matrices is removed is that the algorithm converges to a ``policy'' that maximizes the sum of reward functions of all the agents. Instead, the convergence is to a set of policies that optimize a convex combination of the network-wise accumulative reward, with the exact combination depending on the limit product of the infinite sequence of stochastic matrices. Nonetheless, by defining the error as the deviation of the output of the algorithm from the eventual equilibrium point, we derive finite-time bounds on the mean-square error. We consider both the cases with constant and time-varying step sizes. In the important special case where the goal is to optimize the average of the individual accumulative rewards of all the agents, we provide a distributed stochastic approximation algorithm, which builds on the push-sum idea \cite{pushsum} that has been used to solve distributed averaging problem over strongly connected graphs,
%This paper proposes a novel way of embedding the push-sum idea in distributed linear stochastic approximation and TD learning, which enables us to establish finite-time performance bounds for these push-type algorithms by leveraging our analysis tools for consensus-type algorithms with general stochastic matrices. 
%Our first contribution is to provide a convergence proof for the algorithm when a stochastic matrix is being used due to unidirectional communication. Unfortunately, it turns out that we are no longer able to guarantee convergence to a policy that maximizes the sum of the rewards of the agents. Instead, we show that the resulting policy maximizes a convex combination of these rewards. Our major contribution is to provide an alternate algorithm, based on push-type distributed temporal difference learning, that converges to a policy that maximizes the sum of the rewards of the agents, in spite of unidirectional communication. We also
and characterize its finite-time performance. Thus, this paper provides the first distributed algorithm that can be applied (e.g., in TD learning) to converge to the policy maximizing the team objective of the sum of the individual utility functions over time-varying, uni-directional, communication graphs, 
%when a stochastic matrix characterizes the communication among the agents, 
and characterizes the finite-time bounds on the mean-square error of the algorithm output from the equilibrium point under appropriate assumptions.

{\bf Technical Innovation and Contributions \;}
There are two main technical challenges in removing the assumption of doubly stochastic matrices being used in the analysis of distributed stochastic approximation algorithms. The first is in the direction of finite-time analysis. For distributed RL algorithms, finite-time performance analysis essentially boils down to two parts, namely bounding the consensus error and bounding the ``single-agent’’ mean-square error. For the case when consensus interaction matrices are all doubly stochastic, the consensus error bound can be derived by analyzing the square of the 2-norm of the deviation of the current state of each agent from the average of the states of the agents. With consensus in the presence of doubly stochastic matrices, the average of the states of the agents remains invariant. Thus, it is possible to treat the average value as the state of a fictitious agent to derive the mean-square consensus error bound with respect to the limiting point. More formally, this process relies on two properties of a doubly stochastic matrix $W$, namely that (1) $\1^\top W =\1^\top$, and (2) if $x_{t+1}=Wx_t$, then $\|x_{t+1} - (\1^\top x_{t+1})\1\|_2 \le \sigma_2(W) \|x_{t} - (\1^\top x_{t})\1\|_2$ where $\sigma_2(W)$ denotes the second largest singular value of $W$ (which is strictly less than one if $W$ is irreducible). Even if the doubly stochastic matrix is time-varying (denoted by $W_t$), property (1) still holds and property (2) can be generalized as in \cite{nedic2018network}. Thus, the square of the 2-norm $\|x_{t} - (\1^\top x_{t})\1\|_2^2$ is a quadratic Lyapunov function for the average consensus processes. Doubly stochastic matrices in expectation can be treated in the same way by looking at the expectation. 
This is the core on which all the existing finite-time analyses of distributed RL algorithms are based. 

However, if each consensus interaction matrix is stochastic, and not necessarily doubly stochastic, the above two properties may not hold. In fact, it is well known that quadratic Lyapunov functions for general consensus processes $x_{t+1}=S_tx_t$, with $S_t$ being stochastic, do not exist \cite{olshevsky2008nonexistence}. This breaks down all the existing analyses and provides the first technical challenge that we tackle in this paper. Specifically, we appeal to the idea of quadratic comparison functions for general consensus processes. This was first proposed in \cite{touri2012product} and makes use of the concept of ``absolute probability sequences''. We provide a general analysis methodology and results that subsume the existing finite-time analyses for single-timescale distributed linear stochastic approximation and TD learning as special cases. 

The second technical challenge arises from the fact that with stochastic matrices, the distributed RL algorithms may not converge to the policies that maximize the average of the utility functions of the agents. To regain this property, we propose a new algorithm that utilizes a push-sum protocol for consensus. However, finite-time analysis for such a push-based distributed algorithm is challenging. 
Almost all, if not all, the existing push-based distributed optimization works build on the analysis in \cite{nedic}; however, that analysis assumes that a convex combination of the entire history of the states of each agent (and not merely the current state of the agent) is being calculated. This assumption no longer holds in our case.  To obtain a direct finite-time error bound without this assumption, we propose a new approach to analyze our push-based distributed algorithm by leveraging our consensus-based analyses to establish direct finite-time error bounds for stochastic approximation. 
Specifically, we tailor an ``absolute probability sequence'' for the push-based stochastic approximation algorithm and exploit its properties. Such properties have never been found in the existing literature and may be of independent interest for analyzing any push-sum based distributed algorithm. }

We propose a novel consensus-based distributed linear stochastic approximation algorithm driven by Markovian noise in which each agent evolves according to its local stochastic approximation process and the information from its neighbors. We assume only a (possibly time-varying) stochastic matrix being used during the consensus phase, which is a more practical assumption when only unidirectional communication is possible among agents. We establish both convergence guarantees and finite-time bounds on the mean-square error, defined as the deviation of the output of the algorithm from the unique equilibrium point of the associated ordinary differential equation. The equilibrium point can be an ``uncontrollable'' convex combination of the local equilibria of all the agents in the absence of communication. We consider both the cases of constant and time-varying step-sizes. Our results subsume the existing results on convergence and finite-time analysis of distributed RL algorithms that assume doubly stochastic matrices and bi-directional communication as special cases. 
%The algorithm and analysis are then applied to characterize finite-time performance of new distributed TD learning algorithms, including both TD(0) and TD($\lambda$), with linear function approximation, without requiring a projection step or an i.i.d. noise assumption, which estimates an unspecified convex combination of the network-wise discounted accumulative reward. 
In the case when the convex combination is required to be a straight average and interaction between any pair of neighboring agents may be uni-directional, we propose a push-type distributed stochastic approximation algorithm and establish its finite-time performance bound. 
It is worth emphasizing that it is straightforward to extend our algorithm from the straight average point to any pre-specified convex combination.
Since it is well known that TD algorithms can be viewed as a special case of linear stochastic approximation \cite{tsitsiklis1997analysis}, our distributed linear stochastic approximation algorithms and their finite-time bounds can be  applied to TD algorithms in a straightforward manner.

 \label{sec:introdcution}

%\section{Notation} \label{sec:notation}
%\subsection{Notation} \label{sec:notation}
%{\color{red}add notation used throughout the paper, confirm with me whenever change notation}
%For any positive integer $n$, we use $[n]$ to denote the set $\{1,2,\ldots,n\}$.

%\paragraph{Notation}
{\bf Notation \;}
We use $X_t$ to represent that a variable $X$ is time-dependent and $t\in\{0,1,2,\ldots\}$ is the discrete time index.
The $i$th entry of a vector $x$ will be denoted by $x^i$ and, also, by $(x)^{i}$ when convenient.
The $ij$th entry of a matrix $A$ will be denoted by $a^{ij}$ and, also, by $(A)^{ij}$ when convenient.
We use $\1_n$ to denote the vectors in $\R^n$ whose entries all equal to $1$'s,
and $I$ to denote the identity matrix, whose dimension is to be understood from the context.
%{\color{red} 
%For any vector $x\in\R^n$, we use ${\rm diag}(x)$ to denote the $n\times n$ diagonal matrix whose $i$th diagonal entry equals $x^i$.}
%For any two sets $\mathcal{A}$ and $\mathcal{B}$,
%we use $\mathcal{A}\setminus \mathcal{B}$ to denote the set of elements that are in $\mathcal{A}$ but not in $\mathcal{B}$, and $\mathcal{A} \subset \mathcal{B}$ to denote equality or a proper subset.
Given a set $\mathcal S$ with finitely many elements, we use $|\mathcal S|$ to denote the cardinality of $\mathcal S$. %We use $\|\cdot\|_2$ to denote the 2-norm for vectors and the induced 2-norm for matrices.
%, {\color{red}  and $\|\cdot\|_F$ to denote the Frobenius norm.
%For any positive diagonal matrix $W\in\R^{n\times n}$, we use $\|A\|_W$ to denote the weighted Frobenius norm for $A\in\R^{n\times m}$, defined as $\|A\|_W = \|W^{\frac{1}{2}}A\|_F$. It is easy to see that $\|\cdot\|_W$ is a matrix norm.
%We use $\mathbf{P}(\cdot)$ to denote the probability of an event} and 
%We use $\mathbf{E}(X)$ to denote the expected value of a random variable $X$.
We use $\ceil{\cdot}$ to denote the ceiling function.

A vector is called a stochastic vector if
its entries are nonnegative and sum to one. 
A square nonnegative matrix is called a row stochastic matrix, or simply stochastic matrix, if its row sums all equal one. Similarly, a square nonnegative matrix is called a column stochastic matrix if its column sums all equal one.
A square nonnegative matrix is called a doubly stochastic matrix if its row sums and column sums all equal one.
The graph of an $n\times n$ matrix is a direct graph with $n$ vertices and a directed edge from vertex $i$ to vertex $j$ whenever the $ji$-th entry of the matrix is nonzero.
A directed graph is strongly connected if it has a directed path from any vertex to any other vertex. 
For a strongly connected graph $\bbb G$, the distance from vertex $i$ to another vertex $j$ is the length of the shortest directed path from $i$ to $j$; the longest distance among all ordered pairs of distinct vertices $i$ and $j$ in $\bbb G$ is called the diameter of $\bbb G$.
%The union of two directed graphs, $\bbb G_p$ and $\bbb G_q$, with the same vertex set, written $\bbb G_p \cup \bbb G_q$, is meant the directed graph with the same vertex set and edge set being the union of the edge set of $\bbb G_p$ and $\bbb G_q$. Since this union is a commutative and associative binary operation, the definition extends unambiguously to any finite sequence of directed graphs. 
%with the same vertex set.

\section{Distributed Linear Stochastic Approximation} \label{sec:SA}

The stochastic approximation is a method for approximating the solution of an optimization problem when the objective function is not known, but where only noisy observations are available \cite{harold1997stochastic}. The linear stochastic approximation is a specific form of stochastic approximation that is used to solve linear regression problems with stochastic noise. 

Consider a network consisting of $N$ agents. For the purpose of presentation, we label the agents from $1$ through $N$. The agents are not aware of such a global labeling, but can differentiate between their neighbors. The neighbor relations among the $N$ agents are characterized by a time-dependent directed graph $\bbb{G}_t = (\mathcal{V},\mathcal{E}_t)$ whose
vertices correspond to agents and whose directed edges (or arcs) depict neighbor relations, where $\mathcal{V}=\{1,\ldots, N\}$ is the vertex set and $\mathcal{E}_t = \mathcal{V} \times \mathcal{V}$ is the edge set at time $t$.
Specifically, agent $j$ is an in-neighbor of agent $i$ at time $t$ if $(j,i)\in\mathcal{E}_t$, and similarly, agent $k$ is an out-neighbor of agent $i$ at time $t$ if $(i,k)\in\mathcal{E}_t$.
Each agent can send information to its out-neighbors and receive information from its in-neighbors. Thus, the directions of edges represent the
directions of information flow. For convenience, we assume that each agent is always an in- and out-neighbor of itself, which implies that $\bbb{G}_t$ has self-arcs at all vertices for all time $t$. We use $\mathcal{N}_t^{i}$ and $\mathcal{N}_t^{i-}$ to denote the in- and out-neighbor set of agent $i$ at time $t$, respectively, i.e.,
\begin{align*}
    \mathcal{N}_t^{i} = \{ j \in \mathcal{V}  : ( j, i ) \in \mathcal{E}_t \}, \;\;
    \mathcal{N}_t^{i-} = \{ k \in \mathcal{V}  :  ( i, k ) \in \mathcal{E}_t \}.
\end{align*}
%$$
%    \mathcal{N}_t^{i} = \{ j \in \mathcal{V} \; : \;( j, i ) \in \mathcal{E}_t \}, \;\;\;
%    \mathcal{N}_t^{i-} = \{ k \in \mathcal{V} \; : \; ( i, k ) \in \mathcal{E}_t \}.
%$$
It is clear that $\mathcal{N}_t^{i}$ and $\mathcal{N}_t^{i-}$ are nonempty as they both contain index $i$.

We propose the following distributed linear stochastic approximation over a time-varying neighbor graph sequence $\{\bbb{G}_t\}$.
Each agent $i\in\mathcal{V}$ has control over a random vector $\theta^i_t\in \R^d$ for any $t\in\{0,1,2,\ldots\}$, which is %recursively
updated~by
\begin{align} \label{eq:theta update}
    \theta_{t+1}^i = \sum_{j \in \mathcal{N}_t^i} w_t^{ij} \theta_t^j + \alpha_t \bigg(A(X_t)\sum_{j \in \mathcal{N}_t^i} w_t^{ij}\theta_t^j + b^i(X_t)\bigg),
\end{align}
where $w_t^{ij}$ are consensus weights, $\alpha_t$ is the step-size at time $t$, $A(X_t)\in \R^{d\times d}$ is a random matrix and $b^i(X_t)\in \R^d$ is a random vector, both generated based on the Markov chain $\{ X_t \}$ with state spaces  $\mathcal{X}$. 
It is worth noting that the update \eqref{eq:theta update} of each agent only uses its own and in-neighbors' information and thus is distributed.
%Its comparison with the algorithm in \cite{kushner87} is given in Remark~\ref{remark:ode}.

\begin{remark} \label{remark:ode}
%{\color{red}Yin's distributed stochastic approximation converges to the same limit as ours}
The work of \cite{kushner87} considers a different consensus-based networked linear stochastic approximation for any $i\in\mathcal{V}, t\in\{0,1,2,\ldots\}$ as follows: 
\begin{align}\label{eq:yin}
    \theta_{t+1}^i = \sum_{j \in \mathcal{N}_t^i} w_t^{ij} \theta_t^j + \alpha_t \left(A(X_t)\theta_{t}^i + b^i(X_t)\right),
\end{align}
whose state form is $ \Theta_{t+1} = W_t \Theta_t + \alpha_t \Theta_t A(X_t)^\top + \alpha_t B(X_t)$, and mainly focuses on asymptotically weakly convergence for the fixed step-size case (i.e., $\alpha_t=\alpha$ for all $t$).  
Under the similar set of conditions, with its condition (C3.4') being a stochastic analogy for Assumption~\ref{assum:limit_pi}, Theorem~3.1 in \cite{kushner87} shows that \eqref{eq:yin} has a limit which can be verified to be the same as $\theta^*$, the limit of \eqref{eq:theta update}.
How to apply the finite-time analysis tools in this paper to \eqref{eq:yin} has so far eluded us. 
The two updates 
\eqref{eq:theta update} and \eqref{eq:yin} are analogous to the ``combine-then-adapt'' and ``adapt-then-combine'' diffusion strategies in distributed optimization \cite{chen2012diffusion}.
%They assume that the weight matrix $W_t$ is the aperiodic stochastic matrix and satisfies that the product of $W_t$ will converge to $\pi_{\infty}^\top \1_N$, $\{ \bbb{G}_t \}$ are the repeatedly jointly strongly connected graphs and $\lim_{t-k \to \infty}\mathbf{E}[\Theta A(X_t)^\top + B(X_t)| X_k]  = \Theta A^\top + B$, where $B = [b^1, \cdots, b^N]^\top$. For the case with the fixed step-size, i.e., $\alpha_t=\alpha$, for all $t \ge 0$, define $\Theta^{\alpha}_n = \Theta_t $ for $n\in[ (t-\tau(\alpha))\alpha, (t-\tau(\alpha)+1)\alpha )$. Then, from the Theorem 3.1 in \cite{kushner87}, we know that $\Theta^{\alpha}_n$ will weakly converge to a process $\bar \Theta_t = [\bar \theta_t, \cdots,\bar \theta_t]^\top$, where $\bar \theta_t $ satisfies $ \dot{\bar\theta} = A \bar\theta + \sum_{i=1}^N \pi_\infty^i b^i $ with the initial condition $ \Theta_0^\top \pi_{\infty}$. 
%(2) With a time-varying step-size satisfying Assumption~\ref{assum:step-size}, $\theta_t^i$ will converge to $\theta^*$ w.p.1 for all $i\in\scr V$.
\hfill $\Box$
\end{remark}

We impose the following assumption on the weights $w_t^{ij}$ which has been widely adopted in consensus literature \cite{vicsekmodel,survey,tacrate}. 

\begin{assumption}\label{assum:weighted matrix}
There exists a constant $\beta>0$ such that for all $i,j\in\mathcal V$ and $t$, $w_t^{ij} \ge \beta$ whenever $j\in\mathcal{N}_t^{i}$. For all $i\in\mathcal V$ and $t$, $\sum_{j\in\mathcal{N}_t^{i}} w_t^{ij} = 1$.
\end{assumption}
Let $W_t$ be the $N\times N$ matrix whose $ij$th entry equals $w_t^{ij}$ if $j\in\mathcal{N}_t^i$ and zero otherwise.
From Assumption~\ref{assum:weighted matrix}, each $W_t$ is a stochastic matrix that is compliant with the neighbor graph $\bbb{G}_t$. 
Since each agent $i$ is always assumed to be an in-neighbor of itself, all diagonal entries of $W_t$ are positive. Thus, if $\bbb{G}_t$ is strongly connected, $W_t$ is  irreducible and aperiodic. To proceed, define 
\begin{align*}
    \Theta_t = \left[
    \begin{array}{c}
        (\theta_t^1)^\top  \\
        \vdots \\
        (\theta_t^N)^\top
    \end{array}
    \right], \;\;\;
    B(X_t) = \left[
    \begin{array}{c}
        (b^1(X_t))^\top  \\
        \vdots \\
        (b^N(X_t))^\top
    \end{array}
    \right].
\end{align*}
Then, the $N$ linear stochastic recursions in \eqref{eq:theta update} for any $t\in\{0,1,2,\ldots\}$ can be combined and  written as
\begin{align} \label{eq:updtae_Theta}
    \Theta_{t+1} = W_t \Theta_t + \alpha_t W_t \Theta_t A(X_t)^\top + \alpha_t B(X_t).
\end{align} 
The goal of this section is to characterize the finite-time performance of~\eqref{eq:theta update}, or equivalently~\eqref{eq:updtae_Theta}, with the following standard assumptions, which were adopted e.g. in \cite{Srikant,doan2019convergence}.

\begin{assumption} \label{assum:A and b}
    %Assume the limiting behavior of $\mathbf{E}[A(X_t)]$ and $\mathbf{E}[b^i(X_t)]$ exist, i.e., 
    There exists a matrix $A$ and vectors $b^i$, $i\in\mathcal V$, such that
    \begin{align*}
        \lim_{t\to\infty} \mathbf{E}[A(X_t)] = A, \;\;\;
        \lim_{t\to\infty} \mathbf{E}[b^i(X_t)] = b^i,\;\;\; i\in\mathcal V.
    \end{align*}
    Define $b_{\max} = \max_{i\in\mathcal{V}}\sup_{x\in\mathcal{X}} \| b^i(x) \|_2 < \infty$ and $A_{\max} = \sup_{x\in\mathcal{X}} \| A(x) \|_2 < \infty $.
    Then, $\| A \|_2 \le A_{\max}$ and $\| b^i \|_2 \le b_{\max}$, $i\in\mathcal V$.
\end{assumption}

\begin{assumption} \label{assum:mixing-time}
    Given a positive constant $\alpha$, we use  $\tau(\alpha)$ to denote the mixing time of the Markov chain $\{ X_t \}$ for which
    \begin{align*}
        \left\{ 
        \begin{array}{ll}
            \| \mathbf{E}[A(X_t) - A | X_0 = X] \|_2 \le \alpha, & \forall X, \;\; \forall t \ge \tau(\alpha),\\\\
            \| \mathbf{E}[ b^i(X_t) - b^i | X_0 = X] \|_2 \le \alpha, & \forall X, \;\; \forall t \ge \tau(\alpha), \;\; \forall i\in\mathcal{V}.
        \end{array}
        \right.
    \end{align*}
    The Markov chain $\{ X_t \}$ mixes at a geometric rate, i.e., there exists a constant $C$ such that $\tau(\alpha) \le - C \log \alpha$.
    %In addition, there exists a constant $C$ such that given a small constant $\alpha$ we have
    %\begin{align*}
    %    \tau(\alpha) \le - C \log \alpha.
    %\end{align*}
\end{assumption}

\begin{assumption} \label{assum:lyapunov}
   All eigenvalues of $A$ have strictly negative real parts, i.e., $A$ is a Hurwitz matrix. Then, there exists a symmetric positive definite matrix $P$, such that $A^\top P + P A = - I$.
   %\begin{align*}
   %       A^\top P + P A = - I.
   %\end{align*}
   Let $\gamma_{\max}$ and $\gamma_{\min}$ be the maximum and minimum eigenvalues of $P$, respectively.
\end{assumption}

\begin{assumption} \label{assum:step-size}
    The step-size sequence $\{\alpha_t\}$ is positive, non-increasing, and satisfies $\sum_{t=0}^\infty \alpha_t = \infty$ and $\sum_{t=0}^\infty \alpha_t^2 < \infty$. 
\end{assumption}

To state our first main result, we need the following concepts.

\begin{definition}
     A graph sequence $\{ \bbb{G}_t \}$ is uniformly strongly connected if there exists a positive integer $L$ such that for any $t\ge 0$, the union graph $\cup_{k=t}^{t+L-1} \bbb{G}_k$ is strongly connected.
     If such an integer exists, we sometimes say that $\{ \bbb{G}_t \}$ is uniformly strongly connected by sub-sequences of length $L$.
\end{definition}

%The generality of the above joint connectivity condition is explained in Remark~\ref{remark:uniformly}. 

\begin{remark} \label{remark:uniformly}
Two popular joint connectivity definitions in consensus literature are ``$B$-connected'' \cite{nedic2009distributed_quan} and ``repeatedly jointly strongly connected'' \cite{reachingp1}. 
A graph sequence $\{ \bbb{G}_t \}$ is $B$-connected if there exists a positive integer $B$ such that the union graph $\cup_{t=kB}^{(k+1)B-1} \bbb{G}_t$ is strongly connected for each integer $k\ge 0$. Although the uniformly strongly connectedness looks more restrictive compared with $B$-connectedness at first glance, they are in fact equivalent. To see this, first it is easy to see that if $\{ \bbb{G}_t \}$ is uniformly strongly connected, $\{ \bbb{G}_t \}$ must be $B$-connected; now supposing $\{ \bbb{G}_t \}$ is $B$-connected, for any fix $t$, the union graph $\cup_{k=t}^{t+2B-1} \bbb{G}_k$ must be strongly connected, and thus $\{ \bbb{G}_t \}$ is uniformly strongly connected by sub-sequences of length $2B$. Thus, the two definitions are equivalent. It is also not hard to show that the uniformly strongly connectedness is equivalent to ``repeatedly jointly strongly connectedness'' provided the directed graphs under consideration all have self-arcs at all vertices, with ``repeatedly jointly strongly connectedness'' being defined upon ``graph composition'' \cite{reachingp1}. 
\hfill $\Box$
\end{remark}

\begin{definition}\label{def: absolute prob}
Let $\{ W_t \}$ be a sequence of stochastic matrices. A sequence of stochastic vectors $\{ \pi_t \}$ is an absolute probability sequence for $\{ W_t \}$ if
$\pi_t^\top = \pi_{t+1}^\top W_t$ for all~$t$.
\end{definition}

This definition was first introduced by Kolmogorov who proved that every sequence of stochastic matrices has an absolute probability sequence \cite{kolmogorov}. An alternative proof of this fact was given by Blackwell \cite{blackwell}.
In general, a sequence of stochastic matrices may have more than one absolute probability sequence; when the sequence of stochastic matrices is ``ergodic'', it has a unique absolute probability sequence \cite{tacrate}. It is easy to see that when $W_t$ is a fixed irreducible stochastic matrix $W$, $\pi_t$ is simply the normalized left eigenvector of $W$ for eigenvalue one. 
More can be said.

\begin{lemma} \label{lemma:bound_pi_jointly}
    {\rm (Lemma 5.8 in \cite{touri2012product})}
    %{\rm (Theorem 4.8 in \cite{touri2012product})}
    Suppose that Assumption~\ref{assum:weighted matrix} holds. If $\{\bbb{G}_t\}$ is uniformly strongly connected, then there exists a unique absolute probability sequence $\{ \pi_t \}$ for the matrix sequence $\{W_t\}$ and a constant $\pi_{\min} \in (0,1)$ such that $\pi_t^i \ge \pi_{\min}$ for all $i$ and $t$.
    %where $\pi_t^i$ is the $i$-th entry of vector $\pi_t$.
\end{lemma}

Let $\langle \theta \rangle_t  = \sum_{i=1}^N\pi^i_t \theta^i_t$, which is a column vector and convex combination of all $\theta_t^i$. It is easy to see that 
$\langle \theta \rangle_t  = (\pi_t^\top\Theta_t)^\top=\Theta_t^\top \pi_t$.
From Definition~\ref{def: absolute prob} and \eqref{eq:updtae_Theta}, we have
$    \pi^\top_{t+1}  \Theta_{t+1} = \pi^\top_{t+1}  W_t \Theta_t + \alpha_t \pi^\top_{t+1} W_t \Theta_t A(X_t)^\top + \alpha_t \pi^\top_{t+1} B(X_t)  
     = \pi^\top_t  \Theta_t + \alpha_t \pi^\top_{t} \Theta_t A(X_t)^\top + \alpha_t \pi^\top_{t+1} B(X_t)$,
%\begin{align*}
%    \pi^\top_{t+1}  \Theta_{t+1} &= \pi^\top_{t+1}  W_t \Theta_t + \alpha_t \pi^\top_{t+1} W_t \Theta_t A(X_t)^\top + \alpha_t \pi^\top_{t+1} B(X_t)  \\
%    & = \pi^\top_t  \Theta_t + \alpha_t \pi^\top_{t} \Theta_t A(X_t)^\top + \alpha_t \pi^\top_{t+1} B(X_t),
%\end{align*}
which implies that
\begin{align} \label{eq:update of average_time-varying}
     \langle \theta \rangle_{t+1} &= \langle \theta \rangle_t + \alpha_t A(X_t) \langle \theta \rangle_t + \alpha_t  B(X_t)^\top \pi_{t+1}.
\end{align}

Asymptotic performance of \eqref{eq:theta update} with any uniformly strongly connected neighbor graph sequence is characterized by the following two theorems.

\begin{theorem} \label{thm:consensus_time-varying_jointly}
    Suppose that Assumptions~\ref{assum:weighted matrix}, \ref{assum:A and b} and \ref{assum:step-size} hold. Let $\{ \theta_t^i \}$, $i\in \mathcal{V}$, be generated by \eqref{eq:theta update}. If $\{\bbb{G}_t\}$ is uniformly strongly connected,
    then $\lim_{t\rightarrow{\infty}}\|\theta^i_t-\langle \theta\rangle_t\|_2=0$ for all $i\in\mathcal V$.
\end{theorem}

Theorem~\ref{thm:consensus_time-varying_jointly} only shows that all the sequences $\{ \theta_t^i \}$, $i\in\mathcal V$,  generated by \eqref{eq:theta update} will finally reach a consensus, but not necessarily convergent or bounded. To guarantee the convergence of the sequences, we further need the following assumption, whose validity is discussed in Remark~\ref{remark:on assmption}.

\begin{assumption} \label{assum:limit_pi}
    The absolute probability sequence $\{ \pi_t \}$ for the stochastic matrix sequence $\{W_t\}$ has a limit, i.e., there exists a stochastic vector $\pi_{\infty}$ such that $\lim_{t\to\infty} \pi_t = \pi_{\infty}$.
\end{assumption}

\begin{theorem} \label{thm:theta^*_jointly}
    Suppose that Assumptions~\ref{assum:weighted matrix}--\ref{assum:limit_pi} hold. Let $\{ \theta_t^i \}$, $i\in \mathcal{V}$, be generated by \eqref{eq:theta update} and $\theta^*$ be the unique equilibrium point of the ODE    
    \begin{align} \label{eq:definition theta^*}
        \dot \theta = A \theta + b, \;\;\;  b=\sum_{i=1}^N \pi_{\infty}^i b^i,
   \end{align}
    where $A$ and $b^i$ are defined in Assumption~\ref{assum:A and b} and $\pi_{\infty}$ is defined in Assumption~\ref{assum:limit_pi}.
    If $\{ \bbb{G}_t \}$ is uniformly strongly connected, then all $\theta_t^i$ will converge to $\theta^*$ both with probability 1 and in mean square. %for all $i\in\mathcal V$.
\end{theorem}

\begin{remark} \label{remark:on assmption}
Though Assumption~\ref{assum:limit_pi} may look restrictive at first glance, simple simulations show that the sequences $\{ \theta_t^i \}$, $i\in\mathcal V$, do not converge if the assumption does not hold (e.g., even when $W_t$ changes periodically). It is worth emphasizing that the existence of $\pi_{\infty}$ does not imply the existence of $\lim_{t\rightarrow\infty}W_t$, though the converse is true. Indeed, the assumption subsumes various cases including (a) all $W_t$ are doubly stochastic matrices, and (b) all $W_t$ share the same left eigenvector for eigenvalue 1, which may arise from the scenario when the number of in-neighbors of each agent does not change over time \cite{olshevsky2013degree}. An important implication of Assumption~\ref{assum:limit_pi} is when the consensus interaction among the agents, characterized by $\{W_t\}$, is replaced by resilient consensus algorithms such as \cite{vaidya2012iterative,leblanc2013resilient} in order to attenuate the effect of unknown malicious agents, the resulting dynamics of non-malicious agents, in general, will not converge, because the resulting interaction stochastic matrices among the non-malicious agents depend on the state values transmitted by the malicious agents, which can be arbitrary, and thus the resulting stochastic matrix sequence, in general, does not have a convergent absolute probability sequence; of course, in this case, the trajectories of all the non-malicious agents will still reach a consensus as long as the step-size is diminishing, as implied by Theorem~\ref{thm:consensus_time-varying_jointly}. 
Further discussion on Assumption~\ref{assum:limit_pi} can be found in Appendix~\ref{discussionAss6}.
\hfill $\Box$
\end{remark}

We now study the finite-time performance of the proposed distributed linear stochastic approximation \eqref{eq:theta update} for both fixed and time-varying step-size cases. Its finite-time performance is characterized by the following theorem. 
%\subsubsection{Fixed Step-size} \label{sec:strongly_fixed}

Let $\eta_t = \| \pi_t - \pi_\infty \|_2$ for all $t\ge 0$. From Assumption~\ref{assum:limit_pi}, $\eta_t$ converges to zero as $t\rightarrow\infty$.

\begin{theorem} \label{thm:bound_jointly_SA}
    Let the sequences $\{ \theta_t^i \}$, $i \in \mathcal{V}$, be generated by \eqref{eq:theta update}. Suppose that  Assumptions~\ref{assum:weighted matrix}--\ref{assum:lyapunov},~\ref{assum:limit_pi} hold and $\{ \bbb{G}_t \}$ is uniformly strongly connected by sub-sequences of length $L$.  Let $q_t$ and $m_t$ be the unique integer quotient and remainder of $t$ divided by $L$, respectively.     Let $\delta_t$ be the diameter of $\cup_{k=t}^{t+L-1} \bbb{G}_k$, $ \delta_{\max} = \max_{t\ge 0} \delta_t$, and
    \begin{align}
        \epsilon & =   \bigg(1+\frac{2 b_{\max}}{A_{\max}}-\frac{\pi_{\min} \beta^{2L}}{2 \delta_{\max}} \bigg)( 1 + \alpha A_{\max})^{2L} 
        -  \frac{2 b_{\max}}{ A_{\max}} (1 + \alpha A_{\max})^{L}, \label{eq:define epsilon_jointly}
    \end{align}
where $0 < \alpha < \min \{ K_1 ,\; \frac{\log2}{A_{\max} \tau(\alpha)},\; \frac{0.1}{K_2 \gamma_{\max}} \}$.
    
%{\rm\bf 1) Fixed step-size:}         
%        Let $\alpha_t = \alpha$ for all $t\ge 0$. 
%        For all $ t\ge T_1 $,  
%        \begin{align} \label{eq:bound_jointly_fixed}
%            \sum_{i=1}^N \pi_{t}^i \mathbf{E}\left[\left\|\theta_{t}^i - \theta^*\right\|_2^2\right]
%            &\le 2 \epsilon^{q_{t}} \sum_{i=1}^N \pi_{m_t}^i \mathbf{E}\left[\left\| \theta_{m_t}^i - \langle \theta \rangle_{m_t} \right\|_2^2 \right]  + C_1 \bigg( 1 - \frac{0.9  \alpha}{\gamma_{\max}} \bigg)^{{t}-T_1}  + C_2.
%        \end{align}

{\rm\bf 1) Fixed step-size:}         
        Let $\alpha_t = \alpha$ for all $t\ge 0$. 
        For all $ t\ge T_1 $,  
        \begin{align} \label{eq:bound_jointly_fixed}
            \sum_{i=1}^N \pi_{t}^i \mathbf{E}\left[\left\|\theta_{t}^i - \theta^*\right\|_2^2\right]
            &\le 2 \epsilon^{q_{t}} \sum_{i=1}^N \pi_{m_t}^i \mathbf{E}\left[\left\| \theta_{m_t}^i - \langle \theta \rangle_{m_t} \right\|_2^2 \right]  + C_1 \bigg( 1 - \frac{0.9  \alpha}{\gamma_{\max}} \bigg)^{{t}-T_1}  + C_2 \nonumber \\
            &\;\;\;  + \frac{\gamma_{\max}}{\gamma_{\min}} 2\alpha \zeta_4 \sum_{k={0}}^{t-T_1}  \eta_{t+1-k}  \bigg(1-\frac{0.9  \alpha}{\gamma_{\max}} \bigg)^{k}.
        \end{align}

%old:
%$C_2 =  \frac{2\zeta_2}{1- \epsilon} + \frac{\gamma_{\max}}{\gamma_{\min}}\cdot \frac{ 2 \alpha  \zeta_3 \gamma_{\max} + 4 \gamma_{\max}  \zeta_4}{0.9  } $
%new:
%$C_2 =  \frac{2\zeta_2}{1- \epsilon} + \frac{ 2 \alpha  \zeta_3 \gamma_{\max}^2 }{0.9 \gamma_{\min} } $
%From Appendix~\ref{sec:constants}, $C_2$ is monotonically increasing for $\gamma_{\max}, \delta_{\max}, b_{\max}$ and $L$, and monotonically decreasing for $\gamma_{\min}, \pi_{\min}$ and $\beta$.

{\rm\bf 2) Time-varying step-size:} 
    Let $\alpha_t = \frac{\alpha_0}{t+1}$ with $\alpha_0 \ge \frac{\gamma_{\max}}{0.9}$. For all $t\ge LT_2$,
    \begin{align}
        &\sum_{i=1}^N \pi_{t}^i \mathbf{E}\left[\left\|\theta_{t}^i - \theta^*\right\|_2^2\right]
        \; \le \; 2  \epsilon^{q_{t}-T_2} \sum_{i=1}^N \pi_{LT_2+m_t}^i \mathbf{E}\left[\left\| \theta_{LT_2+m_t}^i - \langle \theta \rangle_{LT_2+m_t} \right\|_2^2\right] \nonumber  \\
        & \;\;\;\;\;\;\;\;\;\;\;\;\;\;\;\;  + C_3 \left( \alpha_0  \epsilon^{\frac{q_t-1}{2}} + \alpha_{\ceil{\frac{q_t-1}{2}}L}\right) +  \frac{1}{t} \bigg(C_4 \log^2\Big(\frac{t}{\alpha_0}\Big)+C_5\sum_{k = LT_2}^{t} \eta_{k} + C_6\bigg). \label{eq:bound_jointly_time-varying}
    \end{align}
    
    Here $T_1, T_2, K_1, K_2, C_1 - C_6$ are finite constants whose definitions are given in Appendix~\ref{sec:thmSA_constant}.
\end{theorem}

Since $\pi_t^i$ is uniformly bounded below by $\pi_{\min}\in(0,1)$ from Lemma~\ref{lemma:bound_pi_jointly}, it is easy to see that the above bound holds for each individual $\mathbf{E}[\|\theta_t^i - \theta^*\|_2^2]$. 
To better understand the theorem, we provide the following remark.

\begin{remark} \label{remark:exists_T_1}
In Appendix~\ref{sec:proof_jointly_fixed}, we show that both $\epsilon$ and $(1-\frac{0.9 \alpha}{\gamma_{\max}})$ lie in the interval $(0,1)$. It is easy to show that $\epsilon$ is monotonically increasing for $\delta_{\max}$ and $L$, monotonically decreasing for $\beta$ and $\pi_{\min}$. Also, 
\begin{align*}
    & \lim_{t\to\infty} \sum_{k=0}^{t-T_1} \eta_{t+1-k} \Big(1-\frac{0.9 \alpha}{\gamma_{\max}}\Big)^k \\
=\; & \lim_{t\to\infty} \sum_{l=0}^{\lfloor \frac{t-T_1}{2} \rfloor} \eta_{T_1+1+l} \Big(1-\frac{0.9 \alpha}{\gamma_{\max}}\Big)^{t-T_1-l}  + \sum_{l=\lceil \frac{t-T_1}{2} \rceil}^{ t-T_1 } \eta_{T_1+1+l} \Big(1-\frac{0.9 \alpha}{\gamma_{\max}}\Big)^{t-T_1-l} \\
\le\; & \lim_{t\to\infty} \frac{\gamma_{\max}}{0.9 \alpha} \bigg( \Big(1-\frac{0.9 \alpha}{\gamma_{\max}}\Big)^{\frac{t-T_1}{2}} \max_{l=0,\ldots, \ceil{\frac{t-T_1}{2}}} \eta_{T_1+1+l}  
+ \max_{l=\lceil \frac{t-T_1}{2} \rceil,\ldots, t-T_1+1} \eta_{l}  \bigg) = 0.
\end{align*}
%$\lim_{t\to\infty} \sum_{k=0}^{t-T_1} \eta_{t+1-k} (1-\frac{0.9 \alpha}{\gamma_{\max}})^k \le \lim_{t\to\infty} \frac{\gamma_{\max}}{0.9 \alpha} [ \eta_{\ceil{\frac{t-T_1}{2}}} + \eta_1 (1-\frac{0.9 \alpha}{\gamma_{\max}})^{\frac{t-T_1}{2}} ] = 0 $.
Therefore, the summands in the finite-time bound \eqref{eq:bound_jointly_fixed} for the fixed step-size case are exponentially decaying except for the constant $C_2$, which implies that  
$\limsup_{t\rightarrow\infty}\sum_{i=1}^N \pi_t^i \mathbf{E}[\|\theta_t^i - \theta^*\|_2^2]
        \le C_2$,
%\begin{align*}
%        \limsup_{t\rightarrow\infty}\sum_{i=1}^N \pi_t^i \mathbf{E}\left[\left\|\theta_t^i - \theta^*\right\|_2^2\right]
%        \le C_2 , 
%\end{align*}
providing a constant limiting bound. 
%From Appendix~\ref{sec:constants}, $C_2$ depends on $L, \gamma_{\min}, \gamma_{\max}, A_{\max}, b_{\max}$. 
From Appendix~\ref{sec:constants}, $C_2$ is monotonically increasing for $\gamma_{\max}, \delta_{\max}, b_{\max}$ and $L$, and monotonically decreasing for $\gamma_{\min}, \pi_{\min}$ and $\beta$.
In Appendix~\ref{sec:proof_jointly_time-varying}, we show that $\lim_{t\rightarrow\infty}\frac{1}{t}\sum_{k=1}^t \eta_k=0$, which implies that the finite-time bound \eqref{eq:bound_jointly_time-varying} for the time-varying step-size case converges to zero as $t\rightarrow\infty$. 
We next comment on $0.1$ in the inequality defining $\alpha$. 
%Note that in the definition of $T_1^*$, we require that for all $t \ge T_1^*$, there hold $36 \sqrt{N}b_{\max} \eta_{t+1} \gamma_{\max} + \Psi_3 \alpha \gamma_{\max} \le 0.1 $. 
Actually, we can replace $0.1$ with any constant $c\in (0,1)$, which will affect the value of $\epsilon$ and the feasible set of $\alpha$, with the latter becoming 
$0 < \alpha < \min \{ K_1 ,\; \frac{\log2}{A_{\max} \tau(\alpha)},\; \frac{c}{K_2 \gamma_{\max}}\}.$
Thus, the smaller the value of $c$ is, the smaller is the feasible set of $\alpha$, though the feasible set is always nonempty. %So, if we fix the $\alpha$ small enough, then we can choose any $c \ge 10^{-6}$. 
For convenience, we simply pick $c = 0.1$ in this paper; that is why we also have $0.9$ in \eqref{eq:bound_jointly_fixed}.
Lastly, we comment on $\alpha_0$ in the time-varying step-size case. We set $\alpha_0 \ge \frac{\gamma_{\max}}{0.9}$ for the purpose of getting a cleaner expression of the finite-time bound. For $\alpha_0 < \frac{\gamma_{\max}}{0.9}$, our approach still works, but will yield a more complicated expression. The same is true for Theorem~\ref{thm:bound_time-varying_step_Push_SA}.
\hfill $\Box$
\end{remark}

{\bf Technical Challenge and Proof Sketch \;}
As described in the introduction, the key challenge of analyzing the finite-time performance of the distributed stochastic approximation \eqref{eq:theta update} lies in the condition that the consensus-based interaction matrix is time-varying and stochastic (not necessarily doubly stochastic). To tackle this, we appeal to the absolute probability sequence $\pi_t$ of the time-varying interaction matrix sequence and introduce the quadratic Lyapunov comparison function $\sum_{i=1}^N \pi_{t}^i \mathbf{E}[\|\theta_{t}^i - \theta^*\|_2^2]$. Then, using the inequality
$\sum_{i=1}^N \pi_{t}^i \mathbf{E}[\|\theta_{t}^i - \theta^*\|_2^2]
         \le 2 \sum_{i=1}^N \pi_{t}^i \mathbf{E}[\|\theta_{t}^i - \langle \theta \rangle_{t} \|_2^2 ] + 2 \mathbf{E} [\| \langle \theta \rangle_{t} - \theta^*\|_2^2]$,
%\begin{align*}
%        \sum_{i=1}^N \pi_{t}^i \mathbf{E}[\|\theta_{t}^i - \theta^*\|_2^2]
%        & \le 2 \sum_{i=1}^N \pi_{t}^i \mathbf{E}[\|\theta_{t}^i - \langle \theta \rangle_{t} \|_2^2 ] + 2 \mathbf{E} [\| \langle \theta \rangle_{t} - \theta^*\|_2^2] .
%\end{align*}
the next step is to find the finite-time bounds of $\sum_{i=1}^N \pi_{t}^i \mathbf{E}[\|\theta_{t}^i - \langle \theta \rangle_{t} \|_2^2 ]$ (Lemmas~\ref{lemma:bound_consensus_jointly} and \ref{lemma:bound_consensus_time-varying_jointly})
and $ \mathbf{E} [\| \langle \theta \rangle_{t} - \theta^*\|_2^2]$ (Lemmas~\ref{lemma:bound_average} and \ref{lemma:bound_average_time-varying_jointly}), respectively. The latter term is essentially the ``single-agent'' mean-square
error. Our main analysis contribution here is to bound the former term for both fixed and time-varying step-size cases.

\vspace{-.1in}

\section{Push-SA} \label{sec:SA_pushsum}
%\section{Push-SA } \label{sec:SA_pushsum}

\vspace{-.1in}

The preceding section shows that the limiting state of consensus-based distributed stochastic approximation depends on $\pi_{\infty}$, which leads to a convex combination of the local equilibria of all the agents in the absence of communication, but the convex combination is in general ``uncontrollable''. Note that this convex combination will correspond to a convex combination of the network-wise accumulative rewards in applications such as distributed TD learning. 
In an important case when the convex combination is desired to be the straight average,
the existing literature e.g.  \cite{doan2019convergence,doan2019finite} relies on doubly stochastic matrices whose corresponding $\pi_{\infty}=(1/N)\1_N$. As mentioned in the introduction, doubly stochastic matrices implicitly require bi-directional communication between any pair of neighboring agents; see e.g. gossiping \cite{boyd052,pieee} and the Metropolis algorithm \cite{metro2}. A popular method to achieve the straight average target while allowing uni-directional communication between neighboring agents is to appeal to the idea so-called ``push-sum'' \cite{pushsum}, which was tailored for solving the distributed averaging problem over directed graphs and has been applied to distributed optimization \cite{nedic}. 
In this section, we will propose a push-based distributed stochastic approximation algorithm tailored for uni-directional communication and establish its finite-time error bound.

%In this section, we focus on the general case when $\{ \bbb{G}_t \}$ is uniformly strongly connected. 
%In addition, we assume it satisfies the following Assumption.
%\begin{assumption} \label{assum:push-sum_jointly_weight_matrix}
%    There exists an integer $ L \ge 1$ such that the union of graphs $\cup_{t=kL}^{kL+L-1} \bbb{G}_t$ is a strongly connected graph for all $k\ge 0$.
%\end{assumption}

Each agent $i$ has control over three variables, namely $y_t^i$, $\tilde{\theta}_t^i$ and $\theta_t^i$, in which $y^{i}_t$ is scalar-valued with initial value 1, $\tilde{\theta}_t^i$ can be arbitrarily initialized, and $\theta_0^i=\tilde{\theta}_0^i$.
At each time $t\ge 0$, each agent $i$ sends its weighted current values $\hat w_t^{ji} y^i_{t}$ and $\hat w_t^{ji}(\tilde\theta^i_{t} + \alpha_t A(X_t) \theta_t + \alpha_t b^i(X_t)) $ to each of its current out-neighbors $j\in\mathcal{N}_t^{i-}$, and updates its variables as follows:
%\begin{empheq}[left = \empheqlbrace]{align}\label{eq:SA_push-sum}
%y^i_{t+1}&=\sum_{j \in \mathcal{N}_t^i} \hat w_t^{ij } y^j_{t}, \nonumber\\
%\hat \theta^i_{t+1} &= \sum_{j \in \mathcal{N}_t^i} \hat w_t^{ij } \hat\theta^j_{t}  + \alpha_t \bigg(A(X_{t})  \sum_{j \in \mathcal{N}_t^i}  \hat w_t^{ij } \hat \theta^j_{t} + b^i(X_{t} ) y^i_{t+1}\bigg),\\ 
%\theta^i_{t+1}&=\frac{\hat \theta^i_{t+1}}{y^i_{t+1}},\nonumber
%\end{empheq}
\begin{empheq}[left = \empheqlbrace]{align}\label{eq:SA_push-sum}
y^i_{t+1}&=\sum_{j \in \mathcal{N}_t^i} \hat w_t^{ij } y^j_{t}, \;\;\;\;\; y^i_0=1,\nonumber\\
\tilde \theta^i_{t+1} &= \sum_{j \in \mathcal{N}_t^i} \hat w_t^{ij } \left[ \tilde\theta^j_{t}  + \alpha_t \left(A(X_{t}) \theta^j_{t} + b^j(X_{t} ) \right) \right],\\ 
\theta^i_{t+1}&=\frac{\tilde \theta^i_{t+1}}{y^i_{t+1}}, \;\;\;\;\; \theta_0^i=\tilde{\theta}_0^i,\nonumber
\end{empheq}
where $\hat w_t^{ij}=1/|\mathcal{N}_t^{j-}|$.
It is worth noting that the algorithm is distributed yet requires that each agent be aware of the number of its out-neighbors.

Asymptotic performance of \eqref{eq:SA_push-sum} with any uniformly strongly connected neighbor graph sequence is characterized by the following theorem.

\begin{theorem} \label{thm:push_meansq}
    Suppose that Assumptions~\ref{assum:A and b}--\ref{assum:step-size} hold. Let $\{ \theta_t^i \}$, $i\in \mathcal{V}$, be generated by \eqref{eq:SA_push-sum} and {\color{black} $\theta^*\in\R^d$} be the unique equilibrium point of the ODE %\footnote{It is not hard to extend our algorithm from the straight average point to any pre-specified convex combination as long as each agent knows its own positive coefficient.} 
\begin{align} \label{eq:ode_pushsum}
    \dot \theta = A \theta + \frac{1}{N} \sum_{i=1}^N  b^i,
\end{align} 
    where $A$ and $b^i$ are defined in Assumption~\ref{assum:A and b}.
    If $\{ \bbb{G}_t \}$ is uniformly strongly connected, then $\theta_t^i$ will converge to $\theta^*$ in mean square for all $i\in\mathcal V$.
\end{theorem}

In this section, we define $\langle \tilde \theta \rangle_t = \frac{1}{N} \sum_{i=1}^N \tilde \theta_t^i$ and $\langle \theta \rangle_t = \frac{1}{N} \sum_{i=1}^N \theta_t^i$.
To help understand these definitions, let $\hat W_t$ be the $N\times N$ matrix whose $ij$-th entry equals $\hat w_t^{ij}$ if $j\in\mathcal{N}_t^i$, otherwise equals zero. 
It is easy to see that each $\hat W_t$ is a column stochastic matrix whose diagonal entries are all positive. Then, $\pi_t = \frac{1}{N}\1_N$ for all $t \ge 0$ can be regarded as an absolute probability sequence of $\{ \hat W_t \}$. Thus, the above two definitions are intuitively consistent with $\langle \theta \rangle_t$ in the previous section.

Finite-time performance of \eqref{eq:SA_push-sum} with any uniformly strongly connected neighbor graph sequence is characterized by the following theorem.

Let $ \mu_t = \|A(X_t) (\langle \theta \rangle_t -  \langle \tilde \theta \rangle_t)\|_2$.
In Appendix~\ref{sec:proof_push}, we show that $\|\langle \theta \rangle_t -  \langle \tilde \theta \rangle_t\|_2$ converges to zero as $t\rightarrow\infty$, so does $ \mu_t$.

\begin{theorem} \label{thm:bound_time-varying_step_Push_SA}
    Suppose that Assumptions~\ref{assum:A and b}--\ref{assum:lyapunov}
    hold and $\{ \bbb{G}_t \}$ is uniformly strongly connected by sub-sequences of length $L$. 
    Let $\{ \theta_t^i \}$, $i\in \mathcal{V}$, be generated by \eqref{eq:SA_push-sum}
    with $\alpha_t = \frac{\alpha_0}{t+1}$ and $\alpha_0 \ge \frac{\gamma_{\max}}{0.9}$. 
Then, there exists a nonnegative  $\bar\epsilon \le (1-\frac{1}{N^{NL}})^{\frac{1}{L}}$ such that
    for all $t\ge \bar T$,
    \begin{align}
        \sum_{i=1}^N \mathbf{E}\left[\left\|\theta_{t+1}^i - \theta^*\right\|_2^2\right] 
        \le \;\; &  C_7 \bar\epsilon^t  +  C_8 \left(  \alpha_0 \bar\epsilon^{\frac{t}{2}}  +  \alpha_{\ceil{\frac{t}{2}}} \right)+ C_9 \alpha_t \nonumber \\
        & + \frac{1}{t}\bigg(C_{10} \log^2\Big(\frac{t}{\alpha_0}\Big) + C_{11}\sum_{k = \bar T}^{t} \mu_{k} +C_{12}\bigg), \label{eq:bound_SA}
    \end{align}
    where $\bar T$ and $C_7 - C_{12}$ are finite constants whose definitions are given in Appendix~\ref{sec:thmPush_constant}.
\end{theorem}

In Appendix~\ref{sec:proof_push}, we show that $\lim_{t\rightarrow\infty}\frac{1}{t}\sum_{k=1}^t \mu_k=0$, which implies that the finite-time bound \eqref{eq:bound_SA} converges to zero as $t\rightarrow\infty$. 
It is worth mentioning that the theorem does not consider the fixed step-size case, as our current analysis approach cannot be directly applied for this case.

%\paragraph{Proof Sketch and Technical Challenge}

{\bf Proof Sketch and Technical Challenge \;}
Using the inequality for any $i$
$$ \mathbf{E}[\|\theta_{t+1}^i - \theta^*\|_2^2]
         \le 2  \mathbf{E}[\|\theta_{t+1}^i - \langle \tilde \theta \rangle_t \|_2^2 ] + 2 \mathbf{E} [\| \langle \tilde \theta \rangle_t - \theta^*\|_2^2],$$
%$$\sum_{i=1}^N \mathbf{E}[\|\theta_{t+1}^i - \theta^*\|_2^2]
%         \le 2 \sum_{i=1}^N  \mathbf{E}[\|\theta_{t+1}^i - \langle \tilde \theta \rangle_t \|_2^2 ] + 2 N \mathbf{E} [\| \langle \tilde \theta \rangle_t - \theta^*\|_2^2],$$
%\begin{align*}
%        \sum_{i=1}^N \mathbf{E}[\|\theta_{t+1}^i - \theta^*\|_2^2]
%        & \le 2 \sum_{i=1}^N  \mathbf{E}[\|\theta_{t+1}^i - \langle \tilde \theta \rangle_t \|_2^2 ] + 2 N \mathbf{E} [\| \langle \tilde \theta \rangle_t - \theta^*\|_2^2] .
%    \end{align*}
our goal is to derive the finite-time bounds of 
$ \mathbf{E}[\|\theta_{t+1}^i - \langle \tilde \theta \rangle_t \|_2^2 ]$ {\color{black} 
(Lemma~\ref{lemma:bound_consensus_time-varying_push_SA}) }
%$\sum_{i=1}^N  \mathbf{E}[\|\theta_{t+1}^i - \langle \tilde \theta \rangle_t \|_2^2 ]$ 
and $\mathbf{E} [\| \langle \tilde \theta \rangle_t - \theta^*\|_2^2]$
{\color{black} (Lemma~\ref{lemma:bound_average_time-varying_push_SA})
}, respectively. Although this looks similar to the proof of Theorem~\ref{thm:bound_jointly_SA}, the derivation is quite different. First, the iteration of $\langle \tilde \theta \rangle_t$ is a single-agent stochastic approximation (SA) plus a disturbance term $\langle \theta \rangle_t-\langle \tilde \theta \rangle_t$, so we cannot directly apply the existing single-agent SA finite-time analyses to bound $\mathbf{E} [\| \langle \tilde \theta \rangle_t - \theta^*\|_2^2]$; instead, we have to show that $\langle \theta \rangle_t-\langle \tilde \theta \rangle_t$ will diminish and quantify the diminishing ``speed''. 
Second, both the proof of showing diminishing $\langle \theta \rangle_t-\langle \tilde \theta \rangle_t$ and derivation of bounding $\sum_{i=1}^N  \mathbf{E}[\|\theta_{t+1}^i - \langle \tilde \theta \rangle_t \|_2^2 ]$ involve a key challenge: to prove the sequence $\{ \theta_t^i \}$ generated from the Push-SA \eqref{eq:SA_push-sum} is bounded almost surely  {\color{black}(Lemma~\ref{lemma:bound_theta})}. %i.e., $\sup_{t} \| \theta_t^i \| < \infty$ almost surely for any $i \in \mathcal{V}$.
To tackle this, we introduce a novel way to constructing an absolute probability sequence for the Push-SA as follows. 
From~\eqref{eq:SA_push-sum}, 
$    \theta^i_{t+1}
    = \sum_{j=1}^N \tilde w_t^{ij}  [ \theta^j_{t} + \alpha_t A(X_t) \frac{ \theta^j_{t}}{y_t^j} + \alpha_t \frac{ b^j(X_t ) }{y_t^j}]$,
%\begin{align*}
%    \theta^i_{t+1}
%    &= \sum_{j=1}^N \tilde w_t^{ij}  \bigg[ \theta^j_{t} + \alpha_t A(X_t) \frac{ \theta^j_{t}}{y_t^j} + \alpha_t \frac{ b^j(X_t ) }{y_t^j}\bigg], %\label{eq:push-sum_ratio}
%\end{align*}
where $\tilde w_t^{ij} = (\hat w_t^{ij} y_t^j)/(\sum_{k=1}^N \hat w_t^{ik } y^k_{t})$.  
We show that each matrix $\tilde W_t = [\tilde w_t^{ij}]$ is stochastic, and there exists a unique absolute  probability  sequence $\{ \tilde \pi_t \} $ for  the  matrix  sequence $\{ \tilde W_t \} $ such that $ \tilde \pi_t^i \ge \tilde  \pi_{\min}$ for all $i\in\mathcal V$ and $t\ge 0$, with the constant $ \tilde \pi_{\min}\in(0,1)$. 
Most importantly, we show two critical properties of $\{ \tilde W_t \} $ and $\{ \tilde \pi_t \} $ {\color{black} in Lemma~\ref{lemma:push-sum_pi_intfty}}, namely $\lim_{t\to\infty} (\Pi_{s=0}^{t} \tilde W_s)= \frac{1}{N}\1_N \1^\top_N$ and $\frac{\tilde \pi_t^i}{y_t^i} = \frac{1}{N}$
for all $i,j\in\mathcal V$ and $t \ge 0$, 
which have never been reported in the literature though push-sum-based distributed algorithms have been extensively studied.

\begin{remark}
It is worth mentioning that the approach for analyzing push-SA here can be leveraged to establish a better convergence rate for the subgradient-push algorithm proposed in \cite{nedic}; see a much more comprehensive development of the novel push-sum based analysis tool and its application in analyzing subgradient-push in \cite{yixuanpush}. 
\hfill $\Box$
\end{remark}

\section{Concluding Remarks} \label{sec:conclusion}

In this paper, we have established both asymptotic and non-asymptotic analyses for a  consensus-based distributed linear stochastic approximation algorithm over uniformly strongly connected graphs, and proposed a push-based variant for coping with uni-directional communication. Both algorithms and their analyses can be directly applied to TD learning. 
One limitation of our finite-time bounds is that they involve quite a few constants which are well defined and characterized but whose values are not easy to compute. 
Future directions include leveraging the analyses for resilience in the presence of malicious agents and extending the tools to more complicated RL.

%\newpage
\appendix

\section{List of Constants} \label{sec:constants}

In this appendix, we list all the constants used in our main results, Theorems~\ref{thm:bound_jointly_SA} and \ref{thm:bound_time-varying_step_Push_SA}. They are finite and their expressions do not affect the understanding of the theorems. Since their expressions are quite long and complicated, we begin with the following set of constants, based on which we will be able to present the constants used in the theorems and the proofs of the theorems in an easier way. We hope that this way can also help the readers to better understand and follow our results and analyses. 

%\vspace{.1in}

The first constant $\zeta_1$ is defined as follows. Recall that $\epsilon$ is given in \eqref{eq:define epsilon_jointly} as 
\begin{align*}
    \epsilon  &=   \bigg(1+\frac{2 b_{\max}}{A_{\max}}-\frac{\pi_{\min} \beta^{2L}}{2 \delta_{\max}} \bigg)( 1 + \alpha A_{\max})^{2L}
     -  \frac{2 b_{\max}}{ A_{\max}} (1 + \alpha A_{\max})^{L}.
\end{align*}
$\zeta_1$ is defined as the unique solution for which $ \epsilon = 1$ if $\alpha = \zeta_1$. 
The following remark shows why $\zeta_1$ uniquely exists.

\begin{remark} \label{remark:exists_Psi9}
%We show here why the $\zeta_1$ defined in the theorem exists and is unique.
From~\eqref{eq:define epsilon_jointly}, it is easy to see that  $ \epsilon$ is monotonically increasing for $\alpha>0$. Define the corresponding monotonic function as 
\begin{align*}
    f(\alpha) &= \bigg(1+\frac{2 b_{\max}}{A_{\max}}-\frac{\pi_{\min} \beta^{2L}}{2 \delta_{\max}} \bigg)( 1 + \alpha A_{\max})^{2L} 
     -  \frac{2 b_{\max}}{ A_{\max}} (1 + \alpha A_{\max})^{L}.
\end{align*}
Note that $0<f(0) < 1$ and $f(+\infty)=+\infty$. Thus, $ f(\alpha) = 1$ has a unique solution $\zeta_1$. 
%which implies that when $0< \alpha <\zeta_1$, we have $0<  \epsilon <1$. 
\hfill$\Box$
\end{remark}

The other constants are defined as follows:
\begin{align}
    \zeta_2 &= \frac{4 b_{\max}^2}{ A_{\max}^2}\left[(1 + \alpha A_{\max})^L-1\right]^2 + 2 b_{\max} \frac{(1 + \alpha A_{\max})^L-1}{ A_{\max}} (1 + \alpha A_{\max})^{L}\label{eq:define Psi10} \\
    \zeta_3 &=   \left( 144 + 4 A_{\max}^2 + 912 \tau(\alpha) A_{\max}^2  + 168  \tau(\alpha) A_{\max}  b_{\max} \right) \| \theta^*\|_2^2  + 2 + 2 b_{\max}^2 + 4 \|\theta^* \|_2^2 +  \frac{48b_{\max}^2}{A_{\max}^2}\nonumber \\
        &\;\;\; + \tau(\alpha) A_{\max}^2 \bigg[152 \bigg(\frac{b_{\max}}{A_{\max}} + \| \theta^* \|_2 \bigg)^2 +  \frac{48 b_{\max}}{A_{\max}} \bigg(\frac{b_{\max}}{A_{\max}} + 1  \bigg)^2 +   \frac{87b_{\max}^2}{A_{\max}^2} +   \frac{12 b_{\max}}{A_{\max}} \bigg] \label{eq:define Psi4}\\
    \zeta_4 &= \sqrt{N}b_{\max} \bigg( 2 + \frac{12 b_{\max}^2}{A_{\max}^2}+ 38 \| \theta^* \|_2^2 \bigg)\label{eq:define Psi5} \\
    \zeta_5  & =  144 + 916 A_{\max}^2  + 168  A_{\max}  b_{\max} \label{eq:define Psi7} \\
    \zeta_6 & = 4 b_{\max}^2 \alpha L^2  ( 1 +  \alpha A_{\max})^{2L-2}   +2 b_{\max}L ( 1 +  \alpha A_{\max})^{2L-1}
        \label{eq:define Psi11} \\
    \zeta_7  & = (148 + 916 A_{\max}^2  + 168  A_{\max}  b_{\max}) \| \theta^*\|_2^ 2 +  2 +  \frac{48b_{\max}^2}{A_{\max}^2} +  152  \bigg(b_{\max} + A_{\max} \| \theta^* \|_2 \bigg)^2   +  12  A_{\max}b_{\max} \nonumber\\
        &\;\;\; + 89 b_{\max}^2   + 48  A_{\max}b_{\max} \bigg(\frac{b_{\max}}{A_{\max}} + 1 \bigg)^2 \label{eq:define Psi8}\\
    \zeta_8 &=  144 + 916 A_{\max}^2  + 168  A_{\max}  b_{\max} + 144 A_{\max} \mu_{\max} \label{eq:definition_Psi12} \\
    \zeta_9 &=  2 + ( 4 + \zeta_8) \|\theta^* \|_2^2 +  48\frac{(b_{\max}+ \mu_{\max})^2}{A_{\max}^2} +  152  \left(b_{\max} + \mu_{\max} + A_{\max} \| \theta^* \|_2 \right)^2 +  12  A_{\max}b_{\max} \nonumber\\
        &\;\;\;  + 48  A_{\max}(b_{\max}+ \mu_{\max}) \bigg(\frac{b_{\max} + \mu_{\max}}{A_{\max}} + 1 \bigg)^2 +  89 (b_{\max}+ \mu_{\max})^2  \label{eq:definition_Psi13}
\end{align}
%------------------------------------------------------------------------------------------------------------------------------------------------------
%\end{strip}

Here $\mu_{\max}=(N+1) A_{\max}C_\theta$, where $C_\theta$ is a finite number defined in Lemma~\ref{lemma:bound_theta} which can be regarded as an upper bound of 2-norm of each agent $i$'s state $\theta^i_t$ generated by the Push-SA algorithm \eqref{eq:SA_push-sum}.

\subsection{Constants used in Theorem~\ref{thm:bound_jointly_SA}} \label{sec:thmSA_constant}
    %%Let $ \delta_t$ be the diameter of $\cup_{k=t}^{t+L-1} \bbb{G}_k$ and $  \delta_{\max} = \max_{t\ge 0} \delta_t$.
    %%\begin{align}
    %%     \epsilon & =  ( 1 + \alpha A_{\max})^{2L} \bigg(1-\frac{\pi_{\min} \beta^{2L}}{2 \delta_{\max}} \bigg) + 2 b_{\max} \frac{(1 + \alpha A_{\max})^L-1}{ A_{\max}} (1 + \alpha A_{\max})^{L} \label{eq:define epsilon_jointly}
    %%\end{align}
    
    %$\zeta_1$ is the unique solution for which $ \epsilon = 1$ if $\alpha = \zeta_1$. 

    \begin{align}
        K_1 &= \min\bigg\{ \zeta_1, \; \frac{\gamma_{\max}}{0.9} \bigg\} \hspace{4in}\nonumber\\
        %\zeta_3 &=   \left( 144 + 4 A_{\max}^2 + 912 \tau(\alpha) A_{\max}^2  + 168  \tau(\alpha) A_{\max}  b_{\max} \right) \| \theta^*\|_2^2  \nonumber \\
        %&\;\;\; + \tau(\alpha) A_{\max}^2 \bigg[152 \bigg(\frac{b_{\max}}{A_{\max}} + \| \theta^* \|_2 \bigg)^2 +  \frac{48 b_{\max}}{A_{\max}} \bigg(\frac{b_{\max}}{A_{\max}} + 1  \bigg)^2 +   \frac{87b_{\max}^2}{A_{\max}^2} +   \frac{12 b_{\max}}{A_{\max}} \bigg] \nonumber\\ 
        %& \;\;\; + 2 + 2 b_{\max}^2 + 4 \|\theta^* \|_2^2 +  \frac{48b_{\max}^2}{A_{\max}^2} \label{eq:define Psi4} \\
        %\zeta_4 &= 2\sqrt{N}b_{\max} \bigg( 1 + \frac{6 b_{\max}^2}{A_{\max}^2}+ 19 \| \theta^* \|_2^2 \bigg)\label{eq:define Psi5} \\
        %\zeta_2 &= \frac{4 b_{\max}^2}{ A_{\max}^2}\left((1 + \alpha A_{\max})^L-1\right)^2
        %+ 2 b_{\max} \frac{(1 + \alpha A_{\max})^L-1}{ A_{\max}} (1 + \alpha A_{\max})^{L}\label{eq:define Psi10} \\
    K_2 &= 144 + 4 A_{\max}^2 + 912 \tau(\alpha) A_{\max}^2  + 168  \tau(\alpha) A_{\max}  b_{\max} \label{eq:define Psi3}\\
    C_1 &= \frac{\gamma_{\max}}{\gamma_{\min}} \left( 8 \exp\left\{ 2 \alpha A_{\max}T_1 \right\}+4 \right) \mathbf{E}\left[\| \langle \theta \rangle_{0} -\theta^* \|_2^2\right] 
    + 8 \frac{\gamma_{\max}}{\gamma_{\min}} \exp\left\{ 2\alpha A_{\max}T_1 \right\} \bigg( \|\theta^*\|_2 + \frac{b_{\max}}{A_{\max}} \bigg)^2 \nonumber \\
    C_2 &=  \frac{2\zeta_2}{1- \epsilon} + \frac{\gamma_{\max}}{\gamma_{\min}}\cdot \frac{ 2 \alpha  \zeta_3 \gamma_{\max}}{0.9  } \nonumber\\
                C_3 &= \frac{2\zeta_6}{1-\epsilon} \nonumber\\
        %\zeta_7  & = (148 + 916 A_{\max}^2  + 168  A_{\max}  b_{\max}) \| \theta^*\|_2^ 2 +  2 +  \frac{48b_{\max}^2}{A_{\max}^2} +  152  \bigg(b_{\max} + A_{\max} \| \theta^* \|_2 \bigg)^2    \nonumber\\
        %&\;\;\; + 89 b_{\max}^2  +  12  A_{\max}b_{\max} + 48  A_{\max}b_{\max} \bigg(\frac{b_{\max}}{A_{\max}} + 1 \bigg)^2 \label{eq:define Psi8}\\
        C_4 &= 2\zeta_7 \alpha_0  C \frac{\gamma_{\max}}{\gamma_{\min}} \nonumber \\ 
        C_5 &= 2 \alpha_0 \zeta_4 \frac{\gamma_{\max}}{\gamma_{\min}}\nonumber \\
        %\zeta_6 & = 4 b_{\max}^2 \alpha L^2  ( 1 +  \alpha A_{\max})^{2L-2}   +2 b_{\max}L ( 1 +  \alpha A_{\max})^{2L-1} \label{eq:define Psi11} \\
        C_6 &= 2LT_2 \frac{\gamma_{\max}}{\gamma_{\min}} \mathbf{E}\left[\| \langle \theta \rangle_{LT_2} -\theta^* \|_2^2 \right] \nonumber
    \end{align}    
    
    $T_1$ is any positive integer such that for all $t \ge T_1$, there hold $t\ge \tau(\alpha)$ and 
    $36 \sqrt{N}b_{\max} \eta_{t+1} \gamma_{\max} + K_2 \alpha \gamma_{\max} \le 0.1 $. %where $\eta_t = \| \pi_t - \pi_\infty \|_2.$
    
\begin{remark} %\label{remark:exists_T_1}
We show that $T_1$ must exist. From $0 < \alpha < \min \{ K_1 ,\; \frac{\log2}{A_{\max} \tau(\alpha)},\; \frac{0.1}{K_2 \gamma_{\max}} \}$, it is easy to see that the feasible set of $\alpha$ is nonempty and $ K_2 \alpha \gamma_{\max} < 0.1 $. Since $\lim_{t\to\infty} \eta_t = 0$ by Lemma~\ref{lemma:eta_sum} and $\tau(\alpha) \le -C\log \alpha$ by Assumption~\ref{assum:mixing-time}, there exists a time instant $ T \ge -C\log \alpha$ such that for any $t\ge T$, there hold $t \ge \tau(\alpha)$ and 
$\eta_{t+1} \le (0.1 - K_2 \alpha \gamma_{\max})/(36 \sqrt{N}b_{\max} \gamma_{\max})$, which implies that $T_1$ must exist. 
\hfill $\Box$
\end{remark}

    $T_2$ is any positive integer such that for all $t\ge LT_2$, there hold $\alpha_t \le \alpha$, $2\tau(\alpha_t) \le t$, $\tau(\alpha_t) \alpha_{t-\tau(\alpha_t)} \le \min \{ \frac{ \log2}{A_{\max}},\; \frac{0.1}{\zeta_5 \gamma_{\max}} \}$ and $$ \zeta_5 \alpha_{t-\tau(\alpha_t)} \tau(\alpha_t) \gamma_{\max} + 36\sqrt{N} b_{\max} \eta_{t+1} \gamma_{\max} \le 0.1 .$$ 
    %where $\alpha$ be any constant satisfying $0 < \alpha < \min \left\{ K_2 ,\; \frac{\log2}{A_{\max} \tau(\alpha)},\; \frac{0.1}{K_2 \gamma_{\max}}\right\}$. 

\begin{remark} \label{remark:exists_T_2}
We explain why $T_2$ must exist.
%Similar to Remark~\ref{remark:exists_T_1}, we can change 0.1 to any constant $c\in(0,1)$. 
Since $\alpha_t = \frac{\alpha_0}{t+1}$ is monotonically decreasing for $t$ and $\tau(\alpha_t) \le -C\log \alpha_t = -C\log \alpha_0 + C\log (t+1) $ from Assumption~\ref{assum:mixing-time}, there exists a positive $S_{1}$ such that for any $t\ge S_{1}$, we have $\alpha_t \le \alpha$ and $t\ge 2\tau(\alpha_t)$ for any constant $0 < \alpha < \min \{ K_1 ,\; \frac{\log2}{A_{\max} \tau(\alpha)},\; \frac{0.1}{K_2 \gamma_{\max}} \}$. Moreover, it is easy to show that
\begin{align*}
    \lim_{t\to\infty} t-\tau(\alpha_t) & \ge \lim_{t\to\infty} t+ C\log \alpha_0 - C\log (t+1) 
    = +\infty, \\
    \lim_{t\to\infty}\tau(\alpha_t) \alpha_{t-\tau(\alpha_t)} & \le \lim_{t\to\infty} \frac{-C \alpha_0 \log \alpha_0 + C\alpha_0\log(t+1)}{t-\tau(\alpha_t)+1} 
    = 0.
\end{align*}
Then, there exists a positive $S_{2}$ such that for any $t\ge S_{2}$, we have $\tau(\alpha_t) \alpha_{t-\tau(\alpha_t)} \le \min \{ \frac{ \log2}{A_{\max}},\; \frac{0.1}{\zeta_5 \gamma_{\max}} \}$. In addition, since $\lim_{t\to\infty} \eta_t = 0$ from Lemma~\ref{lemma:eta_sum}, 
when $\tau(\alpha_t) \alpha_{t-\tau(\alpha_t)} \le \frac{0.1}{\zeta_5 \gamma_{\max}}$, there exists a positive $S_{3}$ such that for any $t\ge S_{3}$, we have $\eta_{t+1} \le (0.1 - \zeta_5 \alpha_{t-\tau(\alpha_t)} \tau(\alpha_t) \gamma_{\max})/(36\sqrt{N} b_{\max} \gamma_{\max})$. Thus, $T_2$ must exist as we can set $T_2 = \max\{ S_{1},S_{2},S_{3} \}$.
\hfill $\Box$    
\end{remark}

\subsection{Constants used in Theorem~\ref{thm:bound_time-varying_step_Push_SA}} \label{sec:thmPush_constant}
    %\begin{align}
    %    \zeta_8 &= 2 \left( 72 + 458 A_{\max}^2  + 84  A_{\max}  b_{\max} + 72 A_{\max} \mu_{\max} \right) \label{eq:definition_Psi12}
    %\end{align}
    \begin{flalign}
        C_7 &= \frac{16}{\epsilon_1}  \mathbf{E}\bigg[ \Big\| \sum_{i=1}^N \tilde \theta_0^i + \alpha_0 A(X_0)\tilde \theta_0^i + \alpha_0 b^i(X_0) \Big\|_2\bigg] \hspace{3in}\nonumber \\
        C_8 &= \frac{16}{\epsilon_1} \cdot\frac{ A_{\max} C_\theta + b_{\max}}{1-\bar\epsilon}\nonumber \\
        C_9 &= 2  A_{\max} C_\theta + 2  b_{\max} \nonumber\\
        %\zeta_9 &= \bigg[ 2 + ( 4 + \zeta_8) \|\theta^* \|_2^2 +  48\frac{(b_{\max}+ \mu_{\max})^2}{A_{\max}^2} +  152  \left(b_{\max} + \mu_{\max} + A_{\max} \| \theta^* \|_2 \right)^2  \nonumber\\
        %&\;\;\;\; +  12  A_{\max}b_{\max} + 48  A_{\max}(b_{\max}+ \mu_{\max}) (\frac{b_{\max} + \mu_{\max}}{A_{\max}} + 1 )^2 +  89 (b_{\max}+ \mu_{\max})^2  \bigg]\label{eq:definition_Psi13}\\
        C_{10} &= {2 N \zeta_9 \alpha_0  C } \frac{\gamma_{\max}}{\gamma_{\min}} \nonumber\\
        C_{11} &= 2 \alpha_0 N \frac{\gamma_{\max}}{\gamma_{\min}} \nonumber\\
        C_{12} &= {2\bar T N} \frac{\gamma_{\max}}{\gamma_{\min}} \mathbf{E}\left[\|\langle \tilde \theta \rangle_{\bar T} -\theta^* \|_2^2 \right]  \nonumber
\end{flalign}
Here $\epsilon_1$ is a positive constant defined as
$\epsilon_1 = \inf_{t\ge 0}  \min_{i\in\mathcal V} (\hat W_t \cdots \hat W_0 \1_N)^i. $ 
%where $\hat W_t$ is the column weighted matrix at time $t$ and $[\hat W_t \cdots \hat W_0 \1_N]^i$ is the $i$th entry of vector $[\hat W_t \cdots \hat W_0 \1_N]^i$. 
From Corollary~2~(b) in \cite{nedic} and the fact that each $\hat W_t$ is column stochastic, $\epsilon_1 \in [\frac{1}{N^{NL}}, 1]$.
See Lemma~\ref{lemma:bound_consensus_time-varying_push_SA} for more~details.

\vspace{.1in}

    $\bar T$ is any positive integer such that for all $t\ge \bar T$, there hold $2\tau(\alpha_t) \le t$, $ \mu_{t}  + \tau(\alpha_t) \alpha_{t-\tau(\alpha_t)} \zeta_8 \le \frac{0.1}{\gamma_{\max}}$ and $\tau(\alpha_t) \alpha_{t-\tau(\alpha_t)} \le \min \{ \frac{ \log2}{A_{\max}},\; \frac{0.1}{\zeta_8 \gamma_{\max}} \}$. %where $ \mu_t = A(X_t) \langle \theta \rangle_t - A(X_t) \langle \tilde \theta \rangle_t$. 

\begin{remark}
From Lemma~\ref{lemma:eta_limit_Push_SA},  $\lim_{t\to\infty} \mu_t = 0$. Then, using the similar arguments as in Remark~\ref{remark:exists_T_2}, we can show the existence of $\bar T$.
\hfill$\Box$
\end{remark}

%\section{Related work}\label{relatedwork}
%\input{related work_iclr} 

\section{Discussion on Assumption~\ref{assum:limit_pi}}\label{discussionAss6}
In this appendix, we contend that Assumption~\ref{assum:limit_pi} has more general applications than the previously known case and that it is in fact necessary.

\subsection{Applications}

First, as mentioned in Remark~\ref{remark:on assmption}, there are at least two cases which satisfy Assumption~\ref{assum:limit_pi}, yet cannot be directly handled by the existing analysis tool, which was developed only for doubly stochastic matrices. Case 1 is when the number of in-neighbors of agents is unchanged over time. This case has an interesting behavioral interpretation in fish biology, and has been adopted in bio-inspired distributed algorithm design \cite{abaid2010consensus}.
Case 2 is when the interaction matrix changes arbitrarily over time during an initial period, after which it finally becomes fixed. As we describe below, Case 2 occurs naturally in certain multi-agent systems.
%In ``Technical Innovation and Contributions" part in the introduction, we have explained why the existing analysis for doubly stochastic matrices is relatively easy and already well-understood within the vast distributed optimization literature.

Case 1 is mathematically equivalent to the situation when all stochastic matrices share the same left dominant eigenvector, which subsumes doubly stochastic matrices as a special case; thus it could be analyzed by carefully choosing a fixed norm. There may be different choices: one choice is to apply our time-varying quadratic Lyapunov comparison function $\sum_{i=1}^N \pi_{t}^i \mathbf{E}[||\theta_{t}^i - \theta^*||_2^2]$ to the time-invariant case (i.e., $\pi_t^i$ does not change over time), which leads to the weighted Frobenius norm defined in the appendix.

The extension to Case 1 just described may be straightforward, but Case 2 is not. As we proved in Theorems~\ref{thm:theta^*_jointly} and \ref{thm:bound_jointly_SA}, when the interaction matrix arbitrarily changes over time for an initial period, say of length $T$, and finally becomes a fixed matrix or enters Case 1, all agents' trajectories determined by (1) will converge in mean square. Also, recall that the corresponding finite-time error bounds in this case were derived using the ``absolute probability sequence" technique. Note that the existing techniques can only be applied to analyze (1) after time $T$; when $T$ is very large, such an analysis is undesirable, since the focus and challenge here are for ``finite" time.

It is important to note that Case 2 provides a realistic model for certain systems. Consider scenarios in which some agents do not function stably and thus they communicate with their neighbors sporadically for a certain period, leading to a time-varying stochastic matrix. Such scenarios occur naturally when there is unstable communication due to environmental changes or movement of agents (e.g., robots or UAVs may need to move into a new formation while continuing computation). After this unstable period, which could be long, the whole system then enters a stable operation status. This satisfies Case 2 and our finite-time analysis can be applied to the whole process, no matter how long the unstable period could be, as long as it is finite. In addition to this example, Case 2 and our analysis can be applied to certain scenarios in the presence of malicious agents. Suppose the system is aware that a small subset of agents have potentially been attacked and are thus behaving maliciously. To protect the system, the consensus interaction among the agents can switch to resilient consensus algorithms such as \cite{vaidya2012iterative,leblanc2013resilient} in order to attenuate the effect of malicious agents. In this situation, the resulting dynamics of the non-malicious agents are in general characterized by a time-varying stochastic matrix. After identifying and/or fixing the malicious agents, which could be a very slow process, the system can switch back to normal operation status. This example again satisfies Case 2, and our analysis can be applied to the whole procedure. As we mentioned in Remark~\ref{remark:on assmption}, if some malicious agents always exist, the non-malicious agents in general will not converge, and thus a finite-time analysis is probably meaningless. The non-convergence issue will be further explained in the next subsection.

Whether Assumption~\ref{assum:limit_pi} can represent more realistic/analytic examples is a very interesting future direction. Though consensus has been extensively studied and the ``absolute probability sequence" was proposed decades ago, this question has never been explored. The development of more advanced analysis tools is an interesting topic as well.

\subsection{Necessity}

We now elaborate on why Assumption~\ref{assum:limit_pi} is not restrictive from a theoretical point of view.

As mentioned in Remark~\ref{remark:on assmption}, distributed SA with time-varying stochastic matrices does not converge, in general, if Assumption~\ref{assum:limit_pi} does not hold. Assumption~\ref{assum:limit_pi} is sufficient to guarantee the convergence of the distributed SA algorithm \eqref{eq:theta update} when the interaction matrix is row stochastic and time-varying. Let us denote the necessary and sufficient condition for convergence of consensus-based distributed SA as Condition A, which is currently unknown. It is possible that there is a large gap between Assumption~\ref{assum:limit_pi} and Condition A. But Assumption~\ref{assum:limit_pi} is (to our knowledge) the most general sufficient condition that has been proposed so far; one indirect justification of this claim is Assumption~\ref{assum:limit_pi} is an analogue of condition (C3.4') in \cite{kushner87}, which is itself a sufficient condition guaranteeing the asymptotic convergence of a different form of distributed SA. While \cite{kushner87} only provided asymptotic analysis, we provided both asymptotic and finite-time analyses using a novel tool. Assumption~\ref{assum:limit_pi} subsumes the existing analysis for doubly stochastic matrices as a special case, and can be used for more general, nontrivial cases (see the examples provided in the discussion of Case 2 above).
Existing analysis tools cannot be applied to Case 2. From a theoretical point of view, our paper reduces the gap between the doubly stochastic matrices assumption and Condition A to the smaller gap between Assumption~\ref{assum:limit_pi} and Condition A, for finite-time analysis of consensus-based distributed~SA.

In addition, the other equally important main contribution of our paper, push-SA, does not need Assumption~\ref{assum:limit_pi}, though its analysis still relies on the ``absolute probability sequence" technique.

\section{Distributed TD Learning} \label{sec:TD}

In this section, we apply our distributed stochastic approximation finite-time analyses to distributed TD learning, as TD($\lambda$) is a special cases of stochastic approximation. To this end, we first introduce the following multi-agent MDP tailored for distributed TD. 
%{\color{red}do we still have TD(0) now?}

The multi-agent MDP can be defined by a tuple 
$(\mathcal{S}, \{ \mathcal{U}^i \}_{i\in\mathcal V}, \{ R^i \}_{i\in\mathcal V}, \bar P, \gamma, \{\bbb{G}_t\}_{t\ge 0})$. Here $\mathcal{S}=\{1,\ldots,S\}$ is the finite set of $S$ states and $\mathcal{U}^i$ is the set of control actions for agent $i$. 
For each agent $i$, 
$R^i :\mathcal{S}\times\mathcal{U}\times\mathcal{S} \to\R$ is
the local reward function, where $\mathcal{U}=\prod_{i=1}^{N}\mathcal{U}^i$ is the joint control action space. In addition, $\bar P:\mathcal{S}\times\mathcal{U}\times\mathcal{S}\to[0,1]$ denotes  the state transition probability matrix of the MDP,
 and $\gamma \in (0,1)$ is the discount factor.
Given a fixed policy, let $\bar P$ be of size $S\times S$ for convenience, and thus its $ij$-th entry $\bar p^{ij}$ equals the probability from state $i$ to state $j$ under the given policy. The multi-agent MDP then evolves as follows. 
At each time $t\ge 0$, each agent~$i$ observes the current state $s_t \in \mathcal{S}$, takes action $u_t^i = \mu^i(s_t) \in \mathcal{U}^i$, and receive a corresponding reward $R^i(s_t,u_t,s_{t+1})$,  where $\mu^i:\mathcal{S} \to \mathcal{U}^i$ is a function mapping a state to a control action in $\mathcal{U}^i$ and $ u_t = \prod_{i=1}^{N} u^i \in \mathcal{U} $. It is worth emphasizing that in such a multi-agent setting, each agent's rewards and reward function are private information, and thus cannot be shared with any other agents.
%Then the network can be characterized by a discount reward MDP: $(\mathcal{S}, \{ \mathcal{U}^i \}, \bar P, \{ R^i \}, \gamma, \{\bbb{G}_t\}_{t\ge 0})$ for $i\in\mathcal{V}$. Here, $\bar P = [\bar p^{ij}]$ is the transition probability matrix in $\R^{n\times n}$, i.e., $\bar p^{ij}$ is the probability from the state $i$ to state $j$, and $\gamma \in (0,1)$ is the discount factor. 
%%%%%%%%%% original goal

The discounted accumulative reward $J: \mathcal{S} \to \R $ associated with the above multi-agent MDP is defined for each $ s\in \mathcal{S}$ as
%$J(s) = \mathbf{E}[ \sum_{t=0}^\infty \gamma^t \sum_{i\in\mathcal{V}} c^i R^i(s_t,u_t,s_{t+1}) \;|\; s_0 = s],$
\begin{align} \label{eq:definition convex J}
    J(s) = \mathbf{E}\Big[ \sum_{t=0}^\infty \gamma^t \sum_{i\in\mathcal{V}} c^i R^i(s_t,u_t,s_{t+1}) \;|\; s_0 = s\Big],
\end{align}
%Note that $ J$ in~\eqref{eq:definition convex J} 
which satisfies the Bellman equation \cite{sutton2018reinforcement}, i.e.,
\begin{align*} 
    J(s) = \sum_{s' = 1}^S \bar p^{s s'}\Big[ \sum_{i\in\mathcal{V}} c^i  R^i(s,s')  + \gamma  J(s') \Big], \;\;\; s\in \mathcal{S},
\end{align*}
where $c^i>0$, $i\in\mathcal V$, is a set of convex combination weights. 
The existing distributed RL algorithms all set $c^i=1/N$ for all $i\in\mathcal V$ (e.g., \cite{kaiqing,doan2019convergence}), and this is why they require interaction matrices all be doubly stochastic. 
We will show that $c^i=\pi_{\infty}^i$ for all $i\in\mathcal V$ for general stochastic matrix sequences. Since for any doubly stochastic matrix sequence, its absolute probability sequence is $\pi_t=(1/N)\1_N$, i.e., $\pi_\infty^i=1/N$ for all $i\in\mathcal V$, our results generalize the existing results, e.g. \cite{doan2019convergence,doan2019finite}.  
In Section~\ref{sec:SA_pushsum}, we will show how to achieve the straight average reward, i.e.,$c^i=1/N$ for all $i\in\mathcal V$, without requiring doubly stochastic matrices.

When the number of the states is very large, 
the computation of exact $J$ may be intractable. To get around this, as did in \cite{tsitsiklis1997analysis}, we use a low-dimensional linear function $\hat J$ to approximate $J$. Specifically, the linear function approximator $\hat J$ takes the form $\hat J(s, \theta) = \sum_{k=1}^K \theta^k \phi_k^s, s\in\mathcal{S},$
%\begin{align*}
%    \hat J(s, \theta) = \sum_{k=1}^K \theta^k \phi_k^s, \;\;\; s\in\mathcal{S},
%\end{align*}
where each $\phi_k^s$ is a fixed scalar function defined on the state space $\mathcal{S}$,  each $\theta^k$ is the associated weight, and 
$K \ll S$. In other words, $\hat J$ is  parameterized by $\theta \in \R^K$, with $\theta^k$ being the $k$-th entry of $\theta$.
To proceed, let $\phi_k \in \R^S$ be the vector whose $j$-th entry is $\phi_k^j$ for all $k\in\{1,\ldots,K\}$, let $\phi(s) \in \R^K$ be the vector whose $j$-th entry is $\phi_j^s$ for all $s\in\mathcal{S}$, and let $\Phi\in \R^{S\times K}$ be the matrix whose $i$-th row is the row
vector $\phi(i)^\top$ and whose $j$-th column is the vector $\phi_j$, i.e., 
$\Phi = [\phi_1, \cdots, \phi_K] = [\phi(1), \cdots, \phi(S)]^\top \in \R^{S\times K}$,
%\begin{align*}
%    \Phi = \left[
%    \begin{array}{ccc}
%        \phi_1 & \cdots & \phi_K 
%    \end{array}
%    \right] = 
%    \left[ 
%    \begin{array}{c}
%        \phi(1)^\top \\
%        \vdots \\
%        \phi(S)^\top \\
%    \end{array}
%    \right] \in \R^{S\times K},
%\end{align*}
which implies $\hat J =\Phi \theta$.
The goal for the multi-agent network is to find an optimal $ \theta^*$ with which the distance between $\hat J$ and $ J$ is minimized, under the following standard assumptions adopted in e.g. \cite{Srikant,doan2019convergence}.

%where $\theta^k$ is the $k$-th entry of the vector $\theta$ and $\phi_k^s$ is the $s$-th entry of the vector  $\phi_k$. 
%Let  $\phi(s) = [ \phi_1^s, \cdots, \phi_K^s ]^\top \in \R^K, s\in\mathcal{S}$ be the feature vector

%Next, we will use TD($\lambda$) method to find the $\bar \theta^*$. 

%\cite{doan2019finite} proposed a distributed TD($\lambda$) algorithm to find the $ \theta^*$ with $\pi_{\infty}^i = \frac{1}{N}$ for all $i$ by using a fixed doubly stochastic matrix as the weight matrix. However, in this paper we will use the update, which has the same form as \eqref{eq:theta update}, to find $\theta^*$. 

\begin{assumption} \label{assum:reward} 
    All the rewards are uniformly bounded, i.e., there exists a positive constant $R$ such that $| R^i(s,s') | \le R$ for all $i\in\mathcal{V}$ and $s,s' \in \mathcal{S}$.
\end{assumption}

\begin{assumption} \label{assum:feature}
    The vectors $\phi_1,\ldots,\phi_K$ are linearly independent, i.e., $\Phi$ has full column rank, and $\| \phi(s) \|_2 \le 1$ for all $s\in\mathcal S$. 
    %Meanwhile, we assume all feature vectors are uniformly bounded, i.e. $\| \phi(s) \|_2 \le 1$.
\end{assumption}

\begin{assumption} \label{assum:tranistion}
    The Markov chain that evolves according to the transition probability matrix $\bar P$ is irreducible and aperiodic.
\end{assumption}

Under Assumption \ref{assum:tranistion}, let $d\in\R^S$ be the unique stationary distribution associated with $\bar P$, i.e., $d^\top \bar P =d^\top$.

\subsection{Distributed TD($\lambda$)} \label{sec:TD_lambda}
In this subsection, we make use of TD($\lambda$) to estimate $\theta^*$ in a distributed manner. Note that TD(0) can be applied in a similar manner.
Each agent $i\in\mathcal V$ updates its own estimator of $\theta^*$, $\theta_t^i$, for all time $t\in\{0,1,2,\ldots\}$ as follows: 
\begin{align} \label{eq:theta update_TD}
    \theta_{t+1}^i = \sum_{j \in \mathcal{N}_t^i} w_t^{ij} \theta_t^j + \alpha_t \Big(A(X_t)\sum_{j \in \mathcal{N}_t^i} w_t^{ij}\theta_t^j + b^i(X_t)\Big),
\end{align}
where $X_t = (s_t, s_{t+1}, z_t)$ is the Markov chain, with $z_t = \sum_{k=0}^t (\gamma \lambda)^{t-k} \phi(s_k)$, and
\begin{align}\label{eq:definition Ab(X_t)}
    A(X_t) = z_t (\gamma \phi(s_{t+1}) - \phi(s_t))^\top, \;\; b^i(X_t) = r^i_t z_t,
\end{align}
with $r_t^i$ being the reward for agent $i$ at time $t$.
It is worth emphasizing that the proposed TD($\lambda$) algorithm is different from that in \cite{doan2019finite}.
%The update \eqref{eq:theta update_TD} with \eqref{eq:definition Ab(X_t)} is a special case of \eqref{eq:theta update}. 

In the sequel, we will show that the update \eqref{eq:theta update_TD} with \eqref{eq:definition Ab(X_t)} is a special case of \eqref{eq:theta update} so that our analysis for \eqref{eq:theta update} can be applied here. 
To this end, let $D = \diag(d)\in\R^{S\times S}$, where $d$ is given right after Assumption~\ref{assum:tranistion}, 
\begin{align}\label{eq:definition A}
    & A = \Phi^\top D (U-I) \Phi,  \;\;\; U = (1-\lambda) \sum_{t=0}^\infty \lambda^t (\gamma \bar P)^{t+1}, \nonumber\\
    %b = \sum_{i=1}^N d_i b^i, & b^i =\Phi^\top D \sum_{t=0}^\infty (\gamma\lambda P)^t r^i,
    & b^i =\Phi^\top D \sum_{t=0}^\infty (\gamma\lambda \bar P)^t r^i, \;\;\;i\in\mathcal V,
\end{align}
where $r^i \in \R^S$ whose $k$-th entry is $r^{ik} = \sum_{s=1}^S \bar p^{ks} R^i(k,s)$, and set  $A_{\max} =\frac{1+\gamma}{1-\gamma\lambda} $ and $b_{\max} = \frac{R}{1-\gamma\lambda}$, where $R$ is given in Assumption~\ref{assum:reward}.

\begin{lemma} \label{lemma:assumption_replacement}
    Let the sequences $\{ \theta_t^i \}$, $i \in \mathcal{V}$, be generated by \eqref{eq:theta update_TD} with \eqref{eq:definition Ab(X_t)}.
    If Assumptions~\ref{assum:reward}--\ref{assum:tranistion} hold, so do Assumptions~\ref{assum:A and b}--\ref{assum:lyapunov}. 
\end{lemma}

\noindent
{\bf Proof of Lemma~\ref{lemma:assumption_replacement}:}
Firstly, under Assumptions~\ref{assum:reward}--\ref{assum:tranistion}, we have
\begin{align*} %\label{eq:limiting A_td0}
    \lim_{t\to\infty} \mathbf{E}[A(X_t)] = A, \;\; \lim_{t\to\infty} \mathbf{E}[ B(X_t) ] = \left[
    \begin{array}{c}
        (b^1)^\top \\
        \vdots\\
        (b^N)^\top
    \end{array}
    \right],
\end{align*}
and
$
    \| A(X_t) \|_2 \le \frac{1+\gamma}{1-\gamma\lambda}, \;\;\; \|b(X_t)\|_2 \le \frac{R}{1-\gamma\lambda},
$
where $A(X_t)$ and $b^i(X_t)$ are defined in \eqref{eq:definition Ab(X_t)}, $A$ and $b^i$ are defined in \eqref{eq:definition A}. Since $A_{\max} =1+\gamma $ and $b_{\max} = R$, then we know that Assumption~\ref{assum:A and b} has been satisfied.   Moreover,
\begin{align*}
    \| \mathbf{E}[ b^i(X_t) - b^i | S_0 = s_0, S_1 = s_1 ] \|_2 
    & = \| \sum_{s=1}^S (\mathbf{P}(S_t = s | S_1 = s_1) - \pi_s ) \phi(s) r^{is} \|_2  \\
    & \le  b_{\max} \sum_{i=s}^S |\mathbf{P}(S_t = i | S_1 = s_1) - \pi_i  |, \\
    \| \mathbf{E}[ A(X_t) - A | S_0 = s_0, S_1 = s_1 ] \|_2  
    & = \| \sum_{s=1}^S (\mathbf{P}(S_t = s | S_1 = s_1) - \pi_s ) \phi(s) (\sum_{j=1}^S \bar p^{sj}\gamma \phi(j)^\top - \phi(s)^\top ) \|_2 \\
    & \le A_{\max} \sum_{i=1}^S |\mathbf{P}(S_t = s | S_1 = s_1) - \pi_i  |.
\end{align*}
Since $\{ S_t \}$ is a finite state, aperiodic and irreducible Markov chain, it has a geometric mixing rate \cite{bremaud2013markov}, which implies that Assumption~\ref{assum:mixing-time} holds.
Lastly, when Assumption~\ref{assum:feature} holds, from the proof of Theorem~1 in \cite{tsitsiklis1997analysis}, $A$ given in \eqref{eq:definition A} is a negative definite matrix, i.e., $x^\top A x < 0$ for all $x \in \R^{K}$, which implies that $A+A^\top$ is a symmetric negative definite matrix. From Theorem~7.11 in \cite{linear}, $A$ is a Hurwitz matrix. 
\hfill$\qed$

%Then, using the similar arguments as in Lemma~\ref{lemma:assumption_replacement}, we can show that Assumptions~\ref{assum:reward}--\ref{assum:tranistion} imply Assumptions~\ref{assum:A and b}--\ref{assum:lyapunov}, and thus 
Lemma \ref{lemma:assumption_replacement} implies that 
our analysis for \eqref{eq:theta update} can be applied here. From the proof of Theorem~1 in \cite{tsitsiklis1997analysis}, $A$ in \eqref{eq:definition A} is a negative definite matrix, which implies that $A+A^\top$ is a symmetric negative definite matrix. 
From Theorem~7.11 in \cite{linear}, $A$ is a Hurwitz matrix. 
Let $\sigma_{\min}>0$ be the smallest eigenvalue of $-\frac{1}{2}(A+A^\top)$. Thus, we can also choose $P=I$ in Assumption~\ref{assum:lyapunov} and use the Lyapunov function $V(\langle \theta \rangle_t ) = \| \langle \theta \rangle_t - \theta^* \|_2^2$ in the analysis, where $\theta^*$ here is the limiting point of \eqref{eq:theta update_TD}. Using the same argument as in Theorem~\ref{thm:theta^*_jointly}, we can show that $\theta^* $ is the unique equilibrium point of the ODE \eqref{eq:definition theta^*} with $A$ and $b^i$ being defined in \eqref{eq:definition A}.

%{\color{blue} Let $\theta^* $ be the unique equilibrium point of the ODE \eqref{eq:definition theta^*} with $A$ and $b^i$ are defined in \eqref{eq:definition A}.}

%Firstly, we have that
%\begin{align} \label{eq:limiting A_td}
%    \lim_{t\to\infty} \mathbf{E}[A(X_t)] = A, \;\;\; \lim_{t\to\infty} \mathbf{E}[ B(X_t) ] = \left[
%    \begin{array}{c}
%        (b^1)^\top \\
%        \vdots\\
%        (b^N)^\top
%    \end{array}
%    \right].
%\end{align}
%and
%\begin{align*}
%    \| A(X_t) \|_2 \le \frac{1+\gamma}{1-\gamma\lambda}, \;\;\; \|b^i(X_t)\|_2 \le \frac{R}{1-\gamma\lambda}.
%\end{align*}
%Let $A_{\max} =\frac{1+\gamma}{1-\gamma\lambda} $ and $b_{\max} = \frac{R}{1-\gamma\lambda}$, then we show that Assumption~\ref{assum:A and b} holds. Moreover, since the $\{ S_t \}$ is a finite state, aperiodic and irreducible Markov chain, it has a geometric mixing rate \cite{bremaud2013markov}, which show that Assumption~\ref{assum:mixing-time} holds.

%In addition, the Assumption~\ref{assum:lyapunov} holds when the Assumption~\ref{assum:feature} holds. 

%In this section, we will still use the analysis under the new Lyapunov function $V(\langle \theta \rangle_t )$ which is defined in \eqref{eq:Lyapunov function}.

The finite-time performance of the distributed TD($\lambda$) algorithm is characterized by the following theorem.

\begin{theorem} \label{thm:bound_constant_step_jointly_td}
    Let the sequences $\{ \theta_t^i \}$, $i \in \mathcal{V}$, be generated by \eqref{eq:theta update_TD} with \eqref{eq:definition Ab(X_t)}. Suppose that Assumptions~\ref{assum:weighted matrix} and \ref{assum:limit_pi}--\ref{assum:tranistion} hold and $\{ \bbb{G}_t \}$ is uniformly strongly connected by sub-sequences of length~$L$. Let $  \delta_t$ be the diameter of $\cup_{k=t}^{t+L-1} \bbb{G}_k$ and $  \delta = \max_{t\ge 0}  \delta_t$. 
    Set $A_{\max} =\frac{1+\gamma}{1-\gamma\lambda} $, $b_{\max} = \frac{R}{1-\gamma\lambda}$, and $\sigma_{\min}>0$ be the smallest eigenvalue of $-\frac{1}{2}(A+A^\top)$, where $A$ is given in \eqref{eq:definition A}. 
    Let $0 < \alpha < \min \left\{ \Psi_9 , \; \frac{ \log2}{A_{\max} \tau(\alpha)}, \;  \frac{\sigma_{\min}}{\Psi_3} 
    , \;  \frac{1}{\sigma_{\min}} \right\}.$
    
    {\rm\bf 1) Fixed step-size:}   Let $\alpha_t = \alpha$ for all $t\ge 0$. For all $ t\ge  \hat T_1$,
    \begin{align}
        \sum_{i=1}^N \pi_{t}^i \mathbf{E} \left[\|\theta_{t}^i - \theta^*\|_2^2 \right]
        \le & 2\epsilon^{q_t} \sum_{i=1}^N \pi_{m_t}^i \mathbf{E}\left[\| \theta_{m_t}^i - \langle \theta \rangle_{m_t} \|_2^2\right] \nonumber\\
        &+ (1-\alpha \sigma_{\min} )^{t-  \hat T_1}  \hat C_1 +  \hat C_2   +  \hat C_3 \sum_{k={0}}^{t- \hat T_1}  \eta_{t+1-k}  (1-  \alpha \sigma_{\min} )^{k}. \label{eq:finite-time bound_TD_fixed}
    \end{align}

    {\rm\bf 2) Time-varying step-size:} Let $\alpha_t = \frac{\alpha_0}{t+1}$ with $\alpha_0 \ge \frac{1}{\sigma_{\min}}$. For all $t\ge   \hat T_1L$,
    \begin{align}
        \sum_{i=1}^N \pi_{t}^i \mathbf{E}[\|\theta_{t}^i & - \theta^*\|_2^2] 
        \le 2 \epsilon^{q_t - { \hat T_1}} \sum_{i=1}^N \pi_{ \hat T_1L+m_t}^i \mathbf{E}\left[ \| \theta_{ \hat T_1L+m_t}^i - \langle \theta \rangle_{ \hat T_1L+m_t} \|_2^2\right] \nonumber \\
        & + \hat C_4  \left( \alpha_0 \epsilon^{\frac{q_t-1}{2}} + \alpha_{\ceil{\frac{q_t-1}{2}}L} \right)   + \frac{1}{t} \left(   \hat C_5 \log^2(\frac{{t}}{\alpha_0})
        +   \hat C_6  \sum_{l =  \hat T_1L}^{t} \eta_{l} 
        +  \hat C_7 \right). \label{eq:finite-time bound_TD_timevarying}
    \end{align}
Here $ \hat T_1,  \hat T_1 -  \hat C_7$ are finite constants whose definitions are given in Appendix~\ref{proofs:sa} with 
$A_{\max} =\frac{1+\gamma}{1-\gamma\lambda} $ and $b_{\max} = \frac{R}{1-\gamma\lambda}$.   
\end{theorem}

\subsection{Push-TD($\lambda$)}
%\subsection{Push-TD(\(\lambda\))}
%\subsection{Push-TD(\texorpdfstring{$\lambda$}{lambda})}

In this subsection, we propose a push-based distributed TD($\lambda$) algorithm and provide its finite-time error bounds. Note that push-based distributed TD(0) can be applied in the similar manner.
Each agent $i\in\mathcal V$ updates its variables at each time $t\ge 0$ as follows:
\begin{empheq}[left = \empheqlbrace]{align*}\label{eq:td_push-sum}
y^i_{t+1}&=\sum_{j \in \mathcal{N}_t^i} \hat w_t^{ij } y^j_{t}, \;\;\; y^{i}_0=1,\nonumber\\
\hat \theta^i_{t+1} &= \sum_{j \in \mathcal{N}_t^i} \hat w_t^{ij } \hat\theta^j_{t}  + \alpha_t \Big(A(X_{t}) \hat w_t^{ij } \theta^j_{t} + b^j(X_{t} )\Big),\\
\theta^i_{t+1}&=\frac{\hat \theta^i_{t+1}}{y^i_{t+1}},\nonumber
\end{empheq}
where $\hat w_t^{ij}=1/|\mathcal{N}_t^{j-}|$, $X_t = (s_t, s_{t+1}, z_t)$ is the Markov chain, with $z_t = \sum_{k=0}^t (\gamma \lambda)^{t-k} \phi(s_k)$, $A(X_t)$ and $b^i(X_t)$ are given in \eqref{eq:definition Ab(X_t)}.
Using the same argument as in Theorem~\ref{thm:push_meansq}, we can show that $\theta^* $ is the unique equilibrium point of the ODE \eqref{eq:ode_pushsum} with $A$ and $b^i$ being defined in \eqref{eq:definition A}.

It is not hard to verify that Theorems~\ref{thm:bound_jointly_SA} and \ref{thm:bound_time-varying_step_Push_SA} can be applied to Distributed TD($\lambda$) and Push-TD($\lambda$) to obtain their finite-time performance bounds, respectively.

\section{Analysis and Proofs}\label{analysis}
In this appendix, we provide the analysis of our two algorithms, \eqref{eq:theta update} and \eqref{eq:SA_push-sum}, and the proofs of all the assertions in the paper. We begin with some notation.

\subsection{Notation}

We use $\0_n$ to denote the vector in $\R^n$ whose entries all equal to $0$'s.
For any vector $x\in\R^n$, we use ${\rm diag}(x)$ to denote the $n\times n$ diagonal matrix whose $i$th diagonal entry equals $x^i$.
We use $\|\cdot\|_F$ to denote the Frobenius norm.
For any positive diagonal matrix $W\in\R^{n\times n}$, we use $\|A\|_W$ to denote the weighted Frobenius norm for $A\in\R^{n\times m}$, defined as $\|A\|_W = \|W^{\frac{1}{2}}A\|_F$. It is easy to see that $\|\cdot\|_W$ is a matrix norm.
We use $\mathbf{P}(\cdot)$ to denote the probability of an event and $\mathbf{E}(X)$ to denote the expected value of a random variable $X$.

\subsection{Distributed Stochastic Approximation}\label{proofs:sa}

In this subsection, we analyze the distributed stochastic approximation algorithm \eqref{eq:theta update} and provide the proofs of the results in Section~\ref{sec:SA}.
We begin with the asymptotic performance.

{\bf Proof of Lemma~\ref{lemma:bound_pi_jointly}:}
Since the uniformly strongly connectedness is equivalent to $B$-connectedness as discussed in Remark~\ref{remark:uniformly}, the existence is proved in Lemma 5.8 of \cite{touri2012product}, and the uniqueness is proved in Lemma 1 of \cite{tacrate}. 
\hfill$\qed$

{\bf Proof of Theorem~\ref{thm:consensus_time-varying_jointly}:}
Without loss of generality, let $\{\bbb{G}_t\}$ be uniformly strongly connected by sub-sequences of length $L$.
Note that for any $i\in\scr V$, we have 
\begin{align}\label{eq:proof_th1_1_jointly}
    0 \le \pi_{\min}\|\theta^i_t-\langle \theta\rangle_t\|_2^2 \le \pi_{\min} \sum_{j=1}^N \|\theta^j_t-\langle \theta\rangle_t\|_2^2 \le \sum_{j=1}^N \pi_t^j \|\theta^j_t-\langle \theta\rangle_t\|_2^2,
\end{align}
where $\pi_{\min}$ is defined in Lemma~\ref{lemma:bound_pi_jointly}. 

From Lemma~\ref{lemma:bound_consensus_time-varying_jointly}, 
\begin{align} \label{eq:proof_th1_2_jointly}
    &\;\;\;\;\lim_{t\to\infty}\sum_{i=1}^N \pi_{t}^i \| \theta_{t}^i - \langle \theta \rangle_{t} \|_2^2 \nonumber\\
    & \le \lim_{t\to\infty}\hat \epsilon^{q_t - {T_4^*}} \sum_{i=1}^N \pi_{T_4^*L+m_t}^i \| \theta_{T_4^*L+m_t}^i - \langle \theta \rangle_{T_4^*L+m_t} \|_2^2   + \lim_{t\to\infty} \frac{\zeta_6}{1-\hat\epsilon} \left( \alpha_0 \hat\epsilon^{\frac{q_t-1}{2}} + \alpha_{\ceil{\frac{q_t-1}{2}}L} \right)  = 0. 
\end{align}
Combining \eqref{eq:proof_th1_1_jointly} and \eqref{eq:proof_th1_2_jointly}, it follows that for all $i\in\scr V$, $\lim_{t\to\infty} \pi_{\min}\|\theta^i_t-\langle \theta\rangle_t\|_2^2 = 0$. 
Since $\pi_{\min}>0$ by Lemma~\ref{lemma:bound_pi_jointly}, $\lim_{t\to\infty} \|\theta^i_t-\langle \theta\rangle_t\|_2 = 0$ for all $i\in\scr V$.
\hfill$\qed$

\noindent
{\bf Proof of Theorem~\ref{thm:theta^*_jointly}:}
%{\color{red}Part (1) can be proved by specialization of Theorem~3.1 in \cite{kushner87}.}
From Theorem~\ref{thm:consensus_time-varying_jointly}, all $\theta_t^i$, $i\in \mathcal{V}$, will reach a consensus with $ \langle \theta \rangle_t $ and the update of $ \langle \theta \rangle_t $ is given in \eqref{eq:update of average_time-varying}, which can be treated as a single-agent linear stochastic approximation whose corresponding ODE is \eqref{eq:definition theta^*}.
From \cite{kushner87, kushner1983averaging},\footnote{On page 1289 of \cite{kushner87}, it says that the idea in \cite{kushner1983averaging} can be adapted to get the w.p.1 convergence result.
} 
we know that $ \langle \theta \rangle_t $ will converge to $\theta^*$ w.p.1, which implies that $\theta_t^i$ will converge to $\theta^*$ w.p.1. 
In addition, from  Theorem~\ref{thm:bound_jointly_SA}-(2) and Lemma~\ref{lemma:eta_sum}, 
$\lim_{\rightarrow\infty}\sum_{i=1}^N \pi_t^i \mathbf{E}[\|\theta_t^i - \theta^*\|_2^2]=0$. Since $\pi_t^i$ is uniformly bounded below by $\pi_{\min}>0$, as shown in  Lemma~\ref{lemma:bound_pi_jointly}, 
it follows that $\theta_t^i$ will converge to $\theta^*$ in mean square for all $i\in\scr V$.
\hfill $\qed$

We now analyze the finite-time performance of \eqref{eq:theta update}. In the sequel, we use $K$ to denote the dimension of each $\theta_t^i$, i.e., $\theta_t^i \in \R^K$ for all $i\in\scr V$.

\subsubsection{Fixed Step-size} \label{sec:proof_jointly_fixed}

We first consider the fixed step-size case and begin with validation of two ``convergence rates'' in Theorem~\ref{thm:bound_jointly_SA}.

\begin{lemma}\label{lemma:ration_in_0_1}
    Both $\epsilon$ and $(1-\frac{0.9 \alpha}{\gamma_{\max}})$ lie in the interval $(0,1)$.
\end{lemma}

{\bf Proof of Lemma~\ref{lemma:ration_in_0_1}:}
Since $0<\alpha < K_1 = \min\{ \zeta_1, \; \frac{\gamma_{\max}}{0.9}\}$ as imposed in Theorem~\ref{thm:bound_jointly_SA}, we have $0< \alpha < \zeta_1$ and $0<\alpha< \frac{\gamma_{\max}}{0.9}$. The latter immediately implies that $1-\frac{0.9 \alpha}{\gamma_{\max}} \in (0,1)$. 
From Remark~\ref{remark:exists_Psi9}, $ \epsilon$ is monotonically increasing for $\alpha>0$. In addition, from the definition of $\zeta_1$ in Section~\ref{sec:constants} that if $\alpha = \zeta_1$, then $\epsilon=1$. Since $0<\alpha<\zeta_1$, it follows that $0<\epsilon <1$.
\hfill$\qed$

To proceed, we need the following derivation and lemmas. 

Let $Y_t = \Theta_t - \1_N \langle \theta \rangle_t^\top = (I - \1_N \pi_t^\top)\Theta_t$. For any $t\ge s\ge 0$, let $W_{s:t} = W_t W_{t-1} \cdots W_s$. Then,
%and $w_{s:t}^{ij}$ be the $(i,j)$-th entry of matrix $W_{s:t}$.
\begin{align} \label{eq:update_y_fixed}
    Y_{t+1} &= \Theta_{t+1} - \1_N \langle \theta \rangle_{t+1}^\top \nonumber \\
    &= W_t \Theta_t + \alpha W_t \Theta_t A^\top(X_t) + \alpha B(X_t) - \1_N (\langle \theta \rangle^\top_t + \alpha \langle \theta \rangle^\top_t A^\top(X_t) + \alpha \pi_{t+1}^\top B(X_t) )\nonumber\\
    &= W_t (I - \1_N \pi_t^\top) \Theta_t + \alpha W_t (I - \1_N \pi_t^\top) \Theta_t A^\top(X_t) + \alpha (I - \1_N \pi_{t+1}^\top) B(X_t)\nonumber \\
    &= W_t Y_t + \alpha W_t Y_t A^\top(X_t) + \alpha (I - \1_N \pi_{t+1}^\top) B(X_t).
\end{align}
For simplicity, let $Y_{t}^i$ be the $i$-th column of matrix $Y_{t}^\top$. Then,
\begin{align}\label{eq:yyy_fixed}
    Y_{t+1}^i = \sum_{j=1}^N w_t^{ij} Y^j_t + \alpha A(X_t) \sum_{j=1}^N w_t^{ij} Y^j_t + \alpha \left( b^i(X_t) - B^\top(X_t) \pi_{t+1}\right).
\end{align}
From \eqref{eq:update_y_fixed}, we have
\begin{align}\label{eq:update_y_constant_jointly}
    Y_{t+L}
    &= W_{t+L-1} Y_{t+L-1} ( I + \alpha A^\top(X_{t+L-1})) + \alpha (I - \1_N \pi_{{t+L}}^\top) B(X_{t+L-1}) \nonumber\\
    &= W_{t+L-1}  W_{t+L-2} Y_{t+L-2} ( I + \alpha A^\top(X_{t+L-2}))  ( I + \alpha A^\top(X_{t+L-1})) \nonumber \\
    &\;\;\; + \alpha W_{t+L-1} (I - \1_N \pi_{{t+L-1}}^\top) B(X_{t+L-2}) ( I + \alpha A^\top(X_{t+L-1}))    + \alpha (I - \1_N \pi_{{t+L}}^\top) B(X_{t+L-1})\nonumber \\
    &= W_{t:t+L-1} Y_{t} ( I + \alpha A^\top(X_{t})) \cdots ( I + \alpha A^\top(X_{t+L-1})) + \alpha (I - \1_N \pi_{{t+L}}^\top) B(X_{t+L-1}) \nonumber \\
    & \;\;\; + \alpha \sum_{k=t}^{t+L-2}  W_{k+1:t+L-1}    (I - \1_N \pi_{{k+1}}^\top) B(X_{k}) \left(\Pi_{j=k+1}^{t+L-1} ( I + \alpha A^\top(X_{j})) \right),
\end{align}
and from \eqref{eq:yyy_fixed},
\eq{
    Y_{t+L}^i = \left(\Pi_{k=t}^{t+L-1}( I + \alpha A(X_k)) \right) \sum_{j=1}^N  w_{t:t+L-1}^{ij} Y_{t}^j + \alpha \hat b_{t+L}^i,
\label{eq:xxx}}
where
\begin{align*}
    \hat b_{t+L}^i & = (b^i(X_{t+L-1}) - B(X_{t+L-1})^\top \pi_{{t+L}}) \\
    & \;\;\; + \sum_{k=t}^{t+L-2}  \left(\Pi_{j=k+1}^{t+L-1} ( I + \alpha A(X_{j})) \right) \sum_{j=1}^N w_{k+1:t+L-1}^{ij}    (b^j(X_{k}) - B(X_{k})^\top \pi_{{k+1}}).
\end{align*}

\begin{lemma} \label{lemma:lower_bound_jointly}
    Suppose that Assumption~\ref{assum:weighted matrix} holds and $\{ \bbb{G}_t \}$ is uniformly strongly connected by sub-sequences of length $L$. Then, for all $t \ge 0$, 
    \begin{align*}
        \sum_{i=1}^N \pi_{t+L}^i \sum_{j=1}^N \sum_{k=1}^N w_{t:t+L-1}^{ij} w_{t:t+L-1}^{ik}\| Y_{t}^j - Y_{t}^k \|_2^2 \ge \frac{\pi_{\min} \beta^{2L}}{\delta_{\max}} \sum_{i=1}^N \pi_{t}^i \| Y_{t}^{i}\|_2^2,
    \end{align*}
where $\beta>0$ and $\pi_{\min}>0$ are given in Assumption~\ref{assum:weighted matrix} and Lemma~\ref{lemma:bound_pi_jointly}, respectively.
\end{lemma}
\noindent
{\bf Proof of Lemma~\ref{lemma:lower_bound_jointly}:}
We first consider the case when $K=1$, i.e., $Y_t^i \in \R$ for all $i$. From Lemma~\ref{lemma:bound_pi_jointly}, 
    \begin{align*} 
        \sum_{i=1}^N \pi_{t+L}^i \sum_{j=1}^N \sum_{l=1}^N w_{t:t+L-1}^{ij} w_{t:t+L-1}^{il}\| Y_{t}^j - Y_{t}^l \|_2^2   \ge \pi_{\min} \sum_{i=1}^N \sum_{j=1}^N \sum_{l=1}^N w_{t:t+L-1}^{ij} w_{t:t+L-1}^{il}\| Y_{t}^j - Y_{t}^l \|_2^2.
    \end{align*}
Let $j^*$ and $l^*$ be the indices such that 
$
     |Y_{t}^{j^*} - Y_{t}^{l^*}| = \max_{1\le j,l \le N} | Y_{t}^j - Y_{t}^l |.
$ 
From the definition of~$Y_t$, $ Y_{t}^j - Y_{t}^l = \theta_{t}^{j} - \theta_{t}^{l} $ for all $j,l\in\scr V$, which implies that 
$$
     |Y_{t}^{j^*} - Y_{t}^{l^*}| = \max_{1\le j,l \le N} | Y_{t}^j - Y_{t}^l |
     =  \max_{1\le j,l \le N} | \theta_{t}^j - \theta_{t}^l |=|\theta_{t}^{j^*} - \theta_{t}^{l^*}|.
$$ 
Since $\cup_{k=t}^{t+L-1} \bbb{G}_k$ is a strongly connected graph for all $t\ge0$, 
%then, following the proof of Lemma~\ref{lemma:lower_bound}, 
we can find a shortest path from agent~$j^*$ to agent $l^*$: $( j_0, j_1 ), \cdots, (j_{p-1}, j_p)$ with $j_0 = j^*$, $j_p = l^*$, and $( j_{m-1}, j_m )$ is the edge of graph $\cup_{k=t}^{t+L-1} \bbb{G}_k$, for $1 \le m \le p$, which implies that
%Then, we have
\begin{align} \label{eq:lemma1_2_jointly}
    \sum_{i=1}^N \sum_{j=1}^N \sum_{l=1}^N w_{t:t+L-1}^{ij} w_{t:t+L-1}^{il}\| Y_{t}^j - Y_{t}^l \|_2^2  \ge \sum_{i=1}^N \sum_{m=1}^p w_{t:t+L-1}^{ij_{m-1}} w_{t:t+L-1}^{ij_m} ( Y_{t}^{j_{m-1}} - Y_{t}^{j_m} )^2.
\end{align}
Moreover, we have
\begin{align} \label{eq:lemma1_21_jointly}
    \sum_{i=1}^N w_{t:t+L-1}^{ij_{m-1}} w_{t:t+L-1}^{ij_m} \ge w_{t:t+L-1}^{j_{m-1} j_{m-1}} w_{t:t+L-1}^{j_{m-1} j_m} + w_{t:t+L-1}^{j_{m} j_{m-1}} w_{t:t+L-1}^{j_{m} j_m} \ge \beta^{2L}. 
\end{align} 
Then, from Jensen's inequality, \eqref{eq:lemma1_2_jointly} and \eqref{eq:lemma1_21_jointly}, we have
\begin{align} \label{eq:lemma1_3_jointly}
    \sum_{i=1}^N \pi_{t+L}^i \sum_{j=1}^N \sum_{l=1}^N w_{t:t+L-1}^{ij} w_{t:t+L-1}^{il}\| Y_{t}^j - Y_{t}^l \|_2^2 
    & \ge \pi_{\min} \sum_{i=1}^N \sum_{m=1}^p w_{t:t+L-1}^{ij_{m-1}} w_{t:t+L-1}^{ij_m} ( Y_{t}^{j_{m-1}} - Y_{t}^{j_m} )^2 \nonumber \\
    & \ge \frac{\pi_{\min} \beta^{2L}}{p} ( Y_{t}^{j^*} - Y_{t}^{l^*} )^2  = \frac{\pi_{\min} \beta^{2L}}{ \delta_t} ( \theta_{t}^{j^*} - \theta_{t}^{l^*} )^2.
\end{align}
For the case when $K > 1$, let $Y_{t}^{ik}$ be the $k$-th entry of vector $Y_{t}^i$. Then,
\begin{align*}
    \sum_{i=1}^N \pi_{t+L}^i \sum_{j=1}^N \sum_{l=1}^N w_{t:t+L-1}^{ij} w_{t:t+L-1}^{il}\| Y_{t}^j - Y_{t}^l \|_2^2   =  \sum_{k=1}^K \sum_{i=1}^N \pi_{t+L}^i \sum_{j=1}^N \sum_{l=1}^N w_{t:t+L-1}^{ij} w_{t:t+L-1}^{il} (Y_{t}^{jk} - Y_{t}^{lk})^2 .
\end{align*}
For each entry $k$, we have
\begin{align} \label{eq:lemma1_4_jointly}
    \sum_{i=1}^N \pi_{t+L}^i \sum_{j=1}^N \sum_{l=1}^N w_{t:t+L-1}^{ij} w_{t:t+L-1}^{il} (Y_{t}^{jk} - Y_{t}^{lk})^2  \ge \frac{\pi_{\min} \beta^{2L}}{\delta_{\max}} \max_{1\le j,l \le N} (\theta_{t}^{jk} - \theta_{t}^{lk})^2 , 
\end{align}
where $\theta_t^{ik}$ is the $k$-th entry of vector $\theta_t^{i}$.
Moreover, let $\Theta_t^{\bm{\cdot} k}$ be the $k$-th column of matrix $\Theta_t$. 
Since $2 x_1 x_2 \le x_1^2 + x_2^2$, we have for any $k \in\{1,\ldots, K\}$,
\begin{align*} 
    \sum_{i=1}^N \pi_{t}^i( Y_{t}^{ik} )^2 &= \sum_{i=1}^N \pi_t^i \| \theta_t^{ik} - \pi_t^\top \Theta_t^{\bm{\cdot} k} \|_2^2 
    \le \max_{1\le i \le N} \left( \theta_t^{ik} - \pi_t^\top \Theta_t^{\bm{\cdot} k} \right)^2  = \max_{1\le i \le N} \left( \pi_t^\top (\1_N \theta_t^{ik} - \Theta_t^{\bm{\cdot} k}) \right)^2 \nonumber\\
    & = \max_{1\le i \le N} \Big( \sum_{j=1}^N \pi_t^j ( \theta_t^{ik} - \theta_t^{jk} ) \Big)^2  = \max_{1\le i \le N}  \sum_{j=1}^N \sum_{l=1}^N  \pi_t^j \pi_t^l ( \theta_t^{ik} - \theta_t^{jk} )( \theta_t^{ik} - \theta_t^{lk} )  \nonumber\\
    & \le \max_{1\le i \le N}  \sum_{j=1}^N  (\pi_t^j)^2 ( \theta_t^{ik} - \theta_t^{jk} )^2   
    \le \max_{1\le i \le N}  \sum_{j=1}^N  \pi_t^j ( \theta_t^{ik} - \theta_t^{jk} )^2    \le \max_{1\le i \le N} \max_{1\le j \le N} ( \theta_t^{ik} - \theta_t^{jk} )^2.
\end{align*}
%Then, from \eqref{eq:lemma1_5}, for each entry $k = 1,\ldots,K$, we have
%\begin{align*}
%    \sum_{i=1}^N \pi_{t}^i( Y_{t}^{ik} )^2 = \sum_{i=1}^N \pi_{t}^i ( \theta_{t}^{ik} - \pi_{t}^\top \Theta_{t}^{\bm{\cdot} k} )^2 \le \max_{1\le i \le N} \max_{1\le j \le N} ( \theta_{t}^{ik} - \theta_{t}^{jk} )^2.
%\end{align*}
Then, combining this inequality with \eqref{eq:lemma1_3_jointly} and \eqref{eq:lemma1_4_jointly}, we have
\begin{align*} 
    & \;\;\; \sum_{k=1}^K \sum_{i=1}^N \pi_{t+L}^i \sum_{j=1}^N \sum_{l=1}^N w_{t:t+L-1}^{ij} w_{t:t+L-1}^{il} (Y_{t}^{jk} - Y_{t}^{lk})^2  \\
    & \ge \frac{\pi_{\min} \beta^{2L}}{\delta_{\max}} \sum_{k=1}^K \max_{1\le j,l \le N} (\theta_t^{jk} - \theta_t^{lk})^2 = \frac{\pi_{\min} \beta^{2L}}{\delta_{\max}} \sum_{k=1}^K \sum_{i=1}^N \pi_{t}^i( Y_t^{ik} )^2 = \frac{\pi_{\min} \beta^{2L}}{\delta_{\max}} \sum_{i=1}^N \pi_{t}^i \| Y_{t}^{i} \|_2^2.
\end{align*}
This completes the proof.
\hfill $\qed$

%From the definition of $\zeta_1$ in Section~\ref{sec:constants} and Remark~\ref{remark:exists_Psi9}, we know $ \epsilon$ is monotonically increasing for $\alpha>0$ and when $0<\alpha<\zeta_1$, we have $0<\epsilon <1$.

\begin{lemma} \label{lemma:bound_consensus_jointly} 
    Suppose that Assumptions~\ref{assum:weighted matrix} and \ref{assum:A and b} hold and $\{ \bbb{G}_t \}$ is uniformly strongly connected by sub-sequences of length $L$. Then, when $\alpha \in (0, \zeta_1)$, we have for all $t \ge \tau(\alpha)$, 
$$
    \sum_{i=1}^N \pi_{t}^i \| \theta_{t}^i - \langle \theta \rangle_{t} \|_2^2 \le \epsilon^{q_{t}} \sum_{i=1}^N \pi_{m_t}^i \| \theta_{m_t}^i - \langle \theta \rangle_{m_t} \|_2^2 + \frac{\zeta_2}{1- \epsilon},
$$
where $\zeta_1$ is defined in Appendix~\ref{sec:constants}, $ \epsilon$ and $\zeta_2$ are defined in \eqref{eq:define epsilon_jointly} and \eqref{eq:define Psi10}, respectively.
\end{lemma}
\noindent
{\bf Proof of Lemma~\ref{lemma:bound_consensus_jointly}:} 
Let $M_t = \diag{(\pi_t)}$. From \eqref{eq:xxx},
%Recall the update of $Y_{t+L}^i$, i.e.,
%$$
%    Y_{t+L}^i = \left(\Pi_{k=t}^{t+L-1}( I + \alpha A(X_k)) \right) \sum_{j=1}^N  w_{t:t+L-1}^{ij} Y_{t}^j + \alpha \hat b_{t+L}^i.
%$$
%Then, we have
\begin{align}
    \| Y_{t+L} \|_{M_{t+L}}^2 
    %& = \sum_{i=1}^N \pi_{t+L}^i \| Y_{t+L}^i \|_2^2  \nonumber\\
    &= \sum_{i=1}^N \pi_{t+L}^i \Big\| \Big(\Pi_{k=t}^{t+L-1}( I + \alpha A(X_k)) \Big) \sum_{j=1}^N  w_{t:t+L-1}^{ij} Y_{t}^j \Big\|_2^2 \label{eq:matrix_1_jointly}\\
    &\;\;\; + \alpha^2  \sum_{i=1}^N \pi_{t+L}^i \| \hat b_{t+L}^i\|_2^2 \label{eq:matrix_2_jointly}\\
    & \;\;\; + 2 \alpha \sum_{i=1}^N \pi_{t+L}^i 
    (\hat b_{t+L}^i)^\top \left(\Pi_{k=t}^{t+L-1}( I + \alpha A(X_k)) \right) \sum_{j=1}^N  w_{t:t+L-1}^{ij} Y_{t}^j. \label{eq:matrix_3_jointly}
\end{align}
We derive bounds for \eqref{eq:matrix_1_jointly}--\eqref{eq:matrix_3_jointly} separately. 

For \eqref{eq:matrix_1_jointly}, since $ 2 (x_1)^\top x_2 = \|x_1\|_2^2+\|x_2\|_2^2 - \| x_1 - x_2 \|_2^2 $ and $\pi_{t}^\top = \pi_{t+L}^\top W_{t:t+L-1}$, we have
\begin{align}
    &\;\;\; \sum_{i=1}^N \pi_{t+L}^i \Big\| \Big(\Pi_{k=t}^{t+L-1}( I + \alpha A(X_k)) \Big) \sum_{j=1}^N  w_{t:t+L-1}^{ij} Y_{t}^j \Big\|_2^2  \le  ( 1 + \alpha A_{\max})^{2L}  \sum_{i=1}^N \pi_{t+L}^i \Big\| \sum_{j=1}^N  w_{t:t+L-1}^{ij} Y_{t}^j \Big\|_2^2 \nonumber\\
    & =  (1 + \alpha A_{\max})^{2L} \sum_{i=1}^N \pi_{t+L}^i \sum_{j=1}^N \sum_{l=1}^N w_{t:t+L-1}^{ij} w_{t:t+L-1}^{il}\frac{1}{2} \left( \|Y_{t}^j\|_2^2+\|Y_{t}^l\|_2^2 - \| Y_{t}^j - Y_{t}^l \|_2^2 \right) \nonumber \\
    & = ( 1 + \alpha A_{\max})^{2L} \Big( \sum_{i=1}^N \pi_{t}^i \|Y_{t}^i\|_2^2 - \frac{1}{2} \sum_{i=1}^N \pi_{t+L}^i \sum_{j=1}^N \sum_{l=1}^N w_{t:t+L-1}^{ij} w_{t:t+L-1}^{il}\| Y_{t}^j - Y_{t}^l \|_2^2 \Big) \nonumber.
\end{align}
From Lemma~\ref{lemma:lower_bound_jointly}, 
$
        \sum_{i=1}^N \pi_{t+L}^i \sum_{j=1}^N \sum_{k=1}^N w_{t:t+L-1}^{ij} w_{t:t+L-1}^{ik}\| Y_{t}^j - Y_{t}^k \|_2^2 \ge \frac{\pi_{\min} \beta^{2L}}{\delta_{\max}} \sum_{i=1}^N \pi_{t}^i \| Y_{t}^{i}\|_2^2,
$
which implies that 
\begin{align}
    \sum_{i=1}^N \pi_{t+L}^i \Big\| \Big(\Pi_{k=t}^{t+L-1}( I + \alpha A(X_k)) \Big) \sum_{j=1}^N  w_{t:t+L-1}^{ij} Y_{t}^j \Big\|_2^2  \le ( 1 + \alpha A_{\max})^{2L} \Big(1-\frac{\pi_{\min} \beta^{2L}}{2 \delta_{\max}}\Big) \sum_{i=1}^N \pi_{t}^i \| Y_{t}^{i}\|_2^2. \label{eq:matrix_proof_1_jointly}
\end{align}
For \eqref{eq:matrix_2_jointly}, since $\| b^i(X_t) - B^\top(X_t) \pi_{t+1} \|_2 \le 2 b_{\max}$ for all $i$, 
\begin{align*} 
    \| \hat b_{t+L}^i\|_2 
    & \le \| (b^i(X_{t+L-1}) - B(X_{t+L-1})^\top \pi_{{t+L}}) \|_2 \\
    & \;\;\; + \sum_{k=t}^{t+L-2}  \Big\|\Big(\Pi_{j=k+1}^{t+L-1} ( I + \alpha A(X_{j})) \Big) \Big\|_2 \sum_{j=1}^N w_{k+1:t+L-1}^{ij}    \| (b^j(X_{k}) - B(X_{k})^\top \pi_{{k+1}})\|_2 \\
    & \le 2 b_{\max} \sum_{j=0}^{L-1} ( 1 + \alpha A_{\max})^j \le 2 b_{\max} ( 1 + \alpha A_{\max})^{L-1} \sum_{j=0}^{L-1} \frac{1}{( 1 + \alpha A_{\max})^j} \\
    & \le 2 b_{\max} \frac{(1 + \alpha A_{\max})^L-1}{\alpha A_{\max}},
\end{align*}
which implies that
\begin{align} \label{eq:eq:matrix_proof_21_jointly}
    \alpha^2  \sum_{i=1}^N \pi_{t+L}^i \| \hat b_{t+L}^i\|_2^2 
    & \le  \frac{4 b_{\max}^2}{ A_{\max}^2}\left((1 + \alpha A_{\max})^L-1\right)^2.
\end{align}
For \eqref{eq:matrix_3_jointly}, since $2\| x \|_2 \le 1+\|x\|_2^2 $ holds for any vector $x$,  
\begin{align} \label{eq:matrix_proof_3_jointly}
    &\;\;\;\; 2 \alpha \sum_{i=1}^N \pi_{t+L}^i 
    (\hat b_{t+L}^i)^\top \left(\Pi_{k=t}^{t+L-1}( I + \alpha A(X_k)) \right) \sum_{j=1}^N  w_{t:t+L-1}^{ij} Y_{t}^j \nonumber\\
    & \le 2 \alpha \sum_{i=1}^N \pi_{t+L}^i 
    \| \hat b_{t+L}^i\|_2 \| \Pi_{k=t}^{t+L-1}( I + \alpha A(X_k)) \|_2 \sum_{j=1}^N  w_{t:t+L-1}^{ij} \| Y_{t}^j \|_2 \nonumber \\
    & \le 4 \alpha b_{\max} \frac{(1 + \alpha A_{\max})^L-1}{\alpha A_{\max}} (1 + \alpha A_{\max})^{L} \sum_{i=1}^N \pi_{t}^i \| Y_{t}^i \|_2 \nonumber \\
    & \le 2 b_{\max} \frac{(1 + \alpha A_{\max})^L-1}{ A_{\max}} (1 + \alpha A_{\max})^{L}\Big( \sum_{i=1}^N \pi_{t}^i \| Y_{t}^i \|_2^2+1 \Big).
\end{align}
From \eqref{eq:matrix_proof_1_jointly}--\eqref{eq:matrix_proof_3_jointly}, we have
\begin{align*} 
    \| Y_{t+L} \|_{M_{t+L}}^2 
    & \le ( 1 + \alpha A_{\max})^{2L} \Big(1-\frac{\pi_{\min} \beta^{2L}}{2 \delta_{\max}}\Big) \sum_{i=1}^N \pi_{t}^i \| Y_{t}^{i}\|_2^2  + \frac{4 b_{\max}^2}{ A_{\max}^2}\left((1 + \alpha A_{\max})^L-1\right)^2 \nonumber \\
    & \;\;\; +  2  b_{\max} \frac{(1 + \alpha A_{\max})^L-1}{ A_{\max}} (1 + \alpha A_{\max})^{L} \Big( \sum_{i=1}^N \pi_{t}^i \| Y_{t}^i \|_2^2+1 \Big) \nonumber \\
    & =  \bigg( ( 1 + \alpha A_{\max})^{2L}\Big(1-\frac{\pi_{\min} \beta^{2L}}{2 \delta_{\max}}\Big) + 2  b_{\max} \frac{(1 + \alpha A_{\max})^L-1}{ A_{\max}} (1 + \alpha A_{\max})^{L} \bigg) \| Y_{t} \|_{M_{t}}^2    \nonumber \\
    & \;\;\; + \frac{4 b_{\max}^2}{ A_{\max}^2}\left((1 + \alpha A_{\max})^L-1\right)^2
    + 2 b_{\max} \frac{(1 + \alpha A_{\max})^L-1}{ A_{\max}} (1 + \alpha A_{\max})^{L}.  \nonumber
\end{align*}
From Lemma~\ref{lemma:ration_in_0_1}, $0 < \epsilon <1$ when $ 0 < \alpha < \zeta_1$. With the definition of $ \epsilon$ and $\zeta_2$ in \eqref{eq:define epsilon_jointly} and \eqref{eq:define Psi10},
\begin{align*} 
    \| Y_{t+L} \|_{M_{t+L}}^2
     \le \epsilon \| Y_{t} \|_{M_{t}}^2 + \zeta_2  \le  \epsilon^{q_{t+L}} \| Y_{m_t}\|_{M_{m_t}}^2 + \zeta_2 \sum_{k=0}^{q_{t+L}-1}  \epsilon^k 
     \le \epsilon^{q_{t+L}} \| Y_{m_t}\|_{M_{m_t}}^2 + \frac{\zeta_2}{1- \epsilon},
\end{align*}
which implies that
$$
    \sum_{i=1}^N \pi_{t}^i \| \theta_{t}^i - \langle \theta \rangle_{t} \|_2^2 \le  \epsilon^{q_{t}} \sum_{i=1}^N \pi_{m_t}^i \| \theta_{m_t}^i - \langle \theta \rangle_{m_t} \|_2^2 + \frac{\zeta_2}{1- \epsilon},
$$
where $q_{t}$ and $m_t$ are defined in Theorem~\ref{thm:bound_jointly_SA}.
This completes the proof.
\hfill $\qed$

%\begin{remark} 
%{\color{blue}
%Note that for different time $T^*$ defined in the following theorems, Lemma~\ref{lemma:bound_timevarying_Ab} could still hold for any $t \ge T^*$.
%}
%\end{remark}

\begin{lemma} \label{lemma:fixed_single_3}
    Suppose that Assumptions~\ref{assum:A and b} and \ref{assum:mixing-time} hold and $\{ \bbb{G}_t \}$ is uniformly strongly connected. If the step-size $\alpha$ and corresponding mixing time $\tau(\alpha)$ satisfies
    $
        0< \alpha\tau(\alpha) < \frac{\log2}{A_{\max}}$,
    %0< \alpha < \frac{\sigma_{\min}}{K_2}
    then for any $t \ge \tau(\alpha)$, 
\begin{align}
     \|  \langle \theta \rangle_t - \langle \theta \rangle_{t-\tau(\alpha)}  \|_2 & \le 2 \alpha A_{\max} \tau(\alpha) \|  \langle \theta \rangle_{t-\tau(\alpha)} \|_2 +  2 \alpha \tau(\alpha) b_{\max} \label{eq:fixed_single_3_1}\\
     \|  \langle \theta \rangle_t - \langle \theta \rangle_{t-\tau(\alpha)}  \|_2 & \le 6 \alpha  \tau(\alpha) A_{\max} \|  \langle \theta \rangle_{t} \|_2 +  5 \alpha \tau(\alpha) b_{\max} \label{eq:fixed_single_3_2}\\
     \|  \langle \theta \rangle_t - \langle \theta \rangle_{t-\tau(\alpha)}  \|_2^2 & \le 72 \alpha^2  \tau^2(\alpha) A_{\max}^2 \|  \langle \theta \rangle_{t} \|_2^2 +  50 \alpha^2 \tau^2(\alpha) b_{\max}^2  \le 8 \|  \langle \theta \rangle_{t} \|_2^2 +   \frac{6b_{\max}^2}{A_{\max}^2}. \label{eq:fixed_single_3_3}
\end{align}

\end{lemma}

\noindent
{\bf Proof of Lemma~\ref{lemma:fixed_single_3}:}
With $\alpha_t = \alpha$ for all $t\ge 0$, the update of $\langle \theta \rangle_t$ in \eqref{eq:update of average_time-varying} becomes
$
    \langle \theta \rangle_{t+1} = \langle \theta \rangle_t + \alpha A(X_t) \langle \theta \rangle_t  + \alpha  B(X_t)^\top\pi_{t+1}.
$
Then, 
$$
    \| \langle \theta \rangle_{t+1} \|_2  \le \| \langle \theta \rangle_t \|_2 + \alpha A_{\max} \| \langle \theta \rangle_t \|_2 + \alpha  b_{\max}
      \le (1+\alpha A_{\max}) \| \langle \theta \rangle_t \|_2 + \alpha  b_{\max}.
$$
Using $(1+x)\le \exp(x)$, we have for $u \in [t-\tau(\alpha), t]$,
\begin{align*}
    \| \langle \theta \rangle_{u} \|_2 
    & \le (1+\alpha A_{\max})^{u-t+\tau(\alpha)} \| \langle \theta \rangle_{t-\tau(\alpha)} \|_2 + \alpha  b_{\max} \sum_{l = t-\tau(\alpha)}^{u-1} (1+\alpha A_{\max})^{u-1-l} \\
    & \le (1+\alpha A_{\max})^{\tau(\alpha)} \| \langle \theta \rangle_{t-\tau(\alpha)} \|_2 + \alpha  b_{\max} \sum_{l = t-\tau(\alpha)}^{u-1} (1+\alpha A_{\max})^{u-1-t+\tau(\alpha)} \\
    & \le \exp(\alpha \tau(\alpha) A_{\max}) \| \langle \theta \rangle_{t-\tau(\alpha)} \|_2 + \alpha \tau(\alpha)  b_{\max} \exp(\alpha \tau(\alpha) A_{\max}).
\end{align*}
Since $\alpha \tau(\alpha) A_{\max} \le \log2 < \frac{1}{3}$,  $\exp(\alpha \tau(\alpha) A_{\max}) \le 2$, which implies that
$
    \| \langle \theta \rangle_{u} \|_2 
     \le 2 \| \langle \theta \rangle_{t-\tau(\alpha)} \|_2 + 2 \alpha \tau(\alpha)  b_{\max}.
$
Thus, we can use this to prove \eqref{eq:fixed_single_3_1} for all $t\ge \tau(\alpha)$, i.e.,
\begin{align*}
    \| \langle \theta \rangle_{t} - \langle \theta \rangle_{t - \tau(\alpha)} \|_2 
    & \le \sum_{u=t-\tau(\alpha)}^{t-1}  \| \langle \theta \rangle_{u+1} - \langle \theta \rangle_{u} \|_2  \le \alpha A_{\max} \sum_{u=t-\tau(\alpha)}^{t-1}   \| \langle \theta \rangle_{u} \|_2 + \alpha \tau(\alpha) b_{\max}  \\
    & \le \alpha A_{\max} \sum_{u=t-\tau(\alpha)}^{t-1}   \left( 2 \| \langle \theta \rangle_{t-\tau(\alpha)} \|_2 + 2 \alpha \tau(\alpha)  b_{\max} \right) + \alpha \tau(\alpha) b_{\max}  \\
    & \le 2 \alpha \tau(\alpha) A_{\max} \| \langle \theta \rangle_{t-\tau(\alpha)} \|_2 + 2 \alpha^2 \tau^2(\alpha) A_{\max}  b_{\max}  + \alpha \tau(\alpha) b_{\max}  \\
    & \le 2 \alpha \tau(\alpha) A_{\max} \| \langle \theta \rangle_{t-\tau(\alpha)} \|_2 +  \frac{5}{3} \alpha \tau(\alpha) b_{\max}  \\
    & \le 2 \alpha \tau(\alpha) A_{\max} \| \langle \theta \rangle_{t-\tau(\alpha)} \|_2 +  2 \alpha \tau(\alpha) b_{\max}.
\end{align*}
Moreover, we can prove \eqref{eq:fixed_single_3_2} using the equation above for all $t\ge \tau(\alpha)$ as follows:
\begin{align*}
    \| \langle \theta \rangle_{t} - \langle \theta \rangle_{t - \tau(\alpha)} \|_2 
    & \le 2 \alpha \tau(\alpha) A_{\max} \| \langle \theta \rangle_{t-\tau(\alpha)} \|_2 +  \frac{5}{3} \alpha \tau(\alpha) b_{\max}  \\
    & \le \frac{2}{3} \| \langle \theta \rangle_{t} - \langle \theta \rangle_{t - \tau(\alpha)} \|_2 + 2 \alpha \tau(\alpha) A_{\max} \| \langle \theta \rangle_{t} \|_2 + \frac{5}{3} \alpha \tau(\alpha)  b_{\max}    \\
    & \le 6 \alpha \tau(\alpha) A_{\max} \| \langle \theta \rangle_{t} \|_2 + 5 \alpha \tau(\alpha)  b_{\max}.
\end{align*}
Next, using the inequality $(x+y)^2 \le 2x^2 + 2 y^2$ for all $x, y$, we can show \eqref{eq:fixed_single_3_3} using \eqref{eq:fixed_single_3_2}, i.e.,
\begin{align*}
    \| \langle \theta \rangle_{t} - \langle \theta \rangle_{t - \tau(\alpha)} \|_2^2
    \le 72 \alpha^2 \tau^2(\alpha) A_{\max}^2 \| \langle \theta \rangle_{t} \|_2^2 + 50 \alpha^2 \tau^2(\alpha)  b_{\max}^2 
    \le 8 \| \langle \theta \rangle_{t} \|_2^2 + \frac{ 6b_{\max}^2}{A_{\max}^2},
\end{align*}
where we use $\alpha \tau(\alpha) A_{\max} < \frac{1}{3}$ in the last inequality.
\hfill $\qed$

\newpage
\begin{lemma} \label{lemma:bound_fixed_Ab}
    Let $\mathcal{F}_t = \sigma( X_k,\; k\le t )$ be a $\sigma$-algebra on $\{X_t\}$.
    %to summarize the history information. 
    Suppose that  Assumptions~\ref{assum:A and b}--\ref{assum:lyapunov} and~\ref{assum:limit_pi} hold. If $\{ \bbb{G}_t \}$ is uniformly strongly connected and 
    $
    0< \alpha < \frac{ \log2}{A_{\max} \tau(\alpha)},
$
    then for any $t \ge \tau(\alpha)$, 
    \begin{align*}
    & \;\;\;\; \left|\mathbf{E} \left[ (\langle \theta \rangle_t  - \theta^* )^\top (P+P^\top) \big( A(X_t) \langle \theta \rangle_t  +   B(X_t)^\top\pi_{t+1} - A\langle \theta \rangle_t - b\big) \;|\; \mathcal{F}_{t-\tau(\alpha)} \right]\right| \nonumber \\
    & \le \alpha  \gamma_{\max} \left( 72 + 456 \tau(\alpha) A_{\max}^2  + 84  \tau(\alpha) A_{\max}  b_{\max}  \right) \mathbf{E}\left[ \| \langle \theta \rangle_{t} \|_2^2 \;|\; \mathcal{F}_{t-\tau(\alpha)} \right] \nonumber \\
    &\;\;\; + \alpha \gamma_{\max} \bigg[ 2 + 4 \|\theta^* \|_2^2 +  \frac{48b_{\max}^2}{A_{\max}^2} + \tau(\alpha) A_{\max}^2 \bigg(152 \Big(\frac{b_{\max}}{A_{\max}} + \| \theta^* \|_2 \Big)^2 +  \frac{48b_{\max}}{A_{\max}} \Big(\frac{b_{\max}}{A_{\max}} + 1 \Big)^2 \nonumber\\
    &\;\;\;  +   \frac{87b_{\max}^2}{A_{\max}^2} +  \frac{12b_{\max}}{A_{\max}} \bigg)\bigg] +  2 \gamma_{\max}  \eta_{t+1}\sqrt{N}b_{\max} \Big( 1 +  9 \mathbf{E}[ \| \langle \theta \rangle_{t} \|_2^2 \;|\; \mathcal{F}_{t-\tau(\alpha)} ] + \frac{6b_{\max}^2}{A_{\max}^2}+ \| \theta^* \|_2^2 \Big). 
    \end{align*}
\end{lemma}

\noindent
{\bf Proof of Lemma~\ref{lemma:bound_fixed_Ab}:}
Note that for $t\ge\tau(\alpha)$, we have
\begin{align}
    & \;\;\;\; |\mathbf{E}[ ( \langle \theta \rangle_t  - \theta^* )^\top (P+P^\top)( A(X_t) \langle \theta \rangle_t  +   B(X_t)^\top\pi_{t+1} - A \langle \theta \rangle_t - b) \; | \; \mathcal{F}_{t-\tau(\alpha)} ]| \nonumber \\
    %& \le |\mathbf{E}[ ( \langle \theta \rangle_t  - \theta^* )^\top (P+P^\top)( A(X_t) - A) \langle \theta \rangle_t \; | \; \mathcal{F}_{t-\tau(\alpha)}]| \nonumber\\
    %&\;\;\; +   |\mathbf{E}[ ( \langle \theta \rangle_t  - \theta^* )^\top (P+P^\top)(B(X_t)^\top\pi_{t+1} - b) \; | \; \mathcal{F}_{t-\tau(\alpha)} ]| \nonumber \\
    & \le |\mathbf{E}[ ( \langle \theta \rangle_{t-\tau(\alpha)}  - \theta^* )^\top (P+P^\top)( A(X_t) - A) \langle \theta \rangle_{t-\tau(\alpha)} \; | \; \mathcal{F}_{t-\tau(\alpha)} ]| \label{eq:fixed_bound_Ab_1} \\
    & \;\;\; + |\mathbf{E}[ ( \langle \theta \rangle_{t-\tau(\alpha)}  - \theta^* )^\top (P+P^\top)( A(X_t) - A) ( \langle \theta \rangle_t - \langle \theta \rangle_{t-\tau(\alpha)}) \; | \; \mathcal{F}_{t-\tau(\alpha)} ]| \label{eq:fixed_bound_Ab_2} \\
    & \;\;\; + |\mathbf{E}[  ( \langle \theta \rangle_t - \langle \theta \rangle_{t-\tau(\alpha)})^\top (P+P^\top)( A(X_t) - A) \langle \theta \rangle_{t-\tau(\alpha)} \; | \; \mathcal{F}_{t-\tau(\alpha)} ]| \label{eq:fixed_bound_Ab_3}\\
    & \;\;\; + |\mathbf{E}[  ( \langle \theta \rangle_t - \langle \theta \rangle_{t-\tau(\alpha)})^\top (P+P^\top)( A(X_t) - A)  ( \langle \theta \rangle_t - \langle \theta \rangle_{t-\tau(\alpha)}) \; | \; \mathcal{F}_{t-\tau(\alpha)} ]| \label{eq:fixed_bound_Ab_4}\\
    &\;\;\; +   |\mathbf{E}[ ( \langle \theta \rangle_t - \langle \theta \rangle_{t-\tau(\alpha)})^\top (P+P^\top)(B(X_t)^\top\pi_{t+1} - b) \; | \; \mathcal{F}_{t-\tau(\alpha)} ]|  \label{eq:fixed_bound_Ab_5}\\
    &\;\;\; +   |\mathbf{E}[ ( \langle \theta \rangle_{t-\tau(\alpha)}  - \theta^* )^\top (P+P^\top)(B(X_t)^\top\pi_{t+1} - b) \; | \; \mathcal{F}_{t-\tau(\alpha)} ]|\label{eq:fixed_bound_Ab_6}.
\end{align}
We derive bounds for \eqref{eq:fixed_bound_Ab_1}--\eqref{eq:fixed_bound_Ab_6} separately. 

First, using the mixing time in Assumption~\ref{assum:mixing-time}, we can get the bounds for \eqref{eq:fixed_bound_Ab_1} and \eqref{eq:fixed_bound_Ab_6} for  $t\ge\tau(\alpha)$ as follows:
\begin{align}
    & \;\;\; |\mathbf{E}[ ( \langle \theta \rangle_{t-\tau(\alpha)}  - \theta^* )^\top (P+P^\top)( A(X_t) - A) \langle \theta \rangle_{t-\tau(\alpha)} \; | \; \mathcal{F}_{t-\tau(\alpha)} ]|\nonumber\\
    & \le |( \langle \theta \rangle_{t-\tau(\alpha)}  - \theta^* )^\top (P+P^\top) \mathbf{E}[A(X_t) - A \; | \; \mathcal{F}_{t-\tau(\alpha)} ] \langle \theta \rangle_{t-\tau(\alpha)} | \nonumber\\
    & \le 2 \alpha \gamma_{\max}   \mathbf{E}[\| \langle \theta \rangle_{t-\tau(\alpha)}  - \theta^* \|_2  \| \langle \theta \rangle_{t-\tau(\alpha)}\|_2 \; | \; \mathcal{F}_{t-\tau(\alpha)} ] \nonumber\\
    & \le \alpha \gamma_{\max}   \mathbf{E}[\| \langle \theta \rangle_{t-\tau(\alpha)}  - \theta^* \|_2^2 +  \| \langle \theta \rangle_{t-\tau(\alpha)}\|_2^2 \; | \; \mathcal{F}_{t-\tau(\alpha)} ] \nonumber \\
    & \le \alpha \gamma_{\max}   \mathbf{E}[ 2 \|\theta^* \|_2^2 + 3 \| \langle \theta \rangle_{t-\tau(\alpha)}\|_2^2 \; | \; \mathcal{F}_{t-\tau(\alpha)} ] \nonumber \\
    & \le 6 \alpha \gamma_{\max} \mathbf{E}[ \| \langle \theta \rangle_{t} - \langle \theta \rangle_{t-\tau(\alpha)}\|_2^2 \; | \; \mathcal{F}_{t-\tau(\alpha)} ] + 6 \alpha \gamma_{\max} \mathbf{E}[ \| \langle \theta \rangle_{t} \|_2^2 \; | \; \mathcal{F}_{t-\tau(\alpha)} ] + 2 \alpha \gamma_{\max} \|\theta^* \|_2^2 \nonumber \\
    & \le 54 \alpha \gamma_{\max} \mathbf{E}[ \| \langle \theta \rangle_{t} \|_2^2 \; | \; \mathcal{F}_{t-\tau(\alpha)} ] + 36 \alpha \gamma_{\max} (\frac{b_{\max}}{A_{\max}})^2 + 2 \alpha \gamma_{\max} \|\theta^* \|_2^2,  \label{eq:fixed_bound_Ab_1_bounded}
\end{align}
where in the last inequality, we use \eqref{eq:fixed_single_3_1} from Lemma~\ref{lemma:fixed_single_3}. 

Then, from the definition of $\pi_{\infty}$ in Assumption~\ref{assum:limit_pi},
\begin{align}
    & \;\;\; |\mathbf{E}[ ( \langle \theta \rangle_{t-\tau(\alpha)}  - \theta^* )^\top (P+P^\top)(B(X_t)^\top\pi_{t+1} - b) \; | \; \mathcal{F}_{t-\tau(\alpha)} ]|\nonumber\\
    & \le |\mathbf{E}[ ( \langle \theta \rangle_{t-\tau(\alpha)}  - \theta^* )^\top (P+P^\top)(\sum_{i=1}^N \pi_{t+1}^i(b^i(X_t) - b^i ) +  \sum_{i=1}^N (\pi_{t+1}^i - \pi_{\infty}^i) b^i ) \; | \; \mathcal{F}_{t-\tau(\alpha)} ]|\nonumber\\
    & \le | ( \langle \theta \rangle_{t-\tau(\alpha)}  - \theta^* )^\top (P+P^\top)(\sum_{i=1}^N \pi_{t+1}^i \mathbf{E}[ b^i(X_t) - b^i \; | \; \mathcal{F}_{t-\tau(\alpha)} ] +  \sum_{i=1}^N (\pi_{t+1}^i - \pi_{\infty}^i) b^i ) |\nonumber\\
    & \le 2 \gamma_{\max} (\alpha + \eta_{t+1}\sqrt{N}b_{\max}) \mathbf{E}[ \| \langle \theta \rangle_{t-\tau(\alpha)}  - \theta^* \|_2 \; | \; \mathcal{F}_{t-\tau(\alpha)} ]\nonumber\\
    & \le 2 \gamma_{\max} (\alpha + \eta_{t+1}\sqrt{N}b_{\max}) \left( \mathbf{E}[ \| \langle \theta \rangle_{t-\tau(\alpha)} \|_2 \; | \; \mathcal{F}_{t-\tau(\alpha)} ] + \| \theta^* \|_2 \right) \nonumber\\
    & \le 2 \gamma_{\max} (\alpha + \eta_{t+1}\sqrt{N}b_{\max}) \big( 1 + \frac{1}{2} \mathbf{E}[ \| \langle \theta \rangle_{t-\tau(\alpha)} \|_2^2 \; | \; \mathcal{F}_{t-\tau(\alpha)} ] + \frac{1}{2} \| \theta^* \|_2^2 \big) \nonumber\\
    & \le 2 \gamma_{\max} (\alpha + \eta_{t+1}\sqrt{N}b_{\max}) \left( 1 +  \mathbf{E}[ \| \langle \theta \rangle_{t} - \langle \theta \rangle_{t-\tau(\alpha)} \|_2^2 + \| \langle \theta \rangle_{t} \|_2^2 \; | \; \mathcal{F}_{t-\tau(\alpha)} ] + \| \theta^* \|_2^2 \right) \nonumber\\
    & \le 2 \gamma_{\max} (\alpha + \eta_{t+1}\sqrt{N}b_{\max}) \big( 1 +  9 \mathbf{E}[ \| \langle \theta \rangle_{t} \|_2^2 \; | \; \mathcal{F}_{t-\tau(\alpha)} ] + 6 (\frac{b_{\max}}{A_{\max}})^2+ \| \theta^* \|_2^2 \big), 
    \label{eq:fixed_bound_Ab_6_bounded}
\end{align}  
where we also use \eqref{eq:fixed_single_3_1} from Lemma~\ref{lemma:fixed_single_3} in the last inequality.

Next, using Assumption~\ref{assum:A and b}, \eqref{eq:fixed_single_3_1} and \eqref{eq:fixed_single_3_3}, we have
\begin{align}
    & \;\;\; |\mathbf{E}[ ( \langle \theta \rangle_{t-\tau(\alpha)}  - \theta^* )^\top (P+P^\top)( A(X_t) - A) ( \langle \theta \rangle_t - \langle \theta \rangle_{t-\tau(\alpha)} ) \; | \; \mathcal{F}_{t-\tau(\alpha)} ]| \nonumber\\
    &\le 4 \gamma_{\max} A_{\max} \mathbf{E}[ \| \langle \theta \rangle_{t-\tau(\alpha)} - \theta^* \|_2 \|  \langle \theta \rangle_t - \langle \theta \rangle_{t-\tau(\alpha)}\|_2 \; | \; \mathcal{F}_{t-\tau(\alpha)} ] \nonumber\\
    &\le 4 \gamma_{\max} A_{\max} \mathbf{E}[ \| \langle \theta \rangle_{t-\tau(\alpha)} \|_2 \| \langle \theta \rangle_t - \langle \theta \rangle_{t-\tau(\alpha)}\|_2 + \| \theta^* \|_2 \| \langle \theta \rangle_t - \langle \theta \rangle_{t-\tau(\alpha)}\|_2 \; | \; \mathcal{F}_{t-\tau(\alpha)} ] \nonumber\\
    &\le 8 \alpha \tau(\alpha) \gamma_{\max} A_{\max}^2 \mathbf{E}[ \| \langle \theta \rangle_{t-\tau(\alpha)} \|_2^2  \; | \; \mathcal{F}_{t-\tau(\alpha)} ] + 8 \alpha \tau(\alpha) \gamma_{\max} A_{\max} b_{\max} \| \theta^* \|_2 \nonumber\\
    & \;\;\; + 8 \alpha \tau(\alpha) \gamma_{\max} A_{\max}^2 \big(\frac{ b_{\max}}{A_{\max}} + \| \theta^* \|_2 \big) \mathbf{E}[ \| \langle \theta \rangle_{t-\tau(\alpha)} \|_2  \; | \; \mathcal{F}_{t-\tau(\alpha)} ] \nonumber\\
    &\le 8 \alpha \tau(\alpha) \gamma_{\max} A_{\max}^2 \mathbf{E}[ \| \langle \theta \rangle_{t-\tau(\alpha)} \|_2^2  \; | \; \mathcal{F}_{t-\tau(\alpha)} ] + 8 \alpha \tau(\alpha) \gamma_{\max} A_{\max} b_{\max} \| \theta^* \|_2 \nonumber\\
    & \;\;\; + 4 \alpha \tau(\alpha) \gamma_{\max} A_{\max}^2 \mathbf{E}[ \| \langle \theta \rangle_{t-\tau(\alpha)} \|_2^2  \; | \; \mathcal{F}_{t-\tau(\alpha)} ] 
    + 4 \alpha \tau(\alpha) \gamma_{\max} A_{\max}^2 \big(\frac{ b_{\max}}{A_{\max}} + \| \theta^* \|_2 \big)^2, \nonumber%\\
    %&\le 12 \alpha \tau(\alpha) \gamma_{\max} A_{\max}^2 \mathbf{E}[ \| \langle \theta \rangle_{t-\tau(\alpha)} \|_2^2  \; | \; \mathcal{F}_{t-\tau(\alpha)} ] + 8 \alpha \tau(\alpha) \gamma_{\max}  \left(b_{\max} + A_{\max} \| \theta^* \|_2 \right)^2 \nonumber\\
    %&\le 24 \alpha \tau(\alpha) \gamma_{\max} A_{\max}^2 \mathbf{E}[ \| \langle \theta \rangle_{t} - \langle \theta \rangle_{t-\tau(\alpha)} \|_2^2  \; | \; \mathcal{F}_{t-\tau(\alpha)} ]  + 8 \alpha \tau(\alpha) \gamma_{\max}  \left(b_{\max} + A_{\max} \| \theta^* \|_2 \right)^2 \nonumber\\
    %& \;\;\; + 24 \alpha \tau(\alpha) \gamma_{\max} A_{\max}^2 \mathbf{E}[ \| \langle \theta \rangle_{t}\|_2^2  \; | \; \mathcal{F}_{t-\tau(\alpha)} ] \nonumber\\
    %&\le 216 \alpha \tau(\alpha) \gamma_{\max} A_{\max}^2 \mathbf{E}[ \| \langle \theta \rangle_{t}\|_2^2  \; | \; \mathcal{F}_{t-\tau(\alpha)} ] + 144 \alpha \tau(\alpha) \gamma_{\max} b_{\max}^2 
    %+ 8 \alpha \tau(\alpha) \gamma_{\max}  \left(b_{\max} + A_{\max} \| \theta^* \|_2 \right)^2 \nonumber\\
    %&\le 216 \alpha \tau(\alpha) \gamma_{\max} A_{\max}^2 \mathbf{E}[ \| \langle \theta \rangle_{t}\|_2^2  \; | \; \mathcal{F}_{t-\tau(\alpha)} ] + 152 \alpha \tau(\alpha) \gamma_{\max}  \left(b_{\max} + A_{\max} \| \theta^* \|_2 \right)^2.
    %\label{eq:fixed_bound_Ab_2_bounded} 
\end{align}
which implies that
\begin{align}
    & \;\;\; |\mathbf{E}[ ( \langle \theta \rangle_{t-\tau(\alpha)}  - \theta^* )^\top (P+P^\top)( A(X_t) - A) ( \langle \theta \rangle_t - \langle \theta \rangle_{t-\tau(\alpha)} ) \; | \; \mathcal{F}_{t-\tau(\alpha)} ]| \nonumber\\
    %&\le 4 \gamma_{\max} A_{\max} \mathbf{E}[ \| \langle \theta \rangle_{t-\tau(\alpha)} - \theta^* \|_2 \|  \langle \theta \rangle_t - \langle \theta \rangle_{t-\tau(\alpha)}\|_2 \; | \; \mathcal{F}_{t-\tau(\alpha)} ] \nonumber\\
    %&\le 4 \gamma_{\max} A_{\max} \mathbf{E}[ \| \langle \theta \rangle_{t-\tau(\alpha)} \|_2 \| \langle \theta \rangle_t - \langle \theta \rangle_{t-\tau(\alpha)}\|_2 + \| \theta^* \|_2 \| \langle \theta \rangle_t - \langle \theta \rangle_{t-\tau(\alpha)}\|_2 \; | \; \mathcal{F}_{t-\tau(\alpha)} ] \nonumber\\
    %&\le 8 \alpha \tau(\alpha) \gamma_{\max} A_{\max}^2 \mathbf{E}[ \| \langle \theta \rangle_{t-\tau(\alpha)} \|_2^2  \; | \; \mathcal{F}_{t-\tau(\alpha)} ] + 8 \alpha \tau(\alpha) \gamma_{\max} A_{\max} b_{\max} \| \theta^* \|_2 \nonumber\\
    %& \;\;\; + 8 \alpha \tau(\alpha) \gamma_{\max} A_{\max}^2 \big(\frac{ b_{\max}}{A_{\max}} + \| \theta^* \|_2 \big) \mathbf{E}[ \| \langle \theta \rangle_{t-\tau(\alpha)} \|_2  \; | \; \mathcal{F}_{t-\tau(\alpha)} ] \nonumber\\
    %&\le 8 \alpha \tau(\alpha) \gamma_{\max} A_{\max}^2 \mathbf{E}[ \| \langle \theta \rangle_{t-\tau(\alpha)} \|_2^2  \; | \; \mathcal{F}_{t-\tau(\alpha)} ] + 8 \alpha \tau(\alpha) \gamma_{\max} A_{\max} b_{\max} \| \theta^* \|_2 \nonumber\\
    %& \;\;\; + 4 \alpha \tau(\alpha) \gamma_{\max} A_{\max}^2 \mathbf{E}[ \| \langle \theta \rangle_{t-\tau(\alpha)} \|_2^2  \; | \; \mathcal{F}_{t-\tau(\alpha)} ] 
    %+ 4 \alpha \tau(\alpha) \gamma_{\max} A_{\max}^2 \big(\frac{ b_{\max}}{A_{\max}} + \| \theta^* \|_2 \big)^2 \nonumber\\
    &\le 12 \alpha \tau(\alpha) \gamma_{\max} A_{\max}^2 \mathbf{E}[ \| \langle \theta \rangle_{t-\tau(\alpha)} \|_2^2  \; | \; \mathcal{F}_{t-\tau(\alpha)} ] + 8 \alpha \tau(\alpha) \gamma_{\max}  \left(b_{\max} + A_{\max} \| \theta^* \|_2 \right)^2 \nonumber\\
    &\le 24 \alpha \tau(\alpha) \gamma_{\max} A_{\max}^2 \mathbf{E}[ \| \langle \theta \rangle_{t} - \langle \theta \rangle_{t-\tau(\alpha)} \|_2^2  \; | \; \mathcal{F}_{t-\tau(\alpha)} ]  + 8 \alpha \tau(\alpha) \gamma_{\max}  \left(b_{\max} + A_{\max} \| \theta^* \|_2 \right)^2 \nonumber\\
    & \;\;\; + 24 \alpha \tau(\alpha) \gamma_{\max} A_{\max}^2 \mathbf{E}[ \| \langle \theta \rangle_{t}\|_2^2  \; | \; \mathcal{F}_{t-\tau(\alpha)} ] \nonumber\\
    &\le 216 \alpha \tau(\alpha) \gamma_{\max} A_{\max}^2 \mathbf{E}[ \| \langle \theta \rangle_{t}\|_2^2  \; | \; \mathcal{F}_{t-\tau(\alpha)} ] + 144 \alpha \tau(\alpha) \gamma_{\max} b_{\max}^2 
    + 8 \alpha \tau(\alpha) \gamma_{\max}  \left(b_{\max} + A_{\max} \| \theta^* \|_2 \right)^2 \nonumber\\
    &\le 216 \alpha \tau(\alpha) \gamma_{\max} A_{\max}^2 \mathbf{E}[ \| \langle \theta \rangle_{t}\|_2^2  \; | \; \mathcal{F}_{t-\tau(\alpha)} ] + 152 \alpha \tau(\alpha) \gamma_{\max}  \left(b_{\max} + A_{\max} \| \theta^* \|_2 \right)^2.
    \label{eq:fixed_bound_Ab_2_bounded} 
\end{align}
In additional, using \eqref{eq:fixed_single_3_1} and \eqref{eq:fixed_single_3_3}, we have
\begin{align}
    & \;\;\; |\mathbf{E}[ ( \langle \theta \rangle_t - \langle \theta \rangle_{t-\tau(\alpha)} )^\top (P+P^\top)( A(X_t) - A) \langle \theta \rangle_{t-\tau(\alpha)} \; | \; \mathcal{F}_{t-\tau(\alpha)} ]|\nonumber\\
    & \le 4 \gamma_{\max} A_{\max} \mathbf{E}[ \| \langle \theta \rangle_t - \langle \theta \rangle_{t-\tau(\alpha)}\|_2 \| \langle \theta \rangle_{t-\tau(\alpha)}\|_2 \; | \; \mathcal{F}_{t-\tau(\alpha)} ]\nonumber\\
    & \le 8 \alpha \tau(\alpha) \gamma_{\max} A_{\max} \mathbf{E}[ A_{\max } \|  \langle \theta \rangle_{t-\tau(\alpha)} \|_2^2 +  b_{\max}  \|  \langle \theta \rangle_{t-\tau(\alpha)}\|_2 \; | \; \mathcal{F}_{t-\tau(\alpha)} ]\nonumber\\
    & \le 4 \alpha \tau(\alpha) \gamma_{\max} A_{\max} (2 A_{\max }+  b_{\max} ) \mathbf{E}[  \| \langle \theta \rangle_{t-\tau(\alpha)} \|_2^2 \; | \; \mathcal{F}_{t-\tau(\alpha)} ]  + 4 \alpha \tau(\alpha) \gamma_{\max} A_{\max} b_{\max} \nonumber\\
    & \le 8 \alpha \tau(\alpha) \gamma_{\max} A_{\max} (2 A_{\max }+  b_{\max} ) \mathbf{E}[  \|  \langle \theta \rangle_{t} - \langle \theta \rangle_{t-\tau(\alpha)} \|_2^2 \; | \; \mathcal{F}_{t-\tau(\alpha)} ]   \nonumber\\
    &\;\;\; + 8 \alpha \tau(\alpha) \gamma_{\max} A_{\max} (2 A_{\max }+  b_{\max} ) \mathbf{E}[  \|  \langle \theta \rangle_{t} \|_2^2 \; | \; \mathcal{F}_{t-\tau(\alpha)} ] + 4 \alpha \tau(\alpha) \gamma_{\max} A_{\max} b_{\max} \nonumber \\
    & \le 72 \alpha \tau(\alpha) \gamma_{\max} A_{\max} (2 A_{\max }  +  b_{\max} ) \mathbf{E}[  \| \langle \theta \rangle_{t} \|_2^2 \; | \; \mathcal{F}_{t-\tau(\alpha)} ]   48 \alpha \tau(\alpha) \gamma_{\max} A_{\max} b_{\max} (\frac{b_{\max}}{A_{\max}} + 1 )^2.
    \label{eq:fixed_bound_Ab_3_bounded}
\end{align}
Moreover, we can get the bound for \eqref{eq:fixed_bound_Ab_4} using \eqref{eq:fixed_single_3_3} as follows:
\begin{align}
    & \;\;\; |\mathbf{E}[  ( \langle \theta \rangle_t  - \langle \theta \rangle_{t-\tau(\alpha)} )^\top (P+P^\top)( A(X_t) - A)  ( \langle \theta \rangle_t - \langle \theta \rangle_{t-\tau(\alpha)} ) \; | \; \mathcal{F}_{t-\tau(\alpha)} ]| \nonumber\\
    & \le 4 \gamma_{\max} A_{\max}  \mathbf{E}[ \| \langle \theta \rangle_t - \langle \theta \rangle_{t-\tau(\alpha)}\|_2^2 \; | \; \mathcal{F}_{t-\tau(\alpha)} ]| \nonumber\\
    & \le 4 \gamma_{\max} A_{\max}  \mathbf{E}[ 72 \alpha^2  \tau^2(\alpha) A_{\max}^2 \| \langle \theta \rangle_{t} \|_2^2 +  50 \alpha^2 \tau^2(\alpha) b_{\max}^2 \; | \; \mathcal{F}_{t-\tau(\alpha)} ] \nonumber\\
    & \le   96 \alpha  \tau(\alpha) A_{\max}^2 \gamma_{\max} \mathbf{E}[ \| \langle \theta \rangle_{t} \|_2^2 \; | \; \mathcal{F}_{t-\tau(\alpha)} ] + 67 \alpha \tau(\alpha) b_{\max}^2 \gamma_{\max}.
    \label{fixed_bound_Ab_4_bounded}
\end{align}
Finally, using \eqref{eq:fixed_single_3_2} we can get the bound for \eqref{eq:fixed_bound_Ab_5}:
\begin{align}
    &\;\;\; |\mathbf{E}[ ( \langle \theta \rangle_t  - \langle \theta \rangle_{t-\tau(\alpha)} )^\top (P+P^\top)(B(X_t)^\top\pi_{t+1} - b) \; | \; \mathcal{F}_{t-\tau(\alpha)} ]| \nonumber \\
    & \le 4  \gamma_{\max} b_{\max} \mathbf{E}[ \|\langle \theta \rangle_t - \langle \theta \rangle_{t-\tau(\alpha)}\|_2 \; | \; \mathcal{F}_{t-\tau(\alpha)} ] \nonumber \\
    & \le 4  \gamma_{\max} b_{\max} \mathbf{E}[ 6 \alpha  \tau(\alpha) A_{\max} \| \langle \theta \rangle_{t} \|_2 +  5 \alpha \tau(\alpha) b_{\max} \; | \; \mathcal{F}_{t-\tau(\alpha)} ] \nonumber \\
    & \le 12 \alpha  \tau(\alpha) \gamma_{\max} A_{\max}  b_{\max} \mathbf{E}[\| \langle \theta \rangle_{t} \|_2^2  \; | \; \mathcal{F}_{t-\tau(\alpha)} ] +  12 \alpha  \tau(\alpha) \gamma_{\max} A_{\max}  b_{\max} + 20 \alpha \tau(\alpha) b_{\max}^2 \gamma_{\max}. 
    \label{eq:fixed_bound_Ab_5_bounded}
\end{align}
Then, using \eqref{eq:fixed_bound_Ab_1_bounded}--\eqref{eq:fixed_bound_Ab_5_bounded}, we have
\begin{align*}
    & \;\;\;\; |\mathbf{E}[ ( \langle \theta \rangle_t^\top  - \theta^* )^\top (P+P^\top)( A(X_t) \langle \theta \rangle_t   +   B(X_t)^\top\pi_{t+1} - A \langle \theta \rangle_t - b) \; | \; \mathcal{F}_{t-\tau(\alpha)} ]| \nonumber \\
    & \le 54 \alpha \gamma_{\max} \mathbf{E}[ \| \langle \theta \rangle_{t} \|_2^2 \; | \; \mathcal{F}_{t-\tau(\alpha)} ] + 36 \alpha \gamma_{\max} (\frac{b_{\max}}{A_{\max}})^2 + 2 \alpha \gamma_{\max} \|\theta^* \|_2^2+  12 \alpha  \tau(\alpha) \gamma_{\max} A_{\max}  b_{\max} \nonumber \\
    &\;\;\;+ 216 \alpha \tau(\alpha) \gamma_{\max} A_{\max}^2 \mathbf{E}[ \| \langle \theta \rangle_{t}\|_2^2  \; | \; \mathcal{F}_{t-\tau(\alpha)} ] + 152 \alpha \tau(\alpha) \gamma_{\max}  \left(b_{\max} + A_{\max} \| \theta^* \|_2 \right)^2\nonumber \\
    & \;\;\;+ 72 \alpha \tau(\alpha) \gamma_{\max} A_{\max} (2 A_{\max }  +  b_{\max} ) \mathbf{E}[  \|  \langle \theta \rangle_{t} \|_2^2 \; | \; \mathcal{F}_{t-\tau(\alpha)} ] + 48 \alpha \tau(\alpha) \gamma_{\max} A_{\max} b_{\max} (\frac{b_{\max}}{A_{\max}} + 1 )^2 \nonumber\\
    & \;\;\;+   96 \alpha  \tau(\alpha) A_{\max}^2 \gamma_{\max} \mathbf{E}[ \|  \langle \theta \rangle_{t} \|_2^2 \; | \; \mathcal{F}_{t-\tau(\alpha)} ]  + 12 \alpha  \tau(\alpha) \gamma_{\max} A_{\max}  b_{\max} \mathbf{E}[\|  \langle \theta \rangle_{t} \|_2^2  \; | \; \mathcal{F}_{t-\tau(\alpha)} ]\nonumber \\
    & \;\;\;+  2 \gamma_{\max} (\alpha + \eta_{t+1}\sqrt{N}b_{\max}) \big( 1 +  9 \mathbf{E}[ \| \langle \theta \rangle_{t} \|_2^2 \; | \; \mathcal{F}_{t-\tau(\alpha)}  ] + 6 (\frac{b_{\max}}{A_{\max}})^2+ \| \theta^* \|_2^2 \big) + 87 \alpha \tau(\alpha) b_{\max}^2 \gamma_{\max},\nonumber%\\
    %& \le \alpha  \gamma_{\max} \left( 72 + 456 \tau(\alpha) A_{\max}^2  + 84  \tau(\alpha) A_{\max}  b_{\max}  \right) \mathbf{E}[ \| \langle \theta \rangle_{t} \|_2^2 \; | \; \mathcal{F}_{t-\tau(\alpha)} ] \nonumber \\
    %&\;\;\; + \alpha \gamma_{\max} \Big[ 2 + 4 \|\theta^* \|_2^2 +  48(\frac{b_{\max}}{A_{\max}})^2 + \tau(\alpha) A_{\max}^2 \Big(152 \big(\frac{b_{\max}}{A_{\max}} + \| \theta^* \|_2 \big)^2 + 48 \frac{b_{\max}}{A_{\max}} (\frac{b_{\max}}{A_{\max}} + 1 )^2 \nonumber\\
    %&\;\;\;  +  87 (\frac{b_{\max}}{A_{\max}})^2 +  12 \frac{b_{\max}}{A_{\max}} \Big)\Big] +  2 \gamma_{\max}  \eta_{t+1}\sqrt{N}b_{\max} \Big( 1 +  9 \mathbf{E}[ \| \langle \theta \rangle_{t}\|_2^2 \; | \; \mathcal{F}_{t-\tau(\alpha)} ] + 6 (\frac{b_{\max}}{A_{\max}})^2+ \| \theta^* \|_2^2 \Big).
\end{align*}
which implies that
\begin{align*}
    & \;\;\;\; |\mathbf{E}[ ( \langle \theta \rangle_t^\top  - \theta^* )^\top (P+P^\top)( A(X_t) \langle \theta \rangle_t   +   B(X_t)^\top\pi_{t+1} - A \langle \theta \rangle_t - b) \; | \; \mathcal{F}_{t-\tau(\alpha)} ]| \nonumber \\
    %& \le 54 \alpha \gamma_{\max} \mathbf{E}[ \| \langle \theta \rangle_{t} \|_2^2 \; | \; \mathcal{F}_{t-\tau(\alpha)} ] + 36 \alpha \gamma_{\max} (\frac{b_{\max}}{A_{\max}})^2 + 2 \alpha \gamma_{\max} \|\theta^* \|_2^2+  12 \alpha  \tau(\alpha) \gamma_{\max} A_{\max}  b_{\max} \nonumber \\
    %&\;\;\;+ 216 \alpha \tau(\alpha) \gamma_{\max} A_{\max}^2 \mathbf{E}[ \| \langle \theta \rangle_{t}\|_2^2  \; | \; \mathcal{F}_{t-\tau(\alpha)} ] + 152 \alpha \tau(\alpha) \gamma_{\max}  \left(b_{\max} + A_{\max} \| \theta^* \|_2 \right)^2\nonumber \\
    %& \;\;\;+ 72 \alpha \tau(\alpha) \gamma_{\max} A_{\max} (2 A_{\max }  +  b_{\max} ) \mathbf{E}[  \|  \langle \theta \rangle_{t} \|_2^2 \; | \; \mathcal{F}_{t-\tau(\alpha)} ] + 48 \alpha \tau(\alpha) \gamma_{\max} A_{\max} b_{\max} (\frac{b_{\max}}{A_{\max}} + 1 )^2 \nonumber\\
    %& \;\;\;+   96 \alpha  \tau(\alpha) A_{\max}^2 \gamma_{\max} \mathbf{E}[ \|  \langle \theta \rangle_{t} \|_2^2 \; | \; \mathcal{F}_{t-\tau(\alpha)} ]  + 12 \alpha  \tau(\alpha) \gamma_{\max} A_{\max}  b_{\max} \mathbf{E}[\|  \langle \theta \rangle_{t} \|_2^2  \; | \; \mathcal{F}_{t-\tau(\alpha)} ]\nonumber \\
    %& \;\;\;+  2 \gamma_{\max} (\alpha + \eta_{t+1}\sqrt{N}b_{\max}) \big( 1 +  9 \mathbf{E}[ \| \langle \theta \rangle_{t} \|_2^2 \; | \; \mathcal{F}_{t-\tau(\alpha)}  ] + 6 (\frac{b_{\max}}{A_{\max}})^2+ \| \theta^* \|_2^2 \big) + 87 \alpha \tau(\alpha) b_{\max}^2 \gamma_{\max}\nonumber\\
    & \le \alpha  \gamma_{\max} \left( 72 + 456 \tau(\alpha) A_{\max}^2  + 84  \tau(\alpha) A_{\max}  b_{\max}  \right) \mathbf{E}[ \| \langle \theta \rangle_{t} \|_2^2 \; | \; \mathcal{F}_{t-\tau(\alpha)} ] \nonumber \\
    &\;\;\; + \alpha \gamma_{\max} \Big[ 2 + 4 \|\theta^* \|_2^2 +  48(\frac{b_{\max}}{A_{\max}})^2 + \tau(\alpha) A_{\max}^2 \Big(152 \big(\frac{b_{\max}}{A_{\max}} + \| \theta^* \|_2 \big)^2 + 48 \frac{b_{\max}}{A_{\max}} (\frac{b_{\max}}{A_{\max}} + 1 )^2 \nonumber\\
    &\;\;\;  +  87 (\frac{b_{\max}}{A_{\max}})^2 +  12 \frac{b_{\max}}{A_{\max}} \Big)\Big] +  2 \gamma_{\max}  \eta_{t+1}\sqrt{N}b_{\max} \Big( 1 +  9 \mathbf{E}[ \| \langle \theta \rangle_{t}\|_2^2 \; | \; \mathcal{F}_{t-\tau(\alpha)} ] + 6 (\frac{b_{\max}}{A_{\max}})^2+ \| \theta^* \|_2^2 \Big).
\end{align*}
This completes the proof.
\hfill $\qed$

\begin{lemma} \label{lemma:bound_average}
    Suppose that Assumptions~\ref{assum:A and b}--\ref{assum:lyapunov} and \ref{assum:limit_pi} hold. Then, when 
    $
    0< \alpha < \min\{\frac{ \log2}{A_{\max} \tau(\alpha)},\; \frac{0.1}{K_2 \gamma_{\max}}\},
$
    we have for any $t \ge T_1$,
\begin{align*}
    &\;\;\;\; \mathbf{E}[\|\langle \theta \rangle_{t+1} -\theta^* \|_2^2 ] \\
    & \le \Big(1 - \frac{0.9  \alpha}{\gamma_{\max}} \Big)^{t-T_1}  \frac{\gamma_{\max}}{\gamma_{\min}} \mathbf{E}\left[ \| \langle \theta \rangle_{T_1} - \theta^* \|_2^2 \right] +  \frac{\alpha  \zeta_3 \gamma_{\max}^2}{0.9 \gamma_{\min} }  \\
    &\;\;\; + \frac{\gamma_{\max}}{\gamma_{\min}} \alpha \zeta_4 \sum_{k={0}}^{t-T_1}  \eta_{t+1-k}  \Big(1-\frac{0.9  \alpha}{\gamma_{\max}} \Big)^{k}  \\
    & \le \Big(1 - \frac{0.9  \alpha}{\gamma_{\max}} \Big)^{t+1-T_1} \frac{C_1}{2}  +  \frac{\alpha  \zeta_3 \gamma_{\max}^2}{0.9 \gamma_{\min} }\\
    %& \le \Big(1 - \frac{0.9  \alpha}{\gamma_{\max}} \Big)^{t+1-T_1} \frac{C_1}{2} \frac{\gamma_{\max}}{\gamma_{\min}} ( 4 \exp\left\{ 2 \alpha A_{\max}T_1 \right\}+2)  \mathbf{E}[\|\langle \theta \rangle_{0} -\theta^* \|_2^2] \nonumber\\
    %& + 4 \Big(1 - \frac{0.9  \alpha}{\gamma_{\max}} \Big)^{t+1-T_1} \frac{\gamma_{\max}}{\gamma_{\min}} \exp\left\{ 2\alpha A_{\max}T_1 \right\} \Big( \|\theta^*\|_2 + \frac{b_{\max}}{A_{\max}} \Big)^2 \\
    &\;\;\;  + \frac{\gamma_{\max}}{\gamma_{\min}} \alpha \zeta_4 \sum_{k={0}}^{t-T_1}  \eta_{t+1-k}  \Big(1-\frac{0.9  \alpha}{\gamma_{\max}} \Big)^{k} . 
\end{align*}
where $C_1$, $\zeta_3$, $\zeta_4$ and $K_2$ are defined in Appendix~\ref{sec:thmSA_constant}, \eqref{eq:define Psi4}, \eqref{eq:define Psi5} and \eqref{eq:define Psi3}, respectively.
\end{lemma}

\noindent
{\bf Proof of Lemma~\ref{lemma:bound_average}:}
Let $H(\langle \theta \rangle_t ) = ( \langle \theta \rangle_t - \theta^* )^\top P ( \langle \theta \rangle_t  - \theta^* ) $. From Assumption~\ref{assum:lyapunov}, 
$$
    \gamma_{\min} \| \langle \theta \rangle_t - \theta^* \|_2^2 \le H(\langle \theta \rangle_t ) \le \gamma_{\max} \| \langle \theta \rangle_t - \theta^* \|_2^2.
$$
Moreover, from Assumption~\ref{assum:A and b}, we have for all $t\ge 0$,
\begin{align}
    &\;\;\;\; H( \langle \theta \rangle_{t+1} )  = ( \langle \theta \rangle_{t+1} - \theta^* )^\top P ( \langle \theta \rangle_{t+1} - \theta^* ) \nonumber\\
    & = ( \langle \theta \rangle_t  + \alpha A(X_t) \langle \theta \rangle_t   + \alpha  B(X_t)^\top\pi_{t+1} - \theta^* )^\top P (\langle \theta \rangle_t + \alpha A(X_t) \langle \theta \rangle_t  + \alpha  B(X_t)^\top\pi_{t+1} - \theta^* ) \nonumber\\
    & = ( \langle \theta \rangle_t  - \theta^* )^\top P (\langle \theta \rangle_t - \theta^* ) + \alpha^2 ( A(X_t) \langle \theta \rangle_t )^\top P ( A(X_t) \langle \theta \rangle_t ) \nonumber \\
    & \;\;\;\; + \alpha^2  (B(X_t)^\top\pi_{t+1})^\top P (B(X_t)^\top\pi_{t+1}) + \alpha^2 ( A(X_t) \langle \theta \rangle_t )^\top (P+P^\top)(B(X_t)^\top\pi_{t+1}) \nonumber\\
    & \;\;\;\; + \alpha  ( \langle \theta \rangle_t  - \theta^* )^\top (P+P^\top)( A(X_t) \langle \theta \rangle_t +   B(X_t)^\top\pi_{t+1} - A\langle \theta \rangle_t  - b) \nonumber\\
    & \;\;\;\; + \alpha  ( \langle \theta \rangle_t - \theta^* )^\top P( A\langle \theta \rangle_t + b) + \alpha  ( A\langle \theta \rangle_t + b)^\top P( \langle \theta \rangle_t  - \theta^* )  \nonumber\\
    & = H( \langle \theta \rangle_t ) + \alpha^2 ( A(X_t) \langle \theta \rangle_t )^\top P ( A(X_t) \langle \theta \rangle_t ) + \alpha  ( \langle \theta \rangle_t - \theta^* )^\top (PA+A^\top P ) (\langle \theta \rangle_t -\theta^*) \nonumber \\
    & \;\;\;\; + \alpha^2  (B(X_t)^\top\pi_{t+1})^\top P (B(X_t)^\top\pi_{t+1}) + \alpha^2 ( A(X_t) \langle \theta \rangle_t )^\top (P+P^\top)(B(X_t)^\top\pi_{t+1}) \nonumber\\
    & \;\;\;\; + \alpha  ( \langle \theta \rangle_t - \theta^* )^\top (P+P^\top)( A(X_t) \langle \theta \rangle_t +   B(X_t)^\top\pi_{t+1} - A \langle \theta \rangle_t - b) \label{eq:fixed_proof_1},
\end{align}
where we use the fact that $A\theta^* +b =0 $ on the last equality. 

Next, we can take expectation on both sides of \eqref{eq:fixed_proof_1}. 

From Assumption~\ref{assum:lyapunov} and Lemma~\ref{lemma:bound_fixed_Ab}, we have for $t\ge T_1$,
\begin{align}
    &\;\;\;\; \mathbf{E}[H( \langle \theta \rangle_{t+1} )] \nonumber\\
    & = \mathbf{E}[H( \langle \theta \rangle_t )] + \alpha^2 \mathbf{E}[( A(X_t) \langle \theta \rangle_t )^\top P ( A(X_t) \langle \theta \rangle_t )] - \alpha   \mathbf{E}[\| \langle \theta \rangle_t - \theta^* \|_2^2] \nonumber \\
    & \;\;\;\; + \alpha^2  \mathbf{E}[(B(X_t)^\top\pi_{t+1})^\top P (B(X_t)^\top\pi_{t+1})] + \alpha^2 \mathbf{E}[( A(X_t) \langle \theta \rangle_t )^\top (P+P^\top)(B(X_t)^\top\pi_{t+1})] \nonumber\\
    & \;\;\;\; + \alpha  \mathbf{E}[( \langle \theta \rangle_t   - \theta^* )^\top (P+P^\top)( A(X_t) \langle \theta \rangle_t   +   B(X_t)^\top\pi_{t+1} - A\langle \theta \rangle_t  - b)] \nonumber\\
    & \le \mathbf{E}[H( \langle \theta \rangle_t )] - \alpha\mathbf{E}[\| \langle \theta \rangle_t - \theta^*\|_2^2 ]  + \alpha^2 A_{\max}^2 \gamma_{\max} \mathbf{E}[\| \langle \theta \rangle_t\|_2^2 ] + 2 \alpha^2 A_{\max} b_{\max} \gamma_{\max} \mathbf{E}[\| \langle \theta \rangle_t\|_2 ]  \nonumber\\
    & \;\;\; + \alpha^2 b_{\max}^2 \gamma_{\max} + \alpha^2  \gamma_{\max} \left( 72 + 456 \tau(\alpha) A_{\max}^2  + 84  \tau(\alpha) A_{\max}  b_{\max}  \right) \mathbf{E}[ \| \langle \theta \rangle_{t} \|_2^2 ] \nonumber \\
    &\;\;\; + \alpha^2 \gamma_{\max} \Big[ 2 + 4 \|\theta^* \|_2^2 +  48(\frac{b_{\max}}{A_{\max}})^2 + \tau(\alpha) A_{\max}^2 \Big(152 \big(\frac{b_{\max}}{A_{\max}} + \| \theta^* \|_2 \big)^2 + 48 \frac{b_{\max}}{A_{\max}} (\frac{b_{\max}}{A_{\max}} + 1 )^2  \nonumber\\
    &\;\;\; +  87 (\frac{b_{\max}}{A_{\max}})^2 +  12 \frac{b_{\max}}{A_{\max}} \Big)\Big] +  2 \alpha \gamma_{\max}  \eta_{t+1}\sqrt{N}b_{\max} \Big( 1 +  9 \mathbf{E}[ \| \langle \theta \rangle_{t} \|_2^2] + 6 (\frac{b_{\max}}{A_{\max}})^2+ \| \theta^* \|_2^2 \Big) \nonumber \\
    & \le \mathbf{E}[H( \langle \theta \rangle_t )] - \alpha\mathbf{E}[\| \langle \theta \rangle_t - \theta^*\|_2^2 ] + 2 \alpha^2 A_{\max}^2 \gamma_{\max} \mathbf{E}[\| \langle \theta \rangle_t\|_2^2 ] + 2 \alpha^2 b_{\max}^2 \gamma_{\max} \nonumber\\
    & \;\;\; + \alpha^2  \gamma_{\max} \left( 72 + 456 \tau(\alpha) A_{\max}^2  + 84  \tau(\alpha) A_{\max}  b_{\max}  \right) \mathbf{E}[ \| \langle \theta \rangle_{t} \|_2^2 ] \nonumber \\
    &\;\;\; + \alpha^2 \gamma_{\max} \Big[ 2 + 4 \|\theta^* \|_2^2 +  48(\frac{b_{\max}}{A_{\max}})^2 + \tau(\alpha) A_{\max}^2 \Big(152 \big(\frac{b_{\max}}{A_{\max}} + \| \theta^* \|_2 \big)^2 + 48 \frac{b_{\max}}{A_{\max}} (\frac{b_{\max}}{A_{\max}} + 1 )^2  \nonumber\\
    &\;\;\; +  87 (\frac{b_{\max}}{A_{\max}})^2 +  12 \frac{b_{\max}}{A_{\max}} \Big)\Big] +  2 \alpha \gamma_{\max}  \eta_{t+1}\sqrt{N}b_{\max} \Big( 1 +  9 \mathbf{E}[ \| \langle \theta \rangle_{t}\|_2 ] + 6 (\frac{b_{\max}}{A_{\max}})^2+ \| \theta^* \|_2^2 \Big). \nonumber
\end{align}
Since $ \mathbf{E}[ \| \langle \theta \rangle_{t} \|_2^2 ] \le 2 \mathbf{E}[ \| \langle \theta \rangle_{t}- \theta^*  \|_2^2 ] + 2 \| \theta^*\|_2^ 2 $, we have
\begin{align}
    &\;\;\;\; \mathbf{E}[H( \langle \theta \rangle_{t+1} )] \nonumber\\
    & \le \mathbf{E}[H( \langle \theta \rangle_t )] - \alpha\mathbf{E}[\| \langle \theta \rangle_t - \theta^*\|_2^2 ] + 2 \alpha^2 b_{\max}^2 \gamma_{\max} \nonumber\\
    & \;\;\; + \alpha^2  \gamma_{\max} \left( 72 + 2 A_{\max}^2 + 456 \tau(\alpha) A_{\max}^2  + 84  \tau(\alpha) A_{\max}  b_{\max}  \right) (2 \mathbf{E}[ \| \langle \theta \rangle_{t}- \theta^*  \|_2^2 ] + 2 \| \theta^*\|_2^ 2) \nonumber \\
    &\;\;\; + \alpha^2 \gamma_{\max} \Big[ 2 + 4 \|\theta^* \|_2^2 +  48(\frac{b_{\max}}{A_{\max}})^2 + \tau(\alpha) A_{\max}^2 \Big(152 \big(\frac{b_{\max}}{A_{\max}} + \| \theta^* \|_2 \big)^2 + 48 \frac{b_{\max}}{A_{\max}} (\frac{b_{\max}}{A_{\max}} + 1 )^2  \nonumber\\
    &\;\;\; +  87 (\frac{b_{\max}}{A_{\max}})^2 +  12 \frac{b_{\max}}{A_{\max}} \Big)\Big] +  2 \alpha \gamma_{\max}  \eta_{t+1}\sqrt{N}b_{\max} \Big( 1 +  18 \mathbf{E}[ \| \langle \theta \rangle_{t}- \theta^*  \|_2^2 ] + 6 (\frac{b_{\max}}{A_{\max}})^2+ 19 \| \theta^* \|_2^2 \Big) \nonumber \\
    & \le \mathbf{E}[H( \langle \theta \rangle_t )]  + \left( - \alpha + 2\alpha^2  \gamma_{\max} \left( 72 + 2 A_{\max}^2 + 456 \tau(\alpha) A_{\max}^2  + 84  \tau(\alpha) A_{\max}  b_{\max}  \right)  \right) \mathbf{E}[\| \langle \theta \rangle_t - \theta^*\|_2^2 ] \nonumber\\
    & \;\;\; + 2 \alpha^2  \gamma_{\max} \left( 72 + 2 A_{\max}^2 + 456 \tau(\alpha) A_{\max}^2  + 84  \tau(\alpha) A_{\max}  b_{\max}  \right) \| \theta^*\|_2^2\nonumber \\
    &\;\;\; + \alpha^2 \gamma_{\max} \Big[ 2 + 2 b_{\max}^2 + 4 \|\theta^* \|_2^2 +  48(\frac{b_{\max}}{A_{\max}})^2 + \tau(\alpha) A_{\max}^2 \Big(152 \big(\frac{b_{\max}}{A_{\max}} + \| \theta^* \|_2 \big)^2 \nonumber\\
    &\;\;\; + 48 \frac{b_{\max}}{A_{\max}} (\frac{b_{\max}}{A_{\max}} + 1 )^2 +  87 (\frac{b_{\max}}{A_{\max}})^2 +  12 \frac{b_{\max}}{A_{\max}} \Big)\Big] \nonumber \\
    & \;\;\;+  2 \alpha \gamma_{\max}  \eta_{t+1}\sqrt{N}b_{\max} \Big( 1 +  18 \mathbf{E}[ \| \langle \theta \rangle_{t}- \theta^*  \|_2^2 ] + 6 (\frac{b_{\max}}{A_{\max}})^2+ 19 \| \theta^* \|_2^2 \Big) 
    \nonumber \\
    &\le \mathbf{E}[H( \langle \theta \rangle_t )] + (-\alpha + \alpha^2  \gamma_{\max} K_2 + 36\alpha \eta_{t+1}\sqrt{N}b_{\max} \gamma_{\max}  )\mathbf{E}[\| \langle \theta \rangle_t - \theta^*\|_2^2 ]  + \alpha^2  \zeta_3 \gamma_{\max} + \alpha \gamma_{\max}  \eta_{t+1}\zeta_4. \nonumber
\end{align}
%Since $0< \alpha < K_1 = \min\bigg\{ \zeta_1, \; \frac{\gamma_{\max}}{0.9} \bigg\}$, then 
By Lemma~\ref{lemma:ration_in_0_1}, we know $1-\frac{0.9 \alpha}{\gamma_{\max}} \in (0,1)$. From the definition of $T_1$ and $\alpha < \frac{0.1}{K_2 \gamma_{\max}}$, we have 
%{\color{red} 
%\begin{align}
%    &\;\;\;\; \mathbf{E}[H( \langle \theta \rangle_{t+1} )] \nonumber\\
%    & \le \mathbf{E}[H( \langle \theta \rangle_t )] - 0.9 \alpha \mathbf{E}[\| \langle \theta \rangle_t - \theta^*\|_2^2 ] + \alpha^2  \zeta_3 \gamma_{\max} + \alpha \gamma_{\max}  \eta_{t+1}\zeta_4  \nonumber \\
%    & \le \left(1 - \frac{0.9  \alpha}{\gamma_{\max}} \right)\mathbf{E}[H( \langle \theta \rangle_t )] + \alpha^2  \zeta_3 \gamma_{\max} + \alpha \gamma_{\max}  \eta_{t+1}\zeta_4 \nonumber \\
%    & \le \left(1 - \frac{0.9  \alpha}{\gamma_{\max}} \right)^{t+1-T_1} \mathbf{E}[H( \langle \theta \rangle_{T_1} )] + \left( \alpha^2  \zeta_3 \gamma_{\max} + 2 \alpha \gamma_{\max}  \zeta_4 \right) \sum_{k={T_1}}^t (1-\frac{0.9  \alpha}{\gamma_{\max}} )^{t-k}  \nonumber \\
%    & \le \left(1 - \frac{0.9  \alpha}{\gamma_{\max}} \right)^{t+1-T_1} \mathbf{E}[H( \langle \theta \rangle_{T_1} )] + \left( \alpha  \zeta_3 \gamma_{\max} + 2  \gamma_{\max}  \zeta_4 \right)  \frac{\gamma_{\max}}{0.9  },   
%\end{align}
%which implies that
%\begin{align}
%    &\;\;\;\; \mathbf{E}[\|\langle \theta \rangle_{t+1} -\theta^* \|_2^2 ] \nonumber\\
%    & \le \frac{1}{\gamma_{\min}} \mathbf{E}[H( \langle \theta \rangle_{t+1} )] \nonumber\\
%    & \le \left(1 - \frac{0.9  \alpha}{\gamma_{\max}} \right)^{t+1-T_1} \frac{\gamma_{\max}}{\gamma_{\min}} \mathbf{E}[\| \langle \theta \rangle_{T_1} -\theta^* \|_2^2 ]
%    +  \frac{\gamma_{\max}}{\gamma_{\min}} \frac{\alpha  \zeta_3 \gamma_{\max} + 2  \gamma_{\max}  \zeta_4}{0.9  }.  \label{eq:lemma_fix_1}
%\end{align}
%==================================
\begin{align*}
    \mathbf{E}[H( \langle \theta \rangle_{t+1} )]
    & \le \mathbf{E}[H( \langle \theta \rangle_t )] - 0.9 \alpha \mathbf{E}[\| \langle \theta \rangle_t - \theta^*\|_2^2 ] + \alpha^2  \zeta_3 \gamma_{\max} + \alpha \gamma_{\max}  \eta_{t+1}\zeta_4  \nonumber \\
    & \le \Big(1 - \frac{0.9  \alpha}{\gamma_{\max}} \Big)\mathbf{E}[H( \langle \theta \rangle_t )] + \alpha^2  \zeta_3 \gamma_{\max} + \alpha \gamma_{\max}  \eta_{t+1}\zeta_4 ,
\end{align*}
which implies that
\begin{align*}
    \mathbf{E}[H( \langle \theta \rangle_{t+1} )]
    & \le \Big(1 - \frac{0.9  \alpha}{\gamma_{\max}} \Big)^{t+1-T_1} \mathbf{E}[H( \langle \theta \rangle_{T_1} )] + \alpha^2  \zeta_3 \gamma_{\max} \sum_{k={T_1}}^t \Big(1-\frac{0.9  \alpha}{\gamma_{\max}} \Big)^{t-k} \nonumber \\
    &\;\;\; +  \alpha \gamma_{\max} \zeta_4 \sum_{k={0}}^{t-T_1}  \eta_{t+1-k}  \Big(1-\frac{0.9  \alpha}{\gamma_{\max}} \Big)^{k}  \nonumber \\
    & \le \Big(1 - \frac{0.9  \alpha}{\gamma_{\max}} \Big)^{t+1-T_1} \mathbf{E}[H( \langle \theta \rangle_{T_1} )] + \frac{\alpha  \zeta_3 \gamma_{\max}^2}{0.9} +  \alpha \gamma_{\max} \zeta_4 \sum_{k={0}}^{t-T_1}  \eta_{t+1-k}  \Big(1-\frac{0.9  \alpha}{\gamma_{\max}} \Big)^{k}. 
\end{align*}
In addition, 
\begin{align}
    &\;\;\;\; \mathbf{E}[\|\langle \theta \rangle_{t+1} -\theta^* \|_2^2 ] 
     \le \frac{1}{\gamma_{\min}} \mathbf{E}[H( \langle \theta \rangle_{t+1} )] \nonumber\\
    & \le \Big(1 - \frac{0.9  \alpha}{\gamma_{\max}} \Big)^{t+1-T_1} \frac{\gamma_{\max}}{\gamma_{\min}} \mathbf{E}[\| \langle \theta \rangle_{T_1} -\theta^* \|_2^2 ]  +  \frac{\alpha  \zeta_3 \gamma_{\max}^2}{0.9 \gamma_{\min} } + \frac{\gamma_{\max}}{\gamma_{\min}} \alpha \zeta_4 \sum_{k={0}}^{t-T_1}  \eta_{t+1-k}  \Big(1-\frac{0.9  \alpha}{\gamma_{\max}} \Big)^{k} .  \label{eq:lemma_fix_1}
\end{align}
%where $K_2, \zeta_3$ and $\zeta_4$ are defined in \eqref{eq:define Psi3}, \eqref{eq:define Psi4} and \eqref{eq:define Psi5}.
Next, we consider the bound for $\mathbf{E}[\| \langle \theta \rangle_{T_1} -\theta^* \|_2^2 ]$. Since $1+x \le \exp \{x\}$ for any $x$, we have for any~$t$,
\begin{align*} 
    \| \langle \theta \rangle_{t+1} - \langle \theta \rangle_{0} \|_2
    & = \|  \langle \theta \rangle_t - \langle \theta \rangle_{0} + \alpha A(X_t) (\langle \theta \rangle_t - \langle \theta \rangle_{0} )  + \alpha  B(X_t)^\top\pi_{t+1} + \alpha A(X_t) \langle \theta \rangle_{0} \|_2 \nonumber \\
    & \le (1+ \alpha A_{\max}) \|  \langle \theta \rangle_t - \langle \theta \rangle_{0} \|_2  + \alpha\left( A_{\max} \|\langle \theta \rangle_{0} \|_2 + b_{\max} \right)\nonumber \\
    & \le \alpha\left( A_{\max} \| \langle \theta \rangle_{0} \|_2 + b_{\max} \right) \sum_{l=0}^t\left(1+  \alpha A_{\max}\right)^l\nonumber \\
    & \le \left( A_{\max} \|\langle \theta \rangle_{0} \|_2 + b_{\max} \right) \frac{\left(1+  \alpha A_{\max}\right)^{t+1}}{A_{\max}}\nonumber \\
    & \le \Big( \|\langle \theta \rangle_{0} - \theta^* \|_2 + \|\theta^* \|_2 + \frac{b_{\max}}{A_{\max}} \Big) \exp\left\{  \alpha A_{\max}(t+1)\right\},
\end{align*}
which implies that
$
    \| \langle \theta \rangle_{T_1} - \langle \theta \rangle_{0} \|_2 
     \le ( \|\langle \theta \rangle_{0} - \theta^* \|_2 + \|\theta^* \|_2 + \frac{b_{\max}}{A_{\max}} ) \exp\{  \alpha A_{\max}T_1\}.
$
Then, 
\begin{align}\label{eq:lemma_fix_2}
    &\;\;\;\; \mathbf{E}[\| \langle \theta \rangle_{T_1}  - \theta^* \|_2^2 ] 
     \le 2 \| \langle \theta \rangle_{T_1}  - \langle \theta \rangle_{0}  \|_2^2 + 2 \| \langle \theta \rangle_{0}  -\theta^* \|_2^2\nonumber\\
    & \le ( 4 \exp\left\{ 2 \alpha A_{\max}T_1 \right\}+2)  \mathbf{E}[ \|\langle \theta \rangle_{0} -\theta^* \|_2^2] + 4 \exp\left\{ 2\alpha A_{\max}T_1 \right\} \Big( \|\theta^*\|_2 + \frac{b_{\max}}{A_{\max}} \Big)^2.
\end{align}
From \eqref{eq:lemma_fix_1} and \eqref{eq:lemma_fix_2}, we have 
%{\color{red}
%\begin{align*}
%    \mathbf{E}[\|\langle \theta \rangle_{t+1} -\theta^* \|_2^2 ] 
%    & \le \left(1 - \frac{0.9  \alpha}{\gamma_{\max}} \right)^{t+1-T_1} \frac{\gamma_{\max}}{\gamma_{\min}} ( 4 \exp\left\{ 2 \alpha A_{\max}T_1 \right\}+2)  \mathbf{E}[\|\langle \theta \rangle_{0} -\theta^* \|_2^2] \nonumber\\
%    &\;\;\; + 4 \left(1 - \frac{0.9  \alpha}{\gamma_{\max}} \right)^{t+1-T_1} \frac{\gamma_{\max}}{\gamma_{\min}} \exp\left\{ 2\alpha A_{\max}T_1 \right\} ( \|\theta^*\|_2 + \frac{b_{\max}}{A_{\max}} )^2 \\
%    & \;\;\; + \frac{\gamma_{\max}}{\gamma_{\min}} \frac{\alpha  \zeta_3 \gamma_{\max} + 2  \gamma_{\max}  \zeta_4}{0.9  }. 
%\end{align*}
%===========================
\begin{align*}
    \mathbf{E}[\|\langle \theta \rangle_{t+1} -\theta^* \|_2^2 ] 
    & \le \Big(1 - \frac{0.9  \alpha}{\gamma_{\max}} \Big)^{t+1-T_1} \frac{\gamma_{\max}}{\gamma_{\min}} ( 4 \exp\left\{ 2 \alpha A_{\max}T_1 \right\}+2)  \mathbf{E}[\|\langle \theta \rangle_{0} -\theta^* \|_2^2] \nonumber\\
    &\;\;\; + 4 \Big(1 - \frac{0.9  \alpha}{\gamma_{\max}} \Big)^{t+1-T_1} \frac{\gamma_{\max}}{\gamma_{\min}} \exp\left\{ 2\alpha A_{\max}T_1 \right\} \Big( \|\theta^*\|_2 + \frac{b_{\max}}{A_{\max}} \Big)^2 \\
    & \;\;\; +  \frac{\alpha  \zeta_3 \gamma_{\max}^2}{0.9 \gamma_{\min} }  + \frac{\gamma_{\max}}{\gamma_{\min}} \alpha \zeta_4 \sum_{k={0}}^{t-T_1}  \eta_{t+1-k}  \Big(1-\frac{0.9  \alpha}{\gamma_{\max}} \Big)^{k} . 
\end{align*}
This completes the proof.
\hfill $\qed$

%\vspace{.1in}

We are now in a position to prove the fixed step-size case in Theorem~\ref{thm:bound_jointly_SA}.

%\vspace{.1in}

\noindent
{\bf Proof of Case 1) in Theorem~\ref{thm:bound_jointly_SA}:}
%{\color{red}
%Note that Lemma~\ref{lemma:bound_average} still holds when $\{\bbb{G}_t\}$ is uniformly strongly connected.
%Using the same argument as in the proof of Lemma~\ref{lemma:bound_average}, the lemma still holds here when $T^*_1$ is replaced by $T^*_3$. 
%} 
From Lemmas~\ref{lemma:bound_consensus_jointly} and \ref{lemma:bound_average}, we have for any $t \ge T_1$,
%\begin{align*}
%        & \;\;\;\; \sum_{i=1}^N \pi_{t}^i \mathbf{E}[\|\theta_{t}^i - \theta^*\|_2^2] \nonumber\\
%        & \le 2 \sum_{i=1}^N \pi_{t}^i \mathbf{E}[\|\theta_{t}^i - \langle \theta \rangle_{t} \|_2^2 ] + 2 \mathbf{E} [\| \langle \theta \rangle_{t} - \theta^*\|_2^2] \\
%        &\le 2  \epsilon^{q_{t}} \sum_{i=1}^N \pi_{m_t}^i \mathbf{E}[\| \theta_{m_t}^i - \langle \theta \rangle_{m_t} \|_2^2] + \frac{2 \zeta_2}{1- \epsilon} + \frac{\gamma_{\max}}{\gamma_{\min}}\cdot \frac{ 2 \alpha  \zeta_3 \gamma_{\max} + 4 \gamma_{\max}  \zeta_4}{0.9  }  \nonumber\\
%        &\;\;\; + \left(1 - \frac{0.9  \alpha}{\gamma_{\max}} \right)^{{t}-T_1}\frac{\gamma_{\max}}{\gamma_{\min}} ( 8 \exp\left\{ 2 \alpha A_{\max}T_1 \right\}+4)  \mathbf{E}[\| \langle \theta \rangle_{0} -\theta^* \|_2^2] \nonumber\\
%        &\;\;\; + 8 \left(1 - \frac{0.9  \alpha}{\gamma_{\max}} \right)^{{t}-T_1} \frac{\gamma_{\max}}{\gamma_{\min}} \exp\left\{ 2\alpha A_{\max}T_1 \right\} ( \|\theta^*\|_2 + \frac{b_{\max}}{A_{\max}} )^2 \\
%        &\le 2 \epsilon^{q_{t}} \sum_{i=1}^N \pi_{m_t}^i \mathbf{E}\left[\left\| \theta_{m_t}^i - \langle \theta \rangle_{m_t} \right\|_2^2 \right]  + C_1 \bigg( 1 - \frac{0.9  \alpha}{\gamma_{\max}} \bigg)^{{t}-T_1}  + C_2,
%\end{align*}
\begin{align*}
        &\;\;\;\; \sum_{i=1}^N \pi_{t}^i \mathbf{E}[\|\theta_{t}^i - \theta^*\|_2^2] \le 2 \sum_{i=1}^N \pi_{t}^i \mathbf{E}[\|\theta_{t}^i - \langle \theta \rangle_{t} \|_2^2 ] + 2 \mathbf{E} [\| \langle \theta \rangle_{t} - \theta^*\|_2^2] \\
        &\le 2  \epsilon^{q_{t}} \sum_{i=1}^N \pi_{m_t}^i \mathbf{E}[\| \theta_{m_t}^i - \langle \theta \rangle_{m_t} \|_2^2] + \frac{2 \zeta_2}{1- \epsilon} +  \frac{2 \alpha  \zeta_3 \gamma_{\max}^2}{0.9 \gamma_{\min} } + \frac{\gamma_{\max}}{\gamma_{\min}} 2\alpha \zeta_4 \sum_{k={0}}^{t-T_1}  \eta_{t+1-k}  \Big(1-\frac{0.9  \alpha}{\gamma_{\max}} \Big)^{k}  \nonumber\\
        &\;\;\; + \Big(1 - \frac{0.9  \alpha}{\gamma_{\max}} \Big)^{{t}-T_1}\frac{\gamma_{\max}}{\gamma_{\min}} ( 8 \exp\left\{ 2 \alpha A_{\max}T_1 \right\}+4)  \mathbf{E}[\| \langle \theta \rangle_{0} -\theta^* \|_2^2] \nonumber\\
        &\;\;\; + 8 \Big(1 - \frac{0.9  \alpha}{\gamma_{\max}} \Big)^{{t}-T_1} \frac{\gamma_{\max}}{\gamma_{\min}} \exp\left\{ 2\alpha A_{\max}T_1 \right\} \Big( \|\theta^*\|_2 + \frac{b_{\max}}{A_{\max}} \Big)^2 \\
        &\le 2 \epsilon^{q_{t}} \sum_{i=1}^N \pi_{m_t}^i \mathbf{E}[\| \theta_{m_t}^i - \langle \theta \rangle_{m_t} \|_2^2 ]  + C_1 \Big( 1 - \frac{0.9  \alpha}{\gamma_{\max}} \Big)^{{t}-T_1}  + C_2 + \frac{\gamma_{\max}}{\gamma_{\min}} 2\alpha \zeta_4 \sum_{k={0}}^{t-T_1}  \eta_{t+1-k}  \Big(1-\frac{0.9  \alpha}{\gamma_{\max}} \Big)^{k},
\end{align*}
where $C_1$ and $C_2$ are defined in Appendix~\ref{sec:thmSA_constant}.
This completes the proof.
\hfill $\qed$

%{\color{blue} Note that, we have $\lim_{t\to\infty} \sum_{k=0}^{t-T_1} \eta_{t+1-k} (1-\frac{0.9 \alpha}{\gamma_{\max}})^k \le \lim_{t\to\infty} \frac{\gamma_{\max}}{0.9 \alpha} [ \eta_{\ceil{\frac{t+1}{2}}} + \eta_1 (1-\frac{0.9 \alpha}{\gamma_{\max}})^{\frac{t}{2}} ] = 0 $}

\subsubsection{Time-varying Step-size} \label{sec:proof_jointly_time-varying}

In this subsection, we consider the time-varying step-size case and begin with a property of $\eta_t$.

\begin{lemma} \label{lemma:eta_sum}
    Suppose that Assumption~\ref{assum:limit_pi} holds. Then, $\lim_{t \to \infty} \eta_t =0$ and $\lim_{t \to \infty} \frac{1}{t+1} \sum_{k=0}^t \eta_{k} = 0.$
\end{lemma}

\noindent
{\bf Proof of Lemma~\ref{lemma:eta_sum}:}
From Assumption~\ref{assum:limit_pi}, we know that $\pi_t$ will converge to $ \pi_\infty$, and thus $\eta_t$ will converge to 0. 
Next, we will prove that $\lim_{t \to \infty} \frac{1}{t+1} \sum_{k=0}^t \eta_{k} = 0.$ For any positive constant $ c > 0$, there exists a positive integer $ T(c)$, depending on $c$, such that $ \forall t \ge T(c) $, we have $\eta_{t} < c$. Thus,
\begin{align*}
    \frac{1}{t} \sum_{k=0}^{t-1} \eta_{k} 
     = \frac{1}{t}\sum_{k=0}^{T(c)} \eta_k + \frac{1}{t}\sum_{k=T(c)+1}^{t-1} \eta_k 
     \le \frac{1}{t}\sum_{k=0}^{T(c)} \eta_k + \frac{t-1-T(c)}{t} c.
\end{align*}
Let $t \to \infty$ on both sides of the above inequality. Then, we have 
\begin{align*}
    \lim_{t \to \infty}\frac{1}{t} \sum_{k=0}^{t-1} \eta_{k} 
    & \le \lim_{t \to \infty} \frac{1}{t}\sum_{k=0}^{T(c)} \eta_k + \lim_{t \to \infty} \frac{t-1-T(c)}{t} c = c.
\end{align*}
Since the above argument holds for arbitrary positive $c$, then $\lim_{t \to \infty} \frac{1}{t+1} \sum_{k=0}^t \eta_{k} = 0.$
\hfill $\qed$

\vspace{.1in}

%Recall the updates corresponding to the time-varying step-size case given in
From the updates of the time-varying step-size case, given in
\eqref{eq:updtae_Theta} and \eqref{eq:update of average_time-varying}, %namely
%\begin{align*}
%    \Theta_{t+1} &= W_t \Theta_t + \alpha_t W_t \Theta_t A(X_t)^\top + \alpha_t B(X_t),\\
%     \langle \theta \rangle_{t+1} &= \langle \theta \rangle_t + \alpha_t A(X_t) \langle \theta \rangle_t  + \alpha_t B(X_t)^\top \pi_{t+1}.
%\end{align*}
%From 
and using \eqref{eq:update_y_constant_jointly}, the update for $Y_{t}$ in this case can be written as
\begin{align*}
    Y_{t+L}
    &= W_{t:t+L-1} Y_t ( I + \alpha_{t} A^\top(X_{t})) \cdots ( I + \alpha_{t+L-1} A^\top(X_{t+L-1}))  + \alpha_{t+L-1} (I - \1_N \pi_{{t+L}}^\top) B(X_{t+L-1}) \nonumber \\
    & \;\;\; +  \sum_{k=t}^{t+L-2} \alpha_k W_{k+1:t+L-1}    (I - \1_N \pi_{{k+1}}^\top) B(X_{k}) \left(\Pi_{j=k+1}^{t+L-1} ( I + \alpha_j A^\top(X_{j})) \right),
\end{align*}
and 
$
    Y_{t+L}^i = \big(\Pi_{k=t}^{t+L-1}( I + \alpha_k A(X_k)) \big) \sum_{j=1}^N  w_{t:t+L-1}^{ij} Y_{t}^j +  \tilde b_{t+L}^i,
$
where
\begin{align*}
    \tilde b_{t+L}^i & = \alpha_{t+L-1} (b^i(X_{t+L-1}) - B(X_{t+L-1})^\top \pi_{{t+L}}) \\
    & \;\;\; + \sum_{k=t}^{t+L-2} \alpha_k \left(\Pi_{j=k+1}^{t+L-1} ( I + \alpha_j A(X_{j})) \right) \sum_{j=1}^N w_{k+1:t+L-1}^{ij}    (b^j(X_{k}) - B(X_{k})^\top \pi_{{k+1}}).
\end{align*}

To prove the theorem, we need the following lemmas.

\begin{lemma} \label{lemma:bound_consensus_time-varying_jointly} 
Suppose that Assumptions~\ref{assum:weighted matrix} and \ref{assum:A and b} hold and $\{ \bbb{G}_t \}$ is uniformly strongly connected by sub-sequences of length $L$. 
%    Let $q_t$ and $m_t$ be the unique integer quotient and remainder of $t$ divided by $L$, respectively.
    Given $\alpha_t$ and $T_2$ defined in Theorem~\ref{thm:bound_jointly_SA}, for all $t \ge T_2L $,
    \begin{align*} 
        \sum_{i=1}^N \pi_{t}^i \| \theta_{t}^i - \langle \theta \rangle_{t} \|_2^2 
        & \le \epsilon^{q_{t}-{T_2}}  \sum_{i=1}^N \pi_{T_2L + m_t }^i \| \theta_{T_2L + m_t }^i - \langle \theta \rangle_{T_2L+ m_t } \|_2^2 + \frac{\zeta_6}{1-\epsilon} \left(  \epsilon^{\frac{q_t - 1}{2}}  \alpha_{m_t} + \alpha_{\ceil{\frac{q_t - 1}{2}} L+m_t} \right) \\
        & \le  \epsilon^{q_t - {T_2}} \sum_{i=1}^N \pi_{T_2L+m_t}^i \| \theta_{T_2L+m_t}^i - \langle \theta \rangle_{T_2L+m_t} \|_2^2 + \frac{\zeta_6}{1-\epsilon} \left( \alpha_0 \epsilon^{\frac{q_t-1}{2}} + \alpha_{\ceil{\frac{q_t-1}{2}}L} \right),
    \end{align*}
where $\epsilon$ and $\zeta_6$ are defined in \eqref{eq:define epsilon_jointly} and \eqref{eq:define Psi11}, respectively.
\end{lemma}

\noindent
{\bf Proof of Lemma~\ref{lemma:bound_consensus_time-varying_jointly}:} 
Similar to the proof of Lemma~\ref{lemma:bound_consensus_jointly}, we have
\begin{align}
    \| Y_{t+L} \|_{M_{t+L}}^2 
    &= \sum_{i=1}^N \pi_{t+L}^i \| \left(\Pi_{k=t}^{t+L-1}( I + \alpha_k A(X_k)) \right) \sum_{j=1}^N  w_{t:t+L-1}^{ij} Y_{t}^j\|_2^2 \label{eq:matrix_1_time_jointly}\\
    &\;\;\; + \sum_{i=1}^N \pi_{t+L}^i \| \tilde b_{t+L}^i\|_2^2 \label{eq:matrix_2_time_jointly}\\
    & \;\;\; + 2 \sum_{i=1}^N \pi_{t+L}^i 
    (\tilde b_{t+L}^i)^\top \left(\Pi_{k=t}^{t+L-1}( I + \alpha_k A(X_k)) \right) \sum_{j=1}^N  w_{t:t+L-1}^{ij} Y_{t}^j. \label{eq:matrix_3_time_jointly}
\end{align}
%%%%%%%%%%%%%%%%%%%%%
By Lemma~\ref{lemma:lower_bound_jointly}, the item given by \eqref{eq:matrix_1_time_jointly} can be bounded as follows:
\begin{align}
    &\;\;\; \sum_{i=1}^N \pi_{t+L}^i \| \left(\Pi_{k=t}^{t+L-1}( I + \alpha_k A(X_k)) \right) \sum_{j=1}^N  w_{t:t+L-1}^{ij} Y_{t}^j\|_2^2 \nonumber\\
    & \le  \Pi_{k=t}^{t+L-1} ( 1 + \alpha_k A_{\max})^{2} \bigg[ \sum_{i=1}^N \pi_{t}^i \|Y_{t}^i\|_2^2 - \frac{1}{2} \sum_{i=1}^N \pi_{t+L}^i \sum_{j=1}^N \sum_{l=1}^N w_{t:t+L-1}^{ij} w_{t:t+L-1}^{il}\| Y_{t}^j - Y_{t}^l \|_2^2 \bigg] \nonumber \\
    & \le  \Pi_{k=t}^{t+L-1} ( 1 + \alpha_k A_{\max})^{2}   \Big(1-\frac{\pi_{\min} \beta^{2L}}{2 \delta_{\max}}\Big) \sum_{i=1}^N \pi_{t}^i \| Y_{t}^{i}\|_2^2. \label{eq:matrix_proof_1_time_jointly}
\end{align}
Since $\| b^i(X_{t}) - B(X_{t})^\top \pi_{{t+1}} \|_2 \le 2 b_{\max}$ holds for all $i$, 
\begin{align*} 
    \| \tilde b_{t+L}^i\|_2 
    & \le \alpha_{t+L-1} \| (b^i(X_{t+L-1}) - B(X_{t+L-1})^\top \pi_{{t+L}}) \|_2 \\
    & \;\;\; + \sum_{k=t}^{t+L-2} \alpha_k \|\left(\Pi_{j=k+1}^{t+L-1} ( I + \alpha_j A(X_{j})) \right) \|_2 \sum_{j=1}^N w_{k+1:t+L-1}^{ij}    \| (b^j(X_{k}) - B(X_{k})^\top \pi_{{k+1}})\|_2 \\
    & \le 2 b_{\max} \bigg[ \alpha_{t+L-1}  + \sum_{k=t}^{t+L-2} \alpha_k \left(\Pi_{j=k+1}^{t+L-1} ( 1 + \alpha_j A_{\max}) \right)\bigg].
\end{align*}
Then, we can bound the item given by \eqref{eq:matrix_2_time_jointly} as follows:
\begin{align} \label{eq:eq:matrix_proof_21_time_jointly}
    \sum_{i=1}^N \pi_{t+L}^i \| \tilde b_{t+L}^i\|_2^2 
    & \le  4 b_{\max}^2 \bigg( \alpha_{t+L-1}  + \sum_{k=t}^{t+L-2} \alpha_k \left(\Pi_{j=k+1}^{t+L-1} ( 1 + \alpha_j A_{\max}) \right)\bigg)^2.
\end{align}
As for the item given by \eqref{eq:matrix_3_time_jointly}, we have
\begin{align} \label{eq:matrix_proof_3_time_jointly}
    &\;\;\;\; 2 \sum_{i=1}^N \pi_{t+L}^i 
    (\tilde b_{t+L}^i)^\top \left(\Pi_{k=t}^{t+L-1}( I + \alpha_k A(X_k)) \right) \sum_{j=1}^N  w_{t:t+L-1}^{ij} Y_{t}^j \nonumber\\
    & \le 2 \sum_{i=1}^N \pi_{t+L}^i 
    \| \tilde b_{t+L}^i\|_2  \| \Pi_{k=t}^{t+L-1}( I + \alpha_k A(X_k)) \|_2 \sum_{j=1}^N  w_{t:t+L-1}^{ij} \| Y_{t}^j \|_2 \nonumber \\
    & \le 2 b_{\max} \Big( \alpha_{t+L-1}  + \sum_{k=t}^{t+L-2} \alpha_k \left(\Pi_{j=k+1}^{t+L-1} ( 1 + \alpha_j A_{\max}) \right)\Big) \left(\Pi_{k=t}^{t+L-1}( I + \alpha_k A_{\max}) \right)  \Big(\sum_{i=1}^N \pi_{t}^i \| Y_{t}^i \|_2^2 +1 \Big).
\end{align}
From \eqref{eq:matrix_proof_1_time_jointly}--\eqref{eq:matrix_proof_3_time_jointly}, we have 
\begin{align*} 
    & \;\;\; \| Y_{t+L} \|_{M_{t+L}}^2 \\
    & \le \Pi_{k=t}^{t+L-1} ( 1 + \alpha_k A_{\max})^{2}   \Big(1-\frac{\pi_{\min} \beta^{2L}}{2 \delta_{\max}}\Big) \sum_{i=1}^N \pi_{t}^i \| Y_{t}^{i}\|_2^2   \nonumber\\
    & \;\;\;+ 4 b_{\max}^2 \Big( \alpha_{t+L-1}  + \sum_{k=t}^{t+L-2} \alpha_k \Big(\Pi_{j=k+1}^{t+L-1} ( 1 + \alpha_j A_{\max}) \Big)\Big)^2 \nonumber\\
    & \;\;\; +2 b_{\max} \Big( \alpha_{t+L-1}  + \sum_{k=t}^{t+L-2} \alpha_k \Big(\Pi_{j=k+1}^{t+L-1} ( 1 + \alpha_j A_{\max}) \Big)\Big) \Big(\Pi_{k=t}^{t+L-1}( I + \alpha_k A_{\max}) \Big)   \Big(\sum_{i=1}^N \pi_{t}^i \| Y_{t}^i \|_2^2 +1 \Big) \nonumber \\
    & =  \bigg[  2 b_{\max} \Big( \alpha_{t+L-1}  + \sum_{k=t}^{t+L-2} \alpha_k \Big(\Pi_{j=k+1}^{t+L-1} ( 1 + \alpha_j A_{\max}) \Big)\Big) \Big(\Pi_{k=t}^{t+L-1}( I + \alpha_k A_{\max}) \Big) \nonumber\\
    & \;\;\; + \Pi_{k=t}^{t+L-1} ( 1 + \alpha_k A_{\max})^{2}   \Big(1-\frac{\pi_{\min} \beta^{2L}}{2 \delta_{\max}}\Big)  \bigg] \| Y_{t} \|_{M_{t}}^2    \nonumber \\
    & \;\;\;+ 4 b_{\max}^2 \Big( \alpha_{t+L-1}  + \sum_{k=t}^{t+L-2} \alpha_k \Big(\Pi_{j=k+1}^{t+L-1} ( 1 + \alpha_j A_{\max}) \Big)\Big)^2 \nonumber\\
    & \;\;\; +2 b_{\max} \Big( \alpha_{t+L-1}  + \sum_{k=t}^{t+L-2} \alpha_k \Big(\Pi_{j=k+1}^{t+L-1} ( 1 + \alpha_j A_{\max}) \Big)\Big) \Big(\Pi_{k=t}^{t+L-1}( I + \alpha_k A_{\max}) \Big) \\
    & =    \epsilon_{t} \| Y_{t} \|_{M_{t}}^2 + 4 b_{\max}^2 \Big( \alpha_{t+L-1}  + \sum_{k=t}^{t+L-2} \alpha_k \Big(\Pi_{j=k+1}^{t+L-1} ( 1 + \alpha_j A_{\max}) \Big)\Big)^2 \nonumber\\
    & \;\;\; +2 b_{\max} \Big( \alpha_{t+L-1}  + \sum_{k=t}^{t+L-2} \alpha_k \Big(\Pi_{j=k+1}^{t+L-1} ( 1 + \alpha_j A_{\max}) \Big)\Big) \Big(\Pi_{k=t}^{t+L-1}( I + \alpha_k A_{\max}) \Big),
\end{align*}
where
\begin{align*}
     \epsilon_{t} & =  2 b_{\max} \Big( \alpha_{t+L-1}  + \sum_{k=t}^{t+L-2} \alpha_k \Big(\Pi_{j=k+1}^{t+L-1} ( 1 + \alpha_j A_{\max}) \Big)\Big) \Big(\Pi_{k=t}^{t+L-1}( I + \alpha_k A_{\max}) \Big) \nonumber\\
    & \;\;\; + \Pi_{k=t}^{t+L-1} ( 1 + \alpha_k A_{\max})^{2}   \Big(1-\frac{\pi_{\min} \beta^{2L}}{2 \delta_{\max}}\Big).
\end{align*}
Since  $\alpha_{t} \le \alpha$ when $t\ge T_2L$, we have for $t\ge T_2L$, $0 \le  \epsilon_{t} \le  \epsilon \le 1$ and 
\begin{align*} 
    \alpha_{t+L-1}  + \sum_{k=t}^{t+L-2} \alpha_k \big(\Pi_{j=k+1}^{t+L-1} ( 1 + \alpha_j A_{\max}) \big)
     \le \sum_{k=t}^{t+L-1} \alpha_k  ( 1 +  \alpha A_{\max})^{t+L-k-1}
     \le ( 1 +  \alpha A_{\max})^{L-1} \sum_{k=t}^{t+L-1} \alpha_k.
\end{align*}
Since $\sum_{k=t}^{t+L-1} \alpha_k \le L\alpha_{t} \le L \alpha$, 
\begin{align*} 
    \| Y_{t+L} \|_{M_{t+L}}^2 
    & \le    \epsilon \| Y_{t} \|_{M_{t}}^2 + 4 b_{\max}^2 ( 1 +  \alpha A_{\max})^{2L-2} \Big(\sum_{k=t}^{t+L-1} \alpha_k\Big)^2  +2 b_{\max} ( 1 +  \alpha A_{\max})^{2L-1} \Big(\sum_{k=t}^{t+L-1} \alpha_k\Big) \\
    & \le    \epsilon \| Y_{t} \|_{M_{t}}^2 +\left( 4 b_{\max}^2 \alpha L^2  ( 1 +  \alpha A_{\max})^{2L-2}   +2 b_{\max}L ( 1 +  \alpha A_{\max})^{2L-1} \right)   \alpha_{t} \\
    & \le    \epsilon \| Y_{t} \|_{M_{t}}^2 + \zeta_6 \alpha_{t},
\end{align*}
where $ \epsilon$ and $\zeta_6$ are defined in \eqref{eq:define epsilon_jointly} and \eqref{eq:define Psi11}, respectively. Then,
\begin{align*} 
    \| Y_{t+L} \|_{M_{t+L}}^2 
    & \le   \epsilon \| Y_{t} \|_{M_{t}}^2 + \zeta_6 \alpha_{t}  \\
    & \le \epsilon^{q_{t+L}-{T_2}} \| Y_{m_t + T_2L}\|_{M_{m_t + T_2L}}^2 + \zeta_6  \sum_{k=T_2}^{q_t} \epsilon^{q_t-k}  \alpha_{kL+m_t} \\
    & \le  \epsilon^{q_{t+L}-{T_2}} \| Y_{T_2L + m_t}\|_{M_{T_2L + m_t}}^2 + \zeta_6 \Big( \sum_{k=0}^{\floor{\frac{q_t}{2}}}  \epsilon^{q_t-k}  \alpha_{kL+m_t} + \sum_{k=\ceil{\frac{q_t}{2}}}^{q_t}  \epsilon^{q_t-k}  \alpha_{kL+m_t} \Big) \\
    & \le  \epsilon^{q_{t+L}-{T_2}} \| Y_{T_2L + m_t}\|_{M_{T_2L + m_t}}^2 + \frac{\zeta_6}{1-\epsilon} \left(  \epsilon^{\frac{q_t}{2}}  \alpha_{m_t} + \alpha_{\ceil{\frac{q_t}{2}} L+m_t} \right),
\end{align*}
which implies that
\begin{align*} 
    \sum_{i=1}^N \pi_{t}^i \| \theta_{t}^i - \langle \theta \rangle_{t} \|_2^2 
    & \le  \epsilon^{q_{t}-{T_2}}  \sum_{i=1}^N \pi_{T_2L + m_t }^i \| \theta_{T_2L + m_t }^i - \langle \theta \rangle_{T_2L+ m_t } \|_2^2  + \frac{\zeta_6}{1-\epsilon} \left(  \epsilon^{\frac{q_t - 1}{2}}  \alpha_{m_t} + \alpha_{\ceil{\frac{q_t - 1}{2}} L+m_t} \right) \\
    & \le  \epsilon^{q_t - {T_2}} \sum_{i=1}^N \pi_{T_2L+m_t}^i \| \theta_{T_2L+m_t}^i - \langle \theta \rangle_{T_2L+m_t} \|_2^2 + \frac{\zeta_6}{1-\epsilon} \left( \alpha_0 \epsilon^{\frac{q_t-1}{2}} + \alpha_{\ceil{\frac{q_t-1}{2}}L} \right) .
\end{align*}
This completes the proof.
\hfill $\qed$

%{\color{blue}
%\begin{lemma} \label{lemma:bound_consensus_time-varying_jointly_allt} 
%    Suppose that Assumptions~\ref{assum:weighted matrix} and \ref{assum:A and b} hold and $\{ \bbb{G}_t \}$ is uniformly strongly connected by sub-sequences of length $L$. 
%    Let $q_t$ and $m_t$ be the unique integer quotient and remainder of $t$ divided by $L$, respectively.
%    Given $\alpha_t$ and $T_2$ defined in Theorem~\ref{thm:bound_time-varying_step}, for all $t \ge T_2L $,
%    \begin{align*} 
%        \sum_{i=1}^N \pi_{t}^i \| \theta_{t}^i - \langle \theta \rangle_{t} \|_2^2
%        & \le \hat \epsilon^{q_{t}-{T_2}}  \sum_{i=1}^N \pi_{T_2L + m_t }^i \| \theta_{T_2L + m_t }^i - \langle \theta \rangle_{T_2L+ m_t } \|_2^2 + \frac{\zeta_6}{1-\hat\epsilon} \left(  \hat\epsilon^{\frac{q_t - 1}{2}}  \alpha_{m_t} + \alpha_{\ceil{\frac{q_t - 1}{2}} L+m_t} \right) \\
%        & \le \hat \epsilon^{q_t - {T_2}} \sum_{i=1}^N \pi_{T_2L+m_t}^i \| \theta_{T_2L+m_t}^i - \langle \theta \rangle_{T_2L+m_t} \|_2^2 + \frac{\zeta_6}{1-\hat\epsilon} \left( \alpha_0 \hat\epsilon^{\frac{q_t-1}{2}} + \alpha_{\ceil{\frac{q_t-1}{2}}L} \right),
%    \end{align*}
%where $\hat\epsilon$ and $\zeta_6$ is defined in \eqref{eq:define epsilon_jointly} and \eqref{eq:define Psi11}, respectively.
%\end{lemma}}

\newpage

\begin{lemma} \label{lemma:timevarying_single_3}
    Suppose that Assumptions~\ref{assum:A and b} and \ref{assum:mixing-time} hold. When the step-size $\alpha_t$ and corresponding mixing time $\tau(\alpha_t)$ satisfy
    $
    0< \alpha_t \tau(\alpha_t) < \frac{\log2}{A_{\max}},
$
    we have for any $t \ge T_2L$, 
\begin{align}
     \|  \langle \theta \rangle_t - \langle \theta \rangle_{t-\tau(\alpha_{t})} \|_2 & \le 2  A_{\max} \|  \langle \theta \rangle_{t-\tau(\alpha_t)} \|_2 \sum_{k=t-\tau(\alpha_t)}^{t-1} \alpha_k +  2  b_{\max} \sum_{k=t-\tau(\alpha_t)}^{t-1} \alpha_k, \label{eq:timevarying_single_3_1}\\
     \| \langle \theta \rangle_t - \langle \theta \rangle_{t-\tau(\alpha_{t})} \|_2 & \le 6  A_{\max} \| \langle \theta \rangle_{t} \|_2 \sum_{k=t-\tau(\alpha_t)}^{t-1} \alpha_k +  5 b_{\max} \sum_{k=t-\tau(\alpha_t)}^{t-1} \alpha_k, \label{eq:timevarying_single_3_2}\\
     \|  \langle \theta \rangle_t - \langle \theta \rangle_{t-\tau(\alpha_{t})} \|_2^2 & \le 72 \alpha_{t-\tau(\alpha_t)}^2  \tau^2(\alpha_t) A_{\max}^2 \|  \langle \theta \rangle_{t} \|_2^2 +  50 \alpha_{t-\tau(\alpha_t)}^2 \tau^2(\alpha_t) b_{\max}^2 \nonumber\\
     & \le 8 \|  \langle \theta \rangle_{t} \|_2^2 +   \frac{6b_{\max}^2}{A_{\max}^2}. \label{eq:timevarying_single_3_3}
\end{align}
\end{lemma}

%{\color{red}use $\tau^2(\alpha_t)$ instead of $\tau(\alpha_t)^2$} {\color{blue} I will change it.}

\noindent
{\bf Proof of Lemma~\ref{lemma:timevarying_single_3}:}
From the update of $\langle \theta \rangle_t $ in \eqref{eq:update of average_time-varying},
$
    \| \langle \theta \rangle_{t+1} \|_2  \le \| \langle \theta \rangle_t \|_2 + \alpha_{t} A_{\max} \| \langle \theta \rangle_t \|_2 + \alpha_{t}  b_{\max}
      \le (1+\alpha_{t} A_{\max}) \| \langle \theta \rangle_t \|_2 + \alpha_{t}  b_{\max}.
$
Similar to the proof of Lemma~\ref{lemma:fixed_single_3}, for all $u \in [t-\tau(\alpha_{t}), t]$,
\begin{align*}
     \| \langle \theta \rangle_{u} \|_2 
    & \le \Pi_{k = t-\tau(\alpha_{t})}^{u-1}  (1+\alpha_{k} A_{\max})\| \langle \theta \rangle_{t-\tau(\alpha_{t})} \|_2 + b_{\max} \sum_{k = t-\tau(\alpha_{t})}^{u-1} \alpha_k \Pi_{l=k+1}^{u-1}  (1+\alpha_{l} A_{\max}) \\
    & \le \exp\{ \sum_{k = t-\tau(\alpha_{t})}^{u-1}  \alpha_{k} A_{\max}\} \| \langle \theta \rangle_{t-\tau(\alpha_{t})} \|_2 + b_{\max} \sum_{k = t-\tau(\alpha_{t})}^{u-1} \alpha_k \exp\{ \sum_{l=k+1}^{u-1} \alpha_{l} A_{\max}\} \\
    & \le \exp\{ \alpha_{t-\tau(\alpha_{t})} \tau(\alpha_t) A_{\max}\} \| \langle \theta \rangle_{t-\tau(\alpha_{t})} \|_2 + b_{\max} \sum_{k = t-\tau(\alpha_{t})}^{u-1} \alpha_k \exp\{ \alpha_{t-\tau(\alpha_{t})} \tau(\alpha_t) A_{\max}\} \\
    & \le 2 \| \langle \theta \rangle_{t-\tau(\alpha_{t})} \|_2 + 2 b_{\max} \sum_{k = t-\tau(\alpha_{t})}^{u-1} \alpha_k,  
\end{align*}
where we use $\alpha_{t-\tau(\alpha_{t})} \tau(\alpha_t) A_{\max} \le \log2 < \frac{1}{3}$ in the last inequality.
Thus, for all $t\ge T_2L$, we can get \eqref{eq:timevarying_single_3_1} as follows:
\begin{align*}
    \| \langle \theta \rangle_{t} - \langle \theta \rangle_{t - \tau(\alpha_{t})} \|_2 
    & \le \sum_{k=t-\tau(\alpha_{t})}^{t-1}  \| \langle \theta \rangle_{k+1} - \langle \theta \rangle_{k} \|_2  \le A_{\max} \sum_{k=t-\tau(\alpha_{t})}^{t-1} \alpha_{k} \| \langle \theta \rangle_{k} \|_2 + b_{\max} \sum_{k=t-\tau(\alpha_{t})}^{t-1} \alpha_{k} \\
    & \le A_{\max} \sum_{k=t-\tau(\alpha_{t})}^{t-1}  \alpha_k \Big( 2 \| \langle \theta \rangle_{t-\tau(\alpha_{t})} \|_2 + 2 b_{\max} \sum_{l=t-\tau(\alpha_{t})}^{k-1} \alpha_l \Big) + b_{\max} \sum_{k=t-\tau(\alpha_{t})}^{t-1} \alpha_{k}  \\
    & \le 2 A_{\max} \| \langle \theta \rangle_{t-\tau(\alpha_{t})} \|_2 \sum_{k=t-\tau(\alpha_{t})}^{t-1}  \alpha_k + \left( 2 A_{\max} \tau(\alpha_{t}) \alpha_{t-\tau(\alpha_{t})} + 1 \right) b_{\max} \sum_{k=t-\tau(\alpha_{t})}^{t-1} \alpha_{k}  \\
    & \le 2 A_{\max} \| \langle \theta \rangle_{t-\tau(\alpha_{t})} \|_2 \sum_{k=t-\tau(\alpha_{t})}^{t-1}  \alpha_k + \frac{5}{3} b_{\max} \sum_{k=t-\tau(\alpha_{t})}^{t-1} \alpha_{k}  \\
    & \le 2 A_{\max} \| \langle \theta \rangle_{t-\tau(\alpha_{t})} \|_2 \sum_{k=t-\tau(\alpha_{t})}^{t-1}  \alpha_k + 2 b_{\max} \sum_{k=t-\tau(\alpha_{t})}^{t-1} \alpha_{k} .
\end{align*}
Moreover, using the above inequality, we can get \eqref{eq:timevarying_single_3_2} for $t\ge T_2L$ as follows:
\begin{align*}
    &\;\;\;\; \| \langle \theta \rangle_{t} - \langle \theta \rangle_{t - \tau(\alpha_{t})} \|_2 
     \le 2 A_{\max} \| \langle \theta \rangle_{t-\tau(\alpha_{t})} \|_2 \sum_{k=t-\tau(\alpha_{t})}^{t-1}  \alpha_k + \frac{5}{3} b_{\max} \sum_{k=t-\tau(\alpha_{t})}^{t-1} \alpha_{k}  \\
    & \le 2 A_{\max} \tau(\alpha_{t}) \alpha_{t-\tau(\alpha_{t})} 
    \| \langle \theta \rangle_{t} - \langle \theta \rangle_{t-\tau(\alpha_{t})} \|_2 
    + 2 A_{\max} \| \langle \theta \rangle_{t} \|_2 \sum_{k=t-\tau(\alpha_{t})}^{t-1}  \alpha_k   + \frac{5}{3} b_{\max} \sum_{k=t-\tau(\alpha_{t})}^{t-1} \alpha_{k}  \\
    & \le 6 A_{\max} \| \langle \theta \rangle_{t} \|_2 \sum_{k=t-\tau(\alpha_{t})}^{t-1}  \alpha_k 
    + 5 b_{\max} \sum_{k=t-\tau(\alpha_{t})}^{t-1} \alpha_{k}.
\end{align*}
Next, using \eqref{eq:timevarying_single_3_2} and the inequality $(x+y)^2 \le 2x^2 + 2y^2$ for any $x, y$, we can get  \eqref{eq:timevarying_single_3_3} as follows:
\begin{align*}
    \| \langle \theta \rangle_{t} - \langle \theta \rangle_{t - \tau(\alpha_{t})} \|_2^2
    & \le 72 A_{\max}^2 \| \langle \theta \rangle_{t} \|_2^2 (\sum_{k=t-\tau(\alpha_{t})}^{t-1} \alpha_k )^2
    + 50 b_{\max}^2 (\sum_{k=t-\tau(\alpha_{t})}^{t-1} \alpha_{k})^2 \\
    & \le 72 \alpha_{t-\tau(\alpha_{t})}^2 \tau^2(\alpha_t) A_{\max}^2 \| \langle \theta \rangle_{t} \|_2^2 + 50 \alpha_{t-\tau(\alpha_{t})}^2 \tau^2(\alpha_t)  b_{\max}^2 \\
    & \le 8 \| \langle \theta \rangle_{t} \|_2^2 + 6 (\frac{ b_{\max}}{A_{\max}})^2,
\end{align*}
where we use $\alpha_{t-\tau(\alpha_{t})} \tau(\alpha_t) A_{\max} < \frac{1}{3}$ in the last inequality.
\hfill $\qed$

\begin{lemma} \label{lemma:bound_timevarying_Ab}
    Suppose that Assumptions~\ref{assum:A and b}--\ref{assum:limit_pi} hold and  $\{ \bbb{G}_t \}$ is uniformly strongly connected. Then, when  
    $
    0< \alpha_{t - \tau(\alpha_t) } \tau(\alpha_t) < \frac{ \log2}{A_{\max} },
$
    %for any $t \ge T_2$, we have
    we have for any $t \ge T_2L$,
    \begin{align*}
    & \;\;\;\; \left| \mathbf{E}\left[ ( \langle \theta \rangle_t  - \theta^* )^\top (P+P^\top) \Big( A(X_t) \langle \theta \rangle_t +   B(X_t)^\top\pi_{t+1} - A \langle \theta \rangle_t - b \Big) \; | \; \mathcal{F}_{t-\tau(\alpha_{t})} \right] \right| \nonumber \\
    & \le \alpha_{t-\tau(\alpha_t)} \tau(\alpha_t)  \gamma_{\max} \left( 72 + 456 A_{\max}^2  + 84  A_{\max}  b_{\max}  \right) \mathbf{E}[ \| \langle \theta \rangle_{t} \|_2^2 \; | \; \mathcal{F}_{t-\tau(\alpha_t)} ] \nonumber \\
    &\;\;\; + \alpha_{t -\tau(\alpha_t) }  \tau(\alpha_t) \gamma_{\max} \bigg[ 2 + 4 \|\theta^* \|_2^2 +  \frac{48b_{\max}^2}{A_{\max}^2} +  152  \left(b_{\max} + A_{\max} \| \theta^* \|_2 \right)^2 +  12  A_{\max}b_{\max} \nonumber\\
    &\;\;\; + 48  A_{\max}b_{\max} \Big(\frac{b_{\max}}{A_{\max}} + 1\Big)^2 +  87 b_{\max}^2  \bigg] \nonumber \\
    & \;\;\;+  2 \gamma_{\max}  \eta_{t+1}\sqrt{N}b_{\max} \Big( 1 +  9 \mathbf{E}\left[ \| \langle \theta \rangle_{t}\|_2^2 \; | \; \mathcal{F}_{t-\tau(\alpha)} \right] +  \frac{6b_{\max}^2}{A_{\max}^2}+ \| \theta^* \|_2^2 \Big). 
\end{align*}
\end{lemma}

\noindent
{\bf Proof of Lemma~\ref{lemma:bound_timevarying_Ab}:}
Note that for all $t\ge T_2L$, we have
\begin{align}
    & \;\;\;\; |\mathbf{E}[ ( \langle \theta \rangle_t - \theta^* )^\top (P+P^\top)( A(X_t) \langle \theta \rangle_t +   B(X_t)^\top\pi_{t+1} - A \langle \theta \rangle_t - b) \; | \; \mathcal{F}_{t-\tau(\alpha_{t})} ]| \nonumber \\
    & \le |\mathbf{E}[ ( \langle \theta \rangle_t - \theta^* )^\top (P+P^\top)( A(X_t) - A) \langle \theta \rangle_t \; | \; \mathcal{F}_{t-\tau(\alpha_{t})}]| \nonumber\\
    &\;\;\; +   |\mathbf{E}[ ( \langle \theta \rangle_t - \theta^* )^\top (P+P^\top)(B(X_t)^\top\pi_{t+1} - b) \; | \; \mathcal{F}_{t-\tau(\alpha_{t})} ]| \nonumber \\
    & \le |\mathbf{E}[ ( \langle \theta \rangle_{t-\tau(\alpha_{t})} - \theta^* )^\top (P+P^\top)( A(X_t) - A) \langle \theta \rangle_{t-\tau(\alpha_{t})} \; | \; \mathcal{F}_{t-\tau(\alpha_{t})} ]| \label{eq:timevarying_bound_Ab_1} \\
    & \;\;\; + |\mathbf{E}[ ( \langle \theta \rangle_{t-\tau(\alpha_{t})} - \theta^* )^\top (P+P^\top)( A(X_t) - A) ( \langle \theta \rangle_t - \langle \theta \rangle_{t-\tau(\alpha_{t})} ) \; | \; \mathcal{F}_{t-\tau(\alpha_{t})} ]| \label{eq:timevarying_bound_Ab_2} \\
    & \;\;\; + |\mathbf{E}[  ( \langle \theta \rangle_t - \langle \theta \rangle_{t-\tau(\alpha_{t})} )^\top (P+P^\top)( A(X_t) - A) \langle \theta \rangle_{t-\tau(\alpha_{t})} \; | \; \mathcal{F}_{t-\tau(\alpha_{t})} ]| \label{eq:timevarying_bound_Ab_3}\\
    & \;\;\; + |\mathbf{E}[  ( \langle \theta \rangle_t - \langle \theta \rangle_{t-\tau(\alpha_{t})} )^\top (P+P^\top)( A(X_t) - A)  ( \langle \theta \rangle_t - \langle \theta \rangle_{t-\tau(\alpha_{t})} ) \; | \; \mathcal{F}_{t-\tau(\alpha_{t})} ]| \label{eq:timevarying_bound_Ab_4}\\
    &\;\;\; +   |\mathbf{E}[ ( \langle \theta \rangle_t - \langle \theta \rangle_{t-\tau(\alpha_{t})} )^\top (P+P^\top)(B(X_t)^\top\pi_{t+1} - b) \; | \; \mathcal{F}_{t-\tau(\alpha_{t})} ]|  \label{eq:timevarying_bound_Ab_5}\\
    &\;\;\; +   |\mathbf{E}[ ( \langle \theta \rangle_{t-\tau(\alpha_{t})} - \theta^* )^\top (P+P^\top)(B(X_t)^\top\pi_{t+1} - b) \; | \; \mathcal{F}_{t-\tau(\alpha_{t})} ]|\label{eq:timevarying_bound_Ab_6}.
\end{align}
Similar to the proof of Lemma~\ref{lemma:bound_fixed_Ab}, using the mixing time in Assumption~\ref{assum:mixing-time}, we can get the bound for \eqref{eq:timevarying_bound_Ab_1} and \eqref{eq:timevarying_bound_Ab_6} for  $t\ge T_2L$ as follows:
\begin{align}
    & \;\;\; |\mathbf{E}[ ( \langle \theta \rangle_{t-\tau(\alpha_{t})} - \theta^* )^\top (P+P^\top)( A(X_t) - A) \langle \theta \rangle_{t-\tau(\alpha_{t})} \; | \; \mathcal{F}_{t-\tau(\alpha_{t})} ]|\nonumber\\
    & \le |( \langle \theta \rangle_{t-\tau(\alpha_{t})} - \theta^* )^\top (P+P^\top) \mathbf{E}[A(X_t) - A \; | \; \mathcal{F}_{t-\tau(\alpha_{t})} ] \langle \theta \rangle_{t-\tau(\alpha_{t})} | \nonumber\\
    & \le 2 \alpha_{t} \gamma_{\max}   \mathbf{E}[\| \langle \theta \rangle_{t-\tau(\alpha_{t})}  - \theta^* \|_2  \| \langle \theta \rangle_{t-\tau(\alpha_{t})}\|_2 \; | \; \mathcal{F}_{t-\tau(\alpha_{t})} ] \nonumber\\
    & \le \alpha_{t} \gamma_{\max}   \mathbf{E}[\| \langle \theta \rangle_{t-\tau(\alpha_{t})}   - \theta^* \|_2^2 +  \| \langle \theta \rangle_{t-\tau(\alpha_{t})} \|_2^2 \; | \; \mathcal{F}_{t-\tau(\alpha_{t})} ] \nonumber \\
    & \le \alpha_{t} \gamma_{\max}   \mathbf{E}[ 2 \|\theta^* \|_2^2 + 3 \| \langle \theta \rangle_{t-\tau(\alpha_{t})}\|_2^2 \; | \; \mathcal{F}_{t-\tau(\alpha_{t})} ] \nonumber \\
    & \le 6 \alpha_{t} \gamma_{\max} \mathbf{E}[ \| \langle \theta \rangle_{t} - \langle \theta \rangle_{t-\tau(\alpha_{t})}\|_2^2 \; | \; \mathcal{F}_{t-\tau(\alpha_{t})} ] + 6 \alpha_{t} \gamma_{\max} \mathbf{E}[ \| \langle \theta \rangle_{t} \|_2^2 \; | \; \mathcal{F}_{t-\tau(\alpha_{t})} ] + 2 \alpha_{t} \gamma_{\max} \|\theta^* \|_2^2 \nonumber \\
    & \le 54 \alpha_{t} \gamma_{\max} \mathbf{E}[ \| \langle \theta \rangle_{t} \|_2^2 \; | \; \mathcal{F}_{t-\tau(\alpha_{t})} ] + 36 \alpha_{t} \gamma_{\max} (\frac{b_{\max}}{A_{\max}})^2 + 2 \alpha_{t} \gamma_{\max} \|\theta^* \|_2^2,  \label{eq:timevarying_bound_Ab_1_bounded}
\end{align}
where in the last inequality, we use \eqref{eq:timevarying_single_3_3} from Lemma~\ref{lemma:timevarying_single_3}, and
\begin{align}
    & \;\;\; |\mathbf{E}[ ( \langle \theta \rangle_{t-\tau(\alpha_{t})}  - \theta^* )^\top (P+P^\top)(B(X_t)^\top\pi_{t+1} - b) \; | \; \mathcal{F}_{t-\tau(\alpha_{t})} ]|\nonumber\\
    & \le |\mathbf{E}[ ( \langle \theta \rangle_{t-\tau(\alpha_{t})}  - \theta^* )^\top (P+P^\top)\Big(\sum_{i=1}^N \pi_{t+1}^i(b^i(X_t) - b^i ) +  \sum_{i=1}^N (\pi_{t+1}^i - \pi_{\infty}^i) b^i \Big) \; | \; \mathcal{F}_{t-\tau(\alpha_{t})} ]|\nonumber
\end{align}
\begin{align}    
    & \le | ( \langle \theta \rangle_{t-\tau(\alpha_{t})}  - \theta^* )^\top (P+P^\top)\Big(\sum_{i=1}^N \pi_{t+1}^i \mathbf{E}[ b^i(X_t) - b^i \; | \; \mathcal{F}_{t-\tau(\alpha_{t})} ] +  \sum_{i=1}^N (\pi_{t+1}^i - \pi_{\infty}^i) b^i \Big) |\nonumber\\
    & \le 2 \gamma_{\max} (\alpha_{t} + \eta_{t+1}\sqrt{N}b_{\max}) \mathbf{E}[ \| \langle \theta \rangle_{t-\tau(\alpha_{t})} - \theta^* \|_2 \; | \; \mathcal{F}_{t-\tau(\alpha_{t})} ]\nonumber\\
    & \le 2 \gamma_{\max} (\alpha_{t} + \eta_{t+1}\sqrt{N}b_{\max}) \Big( 1 + \frac{1}{2} \mathbf{E}[ \| \langle \theta \rangle_{t-\tau(\alpha_{t})} \|_2^2 \; | \; \mathcal{F}_{t-\tau(\alpha_{t})} ] + \frac{1}{2} \| \theta^* \|_2^2 \Big) \nonumber\\
    & \le 2 \gamma_{\max} (\alpha_{t} + \eta_{t+1}\sqrt{N}b_{\max}) \left( 1 +  \mathbf{E}[ \| \langle \theta \rangle_{t-\tau(\alpha_{t})} - \langle \theta \rangle_{t} \|_2^2 +  \| \langle \theta \rangle_{t} \|_2^2 \; | \; \mathcal{F}_{t-\tau(\alpha_{t})} ] + \| \theta^* \|_2^2 \right) \nonumber\\
    & \le 2 \gamma_{\max} (\alpha_{t} + \eta_{t+1}\sqrt{N}b_{\max}) \Big( 1 +  9 \mathbf{E}[ \| \langle \theta \rangle_{t} \|_2^2 \; | \; \mathcal{F}_{t-\tau(\alpha_{t})} ] + 6 (\frac{b_{\max}}{A_{\max}})^2+ \| \theta^* \|_2^2 \Big), 
    \label{eq:timevarying_bound_Ab_6_bounded}
\end{align}  
where in the last inequality we use \eqref{eq:timevarying_single_3_3}.
Next, by Assumption~\ref{assum:A and b}, \eqref{eq:timevarying_single_3_1} and \eqref{eq:timevarying_single_3_3}, we have
\begin{align}
    & \;\;\; |\mathbf{E}[ ( \langle \theta \rangle_{t-\tau(\alpha_{t})} - \theta^* )^\top (P+P^\top)( A(X_t) - A) ( \langle \theta \rangle_t - \langle \theta \rangle_{t-\tau(\alpha_{t})} ) \; | \; \mathcal{F}_{t-\tau(\alpha_{t})} ]| \nonumber\\
    &\le 4 \gamma_{\max} A_{\max} \mathbf{E}[ \| \langle \theta \rangle_{t-\tau(\alpha_{t})} - \theta^* \|_2 \| \langle \theta \rangle_t - \langle \theta \rangle_{t-\tau(\alpha_{t})}\|_2 \; | \; \mathcal{F}_{t-\tau(\alpha_{t})} ] \nonumber\\
    &\le 4 \gamma_{\max} A_{\max} 
    \mathbf{E}[ \| \langle \theta \rangle_{t-\tau(\alpha_{t})} \|_2 \| \langle \theta \rangle_t - \langle \theta \rangle_{t-\tau(\alpha_{t})}\|_2 + \| \theta^* \|_2 \| \langle \theta \rangle_t - \langle \theta \rangle_{t-\tau(\alpha_{t})}\|_2
    \; | \; \mathcal{F}_{t-\tau(\alpha_{t})} ] \nonumber\\
    %&\;\;\; + 4 \gamma_{\max} A_{\max} \| \theta^* \|_2 \mathbf{E}[  \| \langle \theta \rangle_t - \langle \theta \rangle_{t-\tau(\alpha_{t})}\|_2 \; | \; \mathcal{F}_{t-\tau(\alpha_{t})} ] \nonumber\\
    &\le 8 \gamma_{\max} A_{\max}^2 \mathbf{E}[ \| \langle \theta \rangle_{t-\tau(\alpha_{t})} \|_2^2  \; | \; \mathcal{F}_{t-\tau(\alpha_{t})} ]  \sum_{k=t-\tau(\alpha_t)}^{t-1} \alpha_k + 8 \gamma_{\max} A_{\max} b_{\max} \| \theta^* \|_2 \sum_{k=t-\tau(\alpha_t)}^{t-1} \alpha_k \nonumber\\
    & \;\;\; + 8  \gamma_{\max} A_{\max}^2 \Big(\frac{ b_{\max}}{A_{\max}} + \| \theta^* \|_2 \Big) \mathbf{E}[ \| \langle \theta \rangle_{t-\tau(\alpha_{t})} \|_2  \; | \; \mathcal{F}_{t-\tau(\alpha_{t})} ] \sum_{k=t-\tau(\alpha_t)}^{t-1} \alpha_k \nonumber\\
    &\le  \gamma_{\max} A_{\max}^2 \Big( 12 \mathbf{E}[ \| \langle \theta \rangle_{t-\tau(\alpha_{t})} \|_2^2  \; | \; \mathcal{F}_{t-\tau(\alpha_{t})} ] + 8  \Big(\frac{ b_{\max}}{A_{\max}} + \| \theta^* \|_2 \Big)^2 \Big) \sum_{k=t-\tau(\alpha_t)}^{t-1} \alpha_k ,\nonumber%\\
    %%& \;\;\; + 8  \gamma_{\max} A_{\max}^2 \left(\frac{ b_{\max}}{A_{\max}} + \| \theta^* \|_2 \right)^2  \sum_{k=t-\tau(\alpha_t)}^{t-1} \alpha_k \nonumber\\
    %&\le 24 \gamma_{\max} A_{\max}^2 \mathbf{E}[ \| \langle \theta \rangle_{t} - \langle \theta \rangle_{t-\tau(\alpha_{t})} \|_2^2  \; | \; \mathcal{F}_{t-\tau(\alpha_{t})} ] \sum_{k=t-\tau(\alpha_t)}^{t-1} \alpha_k \nonumber\\
    %& \;\;\; + 24 \gamma_{\max} A_{\max}^2 \mathbf{E}[ \| \langle \theta \rangle_{t}\|_2^2  \; | \; \mathcal{F}_{t-\tau(\alpha_{t})} ] \sum_{k=t-\tau(\alpha_t)}^{t-1} \alpha_k  + 8  \gamma_{\max} A_{\max}^2 \Big(\frac{ b_{\max}}{A_{\max}} + \| \theta^* \|_2 \Big)^2  \sum_{k=t-\tau(\alpha_t)}^{t-1} \alpha_k \nonumber\\
    %&\le  \gamma_{\max} \left( 216 A_{\max}^2 \mathbf{E}[ \| \langle \theta \rangle_{t}\|_2^2  \; | \; \mathcal{F}_{t-\tau(\alpha_{t})} ] + 152  \left(b_{\max} + A_{\max} \| \theta^* \|_2 \right)^2 \right) \sum_{k=t-\tau(\alpha_t)}^{t-1} \alpha_k.
    %%&+ 152 \gamma_{\max}  \left(b_{\max} + A_{\max} \| \theta^* \|_2 \right)^2 \sum_{k=t-\tau(\alpha_t)}^{t-1} \alpha_k.
    \label{eq:timevarying_bound_Ab_2_bounded} 
\end{align}
which implies that
\begin{align}
    & \;\;\; |\mathbf{E}[ ( \langle \theta \rangle_{t-\tau(\alpha_{t})} - \theta^* )^\top (P+P^\top)( A(X_t) - A) ( \langle \theta \rangle_t - \langle \theta \rangle_{t-\tau(\alpha_{t})} ) \; | \; \mathcal{F}_{t-\tau(\alpha_{t})} ]| \nonumber\\
    &\le 24 \gamma_{\max} A_{\max}^2 \mathbf{E}[ \| \langle \theta \rangle_{t} - \langle \theta \rangle_{t-\tau(\alpha_{t})} \|_2^2  \; | \; \mathcal{F}_{t-\tau(\alpha_{t})} ] \sum_{k=t-\tau(\alpha_t)}^{t-1} \alpha_k \nonumber\\
    & \;\;\; + 24 \gamma_{\max} A_{\max}^2 \mathbf{E}[ \| \langle \theta \rangle_{t}\|_2^2  \; | \; \mathcal{F}_{t-\tau(\alpha_{t})} ] \sum_{k=t-\tau(\alpha_t)}^{t-1} \alpha_k  + 8  \gamma_{\max} A_{\max}^2 \Big(\frac{ b_{\max}}{A_{\max}} + \| \theta^* \|_2 \Big)^2  \sum_{k=t-\tau(\alpha_t)}^{t-1} \alpha_k \nonumber\\
    &\le  \gamma_{\max} \left( 216 A_{\max}^2 \mathbf{E}[ \| \langle \theta \rangle_{t}\|_2^2  \; | \; \mathcal{F}_{t-\tau(\alpha_{t})} ] + 152  \left(b_{\max} + A_{\max} \| \theta^* \|_2 \right)^2 \right) \sum_{k=t-\tau(\alpha_t)}^{t-1} \alpha_k.
    %&+ 152 \gamma_{\max}  \left(b_{\max} + A_{\max} \| \theta^* \|_2 \right)^2 \sum_{k=t-\tau(\alpha_t)}^{t-1} \alpha_k.
    %\label{eq:timevarying_bound_Ab_2_bounded} 
\end{align}
In additional, using \eqref{eq:timevarying_single_3_1} and \eqref{eq:timevarying_single_3_3}, we have the following bound for \eqref{eq:timevarying_bound_Ab_3}: 
\begin{align}
    & \;\;\; |\mathbf{E}[ ( \langle \theta \rangle_t -\langle \theta \rangle_{t-\tau(\alpha_{t})} )^\top (P+P^\top)( A(X_t) - A) \langle \theta \rangle_{t-\tau(\alpha_{t})} \; | \; \mathcal{F}_{t-\tau(\alpha_{t})} ]|\nonumber\\
    & \le 4 \gamma_{\max} A_{\max} \mathbf{E}[ \| \langle \theta \rangle_t - \langle \theta \rangle_{t-\tau(\alpha_{t})}\|_2 \| \langle \theta \rangle_{t-\tau(\alpha_{t})}\|_2 \; | \; \mathcal{F}_{t-\tau(\alpha_{t})} ]\nonumber\\
    & \le 8 \gamma_{\max} A_{\max} \mathbf{E}[ A_{\max } \|  \langle \theta \rangle_{t-\tau(\alpha_{t})} \|_2^2 +  b_{\max}  \| \langle \theta \rangle_{t-\tau(\alpha_{t})}\|_2 \; | \; \mathcal{F}_{t-\tau(\alpha_{t})} ] \sum_{k=t-\tau(\alpha_t)}^{t-1} \alpha_k \nonumber\\
    & \le 4  \gamma_{\max} A_{\max} \left( (2 A_{\max }+  b_{\max} ) \mathbf{E}[  \| \langle \theta \rangle_{t-\tau(\alpha_{t})} \|_2^2 \; | \; \mathcal{F}_{t-\tau(\alpha_{t})} ] +  b_{\max} \right) \sum_{k=t-\tau(\alpha_t)}^{t-1} \alpha_k \nonumber
    %& \;\;\; + 4 \gamma_{\max} A_{\max} b_{\max} \sum_{k=t-\tau(\alpha_t)}^{t-1} \alpha_k\nonumber\\
\end{align}
\begin{align}
    & \le 8  \gamma_{\max} A_{\max} (2 A_{\max }+  b_{\max} ) \mathbf{E}[  \| \langle \theta \rangle_{t} -\langle \theta \rangle_{t-\tau(\alpha_{t})} \|_2^2 \; | \; \mathcal{F}_{t-\tau(\alpha_{t})} ] \sum_{k=t-\tau(\alpha_t)}^{t-1} \alpha_k  \nonumber\\
    &\;\;\; + 8  \gamma_{\max} A_{\max} (2 A_{\max }+  b_{\max} ) \mathbf{E}[  \| \langle \theta \rangle_{t} \|_2^2 \; | \; \mathcal{F}_{t-\tau(\alpha_{t})} ] \sum_{k=t-\tau(\alpha_t)}^{t-1} \alpha_k  + 4  \gamma_{\max} A_{\max} b_{\max} \sum_{k=t-\tau(\alpha_t)}^{t-1} \alpha_k \nonumber \\
    & \le 72 \ \gamma_{\max} A_{\max} (2 A_{\max }  +  b_{\max} ) \mathbf{E}[  \| \langle \theta \rangle_{t} \|_2^2 \; | \; \mathcal{F}_{t-\tau(\alpha_{t})} ]  \sum_{k=t-\tau(\alpha_t)}^{t-1} \alpha_k \nonumber\\
    & \;\;\;+ 48  \gamma_{\max} A_{\max} b_{\max} \Big(\frac{b_{\max}}{A_{\max}} + 1 \Big)^2 \sum_{k=t-\tau(\alpha_t)}^{t-1} \alpha_k.
    \label{eq:timevarying_bound_Ab_3_bounded}
\end{align}
Moreover, using \eqref{eq:timevarying_single_3_3}, we can get the bound for \eqref{eq:timevarying_bound_Ab_4} as follows:
\begin{align}
    & \;\;\; |\mathbf{E}[  ( \langle \theta \rangle_t - \langle \theta \rangle_{t-\tau(\alpha_{t})} )^\top (P+P^\top)( A(X_t) - A)  ( \langle \theta \rangle_t - \langle \theta \rangle_{t-\tau(\alpha_{t})} ) \; | \; \mathcal{F}_{t-\tau(\alpha_{t})} ]| \nonumber\\
    & \le 4 \gamma_{\max} A_{\max}  \mathbf{E}[ \| \langle \theta \rangle_t - \langle \theta \rangle_{t-\tau(\alpha_{t})}\|_2^2 \; | \; \mathcal{F}_{t-\tau(\alpha_{t})} ]| \nonumber\\
    & \le 4 \gamma_{\max} A_{\max}  \mathbf{E}[ 72 A_{\max}^2 \| \langle \theta \rangle_{t} \|_2^2 +  50 b_{\max}^2 \; | \; \mathcal{F}_{t-\tau(\alpha_{t})} ] \Big( \sum_{k=t-\tau(\alpha_t)}^{t-1} \alpha_k \Big)^2 \nonumber\\
    & \le   96 A_{\max}^2 \gamma_{\max} \mathbf{E}[ \| \langle \theta \rangle_{t} \|_2^2 \; | \; \mathcal{F}_{t-\tau(\alpha_{t})} ] \sum_{k=t-\tau(\alpha_t)}^{t-1} \alpha_k + 67 b_{\max}^2 \gamma_{\max} \sum_{k=t-\tau(\alpha_t)}^{t-1} \alpha_k.
    \label{timevarying_bound_Ab_4_bounded}
\end{align}
Finally, we can get the bound for \eqref{eq:timevarying_bound_Ab_5} using \eqref{eq:timevarying_single_3_2}: 
\begin{align}
    &\;\;\; |\mathbf{E}[ ( \langle \theta \rangle_t - \langle \theta \rangle_{t-\tau(\alpha_{t})} ) (P+P^\top)(B(X_t)^\top\pi_{t+1} - b) \; | \; \mathcal{F}_{t-\tau(\alpha_{t})} ]| \nonumber \\
    & \le 4  \gamma_{\max} b_{\max} \mathbf{E}[ \| \langle \theta \rangle_t - \langle \theta \rangle_{t-\tau(\alpha_{t})}\|_2 \; | \; \mathcal{F}_{t-\tau(\alpha_{t})} ] \nonumber \\
    & \le 4  \gamma_{\max} b_{\max} \mathbf{E}[ 6 A_{\max} \| \langle \theta \rangle_{t} \|_2 +  5 b_{\max} \; | \; \mathcal{F}_{t-\tau(\alpha_{t})} ] \sum_{k=t-\tau(\alpha_t)}^{t-1} \alpha_k \nonumber \\
    & \le \gamma_{\max}\left( 12  A_{\max}  b_{\max}   \mathbf{E}[\| \langle \theta \rangle_{t} \|_2^2  \; | \; \mathcal{F}_{t-\tau(\alpha_{t})} ] +   12 A_{\max}  b_{\max} + 20 b_{\max}^2  \right)\sum_{k=t-\tau(\alpha_t)}^{t-1} \alpha_k. %\nonumber\\
    %& \;\;\; +   ( 12 A_{\max}  b_{\max} + 20 b_{\max}^2 )\gamma_{\max} \sum_{k=t-\tau(\alpha_t)}^{t-1} \alpha_k.  
    \label{eq:timevarying_bound_Ab_5_bounded}
\end{align}
Then, using \eqref{eq:timevarying_bound_Ab_1_bounded}--\eqref{eq:timevarying_bound_Ab_5_bounded}, we have
\begin{align*}
    & \;\;\;\; |\mathbf{E}[ ( \langle \theta \rangle_t  - \theta^* )^\top (P+P^\top)( A(X_t) \langle \theta \rangle_t +   B(X_t)^\top\pi_{t+1} - A \langle \theta \rangle_t - b) \; | \; \mathcal{F}_{t-\tau(\alpha_{t})} ]| \nonumber \\
    %& \le 54 \alpha_{t} \gamma_{\max} \mathbf{E}[ \| \langle \theta \rangle_{t} \|_2^2 \; | \; \mathcal{F}_{t-\tau(\alpha_{t})} ] + 36 \alpha_{t} \gamma_{\max} (\frac{b_{\max}}{A_{\max}})^2 + 2 \alpha_{t} \gamma_{\max} \|\theta^* \|_2^2  \nonumber \\
    %&\;\;\; + 216 \gamma_{\max} A_{\max}^2 \mathbf{E}[ \| \langle \theta \rangle_{t}\|_2^2  \; | \; \mathcal{F}_{t-\tau(\alpha_{t})} ] \sum_{k=t-\tau(\alpha_t)}^{t-1} \alpha_k  \\
    %& \;\;\; + 152 \gamma_{\max}  \left(b_{\max} + A_{\max} \| \theta^* \|_2 \right)^2 \sum_{k=t-\tau(\alpha_t)}^{t-1} \alpha_k + 67 b_{\max}^2 \gamma_{\max} \sum_{k=t-\tau(\alpha_t)}^{t-1} \alpha_k \nonumber \\
    %& \;\;\; + 12 \ \gamma_{\max} A_{\max} (20 A_{\max }  +  7b_{\max} ) \mathbf{E}[  \| \langle \theta \rangle_{t} \|_2^2 \; | \; \mathcal{F}_{t-\tau(\alpha_{t})} ]  \sum_{k=t-\tau(\alpha_t)}^{t-1} \alpha_k \nonumber\\
    %& \;\;\;+ 48  \gamma_{\max} A_{\max} b_{\max} (\frac{b_{\max}}{A_{\max}} + 1 )^2 \sum_{k=t-\tau(\alpha_t)}^{t-1} \alpha_k +   ( 12 A_{\max}  b_{\max} + 20 b_{\max}^2 )\gamma_{\max} \sum_{k=t-\tau(\alpha_t)}^{t-1} \alpha_k \nonumber \\
    %& \;\;\; + 2 \gamma_{\max} (\alpha_{t} + \eta_{t+1}\sqrt{N}b_{\max}) \left( 1 +  9 \mathbf{E}[ \| \langle \theta \rangle_{t} \|_2^2 \; | \; \mathcal{F}_{t-\tau(\alpha_{t})} ] + 6 (\frac{b_{\max}}{A_{\max}})^2+ \| \theta^* \|_2^2 \right) \nonumber \\
    & \le 54 \alpha_{t} \gamma_{\max} \mathbf{E}[ \| \langle \theta \rangle_{t} \|_2^2 \; | \; \mathcal{F}_{t-\tau(\alpha_{t})} ] + 36 \alpha_{t} \gamma_{\max} (\frac{b_{\max}}{A_{\max}})^2 + 2 \alpha_{t} \gamma_{\max} \|\theta^* \|_2^2  \nonumber \\
    &\;\;\; + \sum_{k=t-\tau(\alpha_t)}^{t-1} \alpha_k \Big[ \left( 216 \gamma_{\max} A_{\max}^2 + 12  \gamma_{\max} A_{\max} (20 A_{\max }  +  7b_{\max} ) \right) \mathbf{E}[ \| \langle \theta \rangle_{t}\|_2^2  \; | \; \mathcal{F}_{t-\tau(\alpha_{t})} ]   + 67 b_{\max}^2 \gamma_{\max} \\
    & \;\;\;    + 152 \gamma_{\max}  \left(b_{\max} + A_{\max} \| \theta^* \|_2 \right)^2
    + 48  \gamma_{\max} A_{\max} b_{\max} (\frac{b_{\max}}{A_{\max}} + 1 )^2 +   ( 12 A_{\max}  b_{\max} + 20 b_{\max}^2 )\gamma_{\max} \Big] \\
    & \;\;\;   + 2 \gamma_{\max} (\alpha_{t} + \eta_{t+1}\sqrt{N}b_{\max}) \left( 1 +  9 \mathbf{E}[ \| \langle \theta \rangle_{t} \|_2^2 \; | \; \mathcal{F}_{t-\tau(\alpha_{t})} ] + 6 (\frac{b_{\max}}{A_{\max}})^2+ \| \theta^* \|_2^2 \right), %\nonumber \\
    %& \le \alpha_{t-\tau(\alpha_t)} \tau(\alpha_t)  \gamma_{\max} \left( 72 + 456 A_{\max}^2  + 84  A_{\max}  b_{\max}  \right) \mathbf{E}[ \| \langle \theta \rangle_{t} \|_2^2 \; | \; \mathcal{F}_{t-\tau(\alpha_t)} ] \nonumber \\
    %&\;\;\; + \alpha_{t -\tau(\alpha_t) }  \tau(\alpha_t) \gamma_{\max} \bigg[ 2 + 4 \|\theta^* \|_2^2 +  48(\frac{b_{\max}}{A_{\max}})^2 +  152  \left(b_{\max} + A_{\max} \| \theta^* \|_2 \right)^2 +  12  A_{\max}b_{\max} \nonumber\\
    %&\;\;\; + 48  A_{\max}b_{\max} (\frac{b_{\max}}{A_{\max}} + 1 )^2 +  87 b_{\max}^2  \bigg] \nonumber \\
    %& \;\;\;+  2 \gamma_{\max}  \eta_{t+1}\sqrt{N}b_{\max} \left( 1 +  9 \mathbf{E}[ \| \langle \theta \rangle_{t}\|_2^2 \; | \; \mathcal{F}_{t-\tau(\alpha)} ] + 6 (\frac{b_{\max}}{A_{\max}})^2+ \| \theta^* \|_2^2 \right), 
\end{align*}
which implies that
\begin{align*}
    & \;\;\;\; |\mathbf{E}[ ( \langle \theta \rangle_t  - \theta^* )^\top (P+P^\top)( A(X_t) \langle \theta \rangle_t +   B(X_t)^\top\pi_{t+1} - A \langle \theta \rangle_t - b) \; | \; \mathcal{F}_{t-\tau(\alpha_{t})} ]| \nonumber \\
    & \le \alpha_{t-\tau(\alpha_t)} \tau(\alpha_t)  \gamma_{\max} \left( 72 + 456 A_{\max}^2  + 84  A_{\max}  b_{\max}  \right) \mathbf{E}[ \| \langle \theta \rangle_{t} \|_2^2 \; | \; \mathcal{F}_{t-\tau(\alpha_t)} ] \nonumber \\
    &\;\;\; + \alpha_{t -\tau(\alpha_t) }  \tau(\alpha_t) \gamma_{\max} \bigg[ 2 + 4 \|\theta^* \|_2^2 +  48(\frac{b_{\max}}{A_{\max}})^2 +  152  \left(b_{\max} + A_{\max} \| \theta^* \|_2 \right)^2 +  12  A_{\max}b_{\max} \nonumber\\
    &\;\;\; + 48  A_{\max}b_{\max} (\frac{b_{\max}}{A_{\max}} + 1 )^2 +  87 b_{\max}^2  \bigg] \nonumber \\
    & \;\;\;+  2 \gamma_{\max}  \eta_{t+1}\sqrt{N}b_{\max} \left( 1 +  9 \mathbf{E}[ \| \langle \theta \rangle_{t}\|_2^2 \; | \; \mathcal{F}_{t-\tau(\alpha)} ] + 6 (\frac{b_{\max}}{A_{\max}})^2+ \| \theta^* \|_2^2 \right), 
\end{align*}
where we use $\alpha_t \le \alpha_{t-\tau(\alpha_t)}$ from Assumption~\ref{assum:step-size} and $\tau(\alpha_t) \ge 1$ in the last inequality.
This completes the proof.
\hfill $\qed$

\begin{lemma} \label{lemma:bound_average_time-varying_jointly}
    Under Assumptions~\ref{assum:weighted matrix}--\ref{assum:limit_pi}, when the $\tau(\alpha_t) \alpha_{t-\tau(\alpha_t)} \le \min \{ \frac{ \log2}{A_{\max}}, \; \frac{0.1}{\zeta_5 \gamma_{\max}} \},$ we have for any $t\ge T_2L$,
%    \begin{align*}
%        \mathbf{E}[\| \langle \theta \rangle_{t+1} -\theta^* \|_2^2 ] 
%        & \le \frac{T_2L}{t+1} \frac{\gamma_{\max}}{\gamma_{\min}} \mathbf{E}[\| \langle \theta \rangle_{T_2L} -\theta^* \|_2^2 ] 
%        +  \frac{\zeta_7 \alpha_0  C \log^2(\frac{t+1}{\alpha_0})}{t+1} \frac{\gamma_{\max}}{\gamma_{\min}}
%     \nonumber\\
%        & \;\;\; +  \alpha_0 \zeta_4 \frac{\gamma_{\max}}{\gamma_{\min}}  \frac{\sum_{l = T_2L}^{t} \eta_{l+1}}{t+1}.
%    \end{align*}
\begin{align*}
    &\;\;\;\; \mathbf{E} \left[\|\langle \theta \rangle_{t} -\theta^* \|_2^2 \right] \le \frac{T_2L}{t} \frac{\gamma_{\max}}{\gamma_{\min}} \mathbf{E}[\| \langle \theta \rangle_{T_2L} -\theta^* \|_2^2 ] 
    +  \frac{\zeta_7 \alpha_0  C \log^2(\frac{t}{\alpha_0})}{t} \frac{\gamma_{\max}}{\gamma_{\min}}  +  \alpha_0 \zeta_4 \frac{\gamma_{\max}}{\gamma_{\min}}  \frac{\sum_{l = T_2L}^{t} \eta_{l}}{t},
\end{align*}
where $T_2$ is defined in Appendix~\ref{sec:thmSA_constant}, and $\zeta_4$, $\zeta_5$, $\zeta_7$ are defined in \eqref{eq:define Psi5}, \eqref{eq:define Psi7}, \eqref{eq:define Psi8}, respectively.
\end{lemma}

\noindent
{\bf Proof of Lemma~\ref{lemma:bound_average_time-varying_jointly}:} 
%Recall the update of
Consider the update of
$\langle \theta \rangle_t $ given in \eqref{eq:update of average_time-varying}.
%\begin{align*} 
%    \langle \theta \rangle_{t+1} &= \langle \theta \rangle_t + \alpha_t A(X_t) \langle \theta \rangle_t + \alpha_t  B(X_t)^\top\pi_{t+1}.
%\end{align*}
Note that $ \mathbf{E}[ \| \langle \theta \rangle_{t} \|_2^2 ] \le 2 \mathbf{E}[ \| \langle \theta \rangle_{t}- \theta^*  \|_2^2 ] + 2 \| \theta^*\|_2^ 2 \le \frac{2}{\gamma_{\min}} \mathbf{E}[H( \langle \theta \rangle_t )] + 2 \| \theta^*\|_2^ 2 $, then from \eqref{eq:fixed_proof_1} and Lemma~\ref{lemma:bound_timevarying_Ab}, for $t\ge T_2L$ we have  
\begin{align}
    &\;\;\;\; \mathbf{E}[ H( \langle \theta \rangle_{t+1} ) ] \nonumber\\
    & \le \mathbf{E}[H( \langle \theta \rangle_t )] - \alpha_t \mathbf{E}[\| \langle \theta \rangle_t - \theta^*\|_2^2 ] + \alpha_t^2 A_{\max}^2 \gamma_{\max} \mathbf{E}[\| \langle \theta \rangle_t\|_2^2 ]   +  2 \alpha_t^2 A_{\max} b_{\max} \gamma_{\max} \mathbf{E}[\| \langle \theta \rangle_t\|_2 ] \nonumber\\
    & \;\;\;\; + \alpha_t  \mathbf{E}[( \langle \theta \rangle_t - \theta^* )^\top (P+P^\top)( A(X_t) \langle \theta \rangle_t +   B(X_t)^\top\pi_{t+1} - A \langle \theta \rangle_t - b) ] + \alpha_t^2 b_{\max}^2 \gamma_{\max}\nonumber\\
    & \le \mathbf{E}[H( \langle \theta \rangle_t )] - \alpha_t \mathbf{E}[\| \langle \theta \rangle_t - \theta^*\|_2^2 ] + 2 \alpha_t^2 A_{\max}^2 \gamma_{\max} \mathbf{E}[\| \langle \theta \rangle_t\|_2^2 ] + 2 \alpha_t^2 b_{\max}^2 \gamma_{\max}  \nonumber\\
    & \;\;\; + \alpha_t  \alpha_{t-\tau(\alpha_t)} \tau(\alpha_t)  \gamma_{\max} \left( 72 + 456 A_{\max}^2  + 84  A_{\max}  b_{\max}  \right) \mathbf{E}[ \| \langle \theta \rangle_{t} \|_2^2 ] \nonumber \\
    &\;\;\; + \alpha_{t} \alpha_{t -\tau(\alpha_t) }  \tau(\alpha_t) \gamma_{\max} \bigg[   152  \left(b_{\max} + A_{\max} \| \theta^* \|_2 \right)^2 + 48  A_{\max}b_{\max} (\frac{b_{\max}}{A_{\max}} + 1 )^2 + 12  A_{\max}b_{\max}  + 2 \nonumber\\
    &\;\;\;    + 4 \|\theta^* \|_2^2   + 48(\frac{b_{\max}}{A_{\max}})^2 + 87 b_{\max}^2   \bigg] +  2 \alpha_t \gamma_{\max}  \eta_{t+1}\sqrt{N}b_{\max} \left( 1 +  9 \mathbf{E}[ \| \langle \theta \rangle_{t}\|_2^2  ] + 6 (\frac{b_{\max}}{A_{\max}})^2+ \| \theta^* \|_2^2 \right) \nonumber\\
    & \le \mathbf{E}[H( \langle \theta \rangle_t )] +  \left( - \alpha_t + 2 \alpha_t  \alpha_{t-\tau(\alpha_t)} \tau(\alpha_t) \gamma_{\max} \left( 72 + 458 A_{\max}^2  + 84  A_{\max}  b_{\max}  \right)  \right) \mathbf{E}[\| \langle \theta \rangle_t - \theta^*\|_2^2 ] \nonumber\\
    & \;\;\; +  \alpha_t  \alpha_{t-\tau(\alpha_t)} \tau(\alpha_t)  \gamma_{\max} \Big[ 2\left( 72 + 458 A_{\max}^2  + 84  A_{\max}  b_{\max}  \right) \| \theta^*\|_2^ 2       +  152  \left(b_{\max} + A_{\max} \| \theta^* \|_2 \right)^2 \nonumber\\
    &\;\;\;  + 2 + 4 \|\theta^* \|_2^2 +  12  A_{\max}b_{\max}  + 48(\frac{b_{\max}}{A_{\max}})^2   + 48  A_{\max}b_{\max} (\frac{b_{\max}}{A_{\max}} + 1 )^2 +  89 b_{\max}^2\bigg]  \nonumber\\
    &\;\;\;     +  2 \alpha_t \gamma_{\max}  \eta_{t+1}\sqrt{N}b_{\max} \left( 1 +  18 \mathbf{E}[\| \langle \theta \rangle_t - \theta^*\|_2^2 ] + 6 (\frac{b_{\max}}{A_{\max}})^2+ 19 \| \theta^* \|_2^2 \right) \nonumber \\
    & \le \mathbf{E}[H( \langle \theta \rangle_t )] +  \left( - \alpha_t +  \alpha_t  \alpha_{t-\tau(\alpha_t)} \tau(\alpha_t) \gamma_{\max} \zeta_5 + 36 \alpha_t \gamma_{\max}  \eta_{t+1}\sqrt{N}b_{\max} \right) \mathbf{E}[\| \langle \theta \rangle_t - \theta^*\|_2^2 ]  \nonumber\\
    & \;\;\; + \alpha_t  \alpha_{t-\tau(\alpha_t)} \tau(\alpha_t)  \gamma_{\max} \zeta_7 \nonumber +   
    \alpha_t \gamma_{\max} \eta_{t+1}\zeta_4, 
\end{align}
where $\zeta_4$, $\zeta_5$ and $\zeta_7$ are defined in \eqref{eq:define Psi5}, \eqref{eq:define Psi7} and \eqref{eq:define Psi8}, respectively.
Moreover, from $\alpha_t = \frac{\alpha_0}{t+1}$, $\alpha_0\ge \frac{\gamma_{max}}{0.9}$ and the definition of $T_2$, we have for all $t \ge T_2L$
\begin{align*}
    \mathbf{E}[H( \langle \theta \rangle_{t+1} )] &\le \left(1 - \frac{0.9  \alpha_t}{\gamma_{\max}} \right) \mathbf{E}[H( \langle \theta \rangle_t )] +  \alpha_t \gamma_{\max} \eta_{t+1}\zeta_4 + \alpha_t  \alpha_{t-\tau(\alpha_t)} \tau(\alpha_t)  \gamma_{\max}  \zeta_7  \nonumber \\
    & \le \frac{t}{t+1}  \mathbf{E}[H( \langle \theta \rangle_t )] +  \alpha_0 \gamma_{\max}  \zeta_4 \frac{\eta_{t+1}}{t+1}  +  \frac{\alpha_0^2 C\log(\frac{t+1}{\alpha_0})  \gamma_{\max}  \zeta_7  }{(t+1)(t-\tau(\alpha_t)+1)} \\
    & \le \frac{T_2L}{t+1}  \mathbf{E}[H( \langle \theta \rangle_{T_2L} )] +  \alpha_0 \gamma_{\max}  \zeta_4 \sum_{l = T_2L}^{t} \frac{\eta_{l+1}}{l+1} \Pi_{u=l+1}^t\frac{u}{u+1} \nonumber\\
    & \;\;\; +  \alpha_0^2  \gamma_{\max}  \zeta_7 
    \sum_{l=T_2L}^{t} \frac{ C\log(\frac{l+1}{\alpha_0})  }{(l+1)(l-\tau(\alpha_l)+1)}  \Pi_{u=l+1}^t\frac{u}{u+1} \nonumber \\
    & \le \frac{T_2L}{t+1}  \mathbf{E}[H( \langle \theta \rangle_{T_2L} )] +  \alpha_0 \gamma_{\max}  \zeta_4  \frac{\sum_{l = T_2L}^{t} \eta_{l+1}}{t+1} +  \frac{\zeta_7 \alpha_0 \gamma_{\max}  C \log^2(\frac{{t+1}}{\alpha_0})}{t+1}\\
    & \le \frac{T_2L}{t+1}  \mathbf{E}[H( \langle \theta \rangle_{T_2L} )] +  \alpha_0 \gamma_{\max}  \zeta_4  \frac{\sum_{l = T_2L}^{t+1} \eta_{l}}{t+1} +  \frac{\zeta_7 \alpha_0 \gamma_{\max}  C \log^2(\frac{{t+1}}{\alpha_0})}{t+1},
\end{align*}
where we use 
$
    \sum_{l=T_2}^t \frac{2 \alpha_0 \log(\frac{l+1}{\alpha_0}) }{l+1} \le \log^2(\frac{t+1}{\alpha_0}) 
$
to get the last inequality. Then, we can get the bound of $ \mathbf{E}[\| \langle \theta \rangle_{t+1} -\theta^* \|_2^2 ]  $ as follows:
%\begin{align*}
%   \mathbf{E}[\| \langle \theta \rangle_{t+L} -\theta^* \|_2^2 ] & \le \frac{1}{\gamma_{\min}} \mathbf{E}[H( \langle \theta \rangle_{t+L} )] \nonumber\\
%    & \le \frac{T_2L}{t+L} \frac{\gamma_{\max}}{\gamma_{\min}} \mathbf{E}[\| \langle \theta \rangle_{T_2L} -\theta^* \|_2^2 ] 
%    +  \frac{\zeta_7 \alpha_0  C \log^2(\frac{t+L}{\alpha_0})}{t+L} \frac{\gamma_{\max}}{\gamma_{\min}}
%     \nonumber\\
%    & \;\;\; +  \alpha_0 \zeta_4 \frac{\gamma_{\max}}{\gamma_{\min}}  \frac{\sum_{l = T_2L}^{t+L-1} \eta_{l+1}}{t+L}.
%\end{align*}
\begin{align*}
    &\;\;\;\; \mathbf{E}[\| \langle \theta \rangle_{t+1} -\theta^* \|_2^2 ]  \le \frac{1}{\gamma_{\min}} \mathbf{E}[H( \langle \theta \rangle_{t+1} )] \nonumber\\
    & \le \frac{T_2L}{t+1} \frac{\gamma_{\max}}{\gamma_{\min}} \mathbf{E}[\| \langle \theta \rangle_{T_2L} -\theta^* \|_2^2 ] 
    +  \frac{\zeta_7 \alpha_0  C \log^2(\frac{t+1}{\alpha_0})}{t+1} \frac{\gamma_{\max}}{\gamma_{\min}}
     +  \alpha_0 \zeta_4 \frac{\gamma_{\max}}{\gamma_{\min}}  \frac{\sum_{l = T_2L}^{t+1} \eta_{l}}{t+1}. 
\end{align*}
This completes the proof.
\hfill $\qed$

\vspace{.1in}

We are now in a position to prove the time-varying step-size case in  Theorem~\ref{thm:bound_jointly_SA}.

\vspace{.1in}

\noindent
{\bf Proof of Case 2) in Theorem~\ref{thm:bound_jointly_SA}:}
From Lemmas~\ref{lemma:bound_consensus_time-varying_jointly} and \ref{lemma:bound_average_time-varying_jointly}, for any $t \ge T_2L$, we have
    \begin{align*}
        & \;\;\;\; \sum_{i=1}^N \pi_{t}^i \mathbf{E}[\|\theta_{t}^i - \theta^*\|_2^2] 
         \le 2 \sum_{i=1}^N \pi_{t}^i \mathbf{E}[\|\theta_{t}^i - \langle \theta \rangle_{t} \|_2^2 ] + 2 \mathbf{E} [\| \langle \theta \rangle_{t} - \theta^*\|_2^2]\\
        & \le 2  \epsilon^{q_{t}-{T_2}} \sum_{i=1}^N \pi_{T_2L+m_t}^i \mathbf{E}[\| \theta_{T_2L+m_t}^i - \langle \theta \rangle_{T_2L+m_t} \|_2^2]  + \frac{2T_2L}{t} \frac{\gamma_{\max}}{\gamma_{\min}} \mathbf{E}[\| \langle \theta \rangle_{T_2L} -\theta^* \|_2^2 ]  \\
        & \;\;\; 
        +  \frac{2\zeta_7 \alpha_0  C \log^2(\frac{t}{\alpha_0})}{t} \frac{\gamma_{\max}}{\gamma_{\min}}
         +  2 \alpha_0 \zeta_4 \frac{\gamma_{\max}}{\gamma_{\min}} \frac{\sum_{l = T_2L}^{t} \eta_{l}}{t}+ \frac{2\zeta_6}{1-\epsilon} ( \alpha_0 \epsilon^{\frac{q_t-1}{2}} + \alpha_{\ceil{\frac{q_t-1}{2}}L}) \\
        & \le  2  \epsilon^{q_{t}-T_2} \sum_{i=1}^N \pi_{LT_2+m_t}^i \mathbf{E}\left[\left\| \theta_{LT_2+m_t}^i - \langle \theta \rangle_{LT_2+m_t} \right\|_2^2\right] + C_3 \left( \alpha_0  \epsilon^{\frac{q_t-1}{2}} + \alpha_{\ceil{\frac{q_t-1}{2}}L}\right) \nonumber  \\
        & \;\;\;   +  \frac{1}{t} \bigg(C_4 \log^2\Big(\frac{t}{\alpha_0}\Big)+C_5\sum_{k = LT_2}^{t} \eta_{k} + C_6\bigg),
    \end{align*}
    where $C_3 - C_6$ are defined in Appendix~\ref{sec:thmSA_constant}.
    This completes the proof.
\hfill $\qed$

{\color{black}
\begin{remark}
For distributed SA algorithms, finite-time performance analysis essentially boils down to two parts, namely bounding the consensus error and bounding the ``single-agent’’ mean-square error. For the case when consensus interaction matrices are all doubly stochastic, the consensus error bound can be derived by analyzing the square of the 2-norm of the deviation of the current state of each agent from the average of the states of the agents. With consensus in the presence of doubly stochastic matrices, the average of the states of the agents remains invariant. Thus, it is possible to treat the average value as the state of a fictitious agent to derive the mean-square consensus error bound with respect to the limiting point. More formally, this process relies on two properties of a doubly stochastic matrix $W$, namely that (1) $\1^\top W =\1^\top$, and (2) if $x_{t+1}=Wx_t$, then $\|x_{t+1} - (\1^\top x_{t+1})\1\|_2 \le \sigma_2(W) \|x_{t} - (\1^\top x_{t})\1\|_2$ where $\sigma_2(W)$ denotes the second largest singular value of $W$ (which is strictly less than one if $W$ is irreducible). Even if the doubly stochastic matrix is time-varying (denoted by $W_t$), property (1) still holds and property (2) can be generalized as in \cite{nedic2018network}. Thus, the square of the 2-norm $\|x_{t} - (\1^\top x_{t})\1\|_2^2$ is a quadratic Lyapunov function for the average consensus processes. Doubly stochastic matrices in expectation can be treated in the same way by looking at the expectation. 
This is the core on which all the existing finite-time analyses of distributed RL algorithms are based. However, if each consensus interaction matrix is stochastic, and not necessarily doubly stochastic, the above two properties may not hold. In fact, it is well known that quadratic Lyapunov functions for general consensus processes $x_{t+1}=S_tx_t$, with $S_t$ being stochastic, do not exist \cite{olshevsky2008nonexistence}. Here we appeal to the idea of quadratic comparison functions for general consensus processes. This was first proposed in \cite{touri2012product} and makes use of the concept of absolute probability sequences. 
We provide a general analysis methodology and results that subsume the existing finite-time analyses for single-timescale distributed linear stochastic approximation (Lemmas~\ref{lemma:bound_consensus_jointly}, \ref{lemma:bound_average}, \ref{lemma:bound_consensus_time-varying_jointly} and \ref{lemma:bound_average_time-varying_jointly}) and TD learning as special cases. 
\hfill$\Box$
\end{remark}
}

\subsection{Push-SA}\label{proofs:push}

In this subsection, we analyze the push-based distributed stochastic approximation algorithm \eqref{eq:SA_push-sum} and provide the proofs of the results in Section~\ref{sec:SA_pushsum}. 

We begin with the analyses of the finite-time performance of algorithm \eqref{eq:SA_push-sum}.
%We next analyze the finite-time performance of \eqref{eq:SA_push-sum}. 

Let $\hat W_t$ be the matrix whose $ij$-th entry is $\hat w_t^{ij}$. Then, from \eqref{eq:SA_push-sum} we have
\begin{align}
    \theta^i_{t+1}&=\frac{\tilde \theta^i_{t+1}}{y^i_{t+1}} = \frac{\sum_{j=1}^N \hat w_t^{ij } ( \tilde \theta^j_{t} + \alpha_t A(X_t) \theta^j_{t} + \alpha_t b^j(X_t ))} {y^i_{t+1}} \nonumber \\
    &= \sum_{j=1}^N \frac{\hat w_t^{ij} y_t^j}{\sum_{k=1}^N \hat w_t^{ik } y^k_{t}} \left[ \frac{\tilde \theta^j_{t}}{y_t^j} + \alpha_t A(X_t) \frac{ \theta^j_{t}}{y_t^j} + \alpha_t \frac{ b^j(X_t ) }{y_t^j}\right]  \nonumber \\
    &= \sum_{j=1}^N \tilde w_t^{ij}  \left[ \theta^j_{t} + \alpha_t A(X_t) \frac{ \theta^j_{t}}{y_t^j} + \alpha_t \frac{ b^j(X_t ) }{y_t^j}\right], \label{eq:push-sum_ratio}
\end{align}
where $ \tilde w_t^{ij} = \frac{\hat w_t^{ij} y_t^j}{\sum_{k=1}^N \hat w_t^{ik } y^k_{t}}$ and $ \tilde W_t = [ \tilde w_t^{ij}]$ is a row stochastic matrix, i.e.,
\begin{align*}
    \sum_{j=1}^N \tilde w_t^{ij} = \frac{\sum_{j=1}^N \hat w_t^{ij} y_t^j}{\sum_{k=1}^N \hat w_t^{ik } y^k_{t}} =1, \;\;\; \forall i.
\end{align*}

Let $\Theta_t = [\theta_t^1, \cdots, \theta_t^N]^\top$ and $\tilde \Theta_t = [\tilde \theta_t^1, \cdots, \tilde \theta_t^N]^\top$. Then \eqref{eq:SA_push-sum} and \eqref{eq:push-sum_ratio} can be written as
\begin{align}
    \tilde\Theta_{t+1} &= \hat W_t \left[\tilde\Theta_{t} + \alpha_t 
    \left[
    \begin{array}{c}
        ( \tilde \theta_t^1)^\top/ y_t^1  \\
        \cdots \\
        ( \tilde \theta_t^N)^\top/ y_t^N
    \end{array}
    \right] A(X_t)^\top
    + \alpha_t B(X_t) \right] \\
    \Theta_{t+1} &=  \tilde W_t \left[\Theta_{t} + \alpha_t 
    \left[
    \begin{array}{c}
        ( \theta_t^1)^\top/ y_t^1 \\
        \cdots \\
        ( \theta_t^N)^\top / y_t^N
    \end{array}
    \right] A(X_t)^\top
    + \alpha_t
    \left[
    \begin{array}{c}
        (b^1(X_t))^\top / y_t^1 \\
        \cdots \\
        (b^N(X_t))^\top / y_t^N
    \end{array}
    \right]
    \right]. \label{eq:update Theta}
\end{align}
%\begin{align}
%    \tilde\Theta_{t+1} &= (\tilde W_t \otimes I_K) \left[\tilde\Theta_{t} + \alpha_t 
%    \left[
%    \begin{array}{c}
%        A(X_t) \tilde \theta_t^1/ y_t^1 \\
%        \cdots \\
%        A(X_t) \tilde \theta_t^N/ y_t^N
%    \end{array}
%    \right]
%    + \alpha_t B(X_t) \right] \\
%    \Theta_{t+1} &= ( W_t \otimes I_K) \left[\Theta_{t} + \alpha_t 
%    \left[
%    \begin{array}{c}
%        A(X_t)  \theta_t^1/ y_t^1 \\
%        \cdots \\
%        A(X_t)  \theta_t^N/ y_t^N
%    \end{array}
%    \right]
%    + \alpha_t
%    \left[
%    \begin{array}{c}
%        b^1(X_t) / y_t^1 \\
%        \cdots \\
%        b^N(X_t) / y_t^N
%    \end{array}
%    \right]
%    \right] \label{eq:update Theta}
%\end{align}
%where $B(X_t) = [(b^1(X_t))^\top \cdots (b^N(X_t))^\top]^\top$.

%\subsubsection{The Boundness}
%{\color{red} Our main goal is to find a constant $C$ such that $ \| \theta_n^i - \frac{\sum_{j=1}^N \tilde \theta_n^j}{N} \| \le C $ for all $n$. }

%\subsection{Kaiqing and Borkar's idea}

Since each matrix $\tilde W_t = [\tilde w_t^{ij}]$ is stochastic, from Lemma~\ref{lemma:bound_pi_jointly}, there exists a unique absolute  probability  sequence $\{ \tilde \pi_t \} $ for  the  matrix  sequence $\{ \tilde W_t \} $ such that $ \tilde \pi_t^i \ge \tilde  \pi_{\min}$ for all $i\in\scr V$ and $t\ge 0$, with the constant $ \tilde \pi_{\min}\in(0,1)$.

\begin{lemma} \label{lemma:pushsum_product}
    Suppose that $\{ \bbb{G}_t \}$ is uniformly strongly connected. Then, $\Pi_{s=0}^t \hat W_s $ will converge to the set $\{v\1_N^\top \; :\; v\in\R^N\}$ exponentially fast as $t\rightarrow\infty$.
\end{lemma}

\noindent
{\bf Proof of Lemma~\ref{lemma:pushsum_product}:}
The lemma is a direct consequence of Theorem~2 in \cite{hajnal}.
\hfill$\qed$

\begin{lemma} \label{lemma:push-sum_pi_intfty}
    Suppose that $\{ \bbb{G}_t \}$ is uniformly strongly connected. Then,
    $ (\Pi_{l=s}^{t} \tilde W_l)^{ij}= \frac{y_s^j}{y_{t+1}^i} (\Pi_{l=s}^{t} \hat W_l)^{ij} $  and $\frac{\tilde \pi_s^i}{y_s^i} = \frac{1}{y_s^i}\lim_{t\to\infty} (\Pi_{l=s}^{t} \tilde W_l)^{ji} = \frac{1}{N}$ for all $i,j\in\scr V$ and $s \ge 0$.
\end{lemma}
\noindent
{\bf Proof of Lemma~\ref{lemma:push-sum_pi_intfty}:}
Note that for all $l \ge 0$, we have $ \tilde w_l^{ij} = \frac{\hat w_l^{ij} y_l^j}{y^i_{l+1}}$. Let $\hat W_{s:t} = \Pi_{l=s}^t \hat W_l$ for all $t \ge s \ge 0$.
We claim that 
$
    (\Pi_{l=s}^{t} \tilde W_l)^{ij} = \frac{ y_s^j \hat w_{s:t}^{ij}}{y_{t+1}^i},
$
where $\hat w_{s:t}^{ij}$ is the $i,j$-th entry of the matrix $\hat W_{s:t}^{ij}$. 
The claim will be proved by induction on $t$. 
When $t=s+1$, 
\begin{align*}
( \tilde W_{s+1} \tilde W_s)^{ij} &= \sum_{k=1}^N  \tilde w_{s+1}^{ik} \cdot \tilde w_s^{kj} = \sum_{k=1}^N \frac{ y_{s+1}^k \hat w_{s+1}^{ik}}{y_{s+2}^i} \frac{ y_s^j \hat w_{s}^{kj}}{y_{s+1}^k} = \frac{ y_s^j }{y_{s+2}^i} \sum_{k=1}^N   \hat w_{s+1}^{ik}\hat w_{s}^{kj} = \frac{  y_s^j }{y_{s+2}^i}  \hat w_{s:s+1}^{ij}.
\end{align*}
Thus, in this case the claim is true.
Now suppose that the claim holds for all $t=\tau \ge s$, where $\tau$ is a positive integer. 
For $t=\tau+1$, we have
\begin{align*}
(\Pi_{s=1}^{\tau+1} \tilde W_s)^{ij} &= \sum_{k=1}^N \tilde w_{\tau+1}^{ik} \cdot \frac{ y_s^j \hat w_{s:\tau}^{kj}}{y_{\tau+1}^k} = \sum_{k=1}^N \frac{\hat w_{\tau+1}^{ik} y_{\tau+1}^k}{y_{\tau+2}^i}  \cdot \frac{ y_s^j \hat w_{s:\tau}^{kj}}{y_{\tau+1}^k} =  \frac{ y_s^j  }{y_{\tau+2}^i}  \sum_{k=1}^N \hat w_{\tau+1}^{ik} \cdot\hat w_{s:\tau}^{kj} 
=  \frac{ y_s^j }{y_{\tau+2}^i}  \hat w_{s:\tau+1}^{ij}, 
\end{align*}
which establishes the claim by induction.

From Lemma~\ref{lemma:pushsum_product}, for given $s\ge 0$, we have $\lim_{t\to\infty}\hat W_{s:t} = v_{s,\infty} \1_N^\top $, with the understanding here that $v_{s,\infty}$ is not a constant vector. Then, since $y_{t+1} = \hat W_t y_t = \Pi_{l=s}^t \hat W_l y_s$ for all $t \ge s$, we have
\begin{align*}
\lim_{t\to\infty}(\Pi_{l=s}^{t} \tilde W_l)^{ij} &= \lim_{t\to\infty} \frac{ y_s^j \hat w_{s:t}^{ij}}{y_{t+1}^i} 
= \lim_{t\to\infty} \frac{ y_s^j \hat w_{s:t}^{ij}}{\sum_{k=1}^N \hat W_{s:t}^{ik} y_s^k} 
= \frac{  y_s^j \lim_{t\to\infty} \hat w_{s:t}^{ij}}{ \lim_{t\to\infty} \sum_{k=1}^N \hat W_{s:t}^{ik} y_s^k}  =  \frac{y_s^j v_{s,\infty}^i }{ \sum_{k=1}^N v_{s,\infty}^i y_s^k}
= \frac{y_s^j}{N},
\end{align*}
where we use the fact that $\1_N^\top y_{s} = N $ for all $s \ge 0$ in the last equality.
This completes the proof.
\hfill $\qed$

%{\color{red}[very good job on the above proof]}

\vspace{.1in}

To proceed, let 
\begin{align*}
    h^j(\Theta_n,y_n) &= A \frac{ \theta^i_{n}}{y_n^i} + \frac{ b^i}{y_n^i}\\
    M^j_n &= \left(A(X_n) - \mathbb{E}[A(X_n)|\mathcal{F}_{n-\tau(\alpha_n)}]\right) \frac{ \theta^j_{n}}{y_n^j} + \frac{ 1 }{y_n^j} \left(b^j(X_n ) - \mathbb{E}[b^j(X_n )|\mathcal{F}_{n-\tau(\alpha_n)}]\right)\\
    G^j_n &= \left(\mathbb{E}[A(X_n)|\mathcal{F}_{n-\tau(\alpha_n)} ] - A\right) \frac{ \theta^j_{n}}{y_n^j} + \frac{ 1 }{y_n^j} \left( \mathbb{E}[b^j(X_n )|\mathcal{F}_{n-\tau(\alpha_n)}] - b^j\right).
\end{align*}
From \eqref{eq:push-sum_ratio}, we have
$
    \theta^i_{n+1}
    = \sum_{j=1}^N  \tilde w_n^{ij}  \left[ \theta^j_{n} + \alpha_n h^j(\theta_n,y_n) + \alpha_n M^j_n + \alpha_n G^j_n \right].
$
Let
$h = [h^1, \cdots, h^N]^\top $, $M = [M^1, \cdots, M^N]^\top $ and $G = [G^1, \cdots, G^N]^\top $. 
Note that 
\begin{align*}
    \mathbb{E}[ M^j_n | \mathcal{F}_n] &= \left( \mathbb{E}[A(X_t) | \mathcal{F}_n] - \mathbb{E}[ \mathbb{E}[A(X_t)|\mathcal{F}_{n-\tau(\alpha_n)}] | \mathcal{F}_n]\right) \frac{ \theta^j_{t}}{y_t^j} \\
    & \;\;\;  + \frac{ 1 }{y_t^j} \left(\mathbb{E}[b^j(X_t ) | \mathcal{F}_n] - \mathbb{E}[\mathbb{E}[b^j(X_t )|\mathcal{F}_{n-\tau(\alpha_n)}]| \mathcal{F}_n]\right) = 0, 
\end{align*} 
and for all $n \ge \tau(\alpha_n)$,
\begin{align*}
    & \;\;\;\; \mathbb{E}[ \| M_n\|_F^2 | \mathcal{F}_n] = \sum_{j=1}^N \mathbb{E}[ \| M^j_n\|_2^2 | \mathcal{F}_n] \\
    &= \sum_{j=1}^N \mathbb{E}[ \| \left(A(X_n) - \mathbb{E}[A(X_n)|\mathcal{F}_{n-\tau(\alpha_n)}]\right) \frac{ \theta^j_{t}}{y_t^j} + \frac{ 1 }{y_t^j} \left(b^j(X_t ) - \mathbb{E}[b^j(X_t )|\mathcal{F}_{n-\tau(\alpha_n)}]\right) \|_2^2  | \mathcal{F}_n]\\
    &\le \sum_{j=1}^N \left( \frac{2A_{\max}+\alpha_0}{\beta}
     \| \theta^j_{t} \|_2 + \frac{2b_{\max}+\alpha_0}{\beta}  \right)^2 
     \le  \frac{2(2A_{\max}+\alpha_0)^2}{\beta^2} \| \Theta_{t} \|_F^2 + \frac{2N}{\beta^2} (2b_{\max}+\alpha_0)^2.
\end{align*} 
Then, $\{ M_n \}$ is a martingale difference sequence satisfying $ \mathbb{E}[ \| M_n\|_F^2 | \mathcal{F}_n] \le \hat C ( 1+ \| \Theta_{t} \|_F )$, where $\hat C = \max\{\frac{2(2A_{\max}+\alpha_0)^2}{\beta^2}, \frac{2N}{\beta^2} (2b_{\max}+\alpha_0)^2 \}$.
Define $h_c : \R^{N\times K}\times \R^{N} \to  \R^{N \times K}$ as $h_c(x, y) = h(cx, y) c^{-1}$ with some $c\ge 1$ and $\tilde h_c(z) :\R^{K} \to  \R^{K}$ as $ \tilde h_c(z) =  h_c(\1_N \cdot z^\top,y_n)^\top \tilde \pi_n$. By Lemma~\ref{lemma:push-sum_pi_intfty}, 
\begin{align*}
    h_c(\Theta_n, y_n) = \left[ 
    \begin{array}{c}
        ( A\frac{ \theta_n^1 }{ y_n^1} + \frac{b^1}{y_n^1 c})^\top \\
        \cdots \\
        (A\frac{ \theta_n^N }{ y_n^N} + \frac{b^N}{y_n^N c})^\top \\
    \end{array}
    \right] , \;\;\;\;\;
    \tilde h_c(z) = A z + \sum_{i=1}^N \frac{b^i}{ N c}.
\end{align*}
Then, $ \tilde h_c(z) \to \tilde h_{\infty}(z)= Az$ as $c \to \infty$ uniformly on compact sets.
Let $\phi_c(z,t)$ and $\phi_\infty(z,t)$ denote the solutions of the ODE:
\begin{align} 
    \dot z(t) =  \tilde h_{c}(z(t)), \;\;\; z(0) = z \label{eq:ODE_infty}\\
    \dot z(t) =  \tilde h_{\infty}(z(t))= Az(t), \;\;\; z(0) = z \nonumber
\end{align}
respectively.
Furthermore, since the origin is the unique globally asymptotically stable equilibrium of the ODE, then we have the following lemma.

\begin{lemma} \label{lemma:bound_z}
There exist constant $c_0 > 0$ and $T > 0$ such that for all initial conditions $z$ with the sphere $\{ z| \| z\|_2 \le \frac{1}{N^{1/2}} \}$ and all $c \ge c_0$, we have $\| \phi_c(z,t) \|_2 < \frac{1-\kappa}{N^{1/2}} $ for $t \in [T, T+1]$ for some $0< \kappa <1$.
\end{lemma}

\noindent
{\bf Proof of Lemma~\ref{lemma:bound_z}:} Similar to the proof of Lemma 5 in \cite{mathkar2016nonlinear}. 
\hfill$\qed$

\vspace{.1in}

Define $t_0 =0$, $t_n = \sum_{i=0}^n \alpha_n$, $n \ge 0$. Define $\bar \Theta(t), t\ge 0 $ as $\bar \Theta(t_n) = \Theta_{n} $ with linear interpolation on each interval $[t_n, t_{n+1}]$. In addition, let $T_0 = 0$ and $T_{n+1} = \min\{ t_m : t_m \ge T_n + T \}$ for all $n \ge 0$. Then, $T_{n+1} \in [T_n + T, T_n + T + \sup_n \alpha_n]$. Let $m(n)$ be the value such that $T_n = t_{m(n)}$ for any $n\ge 0$. Define the piecewise continuous trajectory $\hat \Theta(t) = \bar \Theta(t) \cdot r_n^{-1}$ for $t \in [T_n, T_{n+1})$, where $r_n = \max \{ \| \bar \Theta(T_n)\|_F, 1\}$. 

\begin{lemma} \label{lemma:boundedness of hat theta}
There exists a positive constant $C_{\hat \theta}< \infty$ such that $\sup_{t} \| \hat \Theta(t) \|_F < C_{\hat \theta}$.
%Assume there exists a positive constant $C_{\hat \theta}< \infty$, such that $\sup_{t} \| \hat \Theta(t) \|_F < C_{\hat \theta}$.
\end{lemma}

\noindent
{\bf Proof of Lemma~\ref{lemma:boundedness of hat theta}:}
First, we write the update of $\hat\Theta(t_k)$ for $k \in [ m(n), m(n+1) )$
\begin{align} \label{eq:update_ratio_hat}
    \hat \Theta(t_{k+1}) &= \tilde W_{t_{k}}  \left[\hat \Theta(t_{k}) + \alpha_{t_{k}}
    \left[
    \begin{array}{c}
        (\hat \theta^1(t_{k}))^\top/ y_{t_{k}}^1 \\
        \cdots \\
        (\hat \theta^N(t_{k}))^\top/ y_{t_{k}}^N
    \end{array}
    \right]A(X_{t_{k}})^\top
    + \alpha_{t_{k}} 
    \left[
    \begin{array}{c}
        (b^1(X_{t_{k}}))^\top/( y_{t_{k}}^1 r_n) \\
        \cdots \\
        (b^N(X_{t_{k}}))^\top/( y_{t_{k}}^N r_n)
    \end{array}
    \right]
    \right].
\end{align}
%\begin{align} \label{eq:update_ratio_hat}
%    \hat \Theta(t_{k+1}) &= ( W_{t_{k}} \otimes I_K) \left[\hat \Theta(t_{k}) + \alpha_{t_{k}}
%    \left[
%    \begin{array}{c}
%        A(X_{t_{k}}) \hat \theta^1(t_{k})/ y_{t_{k}}^1 \\
%        \cdots \\
%        A(X_{t_{k}}) \hat \theta^N(t_{k})/ y_{t_{k}}^N
%    \end{array}
%    \right]
%    + \alpha_{t_{k}} 
%    \left[
%    \begin{array}{c}
%        \frac{b^1(X_{t_{k}})}{ y_{t_{k}}^1 r_n} \\
%        \cdots \\
%        \frac{b^N(X_{t_{k}})}{ y_{t_{k}}^N r_n}
%    \end{array}
%   \right]
%    \right].
%\end{align}
Since $ W_{t_k}$ is a column matrix, thus we have 
\begin{align*}
     \| \hat \Theta(t_{k+1}) \|_{\infty}  
    &\le \| \tilde W_{t_{k}} \|_{\infty} \left( \| \hat \Theta(t_{k}) \|_{\infty} + \alpha_{t_{k}}
     \left\| \left[
    \begin{array}{c}
        A(X_{t_{k}}) \hat \theta^1(t_{k})/ y_{t_{k}}^1 \\
        \cdots \\
        A(X_{t_{k}}) \hat \theta^N(t_{k})/ y_{t_{k}}^N
    \end{array}
    \right]  \right\|_{\infty}
    + \alpha_{t_{k}}  \left\| \left[
    \begin{array}{c}
        \frac{b^1(X_{t_{k}})}{ y_{t_{k}}^1 r_n} \\
        \cdots \\
        \frac{b^N(X_{t_{k}})}{ y_{t_{k}}^N r_n}
    \end{array}
    \right] \right\|_{\infty} \right) \\
    &\le \| \hat \Theta(t_{k}) \|_{\infty} + \frac{\alpha_{t_{k}} \sqrt{K} A_{\max}}{\beta} \|
    \hat \Theta(t_{k}) \|_{\infty} + \frac{\alpha_{t_{k}}  \sqrt{K} b_{\max} }{\beta r_n} \\
    &\le \| \hat \Theta(t_{m(n)}) \|_{\infty} + \sqrt{K} \sum_{l=0}^{k-m(n)} \frac{\alpha_{t_{k+l}} A_{\max} }{\beta} \|
    \hat \Theta(t_{k+l}) \|_{\infty} + \frac{\alpha_{t_{k+l}} b_{\max} }{\beta r_n}\\
    &\le \sqrt{K} + \frac{( T + \sup_l \alpha_{l}) \sqrt{K} b_{\max} }{\beta} +  \sum_{l=0}^{k-m(n)} \frac{\alpha_{t_{k+l}}\sqrt{K} A_{\max}}{\beta} \| \hat \Theta(t_{k+l}) \|_{\infty},
\end{align*}
where we use the fact that $\| \hat \Theta(t_{m(n)}) \|_F = 1$ and $r_n \ge 1$ in the last inequality. Therefore, using the discrete-time Gr\"{o}nwall inequality, we have
\begin{align*}
    \sup_{m(n)\le k<m(n+1)} \| \hat \Theta(t_{k+1}) \|_{\infty} 
    &\le \sqrt{K} (1 + ( T + \sup_l \alpha_{l})  b_{\max}) \exp\left\{ \frac{ A_{\max}\sqrt{K} }{\beta} (T+ \sup_l \alpha_l)\right\}.
\end{align*}
Since $ T+ \sup_l \alpha_l < \infty $, we have $\sup_{m(n)\le k<m(n+1)} \| \hat \Theta(t_{k+1}) \|_{\infty}  < \infty$ for all $n$. By equivalence of vector norms, we further obtain that $\sup_{t} \| \hat \Theta(t) \|_F < \infty $.
\hfill$\qed$

\vspace{.1in}

% \subsubsection{Introducing the pi sequence}
For $ n \ge 0 $, let $z^n(t)$ denote the trajectory of $\dot z = \tilde h_c(z)$ with $c = r_n$ and $z^n(T_n) = \sum_{i=1}^N \tilde\pi_{T_n}^i \hat \theta_{T_n} $, for $[T_n, T_{n+1} )$.

\begin{lemma} \label{lemma:hat_to_z}
$ \lim_n \sup_{t \in[T_n, T_{n+1})} \| \hat \Theta_t - \1\otimes z^n(t) \| = 0 $.
\end{lemma}

\noindent
{\bf Proof of Lemma~\ref{lemma:hat_to_z}:}
From \eqref{eq:push-sum_ratio} and \eqref{eq:update_ratio_hat}, for any $k \in [m(n), m(n+1))$, by Lemma~\ref{lemma:push-sum_pi_intfty}, we have
\begin{align*}
    \sum_{i=1}^N \tilde \pi^i_{n+1} \theta^i_{n+1} &= 
    \Theta_{n+1}^\top \tilde \pi_{n+1}
     =  \left(\Theta_{n} + \alpha_n 
    \left[
    \begin{array}{c}
        (A(X_n)  \theta_n^1)^\top/ y_n^1 \\
        \cdots \\
        (A(X_n)  \theta_n^N)^\top/ y_n^N
    \end{array}
    \right]
    + \alpha_n
    \left[
    \begin{array}{c}
        (b^1(X_n))^\top / y_n^1 \\
        \cdots \\
        (b^N(X_n))^\top / y_n^N
    \end{array}
    \right]
    \right)^\top \tilde \pi_{n} \nonumber \\
    &= \sum_{i=1}^N \tilde \pi^i_{n} \theta^i_{n} + \alpha_n \sum_{i=1}^N \tilde \pi_{n}^i (A(X_n)  \theta_n^i/ y_n^i + b^i(X_n) / y_n^i) \nonumber \\
    &= \sum_{i=1}^N \tilde \pi^i_{n} \theta^i_{n} +  \frac{\alpha_n }{N}A(X_n)\sum_{i=1}^N  \theta_n^i +  \frac{\alpha_n}{N}\sum_{i=1}^N b^i(X_n).
\end{align*}
Similarly, we have
\begin{align*}
    & \;\;\;\; \sum_{i=1}^N \tilde \pi^i_{t_{k+1}} \hat \theta^i_{t_{k+1}} 
    = \sum_{i=1}^N \tilde \pi^i_{t_{k}} \hat \theta^i_{t_{k}} + \alpha_t \sum_{i=1}^N \tilde \pi_{{t_{k}}}^i (A(X_{t_{k}})  \hat \theta_{t_{k}}^i/ y_{t_{k}}^i + b^i(X_{t_{k}}) / (y_{t_{k}}^i r_n)) \nonumber \\
    &=\sum_{i=1}^N \tilde \pi^i_{t_{k}} \hat \theta^i_{t_{k}} + \alpha_{t_{k}}\left( A(X_{t_{k}}) \sum_{i=1}^N \tilde \pi^i_{t_{k}} \hat \theta^i_{t_{k}} + \frac{1}{N r_n} \sum_{i=1}^N  b^i(X_{t_{k}}) \right)  + \alpha_{t_{k}} \frac{A(X_{t_{k}})}{N} \sum_{i=1}^N \left( \hat \theta_{t_{k}}^i -  \sum_{i=1}^N \tilde \pi^i_{t_{k}} \hat \theta^i_{t_{k}}  \right)\\
    &= \sum_{i=1}^N \tilde \pi^i_{t_{k}} \hat \theta^i_{t_{k}} + \alpha_{t_{k}}
    \left( A \sum_{i=1}^N \tilde \pi^i_{t_{k}} \hat \theta^i_{t_{k}} + \frac{1}{N r_n} \sum_{i=1}^N  b^i \right) + \alpha_{t_{k}} \frac{A(X_{t_{k}})}{N} \sum_{i=1}^N \left( \hat \theta_{t_{k}}^i -  \sum_{i=1}^N \tilde \pi^i_{t_{k}} \hat \theta^i_{t_{k}}  \right) \nonumber\\
    &\;\;\; + \alpha_{t_{k}}  \left( A(X_{t_{k}}) - \mathbb{E}[A(X_{t_{k}}) | \mathcal{F}_{t_k - \tau(\alpha_{t_k})}] \right)\sum_{i=1}^N \tilde \pi^i_{t_{k}} \hat \theta^i_{t_{k}}  +  \frac{\alpha_{t_{k}}}{N r_n} \sum_{i=1}^N  \left( b^i(X_{t_{k}}) - \mathbb{E}[b^i(X_{t_{k}}) | \mathcal{F}_{t_k - \tau(\alpha_{t_k})}] \right)  \nonumber \\
    &\;\;\; + \alpha_{t_{k}} \left( \left( \mathbb{E}[A(X_{t_{k}}) | \mathcal{F}_{t_k - \tau(\alpha_{t_k})}] - A \right)\sum_{i=1}^N \tilde \pi^i_{t_{k}} \hat \theta^i_{t_{k}} + \frac{1}{N r_n} \sum_{i=1}^N  \left( \mathbb{E}[b^i(X_{t_{k}}) | \mathcal{F}_{t_k - \tau(\alpha_{t_k})}] - b^i \right) \right).
\end{align*}
To proceed, let 
\begin{align*}
    \hat M_{t_k} 
    &= \left(A(X_t) - \mathbb{E}[A(X_{t_k})|\mathcal{F}_{{t_k}-\tau(\alpha_{t_k})}]\right) \sum_{i=1}^N \tilde \pi^i_{t_{k}} \hat \theta^i_{t_{k}} + \frac{ 1 }{N r_n} \sum_{i=1}^N \left(b^i(X_{t_k} ) - \mathbb{E}[b^i(X_{t_k} )|\mathcal{F}_{{t_k}-\tau(\alpha_{t_k})}]\right)\\
    \hat G_{t_k} &= \left(\mathbb{E}[A(X_{t_k})|\mathcal{F}_{{t_k}-\tau(\alpha_{t_k})} ] - A\right) \sum_{i=1}^N \tilde \pi^i_{t_{k}} \hat \theta^i_{t_{k}} + \frac{ 1 }{N r_n} \sum_{i=1}^N \left( \mathbb{E}[b^i(X_{t_k} )|\mathcal{F}_{{t_k}-\tau(\alpha_{t_k})}] - b^i\right) \\
    & \;\;\; + \frac{A(X_{t_{k}})}{N} \sum_{i=1}^N \left( \hat \theta_{t_{k}}^i -  \sum_{i=1}^N \tilde \pi^i_{t_{k}} \hat \theta^i_{t_{k}}  \right).
\end{align*}
It is easy to verify that $\{ \hat M_{t_k} \}$ is a martingale difference sequence satisfying $ \mathbb{E}[\| \hat M_{t_k} \|_2^2 | \mathcal{F}_{t_k} ] \le \bar C (1+\| \sum_{i=1}^N \tilde \pi^i_{t_{k}} \hat \theta^i_{t_{k}} \|_2^2) $ for some $\bar C \le \infty$.
In addition, we have
\begin{align*}
    \hat\theta_{t_k}^i -  \sum_{j=1}^N \tilde \pi^j_{t_{k}} \hat \theta^j_{t_{k}} 
    &=\sum_{j=1}^N (\tilde w_{t_s:t_k}^{ij} - \tilde \pi_{t_s}^j) \hat \theta_{t_s}^j + \sum_{r = s+1}^k \alpha_{t_r} \sum_{i=1}^N (\tilde w_{t_r:t_k}^{ij} - \tilde \pi_{t_r}^j)(A(X_{t_r}) \hat \theta_{t_r}^j/y_{t_r}^j + b^j(X_{t_r}/y_{t_r}^j).
\end{align*}
Since  $\{ \bbb{G}_t \}$ is uniformly strongly connected, then for any $s\ge 0$, $W_{s:t}$ converges to $\1 \pi_s^\top$ exponentially fast as $t \to \infty$ and there exist a finite positive constant $C$ and a constant $0 \le \lambda <1$ such that 
$
    | \tilde w_{s:t}^{ij} - \tilde \pi_s^j  | \le C \lambda^{t-s}
$
for all $i,j \in \mathcal{V}$ and $s \ge 0$. Then,for any $k \in [m(n), m(n+1))$, we have
\begin{align*}
    &\;\;\;\; \| \hat\theta_{t_k}^i -  \sum_{j=1}^N \tilde \pi^j_{t_{k}} \hat \theta^j_{t_{k}} \|_2 \\
    &\le \sum_{j=1}^N \| \tilde w_{t_{m(n)}:t_k}^{ij} - \tilde \pi_{t_{m(n)}}^j\|_2 \| \hat \theta_{t_{m(n)}}^j \|_2 + \sum_{r = m(n)+1}^k \alpha_{t_r} \sum_{i=1}^N \| \tilde w_{t_r:t_k}^{ij} - \tilde \pi_{t_r}^j\|_2 \frac{A_{\max} \| \hat \theta_{t_r}^j\|_2 + b_{\max}}{\beta}\\
    &\le \sum_{j=1}^N C \lambda^{t_k-t_{m(n)}} \| \hat \theta_{t_{m(n)}}^j \|_2 + \sum_{r = m(n)+1}^k \alpha_{t_r} \sum_{i=1}^N C \lambda^{t_k-t_r} (\frac{A_{\max} \| \hat \theta_{t_r}^j\|_2 + b_{\max}}{\beta})\\
    &\le N C \lambda^{t_k-t_{m(n)}} + \frac{\alpha_{t_{m(n)}} N C}{1-\lambda}  \frac{A_{\max} C_{\hat\theta} + b_{\max}}{\beta},
\end{align*}
where in the last inequality, we use the fact that for all $n\ge 0$, we have $\| \hat \Theta(t_{m(n)}) \|_F = 1$, $\alpha_{n+1} \le  \alpha_n$ and the boundedness of $\| \hat \Theta_n \|_F$ from Lemma~\ref{lemma:boundedness of hat theta}. Since $\alpha_{t_k} \to 0$ as $k \to \infty$, then
$
    \lim_{k\to\infty}\| \hat\theta_{t_k}^i -  \sum_{j=1}^N \tilde \pi^j_{t_{k}} \hat \theta^j_{t_{k}} \|_2 = 0,
$
which implies that 
$
    \lim_{k\to\infty}\left\| \frac{A(X_{t_k})}{N} \sum_{i=1}^N (\hat\theta_{t_k}^i -  \sum_{j=1}^N \tilde \pi^j_{t_{k}} \hat \theta^j_{t_{k}} ) \right\|_2 = 0.
$
Then,
\begin{align*}
    \lim_{k\to\infty} \| \hat G_{t_k} \|_2 
    &\le \lim_{k\to\infty} \alpha_{t_k} ( \| \sum_{j=1}^N \tilde \pi^j_{t_{k}} \hat \theta^j_{t_{k}} \|_2 + 1)
    + \lim_{k\to\infty}\left\| \frac{A(X_{t_k})}{N} \sum_{i=1}^N (\hat\theta_{t_k}^i -  \sum_{j=1}^N \tilde \pi^j_{t_{k}} \hat \theta^j_{t_{k}} ) \right\|_2 = 0.
\end{align*}
Therefore, by Corollary~8 and Theorem~9 in Chapter~6 of \cite{borkar2008stochastic}, we obtain that $\sum_{i=1}^N \tilde \pi^i_{t_{k}} \hat \theta^i_{t_{k}} \to z^n(t)$ as $n \to \infty$, namely $k \to \infty$. Furthermore, we obtain that $ \hat \theta_{t_{k+1}}^i \to z^n(t)$ as $n \to \infty$ for all $i \in \mathcal{V}$, which concludes the proof following Theorem~2 in Chapter~2 of \cite{borkar2008stochastic}.
\hfill$\qed$

\begin{lemma} \label{lemma:bound_theta}
The sequence $\{ \Theta_n \}$ generated by \eqref{eq:update Theta} is bounded almost surely, i.e., $ C_\theta = \sup_{n} \| \Theta_n \|_F<\infty$ almost surely.
\end{lemma}

\noindent
{\bf Proof of Lemma~\ref{lemma:bound_theta}:}
In order to prove this lemma, we need to show that $\sup_n \| \bar \Theta(T_n) \|_F < \infty$ first. If this does not hold, there will exist a sequence $T_{n_1}, T_{n_2}, \cdots $ such that $\| \hat \Theta(T_{n_k}) \|_F \to \infty $, i.e., $r_{n_k} \to \infty$. If $r_n > c_0$ and $\| \hat \Theta(T_n) \|_F = 1$, then $\| z^n(T_n) \|_2 = \| \sum_{i=1}^N \tilde \pi_{T_n} \hat \theta^i_{T_n} \|_2 \le N^{-1/2}$. Using Lemma~\ref{lemma:bound_z}, we have $\| \1_N \cdot (z^n(T_{n+1}^-))^\top \|_F = N^{1/2} \| z^n(T_{n+1}^-) \|_2 \le 1 - \kappa $. In addition, using Lemma~\ref{lemma:hat_to_z}, there exists a constant $0 < \kappa' < \kappa$ such that $\| \hat \Theta(T_{n+1}^-) \|_F < 1 - \kappa'$. Hence for $r_n > c_0$ and $n$ sufficiently large,
\begin{align*}
    \frac{\| \bar \Theta(T_{n+1})\|_F}{\| \bar \Theta(T_{n})\|_F} = \frac{\| \hat \Theta(T_{n+1}^-)\|_F}{\| \hat \Theta(T_{n})\|_F} \le 1-\kappa'.
\end{align*}
It shows that if $\| \bar \Theta(T_{n})\|_F > c_0$, $\| \bar \Theta(T_{k})\|_F$ for all $k \ge n $ falls back to the ball of radius $c_0$ at an exponential rate. 

Thus, if $\| \bar \Theta(T_{n})\|_F > c_0$, then $\| \bar \Theta(T_{n-1})\|_F$ is either greater than $\| \bar \Theta(T_{n})\|_F $ or is inside the ball of radius $c_0$. Since we assume $r_{n_k} \to \infty$, then we can find a time $T_n$ such that $\| \bar \Theta(T_{n})\|_F < c_0$ and $\| \bar \Theta(T_{n+1})\|_F = \infty$. However, using the discrete-time Gr\"{o}nwall inequality, we have
\begin{align*}
    \| \bar \Theta(T_{n+1})\|_\infty 
    &\le \| \bar \Theta(T_{n+1}-1)\|_\infty + \alpha_{T_{n+1}-1}\frac{\sqrt{K} A_{\max} }{\beta} \| \bar \Theta(T_{n+1}-1)\|_\infty + \alpha_{T_{n+1}-1}\sqrt{K} \frac{b_{\max} }{\beta} \\
    &\le \| \bar \Theta(T_{n})\|_\infty + \sqrt{K} \sum_{s=0}^{T_{n+1} - T_n} \alpha_{T_{n}+s} \frac{ A_{\max} }{\beta} \| \bar \Theta(T_{n}+s)\|_\infty + \alpha_{T_{n}+s}\frac{b_{\max} }{\beta} \\
    &\le \sqrt{K}c_0 + \sqrt{K}(T+\sup_n \alpha_n)\frac{b_{\max} }{\beta} + \frac{\sqrt{K} A_{\max} }{\beta} \sum_{s=0}^{T_{n+1} - T_n} \alpha_{T_{n}+s}  \| \bar \Theta(T_{n}+s)\|_\infty \\
    &\le \sqrt{K}(c_0 + (T+\sup_n \alpha_n)\frac{b_{\max} }{\beta}) \exp\left\{(T+\sup_n \alpha_n)\frac{\sqrt{K} A_{\max} }{\beta}\right\},
\end{align*}
which implies that $\| \bar \Theta(T_{n+1})\|_F$ can be bounded if $\| \bar \Theta(T_{n})\|_F < c_0$. This leads to a contradiction. 

Moreover, let $C_{\bar \theta} = \sup_n \| \bar \Theta(T_n) \|_F < \infty$, then $C_{\theta} = \sup_n \| \Theta_n \|_F \le C_{\bar \theta} C_{\hat \theta} < \infty$.
\hfill$\qed$

\vspace{.1in}

From \eqref{eq:SA_push-sum},
%\begin{align*}
%    \tilde \theta^i_{t+1} &= \sum_{j =1}^N  \hat w_t^{ij } \left[ \tilde\theta^j_{t}  + \alpha_t \left(A(X_{t}) \theta^j_{t} + b^j(X_{t} ) \right) \right].
%\end{align*}
by using the definition of $\langle \tilde \theta \rangle_t = \frac{1}{N} \sum_{i=1}^N \tilde \theta_t^i$ and $\langle \theta \rangle_t = \frac{1}{N} \sum_{i=1}^N \theta_t^i$, we have
\begin{align} \label{eq:update_average_tilde_theta}
    \langle \tilde \theta \rangle_{t+1}
    &= \langle \tilde \theta \rangle_t + \alpha_t A(X_t) \langle \theta \rangle_t + \frac{\alpha_t}{N}\sum_{i=1}^N b^i(X_t) \nonumber \\
    &= \langle \tilde \theta\rangle_t  + \alpha_t A(X_t) \langle \tilde \theta \rangle_t + \frac{\alpha_t}{N}\sum_{i=1}^N b^i(X_t)  +   \alpha_t \rho_t,
\end{align}
where $ \rho_t = A(X_t) \langle \theta \rangle_t - A(X_t) \langle \tilde \theta \rangle_t$. From Lemma~\ref{lemma:bound_theta}, we have $\| \langle \theta \rangle_t \|_2 \le \max_{i\in\mathcal{V}} \| \theta_t^i \|_2 \le C_\theta$ for all $t \ge 0$, which implies that $\| \langle \tilde \theta \rangle_t \|_2 \le N C_\theta$ and
$
    \mu_t = \| \rho_t \|_2 = \left\| A(X_t) \langle \theta \rangle_t - A(X_t) \langle \tilde \theta  \rangle_t \right\|_2 \le \mu_{\max},
$
where $\mu_{\max} = (N+1) A_{\max} C_\theta$.

 \begin{lemma} \label{lemma:bound_consensus_time-varying_push_SA} 
    Suppose that Assumptions~\ref{assum:A and b} and \ref{assum:step-size} hold and $\{ \bbb{G}_t \}$ is uniformly strongly connected by sub-sequences of length $L$.
    Let $\epsilon_1 = \inf_{t\ge 0}  \min_{i\in\scr V} (\hat W_t \cdots \hat W_0 \1_N)^i $. 
    For all $t \ge 0$ and $i \in \mathcal{V}$,
    \begin{align*}
    \| \theta_{t+1}^i - \langle \tilde \theta \rangle_t \|_2 
    &\le\frac{8}{\epsilon_1}  \bar\epsilon^t \| \sum_{i=1}^N \tilde \theta_0^i + \alpha_0 A(X_0)\theta_0^i + \alpha_0 b^i(X_0) \|_2  \\
    & \;\;\; + \frac{8}{\epsilon_1} \frac{ A_{\max} C_\theta + b_{\max}}{1-\bar\epsilon} \left(  \alpha_0 \bar\epsilon^{t/2}  +  \alpha_{\ceil{\frac{t}{2}}} \right) + \alpha_t  A_{\max} C_\theta + \alpha_t b_{\max},
\end{align*}
    %\begin{align*}
    %\| \theta_{t+1}^i - \langle \tilde \theta \rangle_t \|_2 
   % &\le\frac{8}{\epsilon_1}  \bar\epsilon^t \| \tilde \Theta_0 \|_1 + \frac{8}{\epsilon_1} \frac{N \sqrt{K} (A_{\max} C_\theta + b_{\max})}{1-\bar\epsilon} \left(  \alpha_0 \bar\epsilon^{t/2}  +  \alpha_{\ceil{\frac{t}{2}}} \right),
%\end{align*}
where $\epsilon_1 > 0$ and $\bar\epsilon \in (0,1)$ satisfy $\epsilon_1 \ge \frac{1}{N^{NL}}$ and $\bar\epsilon \le (1-\frac{1}{N^{NL}})^{1/L}$.
\end{lemma}
\noindent
{\bf Proof of Lemma~\ref{lemma:bound_consensus_time-varying_push_SA}:}
Since $\epsilon_1 = \inf_{t\ge 0}  \min_{i\in\scr V} (\hat W_t \cdots \hat W_0 \1_N)^i$ and all weight matrices $\hat W_s$ are column stochastic matrices for all $s\ge 0$, from Corollary~2~(b) in \cite{nedic}, $\epsilon_1 \le \frac{1}{N^{NL}}$. If the weight matrices are doubly stochastic matrices, then $ \epsilon_1 = 1$.
From Assumption~\ref{assum:A and b} and Lemma~\ref{lemma:bound_theta}, $\|A(X_t) \theta_t^i + b^i(X_t)\|_2 \le A_{\max} C_\theta + b_{\max}$. Then, by Lemma~1 in \cite{nedic}, for all $t \ge 0$ and $i \in \mathcal{V}$,
\begin{align*}
    & \;\;\;\; \| \theta_{t+1}^i - \langle \tilde \theta \rangle_t - \alpha_t A(X_t) \langle \theta \rangle_t - \frac{\alpha_t}{N} \sum_{i=1}^N b^i(X_t)\|_2 \\
    &\le\frac{8}{\epsilon_1} ( \bar\epsilon^t \| \sum_{i=1}^N \tilde \theta_0^i + \alpha_0 A(X_0)\theta_0^i + \alpha_0 b^i(X_0) \|_2 + \sum_{s=0}^t  \bar\epsilon^{t-s} \alpha_s (A_{\max} C_\theta + b_{\max}))  \\
    &\le\frac{8}{\epsilon_1}  \bar\epsilon^t \| \sum_{i=1}^N \tilde \theta_0^i + \alpha_0 A(X_0)\theta_0^i + \alpha_0 b^i(X_0) \|_2 + \frac{8}{\epsilon_1}  (A_{\max} C_\theta + b_{\max})\left(\sum_{s=0}^{\floor{\frac{t}{2}}}  \bar\epsilon^{t-s} \alpha_s + \sum_{s=\ceil{\frac{t}{2}}}^{t}  \bar\epsilon^{t-s} \alpha_s \right) \\
    &\le\frac{8}{\epsilon_1}  \bar\epsilon^t \| \sum_{i=1}^N \tilde \theta_0^i + \alpha_0 A(X_0)\theta_0^i + \alpha_0 b^i(X_0) \|_2 + \frac{8}{\epsilon_1} \frac{ A_{\max} C_\theta + b_{\max}}{1-\bar\epsilon} \left(  \alpha_0 \bar\epsilon^{t/2}  +  \alpha_{\ceil{\frac{t}{2}}} \right),
\end{align*}
which implies that
\begin{align*}
    \| \theta_{t+1}^i - \langle \tilde \theta \rangle_t \|_2 
    &\le \| \theta_{t+1}^i - \langle \tilde \theta \rangle_t - \alpha_t A(X_t) \langle \theta \rangle_t - \frac{\alpha_t}{N} \sum_{i=1}^N b^i(X_t)\|_2 + \alpha_t \|  A(X_t) \langle \theta \rangle_t + \frac{1}{N} \sum_{i=1}^N b^i(X_t) \|_2 \\
    &\le\frac{8}{\epsilon_1}  \bar\epsilon^t \| \sum_{i=1}^N \tilde \theta_0^i + \alpha_0 A(X_0)\theta_0^i + \alpha_0 b^i(X_0) \|_2 + \frac{8}{\epsilon_1} \frac{ A_{\max} C_\theta + b_{\max}}{1-\bar\epsilon} \left(  \alpha_0 \bar\epsilon^{t/2}  +  \alpha_{\ceil{\frac{t}{2}}} \right) \\
    & \;\;\; + \alpha_t  A_{\max} C_\theta + \alpha_t b_{\max} .
\end{align*}
    This completes the proof.
\hfill $\qed$

\begin{lemma} \label{lemma:eta_limit_Push_SA}
    $\lim_{t \to \infty} \mu_t = \lim_{t \to \infty} \| \rho_t \|_2 =0$ and $\lim_{t \to \infty} \frac{\sum_{k=0}^t \mu_k}{t+1}   = \lim_{t \to \infty} \frac{\sum_{k=0}^t \| \rho_k \|_2}{t+1}  = 0.$
\end{lemma}
\noindent
{\bf Proof of Lemma~\ref{lemma:eta_limit_Push_SA}:}
From Lemma~\ref{lemma:bound_consensus_time-varying_push_SA}, we have
\begin{align*}
    \mu_t  & = \| \rho_t \|_2
     = \left\| A(X_t) \langle \theta \rangle_t - A(X_t) \langle \tilde \theta \rangle_t \right\|_2 \\
    &\le  \frac{8A_{\max}}{\epsilon_1}  \bar\epsilon^t \| \tilde \Theta_0 \|_1 + \frac{8A_{\max}}{\epsilon_1} \frac{N \sqrt{K} (A_{\max} C_\theta + b_{\max})}{1-\bar\epsilon} \left(  \alpha_0 \bar\epsilon^{t/2}  +  \alpha_{\ceil{\frac{t}{2}}} \right).
\end{align*}
Since $\bar\epsilon \in (0,1)$, then $\lim_{t \to \infty} \| \rho_t \|_2 =0$. 
Next, we will prove that $\lim_{t \to \infty} \frac{1}{t+1} \sum_{k=0}^t \| \rho_k \|_2 = 0.$ For any positive constant $ c > 0$, there exists a positive integer $ T(c)$, depending on $c$, such that $ \forall t \ge T(c) $, we have $\| \rho_t \|_2 < c$. Thus,
\begin{align*}
    \frac{1}{t} \sum_{k=0}^{t-1} \| \rho_k \|_2 
     = \frac{1}{t}\sum_{k=0}^{T(c)} \| \rho_k \|_2 + \frac{1}{t}\sum_{k=T(c)+1}^{t-1} \| \rho_k \|_2
     \le \frac{1}{t}\sum_{k=0}^{T(c)} \| \rho_k \|_2+ \frac{t-1-T(c)}{t} c.
\end{align*}
Let $t \to \infty$ on both sides of the above inequality. Then, we have 
\begin{align*}
    \lim_{t \to \infty}\frac{1}{t} \sum_{k=0}^{t-1} \| \rho_k \|_2
    & \le \lim_{t \to \infty} \frac{1}{t}\sum_{k=0}^{T(c)} \| \rho_k \|_2 + \lim_{t \to \infty} \frac{t-1-T(c)}{t} c = c.
\end{align*}
Since the above argument holds for arbitrary positive $c$, then $\lim_{t \to \infty} \frac{1}{t+1} \sum_{k=0}^t \| \rho_k \|_2 = 0.$
\hfill $\qed$

\begin{lemma} \label{lemma:timevarying_single_3_Push_SA}
    Suppose that Assumptions~\ref{assum:A and b} and \ref{assum:mixing-time} hold. When the step-size $\alpha_t$ and corresponding mixing time $\tau(\alpha_t)$ satisfy
    $
    0< \alpha_t \tau(\alpha_t) < \frac{\log2}{A_{\max}}
$,
    we have for any $t \ge \bar T$, 
\begin{align}
    \|\langle \tilde \theta \rangle_{t} - \langle \tilde \theta \rangle_{t - \tau(\alpha_{t})} \|_2
    & \le 2 A_{\max} \| \langle \tilde \theta \rangle_{t-\tau(\alpha_{t})} \|_2 \sum_{k=t-\tau(\alpha_{t})}^{t-1}  \alpha_k + 2 (b_{\max}+ \mu_{\max} ) \sum_{k=t-\tau(\alpha_{t})}^{t-1} \alpha_{k}, \label{eq:timevarying_single_3_Push_SA_1}\\
    \| \langle \tilde \theta \rangle_{t} - \langle \tilde \theta \rangle_{t - \tau(\alpha_{t})} \|_2 
    & \le 6 A_{\max} \| \langle \tilde \theta \rangle_{t} \|_2 \sum_{k=t-\tau(\alpha_{t})}^{t-1}  \alpha_k 
    + 5 ( b_{\max} + \mu_{\max}) \sum_{k=t-\tau(\alpha_{t})}^{t-1} \alpha_{k}, \label{eq:timevarying_single_3_Push_SA_2}\\
     \| \langle \tilde \theta \rangle_{t} - \langle \tilde \theta \rangle_{t - \tau(\alpha_{t})} \|_2^2
     & \le 72 \alpha_{t-\tau(\alpha_{t})}^2 \tau^2(\alpha_t) A_{\max}^2 \| \langle \tilde \theta \rangle_{t} \|_2^2 + 50 \alpha_{t-\tau(\alpha_{t})}^2 \tau^2(\alpha_t)  ( b_{\max} + \mu_{\max})^2 \nonumber \\
     & \le 8 \| \langle \tilde \theta \rangle_{t} \|_2^2 +  \frac{ 6 (b_{\max} + \mu_{\max})^2}{A_{\max}^2}. \label{eq:timevarying_single_3_Push_SA_3}
\end{align}
\end{lemma}
\noindent
{\bf Proof of Lemma~\ref{lemma:timevarying_single_3_Push_SA}:}
From \eqref{eq:update_average_tilde_theta},
$
    \| \langle \tilde \theta \rangle_{t+1} \|_2 
      \le (1+\alpha_{t} A_{\max}) \|\langle \tilde \theta \rangle_t \|_2 + \alpha_{t}  b_{\max} + \alpha_{t}  \mu_{\max}.
$
In addition, for all $u \in [t-\tau(\alpha_{t}), t]$, we have
\begin{align*}
    \| \langle \tilde \theta \rangle_{u} \|_2 
    & \le \Pi_{k = t-\tau(\alpha_{t})}^{u-1}  (1+\alpha_{k} A_{\max})\| \langle \tilde \theta \rangle_{t-\tau(\alpha_{t})} \|_2 + (b_{\max} + \mu_{\max} ) \sum_{k = t-\tau(\alpha_{t})}^{u-1} \alpha_k \Pi_{l=k+1}^{u-1}  (1+\alpha_{l} A_{\max}) \\
    & \le \exp\{ \sum_{k = t-\tau(\alpha_{t})}^{u-1}  \alpha_{k} A_{\max}\} \|\langle \tilde \theta \rangle_{t-\tau(\alpha_{t})} \|_2  + (b_{\max} + \mu_{\max} ) \sum_{k = t-\tau(\alpha_{t})}^{u-1} \alpha_k \exp\{ \sum_{l=k+1}^{u-1} \alpha_{l} A_{\max}\} \\
    & \le \exp\{ \alpha_{t-\tau(\alpha_{t})} \tau(\alpha_t) A_{\max}\} \| \langle \tilde \theta \rangle_{t-\tau(\alpha_{t})} \|_2  + (b_{\max} + \mu_{\max} ) \sum_{k = t-\tau(\alpha_{t})}^{u-1} \alpha_k \exp\{ \alpha_{t-\tau(\alpha_{t})} \tau(\alpha_t) A_{\max}\} \\
    & \le 2 \| \langle \tilde \theta \rangle_{t-\tau(\alpha_{t})} \|_2 + 2 (b_{\max} + \mu_{\max} ) \sum_{k = t-\tau(\alpha_{t})}^{u-1} \alpha_k,  
\end{align*}
where we use $\alpha_{t-\tau(\alpha_{t})} \tau(\alpha_t) A_{\max} \le \log2 < \frac{1}{3}$ in the last inequality.
Thus, for all $t\ge \bar T$, we have $\| \langle \tilde \theta \rangle_{t} - \langle \tilde \theta \rangle_{t - \tau(\alpha_{t})} \|_2 \le \sum_{k=t-\tau(\alpha_{t})}^{t-1}  \| \langle \tilde \theta \rangle_{k+1} - \langle \tilde \theta \rangle_{k} \|_2$. Then, we
can get \eqref{eq:timevarying_single_3_Push_SA_1} as follows:
\begin{align*}
    &\;\;\;\; \| \langle \tilde \theta \rangle_{t} - \langle \tilde \theta \rangle_{t - \tau(\alpha_{t})} \|_2 
     \le A_{\max} \sum_{k=t-\tau(\alpha_{t})}^{t-1} \alpha_{k} \| \langle \tilde \theta \rangle_{k} \|_2 + (b_{\max}+ \mu_{\max} ) \sum_{k=t-\tau(\alpha_{t})}^{t-1} \alpha_{k} \\
    & \le  \sum_{k=t-\tau(\alpha_{t})}^{t-1}  \alpha_k \left [ A_{\max}\left( 2 \| \langle \tilde \theta \rangle_{t-\tau(\alpha_{t})} \|_2 + 2 (b_{\max}+ \mu_{\max} ) \sum_{l=t-\tau(\alpha_{t})}^{k-1} \alpha_l \right) + (b_{\max}+ \mu_{\max} ) \right]\\
    & \le 2 A_{\max} \| \langle \tilde \theta \rangle_{t-\tau(\alpha_{t})} \|_2 \sum_{k=t-\tau(\alpha_{t})}^{t-1}  \alpha_k + \left( 2 A_{\max} \tau(\alpha_{t}) \alpha_{t-\tau(\alpha_{t})} + 1 \right) (b_{\max}+ \mu_{\max} ) \sum_{k=t-\tau(\alpha_{t})}^{t-1} \alpha_{k}  \\
    & \le 2 A_{\max} \| \langle \tilde \theta \rangle_{t-\tau(\alpha_{t})} \|_2 \sum_{k=t-\tau(\alpha_{t})}^{t-1}  \alpha_k + \frac{5}{3} (b_{\max}+ \mu_{\max} ) \sum_{k=t-\tau(\alpha_{t})}^{t-1} \alpha_{k}  \\
    & \le 2 A_{\max} \| \langle \tilde \theta \rangle_{t-\tau(\alpha_{t})} \|_2 \sum_{k=t-\tau(\alpha_{t})}^{t-1}  \alpha_k + 2 (b_{\max}+ \mu_{\max} ) \sum_{k=t-\tau(\alpha_{t})}^{t-1} \alpha_{k} .
\end{align*}
Moreover, using the above inequality, we can get \eqref{eq:timevarying_single_3_Push_SA_2} for all $t\ge \bar T$ as follows:
\begin{align*}
    & \;\;\;\; \| \langle \tilde \theta \rangle_{t} - \langle \tilde \theta \rangle_{t - \tau(\alpha_{t})} \|_2 
     \le 2 A_{\max} \| \langle \tilde \theta \rangle_{t-\tau(\alpha_{t})} \|_2 \sum_{k=t-\tau(\alpha_{t})}^{t-1}  \alpha_k + \frac{5}{3} ( b_{\max} + \mu_{\max}) \sum_{k=t-\tau(\alpha_{t})}^{t-1} \alpha_{k}  \\
    & \le  2 A_{\max} \tau(\alpha_{t}) \alpha_{t-\tau(\alpha_{t})} 
    \| \langle \tilde \theta  \rangle_{t} - \langle \tilde \theta \rangle_{t-\tau(\alpha_{t})} \|_2 
    + \left[ 2 A_{\max} \| \langle \tilde \theta \rangle_{t} \|_2 + \frac{5}{3} ( b_{\max} + \mu_{\max}) \right] \sum_{k=t-\tau(\alpha_{t})}^{t-1} \alpha_{k}  \\
    & \le 6 A_{\max} \| \langle \tilde \theta \rangle_{t} \|_2 \sum_{k=t-\tau(\alpha_{t})}^{t-1}  \alpha_k 
    + 5 ( b_{\max} + \mu_{\max}) \sum_{k=t-\tau(\alpha_{t})}^{t-1} \alpha_{k}.
\end{align*}
Next, using \eqref{eq:timevarying_single_3_Push_SA_2} and the inequality $(x+y)^2 \le 2x^2 + y^2$ for all $x, y$, we can get  \eqref{eq:timevarying_single_3_Push_SA_3} as follows:
\begin{align*}
    \| \langle \tilde \theta \rangle_{t} - \langle \tilde \theta \rangle_{t - \tau(\alpha_{t})} \|_2^2
    & \le 72 A_{\max}^2 \| \langle \tilde \theta \rangle_{t} \|_2^2 (\sum_{k=t-\tau(\alpha_{t})}^{t-1} \alpha_k )^2
    + 50 ( b_{\max} + \mu_{\max})^2 (\sum_{k=t-\tau(\alpha_{t})}^{t-1} \alpha_{k})^2 \\
    & \le 72 \alpha_{t-\tau(\alpha_{t})}^2 \tau^2(\alpha_t) A_{\max}^2 \| \langle \tilde \theta \rangle_{t} \|_2^2 + 50 \alpha_{t-\tau(\alpha_{t})}^2 \tau^2(\alpha_t)  ( b_{\max} + \mu_{\max})^2 \\
    & \le 8 \| \langle \tilde \theta \rangle_{t} \|_2^2 + \frac{ 6 (b_{\max} + \mu_{\max})^2}{A_{\max}^2},
\end{align*}
where we use $\alpha_{t-\tau(\alpha_{t})} \tau(\alpha_t) A_{\max} < \frac{1}{3}$ in the last inequality.
\hfill $\qed$

\vspace{.1in}

\begin{lemma} \label{lemma:bound_timevarying_Ab_push_SA}
    Suppose that Assumptions~\ref{assum:A and b}--\ref{assum:step-size} hold and $\{ \bbb{G}_t \}$ is uniformly strongly connected by sub-sequences of length $L$. When  
    $
    0< \alpha_{t - \tau(\alpha_t) } \tau(\alpha_t) < \frac{ \log2}{A_{\max} }
$,
    we have for any $t \ge \bar T$,
\begin{align*}
    & \;\;\;\; |\mathbf{E}[ ( \langle \tilde \theta \rangle_t  - \theta^* )^\top (P+P^\top)( A(X_t) \langle \tilde \theta \rangle_t +   B(X_t)^\top\pi_{t+1} - A \langle \tilde \theta \rangle_t - b) \; | \; \mathcal{F}_{t-\tau(\alpha_{t})} ]| \nonumber \\
    & \le \alpha_{t-\tau(\alpha_t)} \tau(\alpha_t)  \gamma_{\max} \left( 72 + 456 A_{\max}^2  + 84  A_{\max}  b_{\max} + 72 A_{\max} \mu_{\max} \right) \mathbf{E}[ \| \langle \tilde \theta \rangle_{t} \|_2^2 \; | \; \mathcal{F}_{t-\tau(\alpha_t)} ] \nonumber \\
    &\;\;\; + \alpha_{t -\tau(\alpha_t) }  \tau(\alpha_t) \gamma_{\max} \bigg[ 2 + 4 \|\theta^* \|_2^2 +  48\frac{(b_{\max}+ \mu_{\max})^2}{A_{\max}^2} +  152  \left(b_{\max} + \mu_{\max} + A_{\max} \| \theta^* \|_2 \right)^2 \nonumber\\
    &\;\;\; +  12  A_{\max}b_{\max} + 48  A_{\max}(b_{\max}+ \mu_{\max}) (\frac{b_{\max} + \mu_{\max}}{A_{\max}} + 1 )^2 +  87 (b_{\max}+ \mu_{\max})^2  \bigg].
\end{align*}
\end{lemma}

%\noindent
{\bf Proof of Lemma~\ref{lemma:bound_timevarying_Ab_push_SA}:}
Note that for all $t\ge \bar T$, we have
\begin{align}
    & \;\;\;\; \mathbf{E}[( \langle \tilde \theta \rangle_t   - \theta^* )^\top (P+P^\top)( A(X_t) \langle \tilde \theta \rangle_t   + \frac{1}{N}   B(X_t)^\top\1_N - A\langle \tilde \theta \rangle_t  - b)\; | \; \mathcal{F}_{t-\tau(\alpha_{t})} ]| \nonumber \\
    %& \le |\mathbf{E}[ ( \langle \tilde \theta \rangle_t - \theta^* )^\top (P+P^\top)( A(X_t) - A) \langle \tilde \theta \rangle_t \; | \; \mathcal{F}_{t-\tau(\alpha_{t})}]| \nonumber\\
    %&\;\;\; +   |\mathbf{E}[ ( \langle \tilde \theta \rangle_t - \theta^* )^\top (P+P^\top)( \frac{1}{N} B(X_t)^\top\1_N - b) \; | \; \mathcal{F}_{t-\tau(\alpha_{t})} ]| \nonumber \\
    & \le |\mathbf{E}[ ( \langle \tilde \theta \rangle_{t-\tau(\alpha_{t})} - \theta^* )^\top (P+P^\top)( A(X_t) - A) \langle \tilde \theta \rangle_{t-\tau(\alpha_{t})} \; | \; \mathcal{F}_{t-\tau(\alpha_{t})} ]| \label{eq:timevarying_bound_Ab_Push_SA1} \\
    & \;\;\; + |\mathbf{E}[ ( \langle \tilde \theta \rangle_{t-\tau(\alpha_{t})} - \theta^* )^\top (P+P^\top)( A(X_t) - A) ( \langle \tilde \theta \rangle_t - \langle \tilde \theta \rangle_{t-\tau(\alpha_{t})} ) \; | \; \mathcal{F}_{t-\tau(\alpha_{t})} ]| \label{eq:timevarying_bound_Ab_Push_SA2} \\
    & \;\;\; + |\mathbf{E}[  ( \langle \tilde \theta \rangle_t - \langle \tilde \theta \rangle_{t-\tau(\alpha_{t})} )^\top (P+P^\top)( A(X_t) - A) \langle \tilde \theta \rangle_{t-\tau(\alpha_{t})} \; | \; \mathcal{F}_{t-\tau(\alpha_{t})} ]| \label{eq:timevarying_bound_Ab_Push_SA3}\\
    & \;\;\; + |\mathbf{E}[  ( \langle \tilde \theta \rangle_t - \langle \tilde \theta \rangle_{t-\tau(\alpha_{t})} )^\top (P+P^\top)( A(X_t) - A)  ( \langle \tilde \theta \rangle_t - \langle \tilde \theta \rangle_{t-\tau(\alpha_{t})} ) \; | \; \mathcal{F}_{t-\tau(\alpha_{t})} ]| \label{eq:timevarying_bound_Ab_Push_SA4}\\
    &\;\;\; +   |\mathbf{E}[ ( \langle \tilde \theta \rangle_t - \langle \tilde \theta \rangle_{t-\tau(\alpha_{t})} )^\top (P+P^\top)(\frac{1}{N} B(X_t)^\top\1_N - b) \; | \; \mathcal{F}_{t-\tau(\alpha_{t})} ]|  \label{eq:timevarying_bound_Ab_Push_SA5}\\
    &\;\;\; +   |\mathbf{E}[ ( \langle \tilde \theta \rangle_{t-\tau(\alpha_{t})} - \theta^* )^\top (P+P^\top)(\frac{1}{N} B(X_t)^\top\1_N - b) \; | \; \mathcal{F}_{t-\tau(\alpha_{t})} ]|\label{eq:timevarying_bound_Ab_Push_SA6}.
\end{align}
Using the mixing time in Assumption~\ref{assum:mixing-time}, we can get the bound for \eqref{eq:timevarying_bound_Ab_Push_SA1} and \eqref{eq:timevarying_bound_Ab_Push_SA6} for all $t\ge \bar T$:
\begin{align}
    & \;\;\; |\mathbf{E}[ ( \langle \tilde \theta \rangle_{t-\tau(\alpha_{t})} - \theta^* )^\top (P+P^\top)( A(X_t) - A) \langle \tilde \theta \rangle_{t-\tau(\alpha_{t})} \; | \; \mathcal{F}_{t-\tau(\alpha_{t})} ]|\nonumber\\
    & \le |( \langle \tilde \theta \rangle_{t-\tau(\alpha_{t})} - \theta^* )^\top (P+P^\top) \mathbf{E}[A(X_t) - A \; | \; \mathcal{F}_{t-\tau(\alpha_{t})} ] \langle \tilde \theta \rangle_{t-\tau(\alpha_{t})} | \nonumber\\
    & \le 2 \alpha_{t} \gamma_{\max}   \mathbf{E}[\| \langle \tilde \theta \rangle_{t-\tau(\alpha_{t})}  - \theta^* \|_2  \| \langle \tilde \theta \rangle_{t-\tau(\alpha_{t})}\|_2 \; | \; \mathcal{F}_{t-\tau(\alpha_{t})} ] \nonumber\\
    & \le \alpha_{t} \gamma_{\max}   \mathbf{E}[\| \langle \tilde \theta  \rangle_{t-\tau(\alpha_{t})}  - \theta^* \|_2^2 +  \| \langle \tilde \theta \rangle_{t-\tau(\alpha_{t})} \|_2^2 \; | \; \mathcal{F}_{t-\tau(\alpha_{t})} ] \nonumber \\
    & \le \alpha_{t} \gamma_{\max}   \mathbf{E}[ 2 \|\theta^* \|_2^2 + 3 \| \langle \tilde \theta \rangle_{t-\tau(\alpha_{t})}\|_2^2 \; | \; \mathcal{F}_{t-\tau(\alpha_{t})} ] \nonumber \\
    & \le 6 \alpha_{t} \gamma_{\max} \mathbf{E}[ \| \langle \tilde \theta \rangle_{t} - \langle \tilde \theta \rangle_{t-\tau(\alpha_{t})}\|_2^2 \; | \; \mathcal{F}_{t-\tau(\alpha_{t})} ] + 6 \alpha_{t} \gamma_{\max} \mathbf{E}[ \| \langle \tilde \theta \rangle_{t} \|_2^2 \; | \; \mathcal{F}_{t-\tau(\alpha_{t})} ] + 2 \alpha_{t} \gamma_{\max} \|\theta^* \|_2^2 \nonumber \\
    & \le 54 \alpha_{t} \gamma_{\max} \mathbf{E}[ \| \langle \tilde \theta  \rangle_{t} \|_2^2 \; | \; \mathcal{F}_{t-\tau(\alpha_{t})} ] + 36 \alpha_{t} \gamma_{\max} \frac{(b_{\max} + \mu_{\max})^2 }{A_{\max}^2} + 2 \alpha_{t} \gamma_{\max} \|\theta^* \|_2^2,  \label{eq:timevarying_bound_Ab_Push_SA1_bounded}
\end{align}
where in the last inequality, we use \eqref{eq:timevarying_single_3_Push_SA_1} from Lemma~\ref{lemma:timevarying_single_3_Push_SA}.
\begin{align}
    & \;\;\; |\mathbf{E}[ ( \langle \tilde \theta \rangle_{t-\tau(\alpha_{t})}  - \theta^* )^\top (P+P^\top)(\frac{1}{N}B(X_t)^\top\1_N - b) \; | \; \mathcal{F}_{t-\tau(\alpha_{t})} ]|\nonumber\\
    & \le | ( \langle \tilde \theta \rangle_{t-\tau(\alpha_{t})}  - \theta^* )^\top (P+P^\top) \frac{1}{N} \sum_{i=1}^N \mathbf{E}[ b^i(X_t) - b^i \; | \; \mathcal{F}_{t-\tau(\alpha_{t})} ] |\nonumber\\
    & \le 2 \gamma_{\max} \alpha_{t} \mathbf{E}[ \| \langle \tilde \theta \rangle_{t-\tau(\alpha_{t})} - \theta^* \|_2 \; | \; \mathcal{F}_{t-\tau(\alpha_{t})} ] \le 2 \gamma_{\max} \alpha_{t}  \left( 1 + \frac{1}{2} \mathbf{E}[ \| \langle \tilde \theta \rangle_{t-\tau(\alpha_{t})} \|_2^2 \; | \; \mathcal{F}_{t-\tau(\alpha_{t})} ] + \frac{1}{2} \| \theta^* \|_2^2 \right) \nonumber\\
    %& \le 2 \gamma_{\max} \alpha_{t}  \left( 1 +  \mathbf{E}[ \| \langle \tilde \theta \rangle_{t-\tau(\alpha_{t})}  - \langle \tilde \theta \rangle_{t} \|_2^2 \; | \; \mathcal{F}_{t-\tau(\alpha_{t})} ] + \mathbf{E}[ \| \langle \tilde  \theta \rangle_{t} \|_2^2 \; | \; \mathcal{F}_{t-\tau(\alpha_{t})} ] + \| \theta^* \|_2^2 \right) \nonumber\\
    & \le 2 \gamma_{\max} \alpha_{t} \left( 1 +  9 \mathbf{E}[ \| \langle \tilde \theta \rangle_{t} \|_2^2 \; | \; \mathcal{F}_{t-\tau(\alpha_{t})} ] + 6 \frac{(b_{\max} +\mu_{\max})^2}{A_{\max}^2}+ \| \theta^* \|_2^2 \right), 
    \label{eq:timevarying_bound_Ab_Push_SA6_bounded}
\end{align}  
where in the last inequality we use \eqref{eq:timevarying_single_3_Push_SA_1}.
Next, using Assumption~\ref{assum:A and b}, \eqref{eq:timevarying_single_3_Push_SA_1} and \eqref{eq:timevarying_single_3_Push_SA_3}, we have
\begin{align}
    & \;\;\; |\mathbf{E}[ ( \langle \tilde \theta \rangle_{t-\tau(\alpha_{t})} - \theta^* )^\top (P+P^\top)( A(X_t) - A) ( \langle \tilde \theta \rangle_t - \langle \tilde \theta \rangle_{t-\tau(\alpha_{t})} ) \; | \; \mathcal{F}_{t-\tau(\alpha_{t})} ]| \nonumber\\
    &\le 4 \gamma_{\max} A_{\max} \mathbf{E}[ \| \langle \tilde \theta \rangle_{t-\tau(\alpha_{t})} - \theta^* \|_2 \| \langle \tilde \theta \rangle_t - \langle \tilde \theta \rangle_{t-\tau(\alpha_{t})}\|_2 \; | \; \mathcal{F}_{t-\tau(\alpha_{t})} ] \nonumber\\
    &\le 4 \gamma_{\max} A_{\max} \left[ \mathbf{E}[ \| \langle \tilde \theta \rangle_{t-\tau(\alpha_{t})} \|_2 \| \langle \tilde \theta \rangle_t - \langle \tilde \theta \rangle_{t-\tau(\alpha_{t})}\|_2   +  \| \theta^* \|_2   \| \langle \tilde \theta \rangle_t - \langle \tilde \theta \rangle_{t-\tau(\alpha_{t})}\|_2 \; | \; \mathcal{F}_{t-\tau(\alpha_{t})} ] \right] \nonumber\\
    &\le \sum_{k=t-\tau(\alpha_t)}^{t-1} \alpha_k \gamma_{\max} \Big[ 8  A_{\max}^2 \mathbf{E}[ \| \langle \tilde \theta \rangle_{t-\tau(\alpha_{t})} \|_2^2  \; | \; \mathcal{F}_{t-\tau(\alpha_{t})} ] + 8  A_{\max} (b_{\max} + \mu_{\max})\| \theta^* \|_2 \nonumber\\
    & \;\;\; + 8  A_{\max}^2 \left(\frac{ b_{\max} + \mu_{\max}}{A_{\max}} + \| \theta^* \|_2 \right) \mathbf{E}[ \| \langle \tilde \theta \rangle_{t-\tau(\alpha_{t})} \|_2  \; | \; \mathcal{F}_{t-\tau(\alpha_{t})} ]  \Big] \nonumber\\
    &\le \sum_{k=t-\tau(\alpha_t)}^{t-1} \alpha_k \gamma_{\max} \left[ 12 A_{\max}^2 \mathbf{E}[ \| \langle \tilde \theta \rangle_{t-\tau(\alpha_{t})} \|_2^2  \; | \; \mathcal{F}_{t-\tau(\alpha_{t})} ]  + 8  A_{\max}^2 \left(\frac{ b_{\max} + \mu_{\max}}{A_{\max}} + \| \theta^* \|_2 \right)^2 \right]
    \nonumber\\
    &\le \sum_{k=t-\tau(\alpha_t)}^{t-1} \alpha_k \gamma_{\max} \Bigg[ 24 A_{\max}^2 \mathbf{E}[ \| \langle \tilde \theta \rangle_{t} - \langle \tilde \theta \rangle_{t-\tau(\alpha_{t})} \|_2^2  \; | \; \mathcal{F}_{t-\tau(\alpha_{t})} ] + 24 A_{\max}^2 \mathbf{E}[ \| \langle \tilde \theta \rangle_{t}\|_2^2  \; | \; \mathcal{F}_{t-\tau(\alpha_{t})} ] 
    \nonumber\\
    & \;\;\; + 8  A_{\max}^2 \left(\frac{ b_{\max} + \mu_{\max} }{A_{\max}} + \| \theta^* \|_2 \right)^2  \Bigg]\nonumber\\
    &\le \sum_{k=t-\tau(\alpha_t)}^{t-1} \alpha_k \left[
    216 \gamma_{\max} A_{\max}^2 \mathbf{E}[ \| \langle \tilde \theta \rangle_{t}\|_2^2  \; | \; \mathcal{F}_{t-\tau(\alpha_{t})} ]   + 152 \gamma_{\max}  \left(b_{\max} + \mu_{\max} + A_{\max} \| \theta^* \|_2 \right)^2 \right].
    \label{eq:timevarying_bound_Ab_Push_SA2_bounded} 
\end{align}

In additional, for the bound of \eqref{eq:timevarying_bound_Ab_Push_SA3}, using \eqref{eq:timevarying_single_3_Push_SA_1} and \eqref{eq:timevarying_single_3_Push_SA_3}, we have
\begin{align}
    & \;\;\; |\mathbf{E}[ ( \langle \tilde \theta \rangle_t -\langle \tilde \theta \rangle_{t-\tau(\alpha_{t})} )^\top (P+P^\top)( A(X_t) - A) \langle \tilde \theta \rangle_{t-\tau(\alpha_{t})} \; | \; \mathcal{F}_{t-\tau(\alpha_{t})} ]|\nonumber\\
    & \le 4 \gamma_{\max} A_{\max} \mathbf{E}[ \| \langle \tilde \theta \rangle_t - \langle \tilde \theta \rangle_{t-\tau(\alpha_{t})}\|_2 \| \langle \tilde \theta \rangle_{t-\tau(\alpha_{t})}\|_2 \; | \; \mathcal{F}_{t-\tau(\alpha_{t})} ]\nonumber\\
    & \le 8 \gamma_{\max} A_{\max} \mathbf{E}[ A_{\max } \|  \langle \tilde \theta \rangle_{t-\tau(\alpha_{t})} \|_2^2 +  (b_{\max} + \mu_{\max} ) \| \langle \tilde \theta \rangle_{t-\tau(\alpha_{t})}\|_2 \; | \; \mathcal{F}_{t-\tau(\alpha_{t})} ] \sum_{k=t-\tau(\alpha_t)}^{t-1} \alpha_k \nonumber\\
    & \le 4  \gamma_{\max} A_{\max} \sum_{k=t-\tau(\alpha_t)}^{t-1} \alpha_k
    \left[ (2 A_{\max }+  b_{\max} + \mu_{\max} ) \mathbf{E}[  \| \langle \tilde \theta \rangle_{t-\tau(\alpha_{t})} \|_2^2 \; | \; \mathcal{F}_{t-\tau(\alpha_{t})} ] + b_{\max} + \mu_{\max}\right] \nonumber\\
    & \le 4  \gamma_{\max} A_{\max} \sum_{k=t-\tau(\alpha_t)}^{t-1} \alpha_k \Big[
    2 (2 A_{\max }+  b_{\max} + \mu_{\max}) \mathbf{E}[  \| \langle \tilde \theta \rangle_{t} -\langle \tilde \theta \rangle_{t-\tau(\alpha_{t})} \|_2^2 \; | \; \mathcal{F}_{t-\tau(\alpha_{t})} ] +  b_{\max} + \mu_{\max} \nonumber\\
    &\;\;\; + 2 (2 A_{\max }+  b_{\max} + \mu_{\max}) \mathbf{E}[  \| \langle \tilde \theta \rangle_{t} \|_2^2 \; | \; \mathcal{F}_{t-\tau(\alpha_{t})} ] \Big] \nonumber\\
    & \le 72 \ \gamma_{\max} A_{\max} (2 A_{\max }  +  b_{\max} + \mu_{\max}) \mathbf{E}[  \| \langle \tilde \theta \rangle_{t} \|_2^2 \; | \; \mathcal{F}_{t-\tau(\alpha_{t})} ]  \sum_{k=t-\tau(\alpha_t)}^{t-1} \alpha_k \nonumber\\
    & \;\;\;+ 48  \gamma_{\max} A_{\max} (b_{\max}+ \mu_{\max}) (\frac{b_{\max} + \mu_{\max}}{A_{\max}} + 1 )^2 \sum_{k=t-\tau(\alpha_t)}^{t-1} \alpha_k.
    \label{eq:timevarying_bound_Ab_Push_SA3_bounded}
\end{align}
Moreover, using \eqref{eq:timevarying_single_3_Push_SA_3}, we can get the bound for \eqref{eq:timevarying_bound_Ab_Push_SA4} as follows:
\begin{align}
    & \;\;\; |\mathbf{E}[  ( \langle \tilde \theta \rangle_t - \langle \tilde \theta \rangle_{t-\tau(\alpha_{t})} )^\top (P+P^\top)( A(X_t) - A)  ( \langle \tilde \theta \rangle_t - \langle \tilde \theta \rangle_{t-\tau(\alpha_{t})} ) \; | \; \mathcal{F}_{t-\tau(\alpha_{t})} ]| \nonumber\\
    & \le 4 \gamma_{\max} A_{\max}  \mathbf{E}[ \| \langle \tilde \theta \rangle_t - \langle \tilde \theta \rangle_{t-\tau(\alpha_{t})}\|_2^2 \; | \; \mathcal{F}_{t-\tau(\alpha_{t})} ]| \nonumber\\
    & \le 4 \gamma_{\max} A_{\max}  \mathbf{E}[ 72 A_{\max}^2 \| \langle \tilde \theta \rangle_{t} \|_2^2 +  50 (b_{\max}+\mu_{\max})^2 \; | \; \mathcal{F}_{t-\tau(\alpha_{t})} ] \left( \sum_{k=t-\tau(\alpha_t)}^{t-1} \alpha_k \right)^2 \nonumber\\
    & \le   96 A_{\max}^2 \gamma_{\max} \mathbf{E}[ \| \langle \tilde \theta \rangle_{t} \|_2^2 \; | \; \mathcal{F}_{t-\tau(\alpha_{t})} ] \sum_{k=t-\tau(\alpha_t)}^{t-1} \alpha_k + 67 (b_{\max}+\mu_{\max})^2 \gamma_{\max} \sum_{k=t-\tau(\alpha_t)}^{t-1} \alpha_k.
    \label{eq:timevarying_bound_Ab_Push_SA4_bounded}
\end{align}
Finally, we can get the bound of \eqref{eq:timevarying_bound_Ab_Push_SA5} using \eqref{eq:timevarying_single_3_Push_SA_2}: 
\begin{align}
    &\;\;\; |\mathbf{E}[ ( \langle \tilde \theta \rangle_t - \langle \tilde \theta \rangle_{t-\tau(\alpha_{t})} ) (P+P^\top)(\frac{1}{N} B(X_t)^\top\1_N - b) \; | \; \mathcal{F}_{t-\tau(\alpha_{t})} ]| \nonumber \\
    & \le 4  \gamma_{\max} b_{\max} \mathbf{E}[ \| \langle \tilde \theta \rangle_t - \langle \tilde \theta \rangle_{t-\tau(\alpha_{t})}\|_2 \; | \; \mathcal{F}_{t-\tau(\alpha_{t})} ] \nonumber \\
    & \le 4  \gamma_{\max} b_{\max} \mathbf{E}[ 6 A_{\max} \| \langle \tilde \theta \rangle_{t} \|_2 +  5 ( b_{\max} + \mu_{\max} ) \; | \; \mathcal{F}_{t-\tau(\alpha_{t})} ] \sum_{k=t-\tau(\alpha_t)}^{t-1} \alpha_k \nonumber \\
    & \le \sum_{k=t-\tau(\alpha_t)}^{t-1} \alpha_k \gamma_{\max}b_{\max} \left( 12  A_{\max}  \mathbf{E}[\| \langle \tilde \theta \rangle_{t} \|_2^2  \; | \; \mathcal{F}_{t-\tau(\alpha_{t})} ]  +   12 A_{\max}   + 20 b_{\max} + 20 \mu_{\max}  \right).  
    \label{eq:timevarying_bound_Ab_Push_SA5_bounded}
\end{align}
Then, using \eqref{eq:timevarying_bound_Ab_Push_SA1_bounded}--\eqref{eq:timevarying_bound_Ab_Push_SA5_bounded}, we have
\begin{align*}
    & \;\;\;\; |\mathbf{E}[ ( \langle \tilde \theta \rangle_t  - \theta^* )^\top (P+P^\top)( A(X_t) \langle \tilde \theta \rangle_t +   B(X_t)^\top\pi_{t+1} - A \langle \tilde \theta \rangle_t - b) \; | \; \mathcal{F}_{t-\tau(\alpha_{t})} ]| \nonumber \\
    & \le  54 \alpha_{t} \gamma_{\max} \mathbf{E}[ \| \langle \tilde \theta  \rangle_{t} \|_2^2 \; | \; \mathcal{F}_{t-\tau(\alpha_{t})} ] + 36 \alpha_{t} \gamma_{\max} \frac{(b_{\max} + \mu_{\max})^2 }{A_{\max}^2} + 2 \alpha_{t} \gamma_{\max} \|\theta^* \|_2^2 \\
    & \;\;\; +  2 \gamma_{\max} \alpha_{t} \left( 1 +  9 \mathbf{E}[ \| \langle \tilde \theta \rangle_{t} \|_2^2 \; | \; \mathcal{F}_{t-\tau(\alpha_{t})} ] + 6 \frac{(b_{\max} +\mu_{\max})^2}{A_{\max}^2}+ \| \theta^* \|_2^2 \right) \\
    %& \;\;\; + 216 \gamma_{\max} A_{\max}^2 \mathbf{E}[ \| \langle \tilde \theta \rangle_{t}\|_2^2  \; | \; \mathcal{F}_{t-\tau(\alpha_{t})} ] \sum_{k=t-\tau(\alpha_t)}^{t-1} \alpha_k  \nonumber\\
    %& \;\;\; + 152 \gamma_{\max}  \left(b_{\max} + \mu_{\max} + A_{\max} \| \theta^* \|_2 \right)^2 \sum_{k=t-\tau(\alpha_t)}^{t-1} \alpha_k\\
    %& \;\;\; + 72 \ \gamma_{\max} A_{\max} (2 A_{\max }  +  b_{\max} + \mu_{\max}) \mathbf{E}[  \| \langle \tilde \theta \rangle_{t} \|_2^2 \; | \; \mathcal{F}_{t-\tau(\alpha_{t})} ]  \sum_{k=t-\tau(\alpha_t)}^{t-1} \alpha_k \nonumber\\
    %& \;\;\;+ 48  \gamma_{\max} A_{\max} (b_{\max}+ \mu_{\max}) (\frac{b_{\max} + \mu_{\max}}{A_{\max}} + 1 )^2 \sum_{k=t-\tau(\alpha_t)}^{t-1} \alpha_k \\
    %& \;\;\; + 96 A_{\max}^2 \gamma_{\max} \mathbf{E}[ \| \langle \tilde \theta \rangle_{t} \|_2^2 \; | \; \mathcal{F}_{t-\tau(\alpha_{t})} ] \sum_{k=t-\tau(\alpha_t)}^{t-1} \alpha_k + 67 (b_{\max}+\mu_{\max})^2 \gamma_{\max} \sum_{k=t-\tau(\alpha_t)}^{t-1} \alpha_k \\
    %& \;\;\; + 12 \gamma_{\max} A_{\max}  b_{\max} \mathbf{E}[\| \langle \tilde \theta \rangle_{t} \|_2^2  \; | \; \mathcal{F}_{t-\tau(\alpha_{t})} ]\sum_{k=t-\tau(\alpha_t)}^{t-1} \alpha_k  \nonumber\\
    %& \;\;\; +   ( 12 A_{\max}   + 20 b_{\max} + 20 \mu_{\max} )\gamma_{\max} b_{\max} \sum_{k=t-\tau(\alpha_t)}^{t-1} \alpha_k, \\
    %& \\
    & \;\;\; + \sum_{k=t-\tau(\alpha_t)}^{t-1} \alpha_k \Big[ 216 \gamma_{\max} A_{\max}^2 \mathbf{E}[ \| \langle \tilde \theta \rangle_{t}\|_2^2  \; | \; \mathcal{F}_{t-\tau(\alpha_{t})} ] + 152 \gamma_{\max}  \left(b_{\max} + \mu_{\max} + A_{\max} \| \theta^* \|_2 \right)^2   \nonumber\\
    & \;\;\; + 72 \ \gamma_{\max} A_{\max} (2 A_{\max }  +  b_{\max} + \mu_{\max}) \mathbf{E}[  \| \langle \tilde \theta \rangle_{t} \|_2^2 \; | \; \mathcal{F}_{t-\tau(\alpha_{t})} ] + 96 A_{\max}^2 \gamma_{\max} \mathbf{E}[ \| \langle \tilde \theta \rangle_{t} \|_2^2 \; | \; \mathcal{F}_{t-\tau(\alpha_{t})} ] \nonumber\\
    & \;\;\;  + 48  \gamma_{\max} A_{\max} (b_{\max}+ \mu_{\max}) (\frac{b_{\max} + \mu_{\max}}{A_{\max}} + 1 )^2 + 67 (b_{\max}+\mu_{\max})^2 \gamma_{\max}   \\
    & \;\;\; + 12 \gamma_{\max} A_{\max}  b_{\max} \mathbf{E}[\| \langle \tilde \theta \rangle_{t} \|_2^2  \; | \; \mathcal{F}_{t-\tau(\alpha_{t})} ]  +   ( 12 A_{\max}   + 20 b_{\max} + 20 \mu_{\max} )\gamma_{\max} b_{\max} \Big] \\%, 
%\end{align*}
%which implies that
%\begin{align*}
%    & \;\;\;\; |\mathbf{E}[ ( \langle \tilde \theta \rangle_t  - \theta^* )^\top (P+P^\top)( A(X_t) \langle \tilde \theta \rangle_t +   B(X_t)^\top\pi_{t+1} - A \langle \tilde \theta \rangle_t - b) \; | \; \mathcal{F}_{t-\tau(\alpha_{t})} ]| \nonumber \\
    & \le \alpha_{t-\tau(\alpha_t)} \tau(\alpha_t)  \gamma_{\max} \left( 72 + 456 A_{\max}^2  + 84  A_{\max}  b_{\max} + 72 A_{\max} \mu_{\max} \right) \mathbf{E}[ \| \langle \tilde \theta \rangle_{t} \|_2^2 \; | \; \mathcal{F}_{t-\tau(\alpha_t)} ] \nonumber \\
    &\;\;\; + \alpha_{t -\tau(\alpha_t) }  \tau(\alpha_t) \gamma_{\max} \bigg[ 2 + 4 \|\theta^* \|_2^2 +  48\frac{(b_{\max}+ \mu_{\max})^2}{A_{\max}^2} +  152  \left(b_{\max} + \mu_{\max} + A_{\max} \| \theta^* \|_2 \right)^2  \nonumber\\
    &\;\;\; +  12  A_{\max}b_{\max} + 48  A_{\max}(b_{\max}+ \mu_{\max}) (\frac{b_{\max} + \mu_{\max}}{A_{\max}} + 1 )^2 +  87 (b_{\max}+ \mu_{\max})^2  \bigg], 
\end{align*}
where we use $\alpha_t \le \alpha_{t-\tau{\alpha_t}}$ from Assumption~\ref{assum:step-size} and $\tau(\alpha_t) \ge 1$ in the last inequality.
This completes the proof.
\hfill $\qed$

\begin{lemma} \label{lemma:bound_average_time-varying_push_SA}
    Suppose that Assumptions~\ref{assum:A and b}--\ref{assum:lyapunov} hold and $\alpha_t = \frac{\alpha_0}{t+1}$. When $ \mu_{t} + \tau(\alpha_t) \alpha_{t-\tau(\alpha_t)} \zeta_8 \le \frac{0.1}{\gamma_{\max}}$ and $\tau(\alpha_t) \alpha_{t-\tau(\alpha_t)} \le \min \{ \frac{ \log2}{A_{\max}},\; \frac{0.1}{\zeta_8 \gamma_{\max}} \}$, we have for $t\ge \bar T$,
%\begin{align*}
%    \mathbf{E}\left[\| \langle \theta \rangle_{t} -\theta^* \|_2^2 \right] 
%    & \le \frac{T_2^*}{t} \frac{\gamma_{\max}}{\gamma_{\min}} \mathbf{E}\left[\|\langle \theta \rangle_{T_2^*} -\theta^* \|_2^2 \right] 
%    +  \frac{\zeta_7 \alpha_0  C \log^2(\frac{t}{\alpha_0})}{t} \frac{\gamma_{\max}}{\gamma_{\min}} +  \alpha_0 \zeta_4 \frac{\gamma_{\max}}{\gamma_{\min}}  \frac{\sum_{l = T_2^*}^{t} \mu_{l}}{t},
%\end{align*}
\begin{align*}
    \mathbf{E}[\|\langle \tilde \theta \rangle_{t+1} -\theta^* \|_2^2 ] 
    & \le \frac{\bar T}{t+1} \frac{\gamma_{\max}}{\gamma_{\min}} \mathbf{E}[\|\langle \tilde \theta \rangle_{\bar T} -\theta^* \|_2^2 ] 
    +  \frac{\zeta_9 \alpha_0  C \log^2(\frac{t+1}{\alpha_0})}{t+1} \frac{\gamma_{\max}}{\gamma_{\min}}
     +  \alpha_0 \frac{\gamma_{\max}}{\gamma_{\min}} \frac{\sum_{l = \bar T}^{t+1}  \mu_{l}  }{t+1},
\end{align*}
where $\bar T$ is defined in Appendix~\ref{sec:thmPush_constant}, $\zeta_8$ and $\zeta_9$ are defined in \eqref{eq:definition_Psi12} and \eqref{eq:definition_Psi13}, respectively.
\end{lemma}

\vspace{.1in}

%\noindent
{\bf Proof of Lemma~\ref{lemma:bound_average_time-varying_push_SA}:}
Let $H(\langle \tilde \theta \rangle_t ) = ( \langle \tilde \theta \rangle_t - \theta^* )^\top P ( \langle \tilde \theta \rangle_t  - \theta^* ) $. From Assumption~\ref{assum:lyapunov}, we know that
$
    \gamma_{\min} \| \langle \tilde \theta \rangle_t - \theta^* \|_2^2 \le H(\langle \tilde \theta \rangle_t ) \le \gamma_{\max} \| \langle \tilde \theta \rangle_t - \theta^* \|_2^2.
$
%Recall the update of $\langle \tilde \theta \rangle_t $ in \eqref{eq:update_average_tilde_theta}:
%\begin{align*} 
%    \langle \tilde \theta \rangle_{t+1}
%    &= \langle \tilde \theta \rangle_t + \alpha_t A(X_t) \langle \tilde \theta \rangle_t + \frac{\alpha_t}{N}\sum_{i=1}^N b^i(X_t)  +   \alpha_t \rho_t.
%\end{align*}
From Assumption~\ref{assum:A and b} and \eqref{eq:update_average_tilde_theta}, for $t\ge \bar T$ we have 
\begin{align}
    &\;\;\;\; H( \langle \tilde \theta \rangle_{t+1} ) 
     = ( \langle \tilde \theta \rangle_{t+1} - \theta^* )^\top P ( \langle \tilde \theta \rangle_{t+1} - \theta^* ) \nonumber\\
    %& = \left( \langle \tilde \theta \rangle_{t} + \alpha_t A(X_t) \langle \tilde \theta \rangle_{t} +  \frac{\alpha_t}{N}\sum_{i=1}^N b^i(X_t) +   \alpha_t \rho_t - \theta^* \right)^\top P \cdot \nonumber\\
    %&\;\;\;\; \left(\langle \tilde \theta \rangle_{t} + \alpha_t A(X_t) \langle \tilde \theta \rangle_{t} +  \frac{\alpha_t}{N}\sum_{i=1}^N b^i(X_t) +   \alpha_t \rho_t - \theta^* \right) \nonumber\\
    & = ( \langle \tilde \theta \rangle_t  - \theta^* )^\top P (\langle \tilde \theta \rangle_t - \theta^* ) + \alpha_t^2 ( A(X_t) \langle \tilde \theta \rangle_t )^\top P ( A(X_t) \langle \tilde \theta \rangle_t )  \nonumber \\
    & \;\;\;\; + \frac{\alpha_t^2}{N^2}  (B(X_t)^\top \1_N)^\top P (B(X_t)^\top\1_N) + \frac{ \alpha_t^2}{N} ( A(X_t) \langle \tilde \theta \rangle_t )^\top (P+P^\top)(B(X_t)^\top\1_N) + \alpha_t^2 \rho_t^\top P \rho_t\nonumber\\
    & \;\;\;\; + \alpha_t^2  ( A(X_t) \langle \tilde \theta \rangle_t + \frac{1}{N}B(X_t)^\top\1_N )^\top (P+P^\top)\rho_t + \alpha_t  ( \langle \tilde \theta \rangle_t  - \theta^* )^\top (P+P^\top) \rho_t \nonumber\\
    & \;\;\;\; + \alpha_t  ( \langle \tilde \theta \rangle_t  - \theta^* )^\top (P+P^\top)( A(X_t) \langle \tilde \theta \rangle_t +   \frac{1}{N}B(X_t)^\top\1_N - A\langle \tilde \theta  \rangle_t - b) \nonumber\\
    & \;\;\;\; + \alpha_t  ( \langle \tilde \theta \rangle_t - \theta^* )^\top P( A\langle \tilde \theta \rangle_t + b) + \alpha_t  ( A\langle \tilde \theta \rangle_t + b)^\top P( \langle \tilde \theta \rangle_t  - \theta^* ),  \nonumber%\\
    %& = H( \langle \tilde \theta \rangle_t )+ \alpha_t^2 ( A(X_t) \langle \tilde \theta \rangle_t )^\top P ( A(X_t) \langle \tilde \theta \rangle_t ) + \frac{\alpha_t^2}{N^2}  (B(X_t)^\top \1_N)^\top P (B(X_t)^\top\1_N) \nonumber \\
    %& \;\;\;\;  + \frac{ \alpha_t^2}{N} ( A(X_t) \langle \tilde \theta \rangle_t )^\top (P+P^\top)(B(X_t)^\top\1_N) + \alpha_t^2 \rho_t^\top P \rho_t\nonumber\\
    %& \;\;\;\; + \alpha_t^2  ( A(X_t) \langle \tilde \theta \rangle_t + \frac{1}{N}B(X_t)^\top\1_N )^\top (P+P^\top)\rho_t + \alpha_t  ( \langle \tilde \theta \rangle_t  - \theta^* )^\top (P+P^\top)\rho_t \nonumber\\
    %& \;\;\;\; + \alpha_t  ( \langle \tilde \theta \rangle_t  - \theta^* )^\top (P+P^\top)( A(X_t) \langle \tilde \theta \rangle_t +   \frac{1}{N}B(X_t)^\top\1_N - A\langle \tilde \theta \rangle_t  - b) \nonumber\\
    %& \;\;\;\; + \alpha_t  ( \langle \tilde \theta \rangle_t - \theta^* )^\top (PA+A^\top P ) (\langle \tilde \theta \rangle_t -\theta^*) \label{eq:push_SA_proof_1},
\end{align}
which implies that
\begin{align}
    H( \langle \tilde \theta \rangle_{t+1} ) 
    & = H( \langle \tilde \theta \rangle_t )+ \alpha_t^2 ( A(X_t) \langle \tilde \theta \rangle_t )^\top P ( A(X_t) \langle \tilde \theta \rangle_t ) + \frac{\alpha_t^2}{N^2}  (B(X_t)^\top \1_N)^\top P (B(X_t)^\top\1_N) \nonumber \\
    & \;\;\;\;  + \frac{ \alpha_t^2}{N} ( A(X_t) \langle \tilde \theta \rangle_t )^\top (P+P^\top)(B(X_t)^\top\1_N) + \alpha_t^2 \rho_t^\top P \rho_t\nonumber\\
    & \;\;\;\; + \alpha_t^2  ( A(X_t) \langle \tilde \theta \rangle_t + \frac{1}{N}B(X_t)^\top\1_N )^\top (P+P^\top)\rho_t + \alpha_t  ( \langle \tilde \theta \rangle_t  - \theta^* )^\top (P+P^\top)\rho_t \nonumber\\
    & \;\;\;\; + \alpha_t  ( \langle \tilde \theta \rangle_t  - \theta^* )^\top (P+P^\top)( A(X_t) \langle \tilde \theta \rangle_t +   \frac{1}{N}B(X_t)^\top\1_N - A\langle \tilde \theta \rangle_t  - b) \nonumber\\
    & \;\;\;\; + \alpha_t  ( \langle \tilde \theta \rangle_t - \theta^* )^\top (PA+A^\top P ) (\langle \tilde \theta \rangle_t -\theta^*) \label{eq:push_SA_proof_1},
\end{align}
where we use the fact that $A\theta^* +b =0 $ on the last equality.

Next, we can take expectation on both sides of \eqref{eq:push_SA_proof_1}. From Assumption~\ref{assum:lyapunov} and Lemma~\ref{lemma:bound_timevarying_Ab_push_SA}, for \;\; $t\ge \bar T$ we have
\begin{align}
    \mathbf{E}[H( \langle \tilde \theta \rangle_{t+1} )] 
    & = \mathbf{E}[H( \langle \tilde \theta \rangle_t )] + \alpha_t^2 \mathbf{E}[( A(X_t) \langle \tilde \theta \rangle_t )^\top P ( A(X_t) \langle \tilde \theta \rangle_t )] - \alpha_t   \mathbf{E}[\| \langle \tilde \theta \rangle_t - \theta^* \|_2^2] + \mathbf{E}[\alpha_t^2 \rho_t^\top P \rho_t]\nonumber \\
    & \;\;\; + \frac{\alpha_t^2 }{N^2} \mathbf{E}[(B(X_t)^\top\1_N)^\top P (B(X_t)^\top\1_N)] + \frac{\alpha_t^2}{N} \mathbf{E}[( A(X_t) \langle \tilde \theta \rangle_t )^\top (P+P^\top)(B(X_t)^\top\1_N)]   \nonumber\\
    & \;\;\; + \alpha_t^2  \mathbf{E}[( A(X_t) \langle \tilde \theta \rangle_t + \frac{1}{N}B(X_t)^\top\1_N )^\top (P+P^\top) \rho_t] + \alpha_t  \mathbf{E}[( \langle \tilde \theta \rangle_t  - \theta^* )^\top (P+P^\top)\rho_t] \nonumber\\
    & \;\;\; + \alpha_t  \mathbf{E}[( \langle \tilde \theta \rangle_t   - \theta^* )^\top (P+P^\top)( A(X_t) \langle \tilde \theta \rangle_t   + \frac{1}{N}   B(X_t)^\top\1_N - A\langle \tilde \theta \rangle_t  - b)], \nonumber %\\
\end{align}
which implies that
\begin{align}
    &\;\;\; \mathbf{E}[H( \langle \tilde \theta \rangle_{t+1} )] \nonumber\\
    %& = \mathbf{E}[H( \langle \tilde \theta \rangle_t )] + \alpha_t^2 \mathbf{E}[( A(X_t) \langle \tilde \theta \rangle_t )^\top P ( A(X_t) \langle \tilde \theta \rangle_t )] - \alpha_t   \mathbf{E}[\| \langle \tilde \theta \rangle_t - \theta^* \|_2^2] + \mathbf{E}[\alpha_t^2 \rho_t^\top P \rho_t]\nonumber \\
    %& \;\;\; + \frac{\alpha_t^2 }{N^2} \mathbf{E}[(B(X_t)^\top\1_N)^\top P (B(X_t)^\top\1_N)] + \frac{\alpha_t^2}{N} \mathbf{E}[( A(X_t) \langle \tilde \theta \rangle_t )^\top (P+P^\top)(B(X_t)^\top\1_N)]   \nonumber\\
    %& \;\;\; + \alpha_t^2  \mathbf{E}[( A(X_t) \langle \tilde \theta \rangle_t + \frac{1}{N}B(X_t)^\top\1_N )^\top (P+P^\top) \rho_t] + \alpha_t  \mathbf{E}[( \langle \tilde \theta \rangle_t  - \theta^* )^\top (P+P^\top)\rho_t] \nonumber\\
    %& \;\;\; + \alpha_t  \mathbf{E}[( \langle \tilde \theta \rangle_t   - \theta^* )^\top (P+P^\top)( A(X_t) \langle \tilde \theta \rangle_t   + \frac{1}{N}   B(X_t)^\top\1_N - A\langle \tilde \theta \rangle_t  - b)] \nonumber \\
    & \le \mathbf{E}[H( \langle \tilde \theta \rangle_t )] + \alpha_t^2 A_{\max}^2 \gamma_{\max} \mathbf{E}[ \| \langle \tilde \theta \rangle_t \|_2^2 ] - \alpha_t   \mathbf{E}[\| \langle \tilde \theta \rangle_t - \theta^* \|_2^2] + 2 \alpha_t \gamma_{\max} \| \rho_{t} \|_2 \mathbf{E}[\|\langle \tilde \theta \rangle_t  - \theta^* \|_2]\nonumber \\
    & \;\;\; + \alpha_t^2 \gamma_{\max} (b_{\max}^2+\mu_{\max}^2) + 2 \alpha_t^2 \gamma_{\max} A_{\max} b_{\max} \mathbf{E}[ \|\langle \tilde \theta \rangle_t\|_2]  + 2 \alpha_t^2 \gamma_{\max} \mu_{\max} ( A_{\max} \mathbf{E}[\| \langle \tilde \theta \rangle_t \|_2] + b_{\max}) \nonumber\\
    & \;\;\; + \alpha_t \alpha_{t-\tau(\alpha_t)} \tau(\alpha_t)  \gamma_{\max} \left( 72 + 456 A_{\max}^2  + 84  A_{\max}  b_{\max} + 72 A_{\max} \mu_{\max} \right) \mathbf{E}[ \| \langle \tilde \theta \rangle_{t} \|_2^2  ] \nonumber \\
    & \;\;\; + \alpha_t  \alpha_{t -\tau(\alpha_t) }  \tau(\alpha_t) \gamma_{\max} \bigg[ 2  +  48\frac{(b_{\max}+ \mu_{\max})^2}{A_{\max}^2} +  152  \left(b_{\max} + \mu_{\max} + A_{\max} \| \theta^* \|_2 \right)^2  \nonumber\\
    &\;\;\; + 4 \|\theta^* \|_2^2 +  12  A_{\max}b_{\max} + 48  A_{\max}(b_{\max}+ \mu_{\max}) (\frac{b_{\max} + \mu_{\max}}{A_{\max}} + 1 )^2 +  87 (b_{\max}+ \mu_{\max})^2  \bigg] \nonumber\\
    & \le \mathbf{E}[H( \langle \tilde \theta \rangle_t )] +  ( - \alpha_t + \alpha_t \gamma_{\max} \| \rho_{t} \|_2 )  \mathbf{E}[\| \langle \tilde \theta \rangle_t- \theta^* \|_2^2] +  \alpha_t \gamma_{\max} \| \rho_{t} \|_2 \nonumber\\
    & \;\;\; + \alpha_t \alpha_{t-\tau(\alpha_t)} \tau(\alpha_t)  \gamma_{\max} \left( 72 + 458 A_{\max}^2  + 84  A_{\max}  b_{\max} + 72 A_{\max} \mu_{\max} \right) \mathbf{E}[ \| \langle \tilde \theta \rangle_{t} \|_2^2  ] \nonumber \\
    & \;\;\;\; + \alpha_t  \alpha_{t -\tau(\alpha_t) }  \tau(\alpha_t) \gamma_{\max} \bigg[ 2 +  48\frac{(b_{\max}+ \mu_{\max})^2}{A_{\max}^2} +  152  \left(b_{\max} + \mu_{\max} + A_{\max} \| \theta^* \|_2 \right)^2  \nonumber\\
    &\;\;\; + 4 \|\theta^* \|_2^2 +  12  A_{\max}b_{\max} + 48  A_{\max}(b_{\max}+ \mu_{\max}) (\frac{b_{\max} + \mu_{\max}}{A_{\max}} + 1 )^2 +  89 (b_{\max}+ \mu_{\max})^2  \bigg]. \nonumber
\end{align}
Using the facts that $ \mathbf{E}[ \| \langle \tilde \theta \rangle_{t} \|_2^2 ] \le 2 \mathbf{E}[ \| \langle \tilde \theta \rangle_{t}- \theta^*  \|_2^2 ] + 2 \| \theta^*\|_2^ 2 $ and $\gamma_{\min} \| \langle \tilde \theta \rangle_t - \theta^* \|_2^2 \le H(\langle \tilde \theta \rangle_t ) \le \gamma_{\max} \| \langle \tilde \theta \rangle_t - \theta^* \|_2^2$, then
\begin{align}
    &\;\;\;\; \mathbf{E}[H( \langle \tilde \theta \rangle_{t+1} )] \nonumber\\
    & \le \mathbf{E}[H( \langle \tilde \theta \rangle_t )] +  ( - \alpha_t + \alpha_t \gamma_{\max} \mu_{t}  )  \mathbf{E}[\| \langle \tilde \theta \rangle_t- \theta^* \|_2^2] +  \alpha_t \gamma_{\max}  \mu_{t}  + 2 \alpha_t^2 \gamma_{\max} (b_{\max} + \mu_{\max})^2 \nonumber\\
    & \;\;\; + 2 \alpha_t \alpha_{t-\tau(\alpha_t)} \tau(\alpha_t)  \gamma_{\max} \left( 72 + 458 A_{\max}^2  + 84  A_{\max}  b_{\max} + 72 A_{\max} \mu_{\max} \right) \mathbf{E}[ \| \langle \tilde \theta \rangle_{t}- \theta^*  \|_2^2 ] \nonumber \\
    & \;\;\; + 2 \alpha_t \alpha_{t-\tau(\alpha_t)} \tau(\alpha_t)  \gamma_{\max} \left( 72 + 458 A_{\max}^2  + 84  A_{\max}  b_{\max} + 72 A_{\max} \mu_{\max} \right) \| \theta^*\|_2^2  \nonumber \\
    & \;\;\; + \alpha_t  \alpha_{t -\tau(\alpha_t) }  \tau(\alpha_t) \gamma_{\max} \bigg[   48\frac{(b_{\max}+ \mu_{\max})^2}{A_{\max}^2} +  152  \left(b_{\max} + \mu_{\max} + A_{\max} \| \theta^* \|_2 \right)^2 + 4 \|\theta^* \|_2^2 \nonumber\\
    &\;\;\;  +2  +  12  A_{\max}b_{\max} + 48  A_{\max}(b_{\max}+ \mu_{\max}) (\frac{b_{\max} + \mu_{\max}}{A_{\max}} + 1 )^2 +  87 (b_{\max}+ \mu_{\max})^2  \bigg] \nonumber \\
    & \le \mathbf{E}[H( \langle \tilde \theta \rangle_t )] +  ( - \alpha_t + \alpha_t \gamma_{\max}  \mu_{t}  +  \alpha_t \alpha_{t-\tau(\alpha_t)} \tau(\alpha_t)  \gamma_{\max} \zeta_8 )  \mathbf{E}[\| \langle \tilde \theta \rangle_t- \theta^* \|_2^2] \nonumber \\
    & \;\;\; + \alpha_t  \alpha_{t -\tau(\alpha_t) }  \tau(\alpha_t) \gamma_{\max} \zeta_9 +  \alpha_t \gamma_{\max}  \mu_{t} \nonumber. 
\end{align}
Moreover, from $\alpha_t = \frac{\alpha_0}{t+1}$, $\alpha_0\ge \frac{\gamma_{max}}{0.9}$ and the definition of $\bar T$, for all $t \ge \bar T$ we have
\begin{align}
    \mathbf{E}[H( \langle \tilde \theta \rangle_{t+1} )] 
    & \le (1-\frac{0.9\alpha_t} {\gamma_{\max}})\mathbf{E}[H( \langle \tilde \theta \rangle_t )] + \alpha_t  \alpha_{t -\tau(\alpha_t) }  \tau(\alpha_t) \gamma_{\max} \zeta_9 +  \alpha_t \gamma_{\max} \mu_{t} \nonumber\\
    & \le \frac{t}{t+1}  \mathbf{E}[H( \langle \tilde \theta \rangle_t )] +  \alpha_0 \gamma_{\max}   \frac{ \mu_{t} }{t+1}  +  \frac{\alpha_0^2 C\log(\frac{t+1}{\alpha_0})  \gamma_{\max}  \zeta_9  }{(t+1)(t-\tau(\alpha_t)+1)}  \nonumber \\
    & \le \frac{\bar T}{t+1}  \mathbf{E}[H( \langle \tilde \theta \rangle_{\bar T} )] +  \alpha_0 \gamma_{\max}  \sum_{l = \bar T}^t \left( \frac{ \mu_{l} }{l+1} + \frac{\alpha_0 \zeta_9 C\log(\frac{l+1}{\alpha_0})  }{(l+1)(l-\tau(\alpha_l)+1)} \right) \Pi_{u=l+1}^t\frac{u}{u+1} \nonumber\\
    & = \frac{\bar T}{t+1}  \mathbf{E}[H( \langle \tilde \theta \rangle_{\bar T} )] +  \alpha_0 \gamma_{\max} \sum_{l = \bar T}^t \frac{ \mu_{l} }{t+1}  +   \frac{\alpha_0^2  \gamma_{\max}  \zeta_9 }{t+1}
    \sum_{l=\bar T}^t \frac{ C\log(\frac{l+1}{\alpha_0})  }{l-\tau(\alpha_l)+1} ,  \nonumber %\\
    %& \le \frac{\bar T}{t+1}  \mathbf{E}[H( \langle \tilde \theta \rangle_{\bar T} )] +  \alpha_0 \gamma_{\max}  \frac{\sum_{l = \bar T}^t  \mu_{l}  }{t+1}  +   \frac{\alpha_0^2  \gamma_{\max}  \zeta_9 }{t+1}
    %\sum_{l=\bar T}^t \frac{ 2 C\log(\frac{l+1}{\alpha_0})  }{l+1}   \nonumber \\
    %& \le \frac{\bar T}{t+1}  \mathbf{E}[H( \langle \tilde \theta \rangle_{\bar T} )] + \alpha_0 \gamma_{\max}  \frac{\sum_{l = \bar T}^{t+1}  \mu_{l}  }{t+1} +  \frac{\zeta_9 \alpha_0 \gamma_{\max}  C \log^2(\frac{t+1}{\alpha_0})}{t+1},  \label{eq:timevarying_singlebound_1_Push_SA}
\end{align}
which implies that
\begin{align}
    \mathbf{E}[H( \langle \tilde \theta \rangle_{t+1} )] 
    %& \le (1-\frac{0.9\alpha_t} {\gamma_{\max}})\mathbf{E}[H( \langle \tilde \theta \rangle_t )] + \alpha_t  \alpha_{t -\tau(\alpha_t) }  \tau(\alpha_t) \gamma_{\max} \zeta_9 +  \alpha_t \gamma_{\max} \mu_{t} \nonumber\\
    %& \le \frac{t}{t+1}  \mathbf{E}[H( \langle \tilde \theta \rangle_t )] +  \alpha_0 \gamma_{\max}   \frac{ \mu_{t} }{t+1}  +  \frac{\alpha_0^2 C\log(\frac{t+1}{\alpha_0})  \gamma_{\max}  \zeta_9  }{(t+1)(t-\tau(\alpha_t)+1)}  \nonumber \\
    %& \le \frac{\bar T}{t+1}  \mathbf{E}[H( \langle \tilde \theta \rangle_{\bar T} )] +  \alpha_0 \gamma_{\max}  \sum_{l = \bar T}^t \left( \frac{ \mu_{l} }{l+1} + \frac{\alpha_0 \zeta_9 C\log(\frac{l+1}{\alpha_0})  }{(l+1)(l-\tau(\alpha_l)+1)} \right) \Pi_{u=l+1}^t\frac{u}{u+1} \nonumber\\
    %& = \frac{\bar T}{t+1}  \mathbf{E}[H( \langle \tilde \theta \rangle_{\bar T} )] +  \alpha_0 \gamma_{\max} \sum_{l = \bar T}^t \frac{ \mu_{l} }{t+1}  +   \frac{\alpha_0^2  \gamma_{\max}  \zeta_9 }{t+1}
    %\sum_{l=\bar T}^t \frac{ C\log(\frac{l+1}{\alpha_0})  }{l-\tau(\alpha_l)+1} ,  \nonumber \\
    & \le \frac{\bar T}{t+1}  \mathbf{E}[H( \langle \tilde \theta \rangle_{\bar T} )] +  \alpha_0 \gamma_{\max}  \frac{\sum_{l = \bar T}^t  \mu_{l}  }{t+1}  +   \frac{\alpha_0^2  \gamma_{\max}  \zeta_9 }{t+1}
    \sum_{l=\bar T}^t \frac{ 2 C\log(\frac{l+1}{\alpha_0})  }{l+1}   \nonumber \\
    & \le \frac{\bar T}{t+1}  \mathbf{E}[H( \langle \tilde \theta \rangle_{\bar T} )] + \alpha_0 \gamma_{\max}  \frac{\sum_{l = \bar T}^{t+1}  \mu_{l}  }{t+1} +  \frac{\zeta_9 \alpha_0 \gamma_{\max}  C \log^2(\frac{t+1}{\alpha_0})}{t+1},  \label{eq:timevarying_singlebound_1_Push_SA}
\end{align}
where we use
$
    \sum_{l=\bar T}^t \frac{2 \alpha_0 \log(\frac{l+1}{\alpha_0}) }{l+1} \le \log^2(\frac{t+1}{\alpha_0})
$
to get the last inequality. Then, we can get the bound of $ \mathbf{E}[\| \langle \tilde \theta \rangle_{t+1} -\theta^* \|_2^2 ]  $ from \eqref{eq:timevarying_singlebound_1_Push_SA} as follows:
\begin{align*}
    &\;\;\;\; \mathbf{E}[\|\langle \tilde \theta \rangle_{t+1} -\theta^* \|_2^2 ] 
     \le \frac{1}{\gamma_{\min}} \mathbf{E}[H( \langle \tilde \theta \rangle_{t+1} )] \nonumber\\
    & \le \frac{\bar T}{t+1} \frac{\gamma_{\max}}{\gamma_{\min}} \mathbf{E}[\|\langle \tilde \theta \rangle_{\bar T} -\theta^* \|_2^2 ] 
    +  \frac{\zeta_9 \alpha_0  C \log^2(\frac{t+1}{\alpha_0})}{t+1} \frac{\gamma_{\max}}{\gamma_{\min}}
    +  \alpha_0 \frac{\gamma_{\max}}{\gamma_{\min}} \frac{\sum_{l = \bar T}^{t+1}  \mu_{l} }{t+1}.
\end{align*}
This completes the proof.
\hfill $\qed$

We are now in a position to prove Theorem~\ref{thm:bound_time-varying_step_Push_SA}.

\noindent
{\bf Proof of Theorem~\ref{thm:bound_time-varying_step_Push_SA}:}
Note that
    $
        \sum_{i=1}^N \mathbf{E}[\|\theta_{t+1}^i - \theta^*\|_2^2]
         \le 2 \sum_{i=1}^N  \mathbf{E}[\|\theta_{t+1}^i - \langle \tilde \theta \rangle_t \|_2^2 ] + 2 N \mathbf{E} [\| \langle \tilde \theta \rangle_t - \theta^*\|_2^2] .
    $
From Lemmas~\ref{lemma:bound_consensus_time-varying_push_SA} and \ref{lemma:bound_average_time-varying_push_SA}, we have for any $t \ge \bar T$,
    \begin{align*}
        \sum_{i=1}^N \mathbf{E}\left[\left\|\theta_{t+1}^i - \theta^*\right\|_2^2\right] 
        & \le \frac{16}{\epsilon_1}  \bar\epsilon^t \mathbf{E}[ \| \sum_{i=1}^N \tilde \theta_0^i + \alpha_0 A(X_0)\theta_0^i + \alpha_0 b^i(X_0) \|_2]  + 2 \alpha_t  A_{\max} C_\theta + 2 \alpha_t b_{\max} \\
        & \;\;\; + \frac{16}{\epsilon_1} \frac{ A_{\max} C_\theta + b_{\max}}{1-\bar\epsilon} \left(  \alpha_0 \bar\epsilon^{t/2}  +  \alpha_{\ceil{\frac{t}{2}}} \right)+ \frac{2\bar TN}{t} \frac{\gamma_{\max}}{\gamma_{\min}} \mathbf{E}[\|\langle \tilde \theta \rangle_{\bar T} -\theta^* \|_2^2 ] \\
        &\;\;\; 
        +  \frac{2 N \zeta_9 \alpha_0  C \log^2(\frac{t}{\alpha_0})}{t} \frac{\gamma_{\max}}{\gamma_{\min}}
        +  2 \alpha_0 N \frac{\gamma_{\max}}{\gamma_{\min}} \frac{\sum_{l = \bar T}^{t}  \mu_{l}  }{t} \\
        &\le   C_7 \bar\epsilon^t  +  C_8 \left(  \alpha_0 \bar\epsilon^{\frac{t}{2}}  +  \alpha_{\ceil{\frac{t}{2}}} \right)+ C_9 \alpha_t + \frac{1}{t}\bigg(C_{10} \log^2\Big(\frac{t}{\alpha_0}\Big) + C_{11}\sum_{l = \bar T}^{t}  \mu_{l}  +C_{12}\bigg).
    \end{align*}
This completes the proof.
\hfill $\qed$

{\color{black}
We next show the asymptotic performance of \eqref{eq:SA_push-sum}.

%\noindent
{\bf Proof of Theorem~\ref{thm:push_meansq}:}
%{\color{red}Part (1) can be proved by specialization of Theorem~3.1 in \cite{kushner87}.}
From Lemma~\ref{lemma:bound_consensus_time-varying_push_SA}, since $\bar \epsilon \in(0,1)$ and $\alpha_t = \frac{\alpha_0}{t}$, it follows that $\lim_{t\to\infty}\| \theta_{t+1}^i - \langle \tilde \theta \rangle_t \|_2 = 0$, which implies that all $\theta_{t+1}^i$, $i\in \mathcal{V}$, will reach a consensus with $ \langle \tilde \theta \rangle_t $. The update of $ \langle \tilde \theta \rangle_t $ is given in \eqref{eq:update_average_tilde_theta}, which can be treated as a single-agent linear stochastic approximation whose corresponding ODE is \eqref{eq:ode_pushsum}.
In addition, from Theorem~\ref{thm:bound_time-varying_step_Push_SA} and Lemma~\ref{lemma:eta_limit_Push_SA}, 
$\lim_{\rightarrow\infty}\sum_{i=1}^N \mathbf{E}[\|\theta_{t+1}^i - \theta^*\|_2^2]=0$, it follows that $\theta_{t+1}^i$ will converge to $\theta^*$ in mean square for all $i\in\mathcal{V}$.
\hfill $\qed$

\begin{remark}
Finite-time analysis for such a push-based distributed algorithm is challenging. 
Almost all, if not all, the existing push-based distributed optimization works build on the analysis in \cite{nedic}; however, that analysis assumes that a convex combination of the entire history of the states of each agent (and not merely the current state of the agent) is being calculated. This assumption no longer holds in our case.  To obtain a direct finite-time error bound without this assumption, we appeal to a new approach to analyze our push-based SA algorithm by leveraging our consensus-based analyses to establish direct finite-time error bounds for stochastic approximation. 
Specifically, we tailor an absolute probability sequence for the push-based stochastic approximation algorithm and exploit its properties (Lemma~\ref{lemma:push-sum_pi_intfty}). 
\hfill$\Box$
\end{remark}

} \label{sec:proof_push}

%==========
%We begin with the proof of asymptotic performance of \eqref{eq:SA_push-sum}.
%\noindent
%{\bf Proof of Theorem~\ref{thm:push_meansq}:}
%%{\color{red}Part (1) can be proved by specialization of Theorem~3.1 in \cite{kushner87}.}
%From Lemma~\ref{lemma:bound_consensus_time-varying_push_SA}, since $\bar \epsilon \in(0,1)$ and $\alpha_t = \frac{\alpha_0}{t}$, we have $\lim_{t\to\infty}\| \theta_{t+1}^i - \langle \tilde \theta \rangle_t \|_2 = 0$, which implies that all $\theta_{t+1}^i$, $i\in \mathcal{V}$, will reach a consensus with $ \langle \tilde \theta \rangle_t $. The update of $ \langle \tilde \theta \rangle_t $ is given in \eqref{eq:update_average_tilde_theta}, which can be treated as a single-agent linear stochastic approximation whose corresponding ODE is \eqref{eq:ode_pushsum}.
%In addition, from  Theorem~\ref{thm:bound_time-varying_step_Push_SA} and Lemma~\ref{lemma:eta_limit_Push_SA}, 
%$\lim_{\rightarrow\infty}\sum_{i=1}^N \mathbf{E}[\|\theta_{t+1}^i - \theta^*\|_2^2]=0$, it follows that $\theta_{t+1}^i$ will converge to $\theta^*$ in mean square for all $i\in\scr V$.
%\hfill $\qed$

\bibliographystyle{unsrt}
\bibliography{references}
%\newpage

%\appendix
%\input{appendix}

\end{document}